\documentclass[12pt]{article}
\usepackage{graphicx} 
\usepackage{amsmath, amsthm, amsfonts,amssymb}
\usepackage{bm}
\usepackage{indentfirst}
\usepackage{amsmath}
\usepackage{xr}
\usepackage{mathrsfs}
\usepackage{subfiles}
\usepackage{verbatim}
\usepackage{float}
\usepackage{caption}
\usepackage{subcaption}
\usepackage{algorithm}
\usepackage{algorithmic}
\usepackage{longtable}
\usepackage{multirow}
\usepackage{multicol}
\usepackage{xcolor}
\usepackage{latexsym}
\usepackage{amsthm}
\usepackage{amsthm,amsmath,amssymb}
\usepackage[left=1in,right=1in,top=1in,bottom=1in]{geometry}
\usepackage{mathrsfs}
\usepackage{natbib}
\usepackage{booktabs}
\usepackage{siunitx}
\usepackage{adjustbox}

\setcitestyle{authoryear,round}
\usepackage[colorlinks, citecolor=blue]{hyperref}
\newtheorem{theorem}{Theorem}

\newtheorem{lemma}{Lemma}

\newtheorem{assumption}{Assumption}
\newcommand{\bE}{\mathbb{E}}

\newcommand{\Err}{\mathrm{Err}}

\newtheorem{proposition}{Proposition}
\newtheorem{remark}{Remark}
\newtheorem{fullthminner}{Theorem}

\newenvironment{fulltheorem}[1]{%
    \renewcommand{\thefullthminner}{\ref*{#1}$'$}%
    \begin{fullthminner}%
}{%
    \end{fullthminner}%
}

\title{Synthetic Data Augmentation for Fairness under Data Imbalance: Theory and Adaptive Allocation}

\author{Zhengchi Ma\thanks{Department of Electrical \& Computer Engineering, Duke University},~ Kaifeng Zhang\thanks{Department of Statistical Science, Duke University} ~ and~ Anru R. Zhang\thanks{Department of Biostatistics \& Bioinformatics and Department of Computer Science, Duke University}}
\date{}

\begin{document}

\maketitle

\begin{abstract}
    Data imbalance across sensitive groups and class labels is common in classification
problems and can contribute to substantial fairness disparities. We study how such
imbalance affects Equal Opportunity and how synthetic data augmentation can mitigate
the resulting unfairness. We first identify two mechanisms, prevalence imbalance and
group-size imbalance, through which changes in the training distribution shift the
learned decision boundary and induce disparities in group true positive rates. Motivated
by this analysis, we study exact full-balance augmentation, which equalizes all
group-label cells. We establish conditions under which full balancing strictly improves
Equal Opportunity together with a prediction-error comparison under the oracle balanced distribution, and show that the Equal
Opportunity gap of the synthetic-data classifier converges to that of the oracle balanced classifier as the sample size grows and the generator error vanishes. The finite-sample convergence rate agrees, up to logarithmic factors and generator-approximation error, with the information-theoretic lower-bound rate under the corresponding regularity regimes. Since exact balancing need not provide the best fairness-accuracy trade-off in
practice, we further propose Accuracy-Constrained Adaptive Synthetic Augmentation
(ACASA), which learns both where and how much to augment by minimizing estimated
fairness disparity subject to a constraint on predictive accuracy. We provide a
finite-sample guarantee comparing the selected augmentation allocation with the
population-optimal accuracy-feasible allocation. Simulation studies and  real data application support the
theoretical findings and demonstrate that synthetic augmentation can substantially
reduce fairness disparities while achieving favorable empirical fairness-accuracy trade-offs.
\end{abstract}

\begin{sloppypar}

\section{Introduction}\label{sec:intro}

Algorithmic fairness has become an important consideration when predictive models are deployed across heterogeneous populations, particularly when performance differs across sensitive groups.
These concerns are particularly consequential in applications such as healthcare \citep{obermeyer2019dissecting}, clinical measurement and diagnosis \citep{sjoding2020racial}, criminal-justice decision-making \citep{kleinberg2018human}, and consumer lending \citep{fuster2022predictably}.
A substantial literature has investigated how to define, assess, and mitigate disparities in algorithmic decisions \citep{mitchell2021algorithmic, mehrabi2021survey, pessach2022review, das2023algorithmic, he2025fairness}.
One important source of unfairness is data imbalance, which can arise from unequal representation of sensitive groups or unequal class-label distributions across groups. Such imbalance may cause overrepresented groups or classes to exert greater influence on model training and contribute to disparities in predictive performance \citep{kleinberg2016inherent, suresh2021framework, chai2022fairness}. A variety of fairness criteria have been proposed to quantify such disparities, including demographic parity, equalized odds, and Equal Opportunity, with Equal Opportunity requiring equal true positive rates across sensitive groups \citep{hardt2016equality, dwork2012fairness, kusner2017counterfactual,kearns2018preventing}. Existing approaches for mitigating unfairness are commonly categorized as pre-processing methods that modify the training data, such as resampling or reweighting \citep{kamiran2012data,zemel2013learning,feldman2015certifying, calmon2017optimized,xu2023fair,chen2024learning}, in-processing methods that incorporate fairness objectives or constraints during model training \citep{zafar2017fairness,agarwal2018reductions,donini2018empirical,scutari2022achieving, do2023joint}, and post-processing methods that adjust model predictions or decision thresholds after training \citep{menon2018cost, chzhen2019leveraging, chen2024post,hou2024finite,zhang2024fair,xian2024unified,zeng2026bayes}.
Several works have explored synthetic data generation as a fairness intervention \citep{xu2018fairgan,sattigeri2019fairness,sonoda2023fair, abroshan2024imposing}. Related work by \citet{zhang2026fda} considers fairness improvement through the generation of synthetic outcome labels, rather than synthetic covariates as studied in the present paper.

Synthetic data augmentation provides a flexible approach for enriching observed datasets by generating additional samples that approximate the underlying data distribution.
A line of statistical research has investigated synthetic-data quality, distributional fidelity, statistical utility, and efficiency \citep{raab2016practical,snoke2018general,raghunathan2021synthetic,awan2020one}. Another substantial literature has studied statistical estimation, uncertainty quantification, and inferential validity based on synthetic data \citep{wilde2021foundations,yu2022semiparametric,gu2023synthetic,mathur2024fully}.
Synthetic observations can be produced using classical parametric and resampling-based methods like Bootstrap \citep{efron1994introduction} or SMOTE \citep{chawla2002smote}, as well as modern generative models such as variational autoencoders (VAEs) \citep{kingma2013auto}, generative adversarial networks (GANs) \citep{goodfellow2014generative}, and diffusion models \citep{ho2020denoising}. Statistical convergence results for learned generative distributions provide complementary guidance on approximation error \citep{liang2021well,huang2022error,lee2023convergence,oko2023diffusion,puchkin2024rates, zhang2024minimax,tang2024conditional,chakraborty2025statistical}. Recent statistical work has begun to study synthetic augmentation itself as a statistical procedure. \citet{shen2023boosting} investigated how synthetic volume affects statistical performance in the presence of generation error, while \citet{tian2025conditional} proposed conditional synthesis to target underrepresented regions. Related theoretical work examines synthetic oversampling for imbalanced learning \citep{nakada2024synthetic}, bias correction under imperfect synthesis \citep{lyu2025bias}, and the effects of augmentation on classification error, score-based evaluation, and optimal augmentation amounts  \citep{ma2026synthetic,ma2026does}.
Complementary statistical approaches analyze imbalanced classification through transfer-learning formulations \citep{xia2026classification}. Beyond prediction, synthetic data have also been studied as a basis for statistical inference \citep{raisa2025consistent}. By generating observations from selected portions of the data distribution, synthetic augmentation offers greater flexibility than simply duplicating or discarding existing observations and provides a natural mechanism for modifying the composition of a training sample.

\begin{figure}
    \centering
    \includegraphics[width=0.9\linewidth]{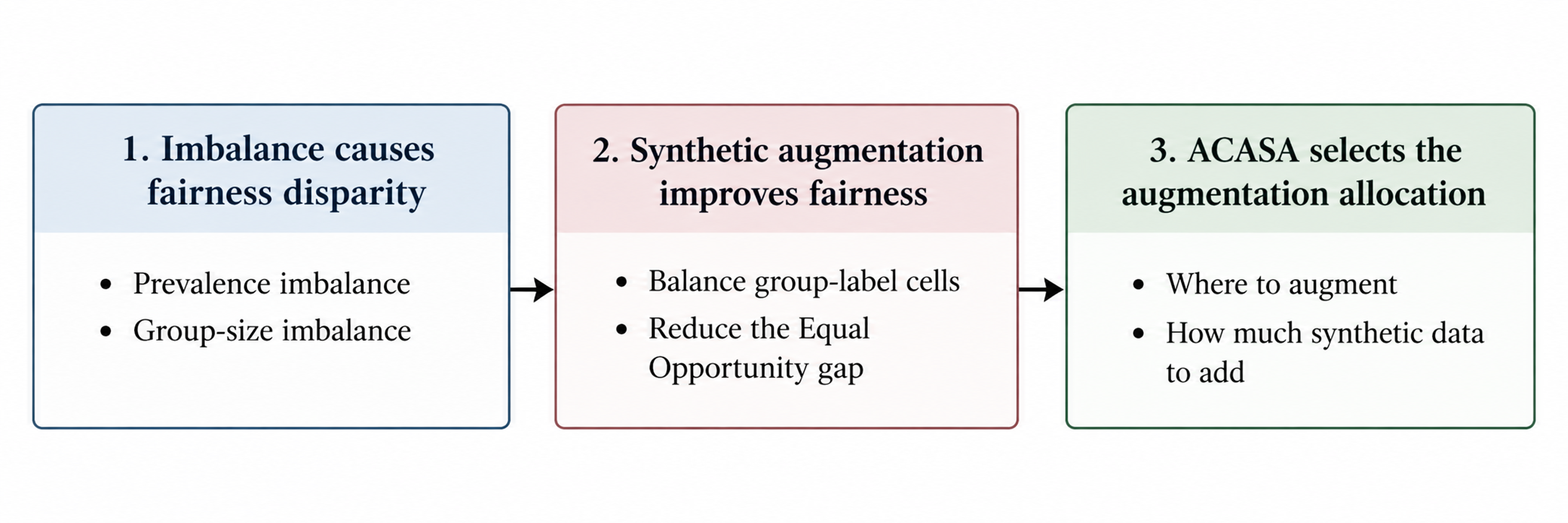}
    \caption{Overview of the main ideas and structure of the paper.}
    \label{fig:flowchartmainidea}
\end{figure}

These two lines of literature intersect naturally when fairness disparities are driven by imbalances across sensitive groups and class labels. By selectively increasing the representation of particular group--label subpopulations, synthetic data augmentation provides a flexible pre-processing mechanism for modifying the training distribution and potentially mitigating such disparities. Its effect on fairness, however, is inherently nontrivial. Augmentation changes the relative weights of different subpopulations in the learning objective and can therefore shift the learned decision boundary, while discrepancies between the synthetic and target distributions may introduce additional bias. Moreover, the augmentation allocation that most effectively reduces a fairness disparity need not correspond to exact balancing, especially when predictive accuracy must be maintained. These considerations motivate a systematic study of how data imbalance gives rise to group-level disparities, when synthetic augmentation can reduce them, and how synthetic samples should be allocated across subpopulations to achieve a favorable fairness--accuracy trade-off. In this paper, we investigate these questions under Equal Opportunity for classification through a cellwise augmentation framework that permits group- and label-specific synthetic generation. Figure \ref{fig:flowchartmainidea} provides a flowchart summarizing the main ideas and overall structure of our paper.

\subsection{Our Contributions}

Our work develops a theoretical and methodological framework for understanding and mitigating fairness disparities induced by data imbalance through synthetic data augmentation. We first identify two distinct mechanisms through which the composition of the training data can affect Equal Opportunity: prevalence imbalance, arising from differences in class-label distributions across sensitive groups, and group-size imbalance, arising from unequal representation of the sensitive groups. For both mechanisms, we characterize how changes in the training distribution shift the learned decision boundary and establish lower bounds on the resulting Equal Opportunity gap. Motivated by this analysis, we then study an exact full-balance intervention that augments each group--label cell to the size of the largest cell. We show that, under suitable conditions, full balancing yields a strictly positive improvement in Equal Opportunity and establish a prediction-error comparison under the oracle balanced distribution, explicitly accounting for imperfect synthetic generation. We further develop a statistical theory for balanced synthetic augmentation by characterizing how the Equal Opportunity gap of a classifier trained on finite synthetic data approaches that of the ideal balanced-population classifier. We also establish the information-theoretic minimax lower bound, showing that the rate cannot in general be improved uniformly over the considered distribution class. Finally, because exact balancing need not yield the most favorable fairness--accuracy trade-off in practice, we propose Accuracy-Constrained Adaptive Synthetic Augmentation (ACASA), which learns both where and how much to augment across group--label cells by minimizing estimated Equal Opportunity disparity subject to a constraint on predictive accuracy. The procedure evaluates candidate augmentation allocations through cross-validation on real validation observations, and we provide a finite-sample guarantee comparing the selected allocation with the population-optimal accuracy-feasible allocation. Simulation studies and real-data analysis further illustrate that synthetic augmentation can substantially reduce fairness disparities while achieving favorable empirical fairness--accuracy trade-offs.

\subsection{Organization}
The remainder of the paper is organized as follows. Section \ref{sec:setup} presents the general problem setup
and cellwise augmentation framework. Section \ref{sec:imbalance} studies how prevalence imbalance and
group-size imbalance can induce Equal Opportunity disparities. Section \ref{sec:full-balance} analyzes
exact full-balance synthetic augmentation and establishes fairness-improvement and balanced-distribution prediction-error guarantees. Section \ref{sec:convergence-minimax} develops the statistical theory of full
balancing, including a finite-sample convergence result and a minimax lower
bound that matches the upper bound up to logarithmic factors and generator-approximation error. 
Section \ref{sec:methodology} introduces the proposed Accuracy-Constrained Adaptive Synthetic
Augmentation (ACASA) procedure and provides a finite-sample guarantee for the
selected augmentation allocation. Section \ref{sec:numerica-studies} presents
simulation studies and real data analysis illustrating the theoretical findings and empirical performance of
the proposed approach. Section~\ref{sec:discussion} concludes with discussions. Detailed assumptions, complete finite sample versions of the main results, auxiliary lemmas, additional discussion, implementation details, and proofs are collected in the Appendix.

\section{Problem Setup}\label{sec:setup}

Consider binary classification with two groups defined by a sensitive attribute. Differences in group sizes and class-label distributions can make a classifier trained on imbalanced data perform differently across groups. We study how synthetic data augmentation can reduce the resulting disparity in true positive rates (Equal Opportunity) while maintaining predictive performance.

\paragraph{Data and classifiers.}
Let $S\in\{A,B\}$ denote the group, $Y\in\{0,1\}$ the label, and $X\in\mathcal X$ the features, with true distribution $P$ of $(X,Y,S)$. For a generic training distribution $Q$, let
\[
\eta_Q(x):=Q(Y=1\mid X=x),\qquad \rho_Q:=\frac{Q(Y=1)}{1-Q(Y=1)}
\]
be the regression function and the positive-class odds. We fit a parametric score $f_\theta:\mathcal X\to[0,1]$, $\theta\in\Theta\subseteq\mathbb R^d$, by minimizing a loss $\ell(\theta;x,y)=\ell_0(f_\theta(x),y)$ such as the cross-entropy; the score uses $X$ only, so the same rule is applied to both groups. Let $\theta^*_Q$ minimize the population loss $L_Q(\theta):=\mathbb E_Q[\ell(\theta;X,Y)]$ and let $\widehat\theta_Q$ minimize the empirical loss $\widehat L_Q$ over $n_Q$ training observations: an i.i.d.\ sample from $Q$ or, for an augmented distribution, the pooled real and synthetic sample. We use three classifiers:
\[
h_Q^\eta(x):=\mathbf 1\{\eta_Q(x)\ge 1/2\},
\qquad
h_Q^*(x):=\mathbf 1\{f_{\theta_Q^*}(x)\ge 1/2\},
\qquad
\widehat h_Q(x):=\mathbf 1\{f_{\widehat\theta_Q}(x)\ge 1/2\}.
\]
The Bayes classifier $h^\eta_Q$ is used to describe mechanisms (Sections~\ref{sec:imbalance}--\ref{sec:full-balance}). The population-risk classifier $h_Q^*$ and the empirical classifier $\widehat h_Q$ enter when we study estimation (Sections~\ref{sec:convergence-minimax}--\ref{sec:methodology}).

\paragraph{Evaluation criteria.}
For a classifier $h$, let $\mathrm{TPR}_s(h):=P(h(X)=1\mid Y=1,S=s)$ and define the Equal Opportunity (EO) gap
\[
\Delta_{\mathrm{EO}}(h):=\bigl|\mathrm{TPR}_A(h)-\mathrm{TPR}_B(h)\bigr|.
\]
Fairness is always evaluated under the true distribution $P$, whatever distribution the classifier was trained on; for a data-dependent classifier, the probabilities are over an independent test observation. The prediction error under a distribution $Q$ is $\Err_Q(h):=Q(h(X)\neq Y)$; which $Q$ is appropriate depends on the goal and is stated with each result.

\paragraph{Imbalance and cells.}
Let $\alpha:=P(S=A)$, $\pi_s:=P(Y=1\mid S=s)$, $\rho_s:=\pi_s/(1-\pi_s)$, and $\bar\pi:=P(Y=1)=\alpha\pi_A+(1-\alpha)\pi_B$; for a generic $Q$ we write $\pi_s(Q):=Q(Y=1\mid S=s)$. The four group--label cells $\mathcal C:=\{A+,A-,B+,B-\}$, where $A+$ denotes $(S,Y)=(A,1)$ and so on, have probabilities $p_c$ (e.g.  $p_{B+}=(1-\alpha)\pi_B$) and conditional distributions $P_c:=P(\cdot\mid(S,Y)=c)$, so that $P=\sum_c p_cP_c$. A real sample of size $n_P$ has $n_c$ observations in cell $c$. Let $n_{\max}:=\max_c n_c$.

\paragraph{Cellwise synthetic augmentation.}
A generator for cell $c$, fitted on the real sample, produces observations from a distribution $P_{\mathrm{syn},c}$ supported on cell $c$, with total variation distance
\(
d_{\mathrm{TV}}(P_{\mathrm{syn},c},P_c)\le\epsilon_c.
\)
For nonnegative weights $\mathbf a=(a_c)_{c\in\mathcal C}$, the oracle and synthetic augmented distributions are
\[
\widetilde P_{\mathbf a}:=\frac{P+\sum_c a_cP_c}{1+\sum_c a_c},
\qquad
\overline P_{\mathbf a}:=\frac{P+\sum_c a_cP_{\mathrm{syn},c}}{1+\sum_c a_c}.
\]
Adding $m_c$ synthetic observations to cell $c$ corresponds to $a_c=m_c/n_P$. On distributions, a tilde denotes oracle augmentation and an overline synthetic augmentation. $\widetilde P_{\mathbf a}$ isolates the effect of reweighting the cells, and the gap between $\widetilde P_{\mathbf a}$ and $\overline P_{\mathbf a}$ is the cost of imperfect generation. Regularity conditions are collected in Appendix~\ref{sec:assumptiondetails}.

\section{How Data Imbalance Leads to Fairness Disparities}\label{sec:imbalance}

\begin{figure}[t]
    \centering
    \begin{subfigure}[t]{0.45\textwidth}
        \centering
        \includegraphics[width=\linewidth]{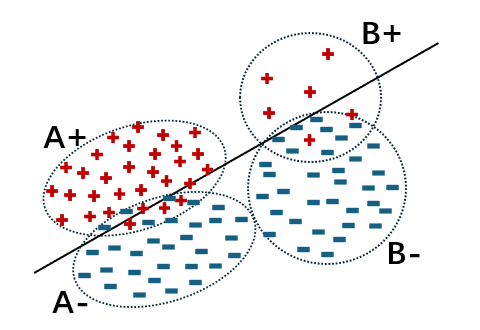}
    \end{subfigure}
    \begin{subfigure}[t]{0.45\textwidth}
        \centering
        \includegraphics[width=\linewidth]{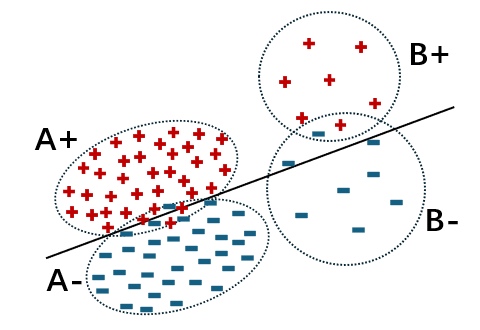}
    \end{subfigure}
    \caption{Illustration of how data imbalance can induce fairness disparities.
Left: prevalence imbalance, with a scarce $B+$ cell. Right: group-size imbalance,
with group $B$ underrepresented. Dotted contours represent the underlying
true distributions, and the solid line represents the learned decision boundary.}
    \label{fig:imbalance-intuition}
\end{figure}

We identify two mechanisms through which imbalance creates an Equal Opportunity gap, labeling as B the group it disadvantages: prevalence imbalance (the groups have different base rates) and group-size imbalance (one group is underrepresented).
Figure~\ref{fig:imbalance-intuition} shows the intuition. When the $B+$ cell (left) or group $B$ as a whole (right) has small probability
mass, the decision boundary is fitted mainly to the other cells, even in large
samples, and can cut through the positive observations of group $B$, lowering
its true positive rate.

\paragraph{Positive-prediction region.}
For a distribution $Q$, let $q_{Q,y}$ be the density of $X$ given $Y=y$ under $Q$
with respect to a common dominating measure, and let
$\Phi_Q(x):=\log\{q_{Q,1}(x)/q_{Q,0}(x)\}$ be the log-likelihood ratio of the two
classes. By Bayes' rule, the posterior log-odds are
\[
\Psi_Q(x):=\log\frac{\eta_Q(x)}{1-\eta_Q(x)}=\log\rho_Q+\Phi_Q(x).
\]
The Bayes classifier predicts positive exactly on the positive-prediction region $\mathcal X_Q^+:=\{x:\Psi_Q(x)\ge 0\}$, so that $h_Q^\eta=\mathbf 1\{x\in\mathcal X_Q^+\}$ and
\[
\mathrm{TPR}_s(h_Q^\eta)=P_{s+}(\mathcal X_Q^+),
\qquad
P_{s+}:=P(X\in\cdot\mid S=s,\,Y=1),
\]
where $P_{A+}$ and $P_{B+}$ are the cell distributions of the positives of each group. 
 
\begin{figure}[htbp]
    \centering

    \begin{subfigure}[t]{0.48\textwidth}
        \centering
        \includegraphics[width=\linewidth]{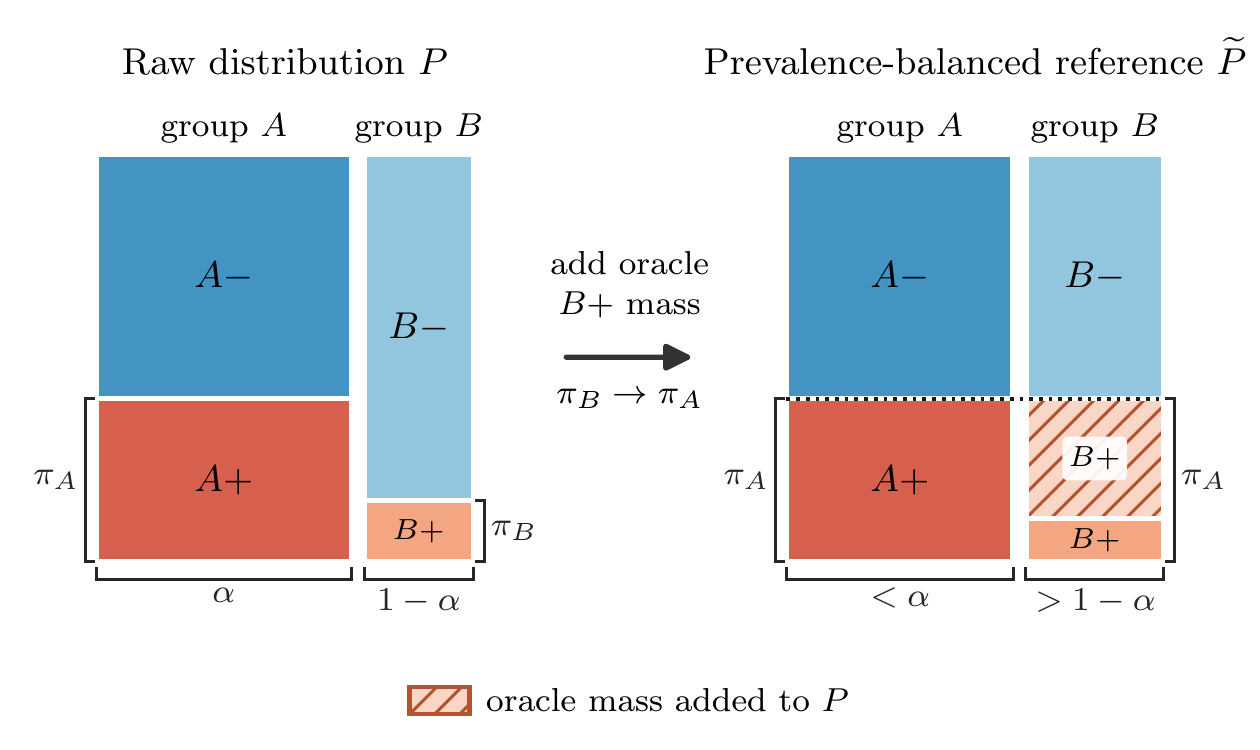}
        \caption{Prevalence balanced reference distribution}
        \label{fig:prevalence}
    \end{subfigure}
    \hfill
    \begin{subfigure}[t]{0.48\textwidth}
        \centering
        \includegraphics[width=\linewidth]{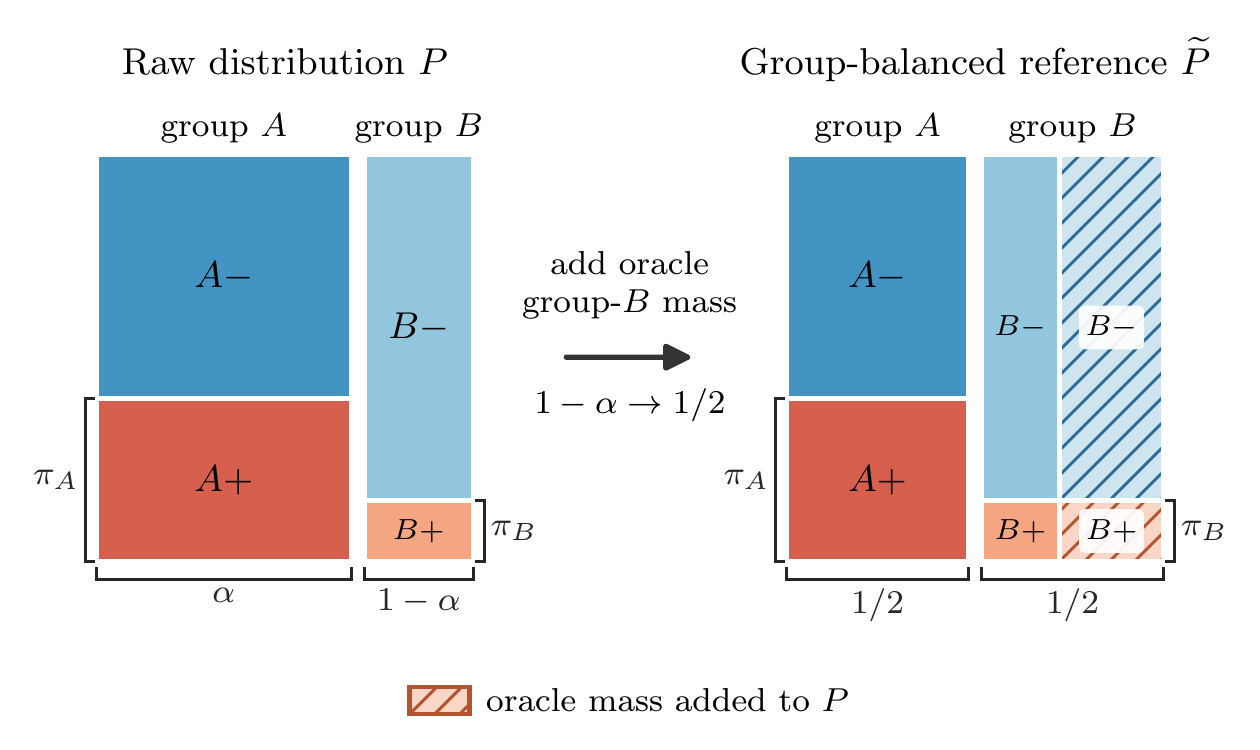}
        \caption{Group-size balanced reference distribution}
        \label{fig:group_size}
    \end{subfigure}

    \caption{Comparison of the raw distribution with two balanced reference distributions.}
    \label{fig:balance_comparison}
\end{figure}

\paragraph{Common mechanism.}
Both results below compare $P$ with an oracle reference $\widetilde P$ in which one source of imbalance is removed, as shown in Figure \ref{fig:balance_comparison}. Each result argues in three steps:
(1)~imbalance shrinks the positive-prediction region, $\mathcal X_P^+\subseteq\mathcal X_{\widetilde P}^+$, so the reference additionally classifies as positive the set $\mathcal D:=\mathcal X_{\widetilde P}^+\setminus\mathcal X_P^+$;
(2)~under a boundary-dominance condition, group $B$ is overrepresented among the positives falling in $\mathcal D$, so the stricter classifier loses true positives of group $B$ at a higher rate than of group $A$,
$P_{B+}(\mathcal D)-P_{A+}(\mathcal D)\gtrsim P(X\in\mathcal D\mid Y=1)$, while a
sign condition, $\mathrm{TPR}_A(h^\eta_{\widetilde P})\ge\mathrm{TPR}_B(h^\eta_{\widetilde P})$,
keeps the sign of the gap;
(3)~$\mathcal D$ contains all observations whose posterior log-odds lie in a band of positive width just below $0$, and under a density lower bound on $\Psi_P(X)\mid Y=1$ this band carries probability proportional to its width.
The results are stated for Bayes classifiers, and finite-sample versions for $\widehat h_P$ are in Section~\ref{sec:finite-sample-bridge} and Appendix~\ref{sec:imbalancemechanisms}.

\subsection{Effect of Prevalence Imbalance}\label{sec:prevalence-imbalance}
 
Suppose $\pi_A>\pi_B$. The prevalence-balanced reference adds oracle $B+$ mass until group $B$ has the prevalence of group $A$:
\[
\widetilde P=(1-\lambda_1^*)P+\lambda_1^*P_{B+},
\qquad
\lambda_1^*=\frac{(1-\alpha)(\pi_A-\pi_B)}{1-\bar\pi},
\]
so that $\pi_A(\widetilde P)=\pi_B(\widetilde P)=\pi_A$ and the overall odds rise to $\rho_A$, as shown in Figure \ref{fig:prevalence}. Let
\[
\omega_B(x):=P(S=B\mid X=x,\,Y=1),
\qquad
\bar\omega_B:=P(S=B\mid Y=1)=\frac{(1-\alpha)\pi_B}{\bar\pi}
\]
be the local and the overall share of group $B$ among the positives, let $\bar\omega_A:=1-\bar\omega_B$, and let $\mathrm{OR}:=\rho_A/\rho_B>1$ be the odds ratio between the two groups. Because $\widetilde P$ only adds $B+$ observations, its posterior log-odds can be compared with those of $P$ exactly (Lemma~\ref{lemma:log-odds-shift}):
\begin{equation}\label{eq:logodds-shift-main}
\Psi_{\widetilde P}(x)-\Psi_P(x)=\log\bigl(1+(\mathrm{OR}-1)\,\omega_B(x)\bigr)\in\bigl[0,\log\mathrm{OR}\bigr].
\end{equation}
The shift is largest where all positives near $x$ belong to group $B$ and vanishes where none do. It is never negative, so balancing only enlarges the positive-prediction region, $\mathcal X_P^+\subseteq\mathcal X_{\widetilde P}^+$. It combines the change in the class odds with the change of $\Phi$ caused by adding $B+$ mass. We then have the following result.
 
\begin{theorem}[Prevalence imbalance]\label{Thm:fairness-improvement}
Under Assumption~\ref{assumption:prevalence-diff} (Appendix~\ref{section:assumption-1}),
\[
\Delta_{\mathrm{EO}}(h_P^\eta)
\;\ge\;
\Delta_{\mathrm{EO}}(h_P^\eta)-\Delta_{\mathrm{EO}}(h_{\widetilde P}^\eta)
\;\ge\;
\frac{\kappa\,c_\Psi}{\bar\omega_A\bar\omega_B}\,
\log\!\Bigl(1+(\mathrm{OR}-1)\bigl(\bar\omega_B+\kappa\bigr)\Bigr).
\]
\end{theorem}
 
Here the constant $\kappa$ measures how strongly group $B$ is overrepresented among the positives in the band just below the decision boundary, and the constant $c_\Psi$ lower-bounds the density of the posterior log-odds of the positives there. The detailed definitions of the constants can be found in Appendix~\ref{section:assumption-1}. The prevalence gap and the share of group $B$ enter through
\[
(\mathrm{OR}-1)\,\bar\omega_B=\frac{(1-\alpha)(\pi_A-\pi_B)}{\bar\pi(1-\pi_A)}\;\ge\;4(1-\alpha)(\pi_A-\pi_B),
\]
so the right-hand side of the theorem is at least
\[
\frac{\kappa\,c_\Psi}{\bar\omega_A\bar\omega_B}\,\log\bigl(1+4(1-\alpha)(\pi_A-\pi_B)\bigr).
\]
That is, the prevalence gap $\pi_A-\pi_B$, weighted by the share $1-\alpha$ of group $B$, creates an Equal Opportunity gap, and oracle prevalence balancing reduces it by at least this bound. Theorem~\ref{Thm:fairness-improvement-full} (Appendix~\ref{sec:imbalancemechanisms}) gives the same lower bound on $\Delta_{\mathrm{EO}}(\widehat h_P)$ for the empirical classifier, up to a calibration and estimation term.

\subsection{Effect of Group-Size Imbalance}\label{sec:group-imbalance}
 
Suppose instead that group $A$ is overrepresented, $\alpha>1/2$. Keep each group's distribution fixed and let $Q_w$ denote $P$ with the share of group $B$ changed to $w$, so that $Q_{1-\alpha}=P$. The group-balanced reference is $\widetilde P:=Q_{1/2}$, shown in Figure \ref{fig:group_size}. Writing $g_s$ for the density of $X\mid S=s$ and $\eta_s(x):=P(Y=1\mid X=x,S=s)$ for the within-group posterior, a direct calculation (Lemma~\ref{lemma:group-shift}) gives
\begin{equation}\label{eq:group-derivative}
\partial_w\Psi_{Q_w}(x)=\frac{g_A(x)\,g_B(x)\,\bigl(\eta_B(x)-\eta_A(x)\bigr)}{q_w(x,0)\,q_w(x,1)},
\end{equation}
where $q_w(x,y)$ is the density of $(X,Y)$ under $Q_w$. Moving mass towards group $B$ therefore raises the posterior log-odds exactly where $\eta_B>\eta_A$: there a group-$B$ observation is more likely to be positive than a group-$A$ observation with the same features. 
The key quantity is the size of this effect just below the decision boundary
of $P$, measured by a constant $c'$ such that
\(
  \partial_w \Psi_{Q_w}(x) \ge c'\)
  for all \(w\in[1-\alpha,\tfrac12]\) and almost every \(x\)
  with \(-\log\tfrac{\alpha}{1-\alpha} \le \Psi_P(x) < 0\).
\begin{theorem}[Group-size imbalance]\label{thm:imbalance-driven-bound}
Under Assumption~\ref{assumption:group-imb-assmp} (Appendix~\ref{sec:second-assumption}), in particular $c'>0$,
\[
\Delta_{\mathrm{EO}}(h_P^\eta)
\;\ge\;
\Delta_{\mathrm{EO}}(h_P^\eta)-\Delta_{\mathrm{EO}}(h_{\widetilde P}^\eta)
\;\ge\;
\frac{\kappa\,c_\Psi\,c'}{\bar\omega_A\bar\omega_B}\Bigl(\alpha-\frac12\Bigr).
\]
\end{theorem}
 
Group-size imbalance alone creates a gap that grows at least linearly in
$\alpha-1/2$ when $\kappa$, $c_\Psi$ and $c'$ stay bounded away from zero as
$\alpha$ varies, even though each group's distribution is unchanged. The classifier is fitted to the pooled distribution, so the group mixture weights move the posterior and hence the positive-prediction region. Because $\eta_s$ already reflects the prevalence of group $s$, \eqref{eq:group-derivative} accounts for the change in the overall class odds as well. 
When $\pi_B<\pi_A$, the lower prevalence of group $B$ pushes $\eta_B$ below
$\eta_A$, which works against the shift, so $c'>0$ requires the local advantage
of group $B$ to outweigh it.
The constants $\kappa$ and $c_\Psi$ are detailed in Appendix \ref{sec:second-assumption}. Theorem~\ref{thm:imbalance-driven-bound-full} gives the finite-sample version.

\section{When Balanced Augmentation Improves Fairness}\label{sec:full-balance}
 \begin{figure}
     \centering
     \includegraphics[width=0.75\linewidth]{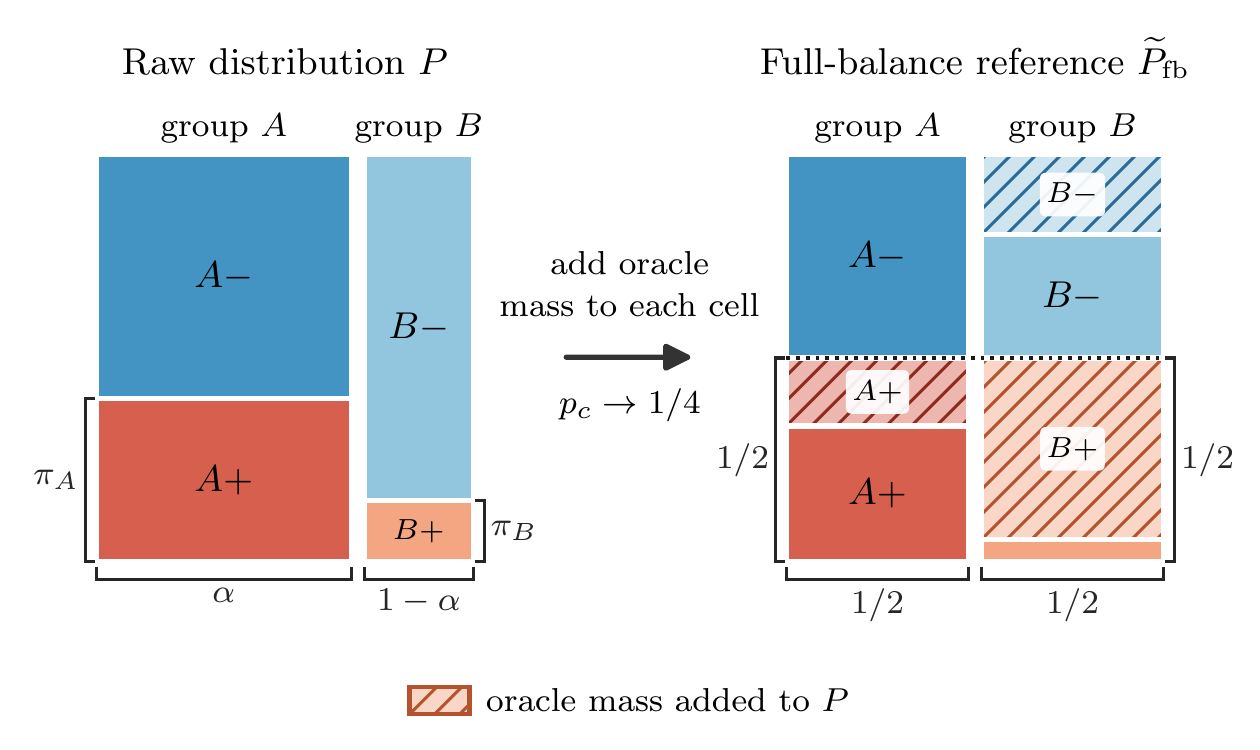}
     \caption{Comparison of the raw distribution with the full-balance reference distribution.}
     \label{fig:fbillustration}
 \end{figure}
We now ask whether removing both imbalances by synthetic augmentation improves fairness. As a benchmark, exact full balancing augments every cell to the size of the largest one, as shown in Figure \ref{fig:fbillustration}. With $p^*:=\max_cp_c$ and $d_c:=p^*-p_c$, the oracle and synthetic full-balance distributions are
\[
\widetilde P_{\mathrm{fb}}:=\frac{1}{4p^*}\Bigl(P+\sum_c d_cP_c\Bigr),
\qquad
\overline P_{\mathrm{fb}}:=\frac{1}{4p^*}\Bigl(P+\sum_c d_cP_{\mathrm{syn},c}\Bigr),
\]
that is, $\widetilde P_{\mathbf a}$ and $\overline P_{\mathbf a}$ with $a_c=d_c$. Both give mass $1/4$ to each cell, equalizing the groups and the classes at once. In practice, cell $c$ receives $n_{\max}-n_c$ synthetic observations, for a pooled sample of size $4n_{\max}$. The generator error enters mainly through
\[
\epsilon_{\mathrm{syn},\mathrm{fb}}
:=\frac{1}{4p^*}\sum_{c\in\mathcal C}d_c\epsilon_c
\;\ge\;
d_{\mathrm{TV}}(\overline P_{\mathrm{fb}},\widetilde P_{\mathrm{fb}}),
\]
the cellwise errors weighted by how much each cell is augmented.
The analysis decomposes full balancing into three oracle steps: prevalence balancing, then group balancing, then class balancing. Let $\alpha_1:=\alpha(1-\pi_A)/(1-\bar\pi)$ be the share of group $A$ after prevalence balancing. Each oracle step shifts the posterior log-odds pointwise, by \eqref{eq:logodds-shift-main}, by \eqref{eq:group-derivative} integrated over the group share, and by the constant $\log(1/\rho_A)$, respectively, and the three shifts add up.
 
\begin{theorem}[Fairness improvement]\label{thm:fullbalance-improvement}
Under Assumption~\ref{assumption:improvement-bound-assumption} (Appendix~\ref{sec:assumptionforfb}), which includes $\alpha_1>1/2$ and $0<\pi_B<\pi_A\le1/2$,
\begin{gather*}
\Delta_{\mathrm{EO}}(h_P^\eta)-\Delta_{\mathrm{EO}}(h_{\overline P_{\mathrm{fb}}}^\eta)
\ge
\frac{\kappa\,c_\Psi}{\bar\omega_A\bar\omega_B}\,h_{\mathrm{fb}}
-C_q\sqrt{\epsilon_{\mathrm{syn},\mathrm{fb}}},\\
h_{\mathrm{fb}}:=
\underbrace{\log\!\bigl(1+(\mathrm{OR}-1)(\bar\omega_B+\kappa)\bigr)}_{\text{prevalence}}
+\underbrace{c'\bigl(\alpha_1-\tfrac12\bigr)}_{\text{group size}}
+\underbrace{\log\tfrac{1}{\rho_A}}_{\text{class balance}},
\end{gather*}
where $C_q>0$ depends only on the margin of the oracle balanced Bayes classifier $h^\eta_{\widetilde P_{\mathrm{fb}}}$.
\end{theorem}
 
Each source of imbalance removed by full balancing contributes a nonnegative term. As in Section~\ref{sec:prevalence-imbalance}, the prevalence term is at least $\log(1+4(1-\alpha)(\pi_A-\pi_B))$, and the class-balance term is at least $4(\frac12-\pi_A)$. The group-size term is the mechanism of Section~\ref{sec:group-imbalance} in its equal-prevalence case, since after prevalence balancing both groups have prevalence $\pi_A$.
Here $\kappa$, $c_\Psi$ and $c'$ are the constants of Assumption~\ref{assumption:improvement-bound-assumption} (Appendix~\ref{sec:assumptionforfb}).
The generator error is the price of synthetic rather than oracle augmentation. It makes the synthetic Bayes classifier disagree with the oracle one, which costs at most $C_q\sqrt{\epsilon_{\mathrm{syn},\mathrm{fb}}}$. Whenever the oracle term dominates this cost, full-balance augmentation strictly improves Equal Opportunity. Theorem~\ref{thm:fullbalance-improvement-full} gives the version for the empirical classifiers $\widehat h_P$ and $\widehat h_{\overline P_{\mathrm{fb}}}$.

Prediction error under $P$ is dominated by the majority cells, so we compare accuracy under the oracle balanced distribution $\widetilde P_{\mathrm{fb}}$, which weights the four cells equally.

\begin{theorem}[Balanced error]\label{thm:maintain-accuracy-fb}
Under Assumption~\ref{assumption:accuracypreserveass} (Appendix~\ref{sec:maintainaccuracyassumption}),
\[
\Err_{\widetilde P_{\mathrm{fb}}}\bigl(h^\eta_{\overline P_{\mathrm{fb}}}\bigr)
\le
\inf_h\Err_{\widetilde P_{\mathrm{fb}}}(h)+2\epsilon_{\mathrm{syn},\mathrm{fb}},
\]
where the infimum is over all classifiers.
\end{theorem}

Training on the synthetic full-balance distribution is nearly Bayes-optimal for balanced error, losing at most twice the generator error. In particular, it is at least as accurate as training on $P$ under this criterion, up to $2\epsilon_{\mathrm{syn},\mathrm{fb}}$. Under $P$ itself, the classifier trained on $P$ may still be more accurate (Appendix~\ref{sec:pop-vs-sampling}). Theorem~\ref{thm:maintain-accuracy-fb-full} gives the finite-sample version.

\begin{remark}[Synthetic augmentation versus reweighting]\label{remark:augvsreweight}
At the population level, oracle synthetic augmentation is equivalent to reweighting the cells, since both induce the same learning objective. In practice, synthetic augmentation can generate new feature configurations that improve coverage of sparse regions, and pretrained generators can bring in external information unavailable to reweighting. Section~\ref{sec:numerica-studies} illustrates that suitable generators can outperform reweighting, although such gains depend on the problem and the generator.
\end{remark}

\section{Convergence and Minimax Theory of Balanced Data Synthesis}\label{sec:convergence-minimax}

\subsection{From Population Mechanisms to Finite Samples}\label{sec:finite-sample-bridge}
 
Sections~\ref{sec:imbalance}--\ref{sec:full-balance} describe mechanisms through Bayes classifiers. A trained classifier $\widehat h_Q$ differs from $h^\eta_Q$ in two ways: the model class may be misspecified, with calibration error $\epsilon_{\mathrm{cal},Q}:=\sup_x|f_{\theta^*_Q}(x)-\eta_Q(x)|$, and $\widehat\theta_Q$ is estimated from $n_Q$ observations. A score error of size $\varepsilon$ changes predictions only where $|\eta_Q-\frac12|\le\varepsilon$, so under a margin condition (Appendix~\ref{section:assumption-1}) both errors move the Equal Opportunity gap by at most
\[
R_Q(\delta)\asymp\epsilon_{\mathrm{cal},Q}+\sqrt{d\log(d/\delta)/n_Q}
\]
with probability at least $1-\delta$. 
Every result in Sections~\ref{sec:imbalance}--\ref{sec:full-balance} therefore holds for the trained classifiers up to such terms. For Theorems~\ref{Thm:fairness-improvement}--\ref{thm:imbalance-driven-bound}, whose reference distributions are oracles, this gives the lower bound on $\Delta_{\mathrm{EO}}(\widehat h_P)$.
The precise statements are Theorems~\ref{Thm:fairness-improvement-full}--\ref{thm:maintain-accuracy-fb-full} in Appendix~\ref{sec:full-statements}. This section studies the estimation error directly.

\subsection{Convergence of Balanced Data Synthesis}\label{sec:balance-convergence}

The natural target for the synthetic full-balance classifier is the population-risk classifier under oracle balancing,
\[
h^*_{\mathrm{bal}}:=h^*_{\widetilde P_{\mathrm{fb}}},
\quad
\theta^*_{\mathrm{bal}}\in\arg\min_\theta L_{\mathrm{bal}}(\theta),
\quad
L_{\mathrm{bal}}(\theta):=L_{\widetilde P_{\mathrm{fb}}}(\theta)=\frac14\sum_{c\in\mathcal C}\mathbb E[\ell(\theta;X,Y)\mid(S,Y)=c].
\]
Using the population-risk target rather than the Bayes classifier removes the calibration term, so the result below holds for a misspecified model. The two targets' Equal Opportunity gaps differ by at most a constant multiple of $\epsilon_{\mathrm{cal},\widetilde P_{\mathrm{fb}}}$.

\begin{theorem}[Full-balance EO convergence]\label{thm:full-balance-convergence}
Under Assumption~\ref{assumption:convergence} (Appendix~\ref{sec:assumptionforconvergencethm}), with probability at least $1-\delta$,
\[
\bigl|\Delta_{\mathrm{EO}}(\widehat h_{\overline P_{\mathrm{fb}}})-\Delta_{\mathrm{EO}}(h^*_{\mathrm{bal}})\bigr|
\le
C_1\left(\sqrt{\frac{d\bigl(d\log(3n_P)+\log(8d/\delta)\bigr)}{n_P}}+\epsilon_{\mathrm{mix}}\right)
+C_2\,\epsilon_{\mathrm{syn},\mathrm{fb}},
\]
where $\epsilon_{\mathrm{mix}}:=\sum_c\bigl|\tfrac{n_c}{4n_{\max}}-\tfrac{p_c}{4p^*}\bigr|\epsilon_c$, and $C_1,C_2$ depend only on $d$ and the regularity constants of Assumption~\ref{assumption:convergence}.
\end{theorem}

The first term is the estimation error, of order $\sqrt{\log n_P/n_P}$ for fixed $d$. The term $\epsilon_{\rm mix}$, the generator error weighted by the sampling fluctuation of the cell proportions, satisfies $\epsilon_{\rm mix}\le 8\max_c\epsilon_c\cdot\max_c|n_c/n_P-p_c|$ because $p^*\ge1/4$ and $n_{\max}\ge n_P/4$. By Hoeffding's inequality over the four cells, it is at most $8\max_c\epsilon_c\sqrt{\log(8/\delta)/(2n_P)}$ with probability at least $1-\delta$, which is of smaller order than the estimation term.
The last term is the generator error, in which cells that need more synthetic data weigh more. If the generator errors vanish and $n_P\to\infty$, then $\Delta_{\mathrm{EO}}(\widehat h_{\overline P_{\mathrm{fb}}})\to\Delta_{\mathrm{EO}}(h^*_{\mathrm{bal}})$ in probability. Explicit constants are given in Theorem~\ref{thm:full-balance-convergence-full}.

\subsection{Minimax Lower Bound}\label{sec:minimaxlowerbound}

Fix strictly positive cell probabilities and let $\mathcal P$ be a class of distributions with these cell probabilities, varying the cell-conditional distributions.
For $P\in\mathcal{P}$, let $L_{\mathrm{bal},P}(\theta):=\frac14\sum_{c\in\mathcal{C}}\mathbb{E}_P[\ell(\theta;X,Y)\mid (S,Y)=c]$, let $\theta^*_{\mathrm{bal},P}$ be a fixed minimizer of $L_{\mathrm{bal},P}$, and let $h^*_{\mathrm{bal},P}(x):=\mathbf{1}\{f_{\theta^*_{\mathrm{bal},P}}(x)\ge 1/2\}$ be the corresponding balanced population-risk classifier.
Given $Z_i=(X_i,Y_i,S_i)$, $i=1,\dots,n$, i.i.d.\ from $P$ and independent auxiliary randomness $U$ whose law does not depend on $P$, a learning procedure is any measurable rule $\widehat q_n(x;Z_1,\dots,Z_n,U)\in[0,1]$, read as the probability of predicting $1$ at $x$. This includes any score with any learned threshold and the synthetic full-balance procedure, whose generator randomness is absorbed into $U$. For a rule $q$, let
\begin{equation}\label{eq:minimax-general-EO}
\Delta_{\mathrm{EO},P}(q):=\bigl|\mathbb E_P[q(X)\mid S=A,Y=1]-\mathbb E_P[q(X)\mid S=B,Y=1]\bigr|,
\end{equation}
which equals $\Delta_{\mathrm{EO}}(h)$ for a deterministic classifier, and define the minimax risk
\begin{equation}\label{eq:minimax-EO-risk}
\mathfrak R_n^{\mathrm{EO}}:=\inf_{\widehat q_n}\sup_{P\in\mathcal P}\mathbb E_P\Bigl[\bigl|\Delta_{\mathrm{EO},P}(\widehat q_n)-\Delta_{\mathrm{EO},P}(h^*_{\mathrm{bal},P})\bigr|\Bigr].
\end{equation}

\begin{theorem}[Minimax lower bound]\label{thm:minimax-EO-lower}
Under Assumption~\ref{assumption:minimax-EO} (Appendix~\ref{sec:minimaxassumption}), there exist constants $t_*,C_{\mathrm{EO}}>0$, depending on the baseline distribution $P_0$ in Assumption~\ref{assumption:minimax-EO}, the model $(f_\theta,\ell)$ and the local submodels in Assumption~\ref{assumption:minimax-EO}(4), such that
\(
\mathfrak R_n^{\mathrm{EO}}\ge C_{\mathrm{EO}}\min\{t_*,n^{-1/2}\}.
\)
\end{theorem}

Theorem \ref{thm:minimax-EO-lower} shows that no procedure can beat the $n^{-1/2}$ rate uniformly over $\mathcal P$. 
Since $n_P\le 4n_{\max}\le 4n_P$ and the loss in (\ref{eq:minimax-EO-risk}) is at most one, when the constants of Assumption~\ref{assumption:convergence} hold uniformly over $\mathcal P$, Theorem~\ref{thm:full-balance-convergence} attains this rate up to a logarithmic factor and the generator error.
The logarithmic factor comes from the uniform concentration needed because the generators are fitted on the same sample. With generators fitted on an independent split it can be removed (Appendix~\ref{sec:disc-independence-minimax}).

\section{Accuracy-Constrained Adaptive Synthetic Augmentation}\label{sec:methodology}

\subsection{Adaptive Augmentation Formulation and Algorithm}\label{sec:adaptive_method}

Full balancing discussed in the previous section is a benchmark, but not necessarily the best allocation. Different cells may benefit differently from augmentation, and full balancing need not remove the disparity: it equalizes the cell probabilities but not the cell-conditional distributions, so $\Delta_{\rm EO}(h^*_{\rm bal})>0$ whenever $P_{A+}$ and $P_{B+}$ give
different probabilities to $\{x:h^*_{\rm bal}(x)=1\}$ (Appendix~\ref{sec:fb-not-fair}), and Theorem~\ref{thm:full-balance-convergence} shows that the synthetic full-balance classifier converges to exactly this gap. We therefore propose Accuracy-Constrained Adaptive Synthetic Augmentation (ACASA), which chooses where and how much to augment by optimizing the estimated fairness--accuracy trade-off.

An allocation $\boldsymbol\lambda=(\lambda_c)_{c\in\mathcal C}\in\mathbb R_+^4$ generates $m_c(\boldsymbol\lambda)=\lfloor\lambda_cn_{\max}\rfloor$ synthetic observations in cell $c$. With $r:=n_{\max}/n_P$, the population counterpart of allocation $\boldsymbol\lambda$ is $\overline P_{\boldsymbol\lambda}:=\overline P_{\mathbf a}$ with $a_c=r\lambda_c$. Full balancing corresponds to $\lambda_c^{\mathrm{fb}}=1-n_c/n_{\max}$, which is one specific point of the search space.

Candidate allocations are evaluated by stratified $K$-fold cross-validation: within each fold, the generators are fitted once on the training portion, the classifier is fitted on the augmented training portion, and the Equal Opportunity gap and the prediction error are computed on the real validation observations. Here $\widehat{\Delta}^{\mathrm{CV}}_{\mathrm{EO}}(\boldsymbol{\lambda})$ and $\widehat{\mathrm{Err}}^{\mathrm{CV}}(\boldsymbol{\lambda})$ denote their averages over the folds. Each grid allocation is evaluated at most once, and its criteria are reused if it is revisited (Appendix~\ref{sec:assumptionforacasa}).

ACASA searches a finite grid $\Lambda_\omega$ of mesh $\omega$ in a bounded region $\Lambda$. Starting from $\boldsymbol\lambda^{(0)}=\mathbf 0$, it considers moves that change one cell's allocation by a prespecified,
possibly negative, increment $\gamma_{c,j}$ ($j=1,\dots,J$) along the unit
vector $\boldsymbol{e}_c$ of cell $c$. A candidate is feasible if
\begin{equation}\label{eq:cv_accuracy_constraint}
\widehat\Err^{\mathrm{CV}}(\boldsymbol\lambda)-\widehat\Err^{\mathrm{CV}}(\mathbf 0)\le\xi_{\mathrm{acc}},
\end{equation}
that is, if it loses at most $\xi_{\mathrm{acc}}$ in accuracy on the deployment population relative to no augmentation (Remark~\ref{remark:choiceofvalidation} discusses alternatives). Among feasible moves, ACASA takes the one with the smallest estimated EO gap if it improves the current EO gap by more than a threshold $\xi_{\mathrm{fair}}$. If no move qualifies, it searches the grid within radius $\rho_{\mathrm{saf}}$ of the current allocation as a safeguard, and stops if this also fails. Finally, the generators are refitted on the full sample and the classifier is trained on the augmented data. Algorithm~\ref{alg:adaptive_augmentation} summarizes the procedure.

\begin{algorithm}[htbp]
\caption{Accuracy-Constrained Adaptive Synthetic Augmentation (ACASA)}
\label{alg:adaptive_augmentation}
\begin{algorithmic}[1]
\small
\REQUIRE Real training data; $K$ folds; $\Lambda_\omega$; increments $\{\gamma_{c,j}\}$; $\xi_{\mathrm{acc}}$, $\xi_{\mathrm{fair}}$, $\rho_{\mathrm{saf}}$.
\ENSURE $\widehat{\boldsymbol\lambda}$ and $\widehat h_{\overline P_{\widehat{\boldsymbol\lambda}}}$.
\STATE Set $\boldsymbol\lambda^{(0)}=\mathbf 0$, $t=1$; fix a stratified $K$-fold partition and fit the generators on each training portion; evaluate $\widehat\Err^{\mathrm{CV}}(\mathbf 0)$.
\WHILE{the search has not terminated}
    \STATE Candidates: $\mathcal M_t=\{\boldsymbol\lambda^{(t-1)}+\gamma_{c,j}e_c\}_{c,j}\cap\Lambda_\omega$; feasible set $\mathcal F_t$: candidates satisfying \eqref{eq:cv_accuracy_constraint}.
    \STATE $\boldsymbol\lambda'\in\arg\min_{\mathcal F_t}\widehat\Delta^{\mathrm{CV}}_{\mathrm{EO}}$; gain $\widehat G_t=\widehat\Delta^{\mathrm{CV}}_{\mathrm{EO}}(\boldsymbol\lambda^{(t-1)})-\widehat\Delta^{\mathrm{CV}}_{\mathrm{EO}}(\boldsymbol\lambda')$ ($-\infty$ if $\mathcal F_t=\emptyset$).
    \IF{$\widehat G_t>\xi_{\mathrm{fair}}$}
        \STATE $\boldsymbol\lambda^{(t)}=\boldsymbol\lambda'$.
    \ELSE
        \STATE Repeat the selection over feasible grid points within distance $\rho_{\mathrm{saf}}$ of $\boldsymbol\lambda^{(t-1)}$; accept if the gain exceeds $\xi_{\mathrm{fair}}$, otherwise terminate.
    \ENDIF
    \STATE $t\leftarrow t+1$.
\ENDWHILE
\STATE Set $\widehat{\boldsymbol\lambda}$ to the current allocation; refit the generators on the full sample, generate $\lfloor\widehat\lambda_cn_{\max}\rfloor$ observations per cell, and fit $\widehat h_{\overline P_{\widehat{\boldsymbol\lambda}}}$.
\RETURN $\widehat{\boldsymbol\lambda}$ and $\widehat h_{\overline P_{\widehat{\boldsymbol\lambda}}}$.
\end{algorithmic}
\end{algorithm}

\begin{remark}[Choice of validation distribution]\label{remark:choiceofvalidation}
We state the criteria under $P$, but the validation distribution can be chosen to suit the application, for example a balanced or reweighted validation set that gives the four cells equal importance. If the cell-conditional distributions are unchanged, such reweighting preserves the Equal Opportunity gap while changing the error criterion.
Likewise, the error in \eqref{eq:cv_accuracy_constraint} may be replaced by another performance measure, such as MCC (Section \ref{sec:realdataapp}).
Our theoretical guarantees concern the $P$-based formulation.
\end{remark}

\subsection{Guarantee for the Selected Allocation}\label{sec:adaptive_theory}

The target we compare with is the fairest allocation that satisfies the accuracy constraint at the population level:
\begin{equation}\label{eq:oracle_allocation}
\boldsymbol\lambda^*\in\arg\min_{\boldsymbol\lambda\in\Lambda_{\mathrm{acc}}}\Delta_{\mathrm{EO}}\bigl(h^*_{\overline P_{\boldsymbol\lambda}}\bigr),
\qquad
\Lambda_{\mathrm{acc}}:=\bigl\{\boldsymbol\lambda\in\Lambda:\Err_P(h^*_{\overline P_{\boldsymbol\lambda}})-\Err_P(h^*_P)\le\xi_{\mathrm{acc}}\bigr\}.
\end{equation}
Because $\overline P_{\boldsymbol\lambda}$ uses the fitted generators, $\boldsymbol\lambda^*$ accounts for generator quality and is defined conditionally on the sample. Whenever $\boldsymbol\lambda^{\mathrm{fb}}\in\Lambda_{\mathrm{acc}}$, the target $\boldsymbol{\lambda}^*$ is at least as fair as full balancing.
The selected allocation $\hat{\boldsymbol{\lambda}}$ differs from $\boldsymbol{\lambda^*}$ for three reasons.
The search stops once no feasible candidate improves the cross-validated EO gap by
more than the acceptance threshold $\xi_{\rm fair}$. It evaluates only the grid
$\Lambda_\omega$, whose mesh is $\omega$. And it replaces both population criteria in
(\ref{eq:oracle_allocation}) by cross-validated estimates. The third source is measured by
$\epsilon_{\rm fair}(\delta)$ and $\epsilon_{\rm acc}(\delta)$, which bound, uniformly
over $\Lambda_\omega$ and on an event of probability at least $1-\delta$, the errors of
$\hat\Delta^{\rm CV}_{\rm EO}(\lambda)$ and of the accuracy loss
$\widehat{\rm Err}^{\rm CV}(\lambda)-\widehat{\rm Err}^{\rm CV}(0)$ relative to their
population counterparts in (\ref{eq:oracle_allocation}). Each is the sum of a validation term, from
evaluating true positive rates or errors on a finite real validation sample, a
training term, from fitting the classifier on finite real and synthetic training
portions, and a generation term, proportional to $\epsilon_{\rm syn,CV}$, which collects
the total-variation errors of the generators and the mismatch between the cell
weights in each training fold and those of $\overline P_\lambda$.
Both are therefore of order
\[
\sqrt{\frac{d^2\log n_P+d\log(KN_\omega/\delta)}{n_{\min}}}+\epsilon_{\rm syn,CV},
\]
where $N_\omega=|\Lambda_\omega|$ and $n_{\min}$ is the smallest training or validation
cell size across folds. Appendix~\ref{sec:appendixadaptivedetail} gives their explicit forms.

\begin{theorem}[Allocation error of ACASA]\label{thm:adaptive_near_optimality}
Under Assumption~\ref{assumption:adaptive_search} (Appendix~\ref{sec:assumptionforacasa}), condition on the group--label cells of the real observations and on the fold partition, which fix the cell counts entering $\epsilon_{\mathrm{fair}}(\delta)$ and $\epsilon_{\mathrm{acc}}(\delta)$, and suppose that $\epsilon_{\mathrm{acc}}(\delta)$ and $\omega$ are small enough and $\rho_{\mathrm{saf}}$ is large enough, as quantified in \eqref{eq:acasa-conditions} in Appendix \ref{sec:appendixadaptivedetail}. Then, with probability at least $1-\delta$,
\[
\|\hat{\boldsymbol{\lambda}}-\boldsymbol{\lambda}^*\|_2\le C_{\boldsymbol{\lambda}}
\sqrt{\xi_{\rm fair}+\epsilon_{\rm fair}(\delta)+\epsilon_{\rm acc}(\delta)+\omega},
\]
where $C_{\boldsymbol{\lambda}}$ depends only on the constants of
Assumption~\ref{assumption:adaptive_search}.
\end{theorem}
The allocation error is driven by the search tolerance $\xi_{\mathrm{fair}}$, the cross-validation estimation errors and the grid mesh.  Theorem~\ref{thm:adaptive_near_optimality-full} gives the bound with explicit constants. Proposition~\ref{prop:excess-eo} in Appendix~\ref{sec:appendixadaptivedetail} complements this result with a bound on the excess Equal Opportunity gap of the selected allocation relative to $\boldsymbol\lambda^*$.

\begin{remark}[Relationship with post-processing]
ACASA is a pre-processing method. It reduces the disparity but does not enforce a prespecified fairness level. When a hard constraint is required, it can be combined with post-processing such as group-specific threshold tuning. Because augmentation first brings the classifier closer to the constraint, less threshold adjustment is needed and the accuracy loss can be smaller, as the numerical studies illustrate.
\end{remark}

\section{Numerical Studies}\label{sec:numerica-studies}

\subsection{Simulation Studies}

We conduct simulations to investigate how data imbalance affects fairness and predictive performance, and to evaluate the effectiveness of full balancing augmentation and ACASA. We consider a binary classification problem with two groups and group–label-specific feature distributions. 
Within each cell, we generate 15-dimensional feature vectors, with the first five coordinates following a mixture of Gaussian and heavy-tailed \(t_3\) distributions. The \(t_3\) mixture weights for cells \(A+\), \(A-\), \(B+\), and \(B-\) are \(0.09\), \(0.04\), \(0.12\), and \(0.06\), respectively, while the remaining ten coordinates are Gaussian.
 An independent test set contains \(5{,}000\) observations per cell, corresponding to a balanced evaluation distribution. We compare raw training, full balancing, and ACASA, with augmentation implemented through reweighting (Appendix~\ref{app:reweighting}), SMOTE \citep{chawla2002smote}, or ADASYN \citep{he2008adasyn}. All methods use the same logistic-regression specification, and results without post-processing use a
standard classification threshold of $0.5$. Further implementation details are provided in  Appendix \ref{sec:simulation-detail}.

\begin{figure*}[!htbp]
  \centering
  \begin{minipage}[t]{0.32\textwidth}
    \centering
    \includegraphics[width=\linewidth]
      {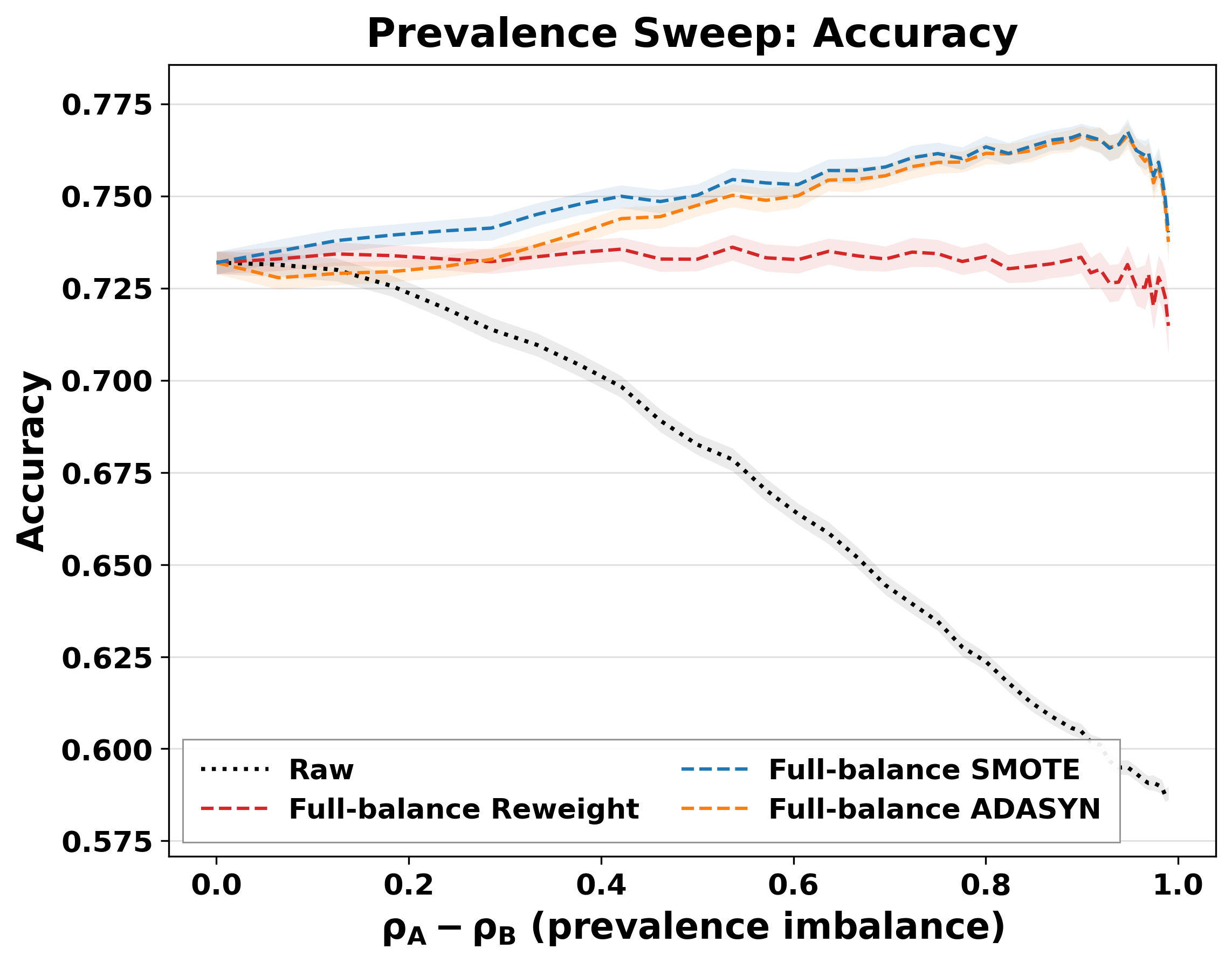}
    \par\smallskip
    {\small (a) Accuracy}
  \end{minipage}\hfill
  \begin{minipage}[t]{0.32\textwidth}
    \centering
    \includegraphics[width=\linewidth]
      {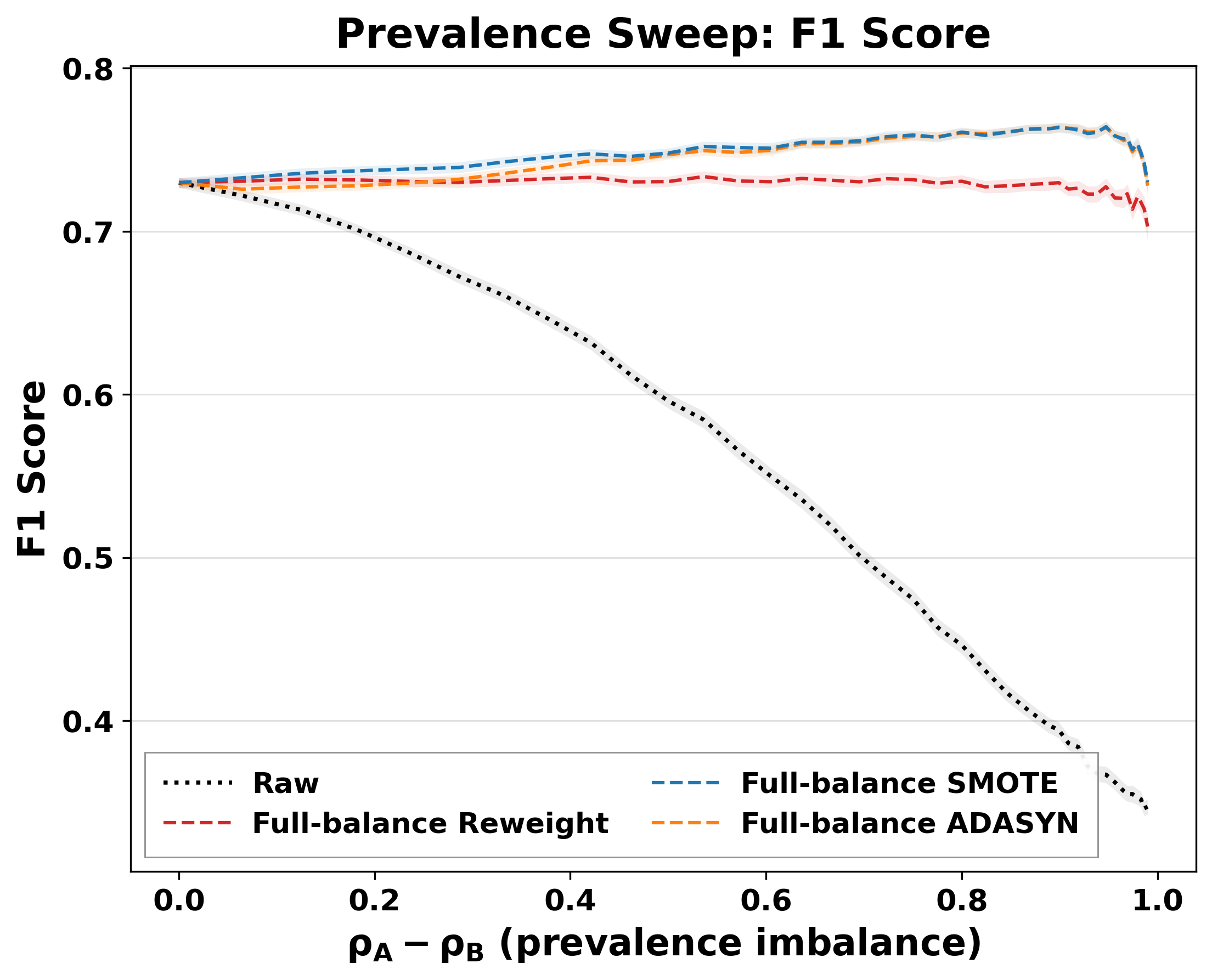}
    \par\smallskip
    {\small (b) F1 score}
  \end{minipage}\hfill
  \begin{minipage}[t]{0.32\textwidth}
    \centering
    \includegraphics[width=\linewidth]
      {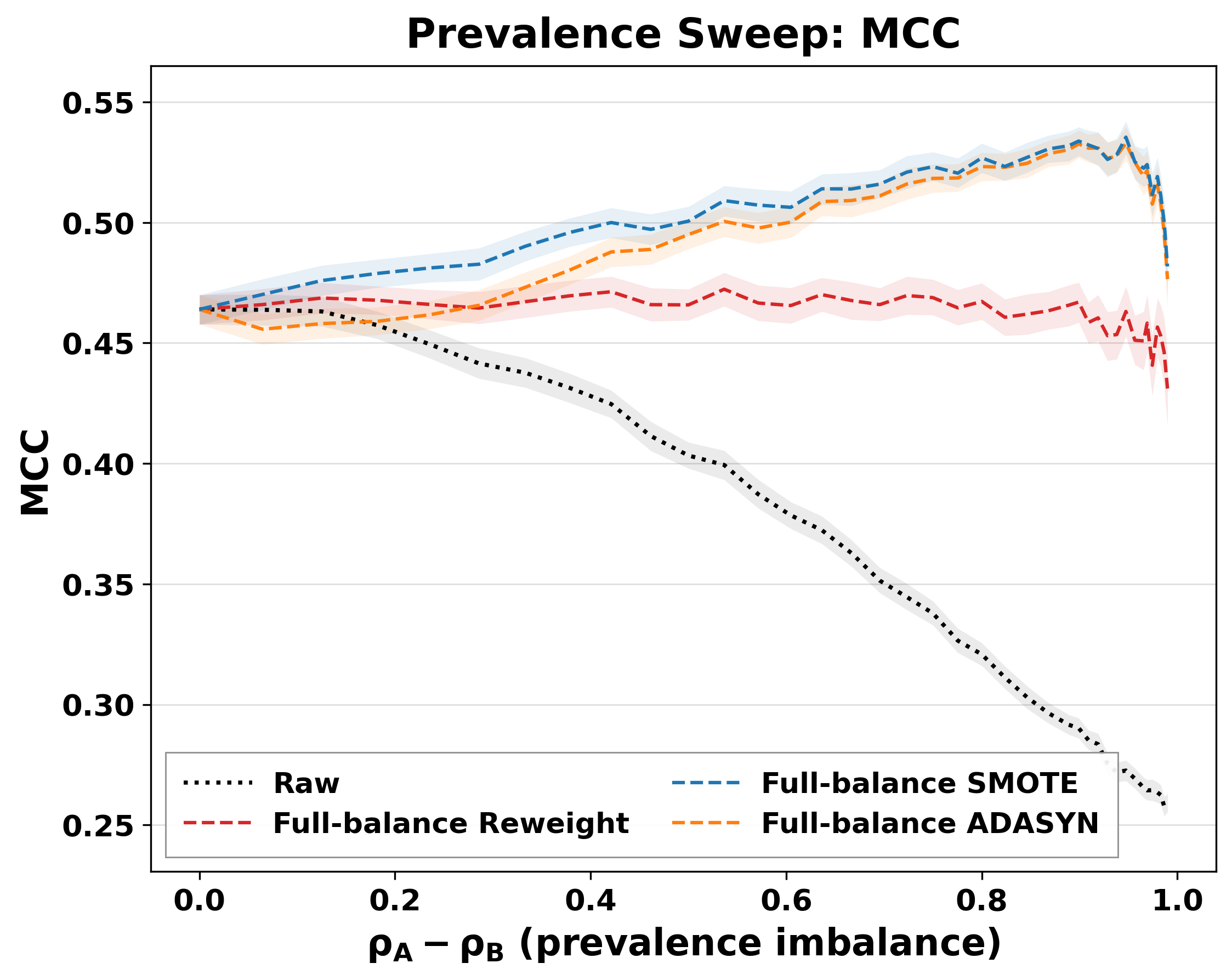}
    \par\smallskip
    {\small (c) MCC}
  \end{minipage}

  \caption{Predictive performance under prevalence imbalance,
  measured by (a) accuracy, (b) F1 score, and
  (c) Matthews correlation coefficient (MCC).
  The horizontal axis shows the positive-class odds
  difference $\rho_A-\rho_B$.  Curves show means,
  and shading denotes pointwise $95\%$ bootstrap
  confidence intervals.}
  \label{fig:prevalence-performance}
\end{figure*}

\begin{figure*}[!htbp]
  \centering
  \begin{minipage}[t]{0.48\textwidth}
    \centering
    \includegraphics[width=\linewidth]
      {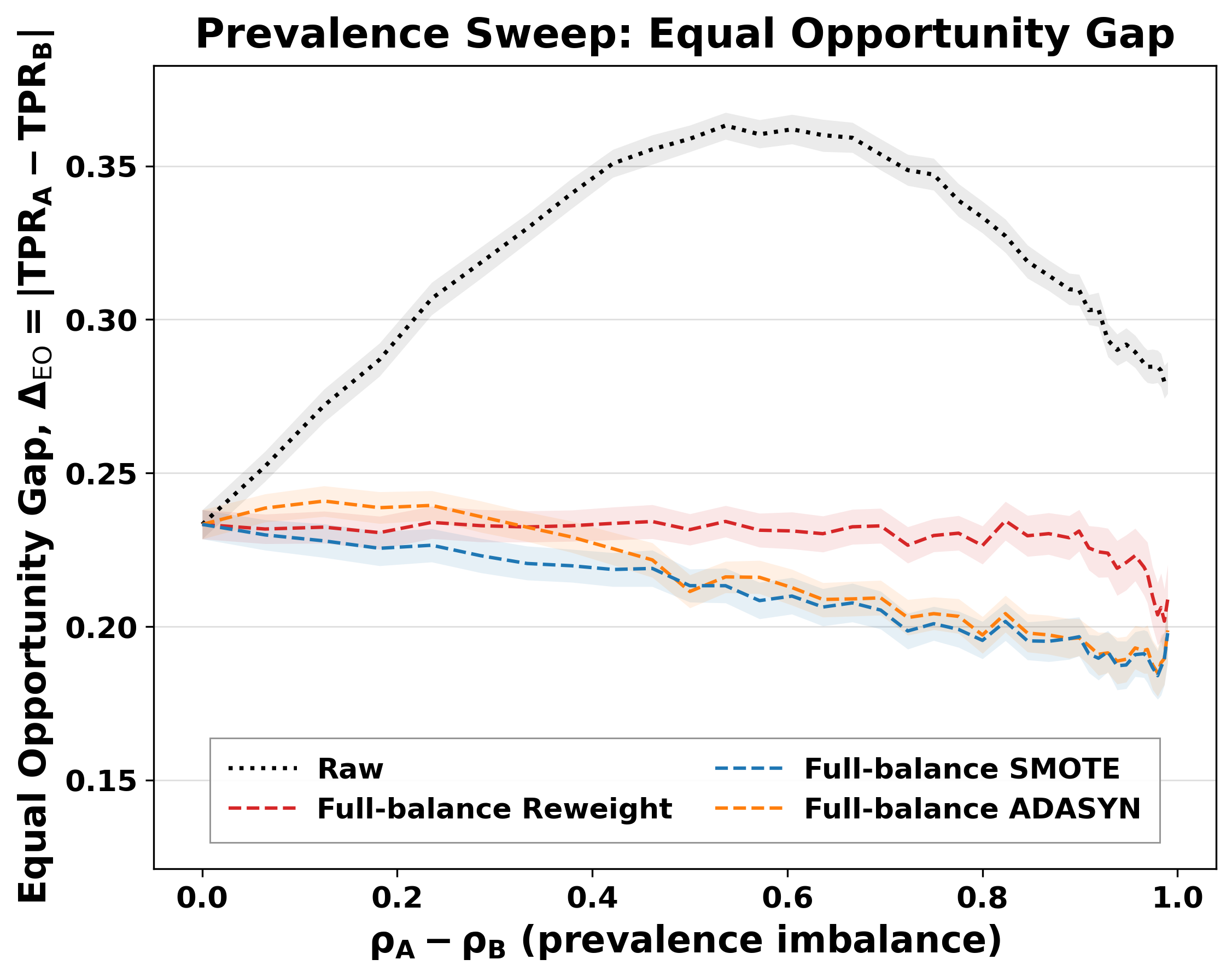}
    \captionof{figure}{Equal Opportunity gap under prevalence
    imbalance. The horizontal axis shows the positive-class
    odds difference $\rho_A-\rho_B$.  Shading denotes pointwise
    $95\%$ bootstrap confidence intervals.}
    \label{fig:prevalence-eogap}
  \end{minipage}\hfill
  \begin{minipage}[t]{0.48\textwidth}
    \centering
    \includegraphics[width=\linewidth]
      {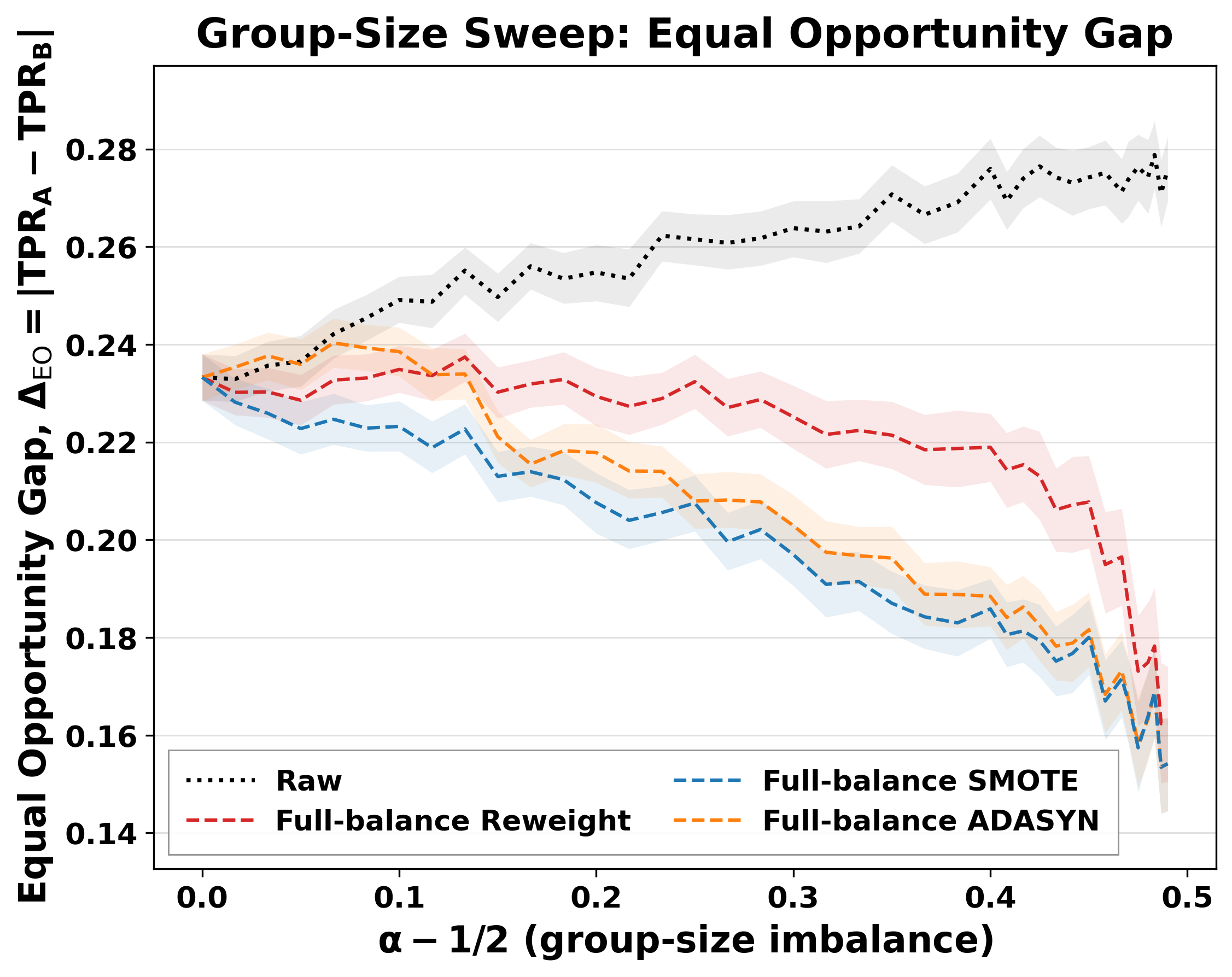}
    \captionof{figure}{Equal Opportunity gap under group-size
    imbalance. The horizontal axis shows $\alpha-1/2$.
    Shading denotes pointwise $95\%$
    bootstrap confidence intervals.}
    \label{fig:groupsize-eogap}
  \end{minipage}
\end{figure*}

\paragraph{Effects of data imbalance.} We first examine the two imbalance mechanisms studied in Section \ref{sec:imbalance}. For prevalence imbalance, we fix the training sample size of each group at $3{,}000$ and the positive-class odds of group A at \(\rho_A=1\). We vary the number of \(B+\) observations from \(1{,}500\) to \(30\), keeping the total size of group \(B\) fixed at \(3{,}000\) by adjusting the number of \(B-\) observations accordingly, while holding \(n_{A+}=n_{A-}=1{,}500\) fixed, thereby sweeping the degree of prevalence imbalance between the two groups. Figures \ref{fig:prevalence-performance} and \ref{fig:prevalence-eogap} show that raw training loses predictive performance as \(\rho_A-\rho_B\) increases. The EO gap under raw training increases initially but decreases under extreme prevalence imbalance, as the classifier's true positive rates deteriorate for both groups. 
Full balancing yields a much smaller gap throughout most of the range and maintains substantially better predictive performance, although the curves become less stable near the most extreme imbalance. Synthetic augmentation  generally outperforms reweighting in both fairness and predictive performance.

\begin{figure*}[!htbp]
  \centering
  \begin{minipage}[t]{0.32\textwidth}
    \centering
    \includegraphics[width=\linewidth]
      {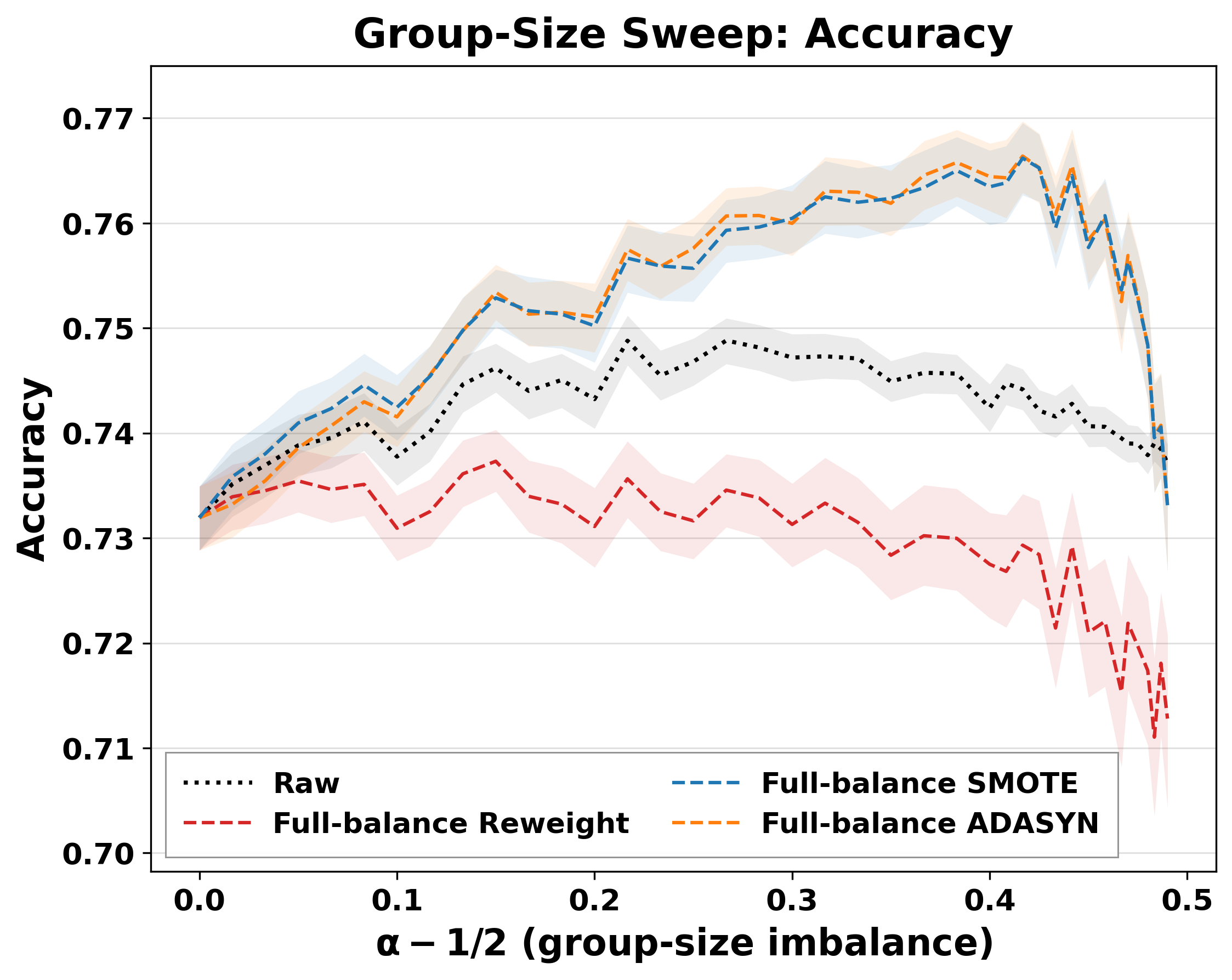}
    \par\smallskip
    {\small (a) Accuracy}
  \end{minipage}\hfill
  \begin{minipage}[t]{0.32\textwidth}
    \centering
    \includegraphics[width=\linewidth]
      {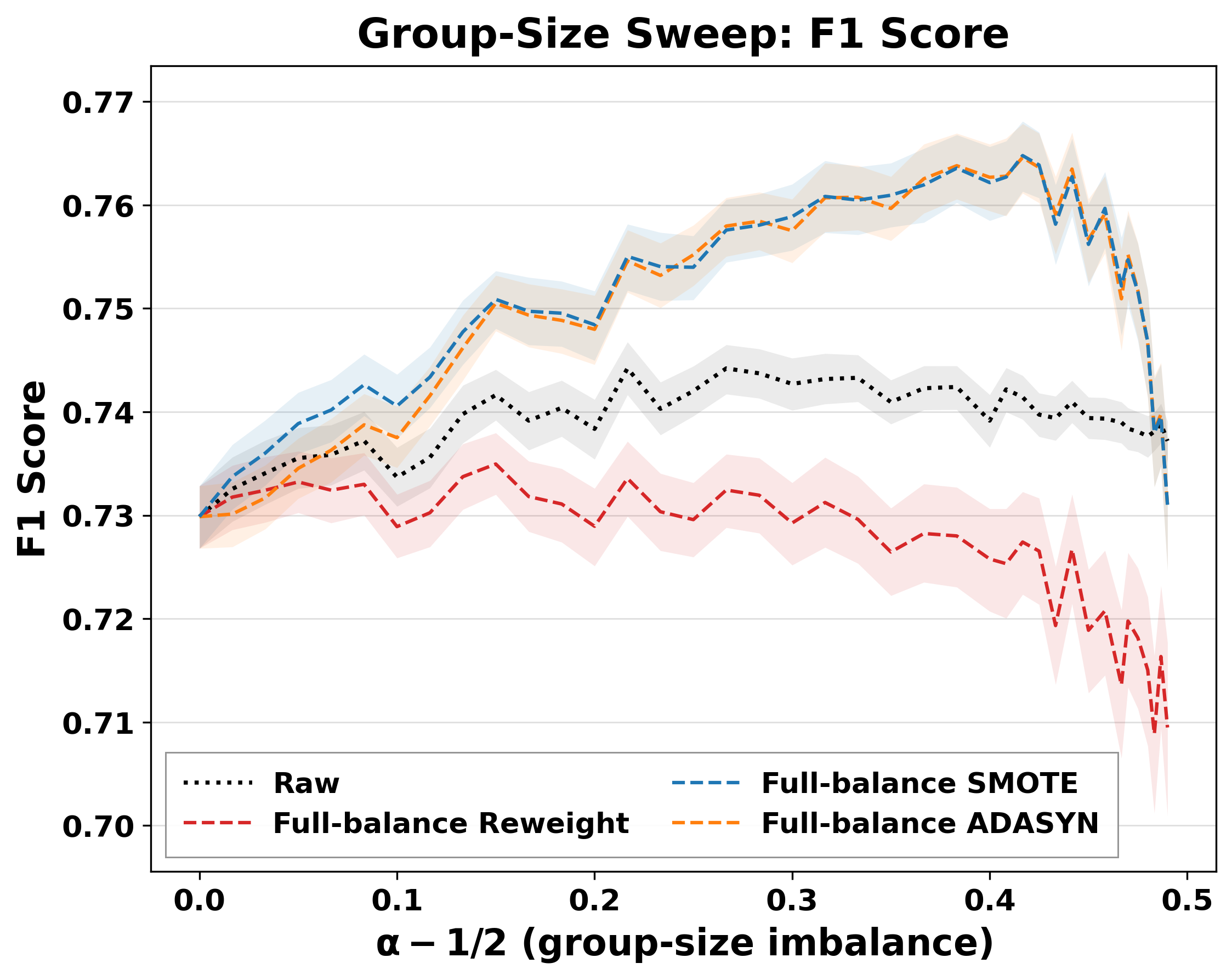}
    \par\smallskip
    {\small (b) F1 score}
  \end{minipage}\hfill
  \begin{minipage}[t]{0.32\textwidth}
    \centering
    \includegraphics[width=\linewidth]
      {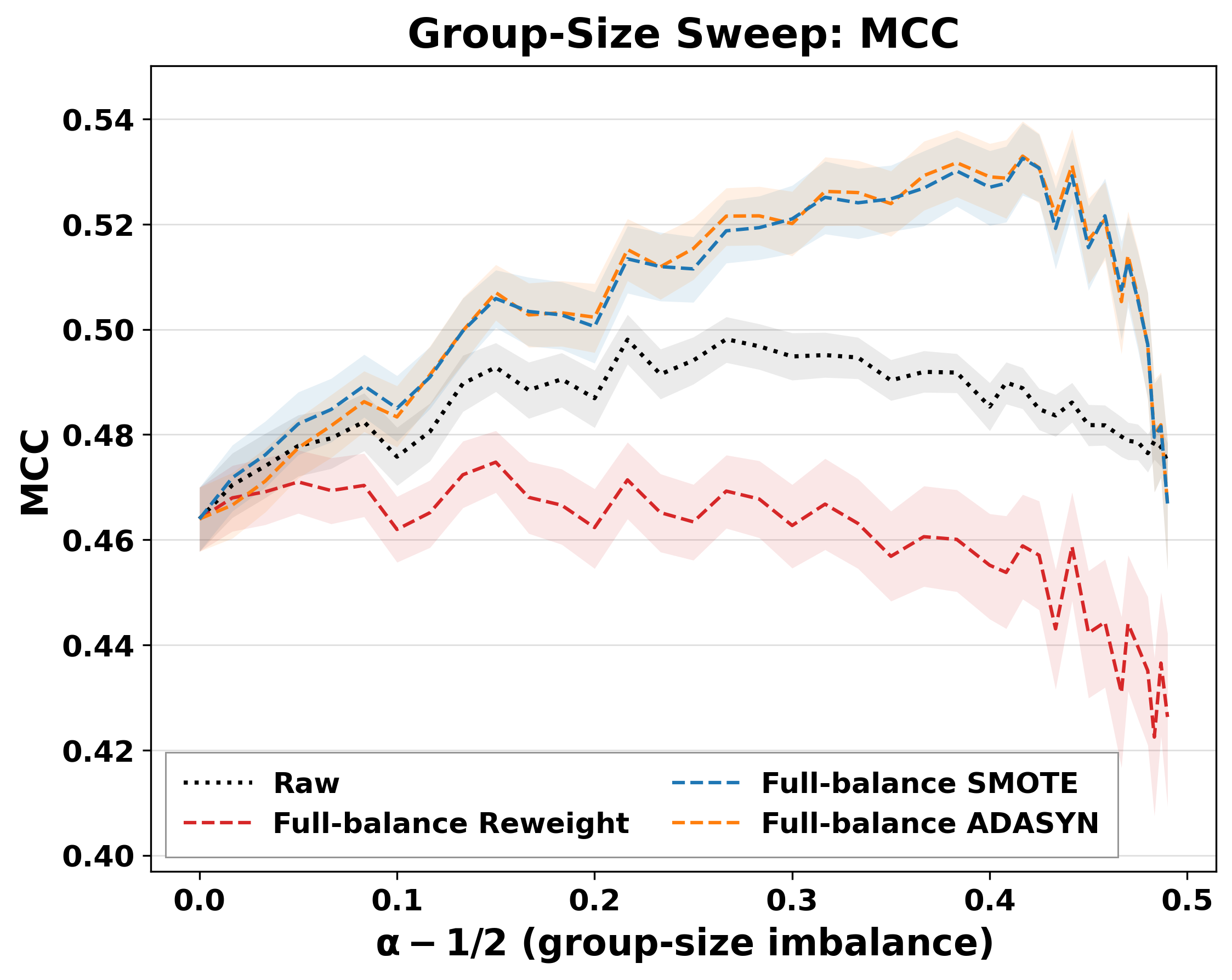}
    \par\smallskip
    {\small (c) MCC}
  \end{minipage}

  \caption{Predictive performance under group-size imbalance,
  measured by (a) accuracy, (b) F1 score, and
  (c) Matthews correlation coefficient (MCC).
  The horizontal axis shows $\alpha-1/2$.
  Curves show means, and shading denotes pointwise
  $95\%$ bootstrap confidence intervals.}
  \label{fig:groupsize-performance}
\end{figure*}

For group-size imbalance, we fix \(\rho_A=\rho_B=1\) and the total training sample size at \(6{,}000\), while decreasing the size of group \(B\) from \(3{,}000\) to \(60\) and increasing the size of group \(A\) accordingly, thereby sweeping the degree of group-size imbalance while maintaining equal class prevalence across groups. Figures \ref{fig:groupsize-performance} and \ref{fig:groupsize-eogap} show that increasing group-size imbalance generally raises the EO gap under raw training, while its predictive performance remains comparatively stable. Full balancing with SMOTE or ADASYN substantially reduces the gap and generally improves predictive performance, while reweighting generally reduces the EO gap less. Near the most extreme imbalance, the augmentation curves become unstable, illustrating the limitations of methods under extreme data scarcity.

\paragraph{Adaptive synthetic augmentation.} We next evaluate ACASA using training cell sizes \((n_{A+},n_{A-},n_{B+},n_{B-})=(2000,3000,1000,3000)\), which introduce both prevalence and group-size imbalance. ACASA selects cell-specific augmentation amounts through $3$-fold cross-validation, subject to an accuracy constraint. We set \(\xi_{\mathrm{acc}}=0.01\), \(\xi_{\mathrm{fair}}=0.003\), and use an allocation grid with spacing $0.2$ and a safeguard radius of $0.6$.

\begin{figure*}[!htbp]
  \centering
  \includegraphics[width=0.55\textwidth]
    {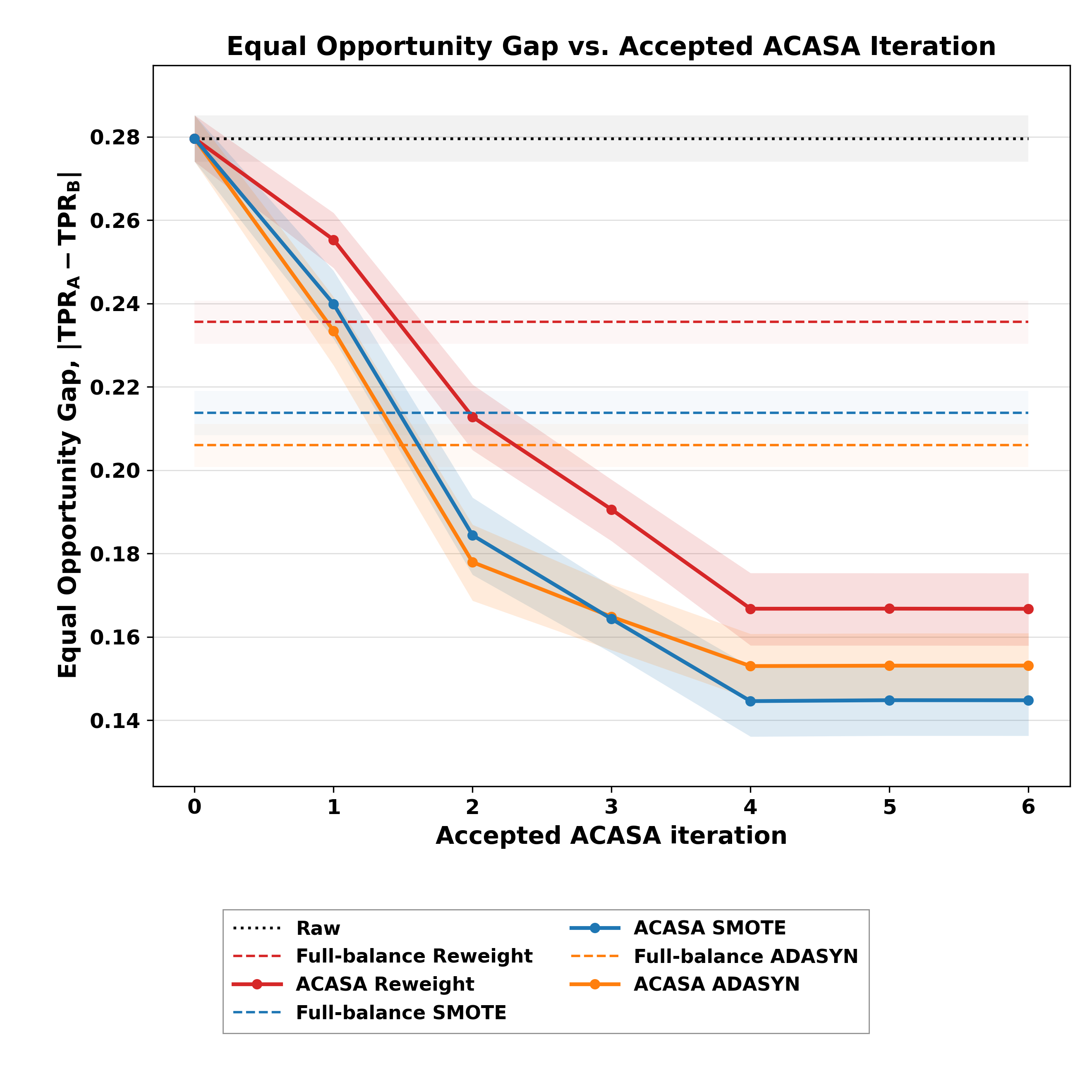}

  \caption{Equal Opportunity gap across accepted ACASA
  iterations.
  Solid curves show mean ACASA trajectories, while dashed
  and dotted lines show full-balancing and raw-training
  benchmarks. Shading denotes
  pointwise $95\%$ bootstrap confidence intervals.
  Runs that stop early retain their final values
  in subsequent plotted iterations.}
  \label{fig:acasa-convergence}
\end{figure*}

Table \ref{tab:simulation-performance} shows that all augmentation methods substantially improve fairness and predictive performance relative to raw training. Compared with full balancing, ACASA further reduces the EO gap with only modest losses in accuracy, while improving AUROC and AUPRC relative to its corresponding full-balancing benchmarks. Figure \ref{fig:acasa-convergence} further illustrates that the EO gap decreases and stabilizes over successive ACASA iterations, with SMOTE and ADASYN achieving smaller final gaps than reweighting.

\begin{table*}[!htbp]
  \centering

  \begingroup
  \footnotesize
  \setlength{\tabcolsep}{5pt}
  \renewcommand{\arraystretch}{1.25}
  \newcommand{\meanCI}[3]{%
    \shortstack{$#1$\\[1pt]{\scriptsize$[#2,\,#3]$}}%
  }

  \begin{adjustbox}{max width=\textwidth}
    \begin{tabular}{@{}lcccc@{}}
      \toprule
      Method & EO gap & Accuracy & AUROC & AUPRC \\
      \midrule

      Raw
      & \meanCI{0.2796}{0.2741}{0.2852}
      & \meanCI{0.6188}{0.6163}{0.6214}
      & \meanCI{0.7708}{0.7679}{0.7735}
      & \meanCI{0.7612}{0.7586}{0.7637} \\
      \midrule

      Full-balance Reweight
      & \meanCI{0.2356}{0.2303}{0.2406}
      & \meanCI{0.7346}{0.7315}{0.7376}
      & \meanCI{0.7915}{0.7884}{0.7945}
      & \meanCI{0.7755}{0.7728}{0.7783} \\
      \addlinespace[3pt]

      Full-balance SMOTE
      & \meanCI{0.2138}{0.2084}{0.2190}
      & \meanCI{0.7555}{0.7529}{0.7582}
      & \meanCI{0.8126}{0.8101}{0.8151}
      & \meanCI{0.7944}{0.7920}{0.7967} \\
      \addlinespace[3pt]

      Full-balance ADASYN
      & \meanCI{0.2060}{0.2008}{0.2111}
      & \meanCI{0.7583}{0.7557}{0.7607}
      & \meanCI{0.8158}{0.8134}{0.8181}
      & \meanCI{0.7973}{0.7951}{0.7994} \\

      \midrule

      ACASA Reweight
      & \meanCI{0.1667}{0.1579}{0.1753}
      & \meanCI{0.7306}{0.7231}{0.7382}
      & \meanCI{0.8060}{0.8021}{0.8099}
      & \meanCI{0.7861}{0.7828}{0.7894} \\
      \addlinespace[3pt]

      ACASA SMOTE
      & \meanCI{0.1448}{0.1363}{0.1531}
      & \meanCI{0.7532}{0.7464}{0.7597}
      & \meanCI{0.8213}{0.8177}{0.8248}
      & \meanCI{0.7995}{0.7964}{0.8025} \\
      \addlinespace[3pt]

      ACASA ADASYN
      & \meanCI{0.1531}{0.1452}{0.1609}
      & \meanCI{0.7520}{0.7445}{0.7588}
      & \meanCI{0.8208}{0.8171}{0.8244}
      & \meanCI{0.7996}{0.7964}{0.8026} \\

      \bottomrule
    \end{tabular}
  \end{adjustbox}
  \endgroup

  \caption{Fairness and predictive performance: means over $200$ runs, with $95\%$ bootstrap confidence intervals.}
  \label{tab:simulation-performance}
\end{table*}

\begin{figure*}[!htbp]
  \centering
  \begin{minipage}[t]{0.32\textwidth}
    \centering
    \includegraphics[width=\linewidth]{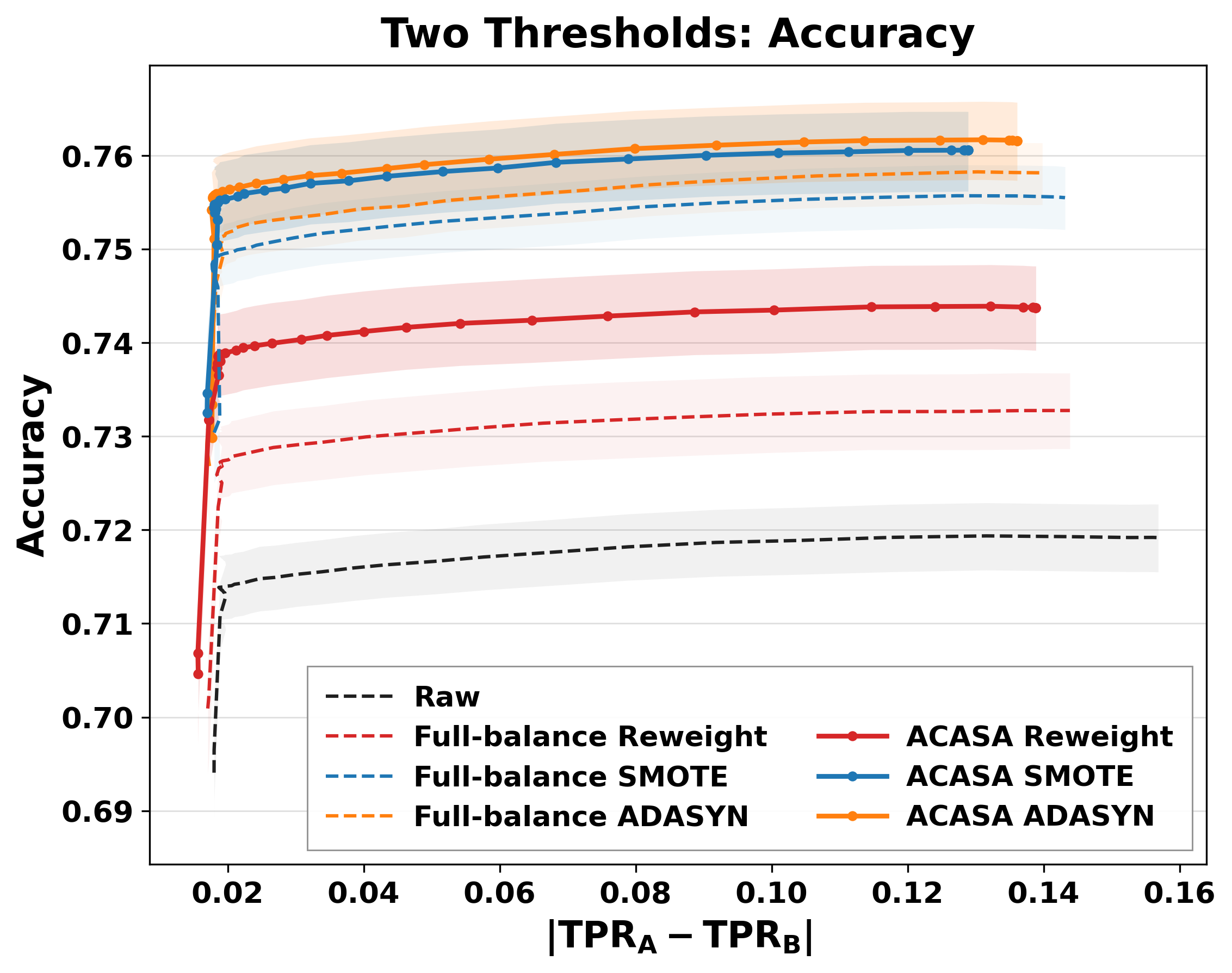}
    \par\smallskip
    {\small (a) Accuracy}
  \end{minipage}\hfill
  \begin{minipage}[t]{0.32\textwidth}
    \centering
    \includegraphics[width=\linewidth]{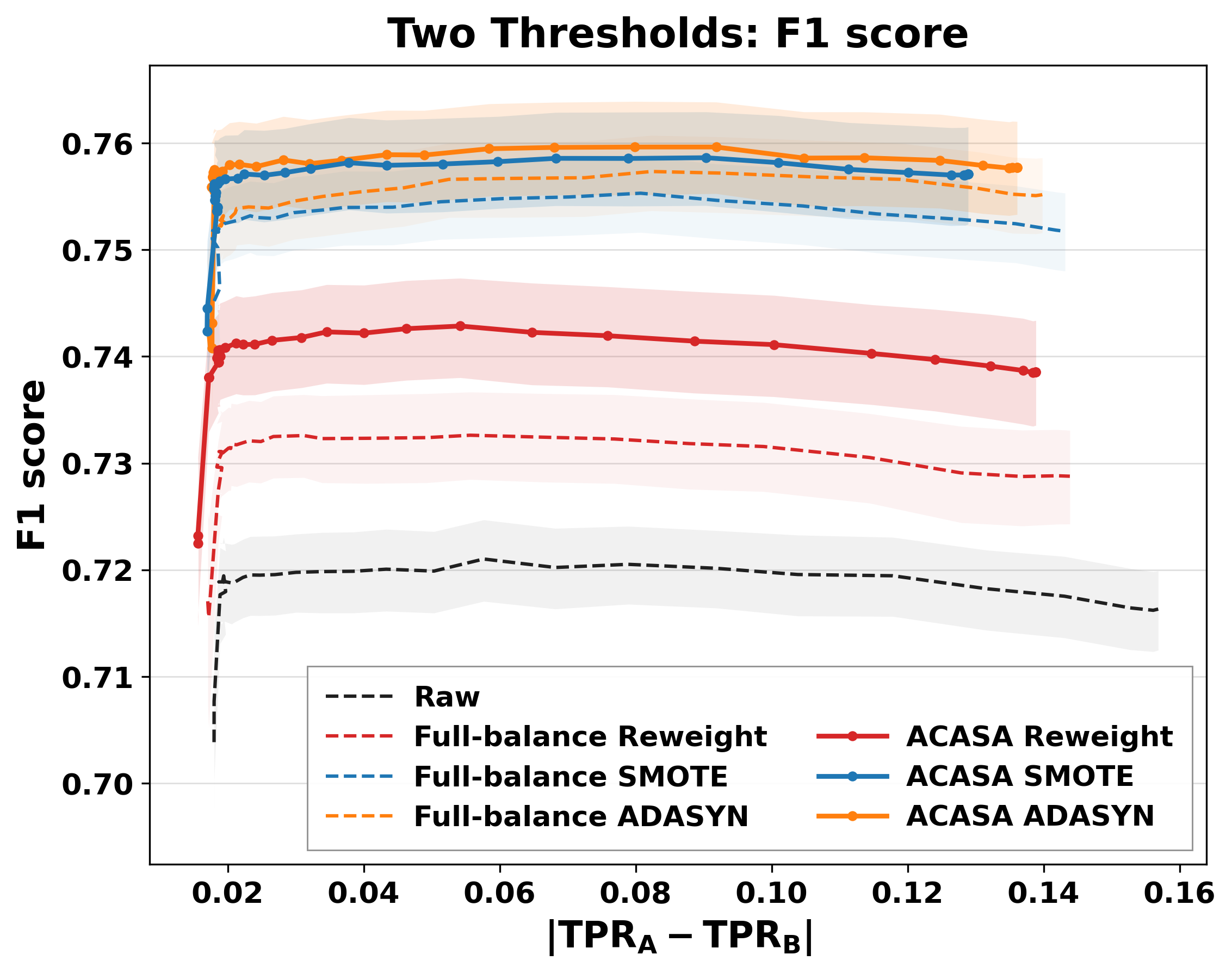}
    \par\smallskip
    {\small (b) F1 score}
  \end{minipage}\hfill
  \begin{minipage}[t]{0.32\textwidth}
    \centering
    \includegraphics[width=\linewidth]{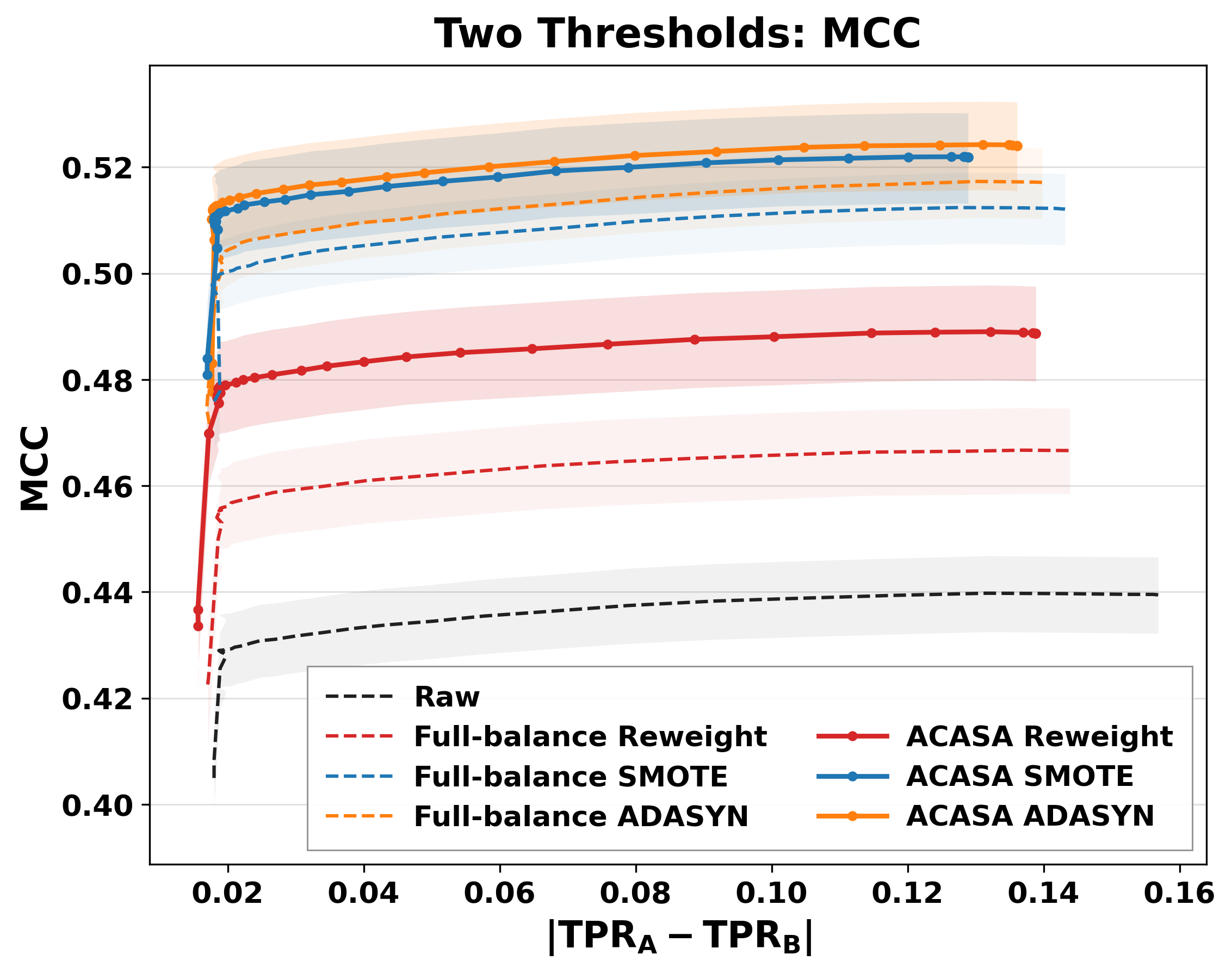}
    \par\smallskip
    {\small (c) MCC}
  \end{minipage}

  \caption{Fairness--performance trade-offs measured by
  (a) accuracy, (b) F1 score, and (c) Matthews correlation
  coefficient (MCC). 
  For each method, we select separate thresholds for groups $A$ and $B$ to minimize prediction error on the calibration set subject to an Equal Opportunity gap at most $\epsilon$, and vary $\epsilon$ to obtain the curves.}
  \label{fig:tradeoff}
\end{figure*}

\paragraph{Fairness–performance trade-off.} When a prespecified fairness level is desired, augmentation can be combined with post-processing to satisfy a fairness constraint. By reducing the fairness disparity beforehand, augmentation may require smaller threshold adjustments, thereby mitigating the associated loss in predictive performance caused by post-processing. We examine this benefit through group-specific threshold adjustment. We divide the training data into fitting and calibration subsets, using the former for augmentation and model fitting and the latter to select thresholds minimizing prediction error subject to an EO-gap constraint. Figure \ref{fig:tradeoff} reports test-set accuracy, F1 score, and MCC against the resulting EO gap as the fairness constraint varies. At comparable EO gaps, ACASA generally outperforms the corresponding full-balancing methods, with SMOTE and ADASYN achieving the strongest predictive performance over much of the common range.

\subsection{Real Data Analysis}\label{sec:realdataapp}

We evaluate ACASA on the COMPAS two-year recidivism dataset \citep{larson2016we, he2025fairness}, using two-year recidivism as the binary outcome and sex as the sensitive attribute to divide groups,  with A = male and B = female. Predictors include age, the number of prior offenses, three juvenile-offense counts, and a felony-charge indicator, excluding sex, race, and the original COMPAS risk score. For each of $100$ random seeds, we stratify observations by group--label cell, allocate $30\%$ to the test set, and further reserve $30\%$ of the remaining observations for calibration, with the rest used for model fitting. We compare raw training, full balancing, and ACASA using logistic regression with reweighting (Appendix~\ref{app:reweighting}), SMOTE, and diffusion model. Two ACASA variants are trained separately to minimize the EO gap subject to constraints on decreases in cross-validated accuracy or MCC score. Here, corresponding to Remark \ref{remark:choiceofvalidation}, we additionally consider an MCC-constrained variant to assess the empirical flexibility of ACASA beyond the prediction-error criterion analyzed theoretically. Further implementation details are provided in  Appendix~\ref{sec:realdata-detail}.

Figure~\ref{fig:real_fairness} reports the test-set EO gap before post-processing at a common classification threshold of $0.5$, tracking successive iterations under accuracy and MCC constraints. Across both constraints, full balancing reduces the EO gap relative to raw training for all three augmentation methods, while ACASA achieves further reductions, primarily at the first accepted update. All three ACASA variants attain lower final mean EO gaps than their full-balancing counterparts. SMOTE and the diffusion model achieve similar final mean EO gaps under the accuracy constraint, while the diffusion model attains the lowest final mean EO gap under the MCC constraint. SMOTE also outperforms reweighting under both constraints.

\begin{figure}[!htbp]
  \centering
  \begin{subfigure}[t]{0.49\linewidth}
    \centering
    \includegraphics[width=\linewidth]{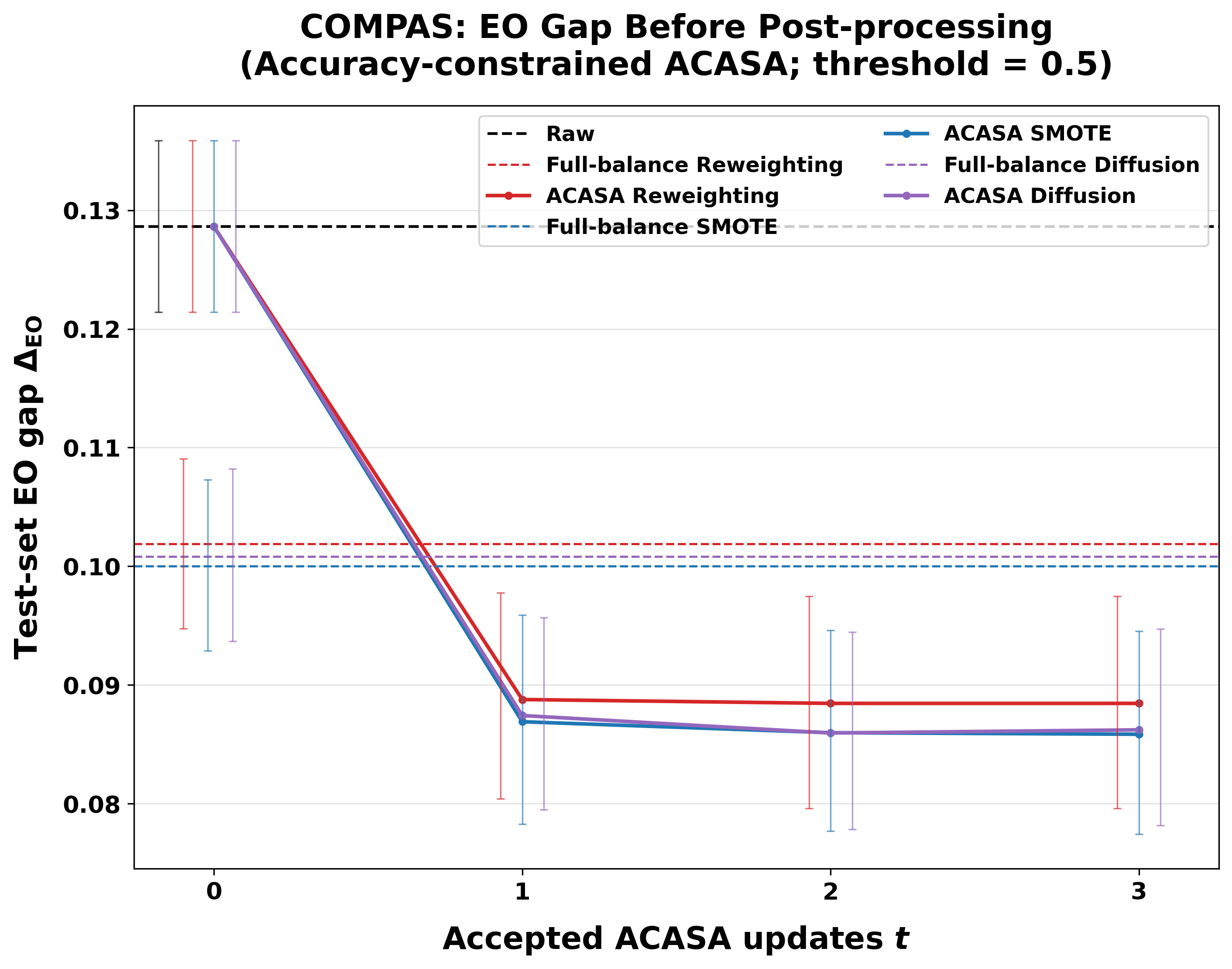}
    \caption{Accuracy-constrained ACASA.}
    \label{fig:real_EOO}
  \end{subfigure}\hfill
  \begin{subfigure}[t]{0.49\linewidth}
    \centering
    \includegraphics[width=\linewidth]{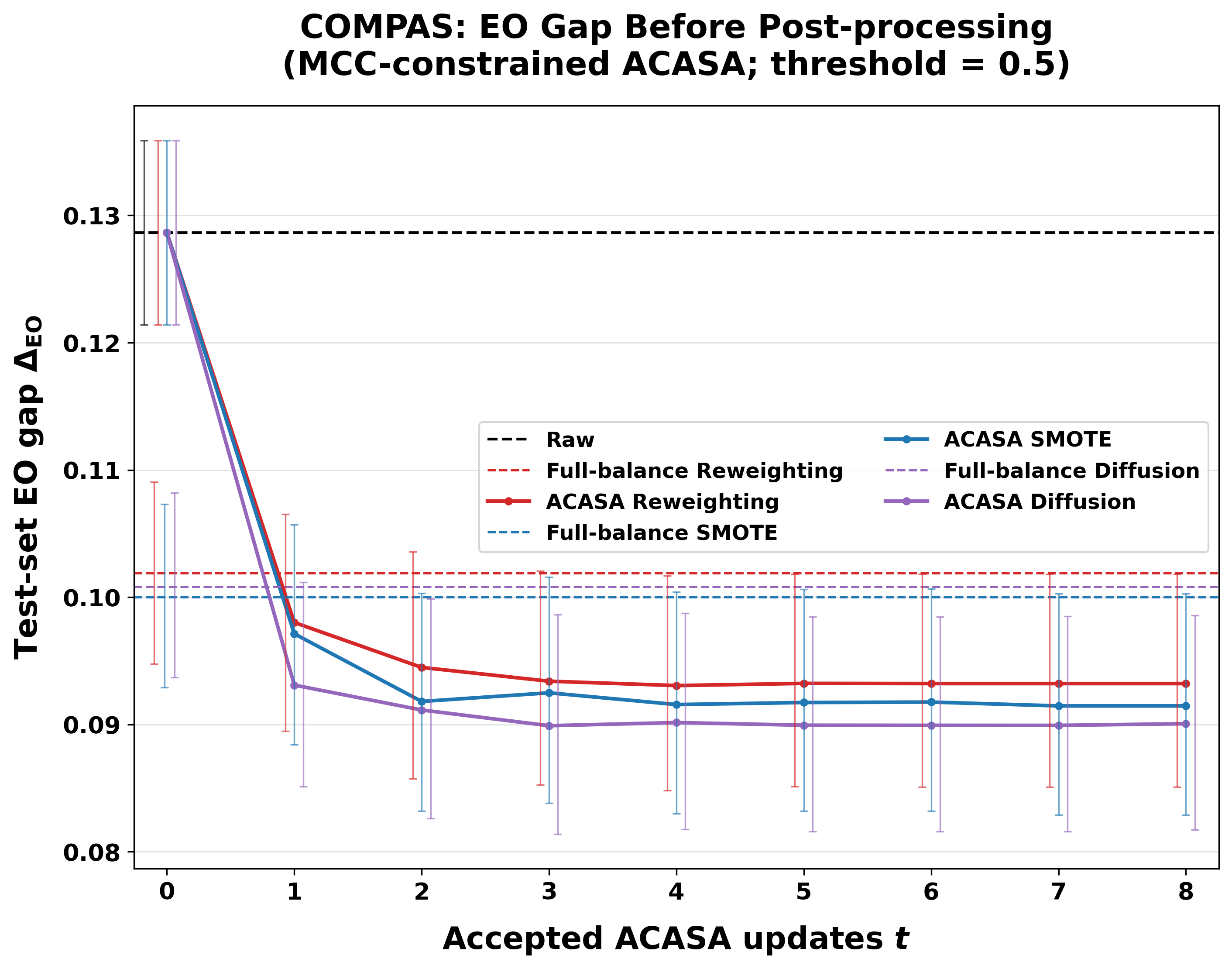}
    \caption{MCC-constrained ACASA.}
    \label{fig:real_EOO_MCC}
  \end{subfigure}

  \caption{Equal Opportunity gap on the COMPAS data before post-processing. Both panels report the test-set EO gap at the common classification threshold of $0.5$. Solid curves show mean ACASA trajectories for reweighting, SMOTE, and diffusion model, while dashed lines show raw training and the corresponding full-balancing methods. Results are averaged over $100$ runs, and error bars give pointwise $95\%$ bootstrap confidence intervals based on $5{,}000$ resamples. Runs that stop early retain their final accepted allocation.}
  \label{fig:real_fairness}
\end{figure}

\begin{figure}[t]
  \centering
  \begin{subfigure}[t]{0.49\linewidth}
    \centering
    \includegraphics[width=\linewidth]{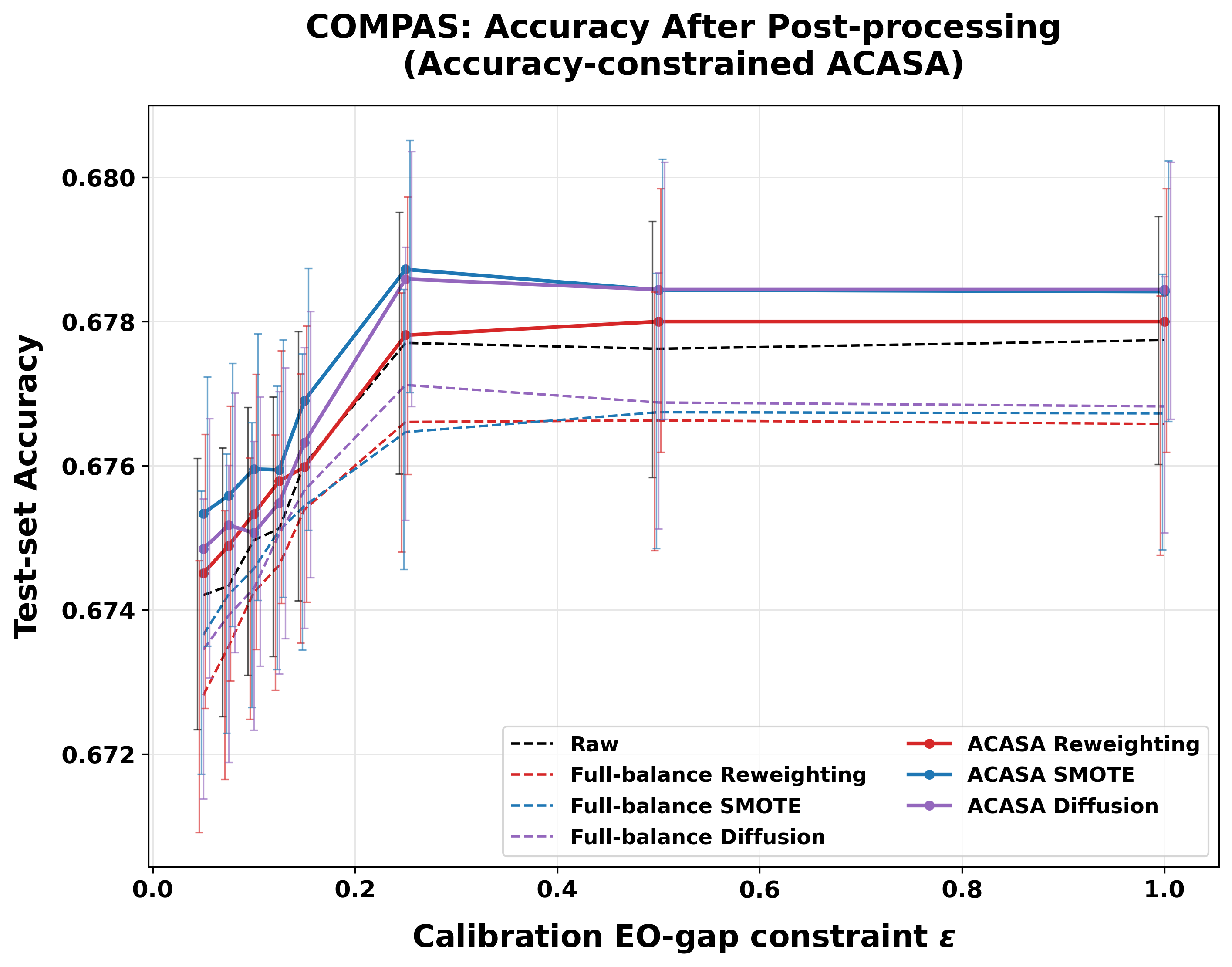}
    \caption{Test-set accuracy.}
    \label{fig:real_iteration}
  \end{subfigure}\hfill
  \begin{subfigure}[t]{0.49\linewidth}
    \centering
    \includegraphics[width=\linewidth]{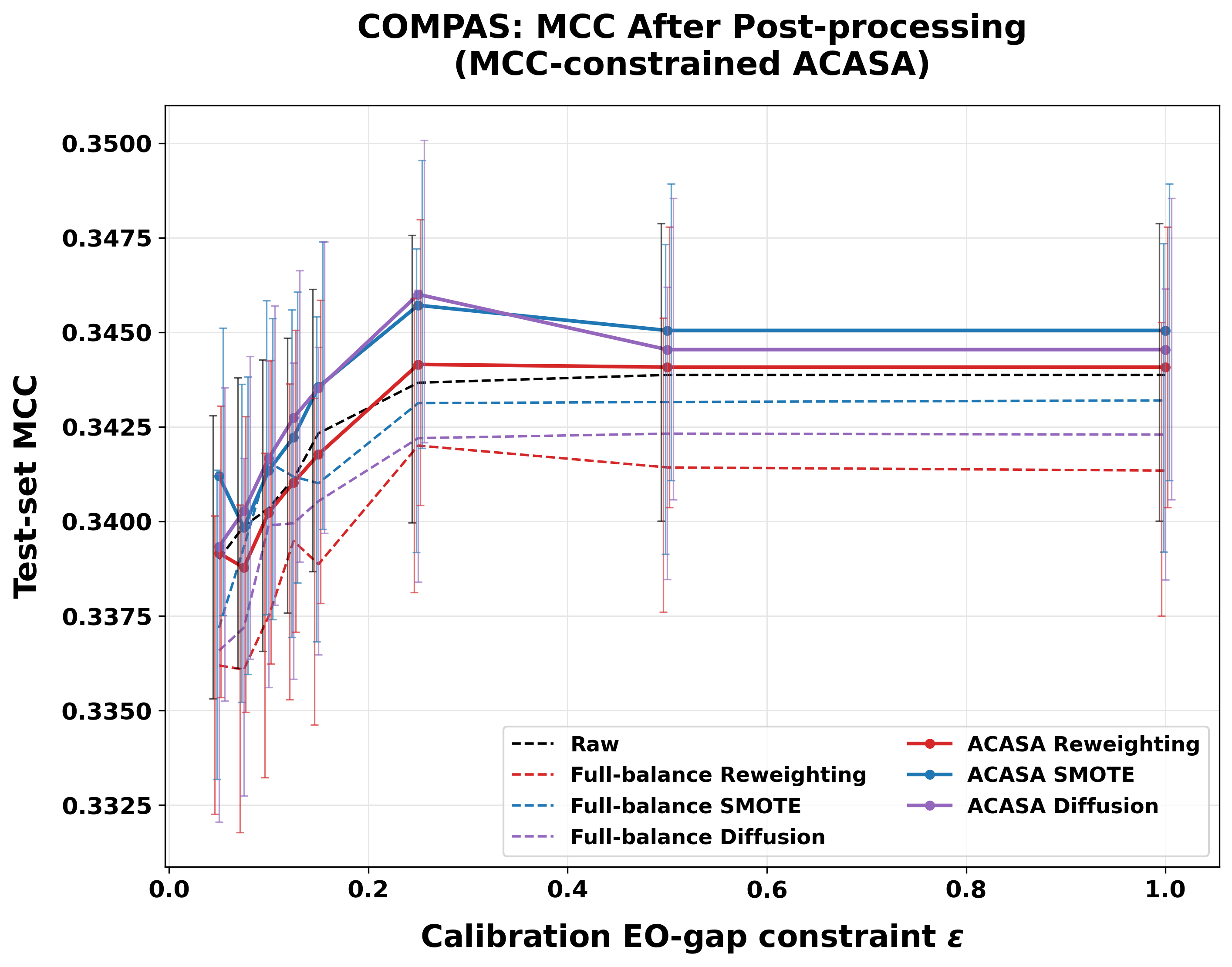}
    \caption{Test-set MCC.}
    \label{fig:real_iteration_MCC}
  \end{subfigure}

  \caption{Predictive performance on the COMPAS data after group-specific threshold post-processing. The left panel reports test-set accuracy under accuracy-constrained ACASA, and the right panel reports test-set MCC under MCC-constrained ACASA. Solid curves show the final ACASA allocations, while dashed curves show raw training and full balancing. Results are averaged over $100$ runs, and error bars give pointwise $95\%$ bootstrap confidence intervals based on $5{,}000$ resamples. The value $\epsilon=1$ corresponds to unconstrained threshold selection, not to the absence of post-processing.}
  \label{fig:real_performance}
\end{figure}

We next evaluate predictive performance after group-specific threshold post-processing. For each calibration-set EO-gap constraint $\epsilon$, thresholds are selected to maximize accuracy or MCC on the calibration set, with performance evaluated on the test set. Figure~\ref{fig:real_performance} shows that predictive performance generally improves as $\epsilon$ is relaxed, with ACASA outperforming the corresponding full-balancing methods across both metrics. For both accuracy- and MCC-constrained ACASA procedures, SMOTE and the diffusion model achieve similar mean performance and generally outperform reweighting. Full balancing can even underperform raw training, whereas the corresponding ACASA procedures improve upon it. These findings highlight the benefit of adapting augmentation allocations to predictive-performance constraints rather than enforcing exact cell balancing.

\section{Discussion}\label{sec:discussion} 

This work develops a theoretical and methodological framework for improving Equal Opportunity through synthetic data augmentation. We characterize how prevalence imbalance and group-size imbalance contribute to fairness disparities and establish conditions under which full balancing improves fairness. We further propose ACASA to adaptively allocate synthetic observations while controlling predictive error. Our theoretical guarantees and numerical results demonstrate the potential of synthetic augmentation to achieve favorable fairness–accuracy trade-offs.

Several promising directions remain for future research. Extending the framework to multiple or intersecting sensitive groups, multiclass classification, and other fairness criteria, such as equalized odds, would broaden its applicability. Because the generator error enters Theorems \ref{thm:fullbalance-improvement}--\ref{thm:full-balance-convergence} weighted by how much each cell is augmented, jointly optimizing the synthetic generation mechanism and augmentation allocation may further improve fairness–accuracy trade-offs, particularly when different generators exhibit heterogeneous performance across group–label cells. Incorporating computational or generation-budget constraints into adaptive allocation could improve efficiency in large-scale applications. Finally, extending the framework to settings with distribution shift or evolving populations would provide a foundation for studying the  effects of synthetic augmentation on algorithmic fairness.

Our theoretical guarantees rely on structural conditions governing the classification boundary, generator approximation quality, and local behavior of the augmentation objective.  Appendix \ref{sec:remarksonassumptions} further discusses the scope of these assumptions and their relationship to practical synthetic-data generators.

\bibliography{ref}
\bibliographystyle{abbrvnat}

\newpage

\appendix
\section{Full Statements of the Main Results}\label{sec:full-statements}

The main text states each result in its simplest form. This appendix gives the full statements, with explicit error terms and constants, that are proved in Appendix~\ref{sec:proofs}, and shows how the main-text versions follow from them. Define $\epsilon_{\mathrm{cal},Q}:=\sup_x|f^*_Q(x)-\eta_Q(x)|$ as the calibration error of the model class, and let $L$, $\mu_Q$, $B$, $L_\nabla$, $D_\Theta$ denote score-Lipschitz, loss-curvature, gradient-bound, gradient-Lipschitz and parameter-diameter constants of the assumptions in Appendix~\ref{sec:assumptiondetails}.

\subsection{Imbalance Mechanisms}\label{sec:imbalancemechanisms}
 
Define the calibration and estimation error of $\widehat f_P$ by
\begin{equation}\label{eq:RP-def}
R_P(\delta)
:=
(C_A+C_B)\,\varepsilon_n(\delta),
\qquad
\varepsilon_n(\delta):=\epsilon_{\mathrm{cal},P}+\frac{L}{\mu_P}\sqrt{\frac{dB^2\log(2d/\delta)}{2n_P}},
\end{equation}
where $C_A,C_B$ are the margin constants of Condition~3 in Assumption~\ref{assumption:prevalence-diff}. Apart from $\epsilon_{\mathrm{cal},P}$, $R_P(\delta)$ vanishes as $n_P\to\infty$.
 
\begin{fulltheorem}{Thm:fairness-improvement}\label{Thm:fairness-improvement-full}
Under Assumption~\ref{assumption:prevalence-diff}, with probability at least $1-\delta$,
\[
\Delta_{\mathrm{EO}}(\widehat h_P)\;\ge\;
\frac{\kappa\,c_\Psi}{\bar\omega_A\bar\omega_B}\,
\log\!\Bigl(1+(\mathrm{OR}-1)\bigl(\bar\omega_B+\kappa\bigr)\Bigr)-R_P(\delta).
\]
\end{fulltheorem}

\begin{fulltheorem}{thm:imbalance-driven-bound}\label{thm:imbalance-driven-bound-full}
Under Assumption~\ref{assumption:group-imb-assmp}, with probability at least $1-\delta$,
\[
\Delta_{\mathrm{EO}}(\widehat h_P)\ge\frac{\kappa\,c_\Psi\,c'}{\bar\omega_A\bar\omega_B}\Bigl(\alpha-\frac12\Bigr)-R_P(\delta).
\]
\end{fulltheorem}
 
The population statements of Theorems~\ref{Thm:fairness-improvement} and~\ref{thm:imbalance-driven-bound} are the display \eqref{eq:eo-lower} in the proof of Theorem~\ref{Thm:fairness-improvement} and \eqref{eq:EO-population-gap} in the proof of Theorem~\ref{thm:imbalance-driven-bound}.

\subsection{Full Balancing}\label{sec:fullbalancingfullstatement}
 
Let $n_{\overline P_{\mathrm{fb}}}:=4n_{\max}$ be the pooled sample size. Write $\Delta_P:=\epsilon_{\mathrm{cal},P}+\frac{L}{\mu_P}\sqrt{dB^2\log(4d/\delta)/(2n_P)}$, so that $R_P(\delta/2)=(C_A+C_B)\Delta_P$. For the augmented classifier, define the mixture discrepancy
\[
\epsilon_{\mathrm{mix}}:=\sum_{c\in\mathcal C}\left|\frac{n_c}{n_{\overline P_{\mathrm{fb}}}}-\frac{p_c}{4p^*}\right|\epsilon_c,
\]
the estimation term
\begin{align*}
e_n(\overline P_{\mathrm{fb}}):=\frac{L}{\mu_{\overline P_{\mathrm{fb}}}}
&\Bigg[
\frac{n_P}{n_{\overline P_{\mathrm{fb}}}}
\left(
B\sqrt{\frac{d\log\!\left(8d(3n_P)^d/\delta\right)}{2n_P}}
+\frac{2\sqrt d L_\nabla D_\Theta}{n_P}
\right)
\\
&\qquad+\frac{1}{n_{\overline P_{\mathrm{fb}}}}
B\sqrt{\frac{(n_{\overline P_{\mathrm{fb}}}-n_P)d\log(8d/\delta)}{2}}
+\sqrt d B\epsilon_{\mathrm{mix}}
\Bigg],
\end{align*}
whose three terms correspond to the real observations, the synthetic observations and the mixture discrepancy, and $\Delta_{\overline P_{\mathrm{fb}}}:=\epsilon_{\mathrm{cal},\overline P_{\mathrm{fb}}}+e_n(\overline P_{\mathrm{fb}})$. Finally, with $C_{\mathrm{fb}}$ and $\overline C_A,\overline C_B$ the margin constants of Condition~2 in Assumption~\ref{assumption:improvement-bound-assumption} and $h_{\mathrm{fb}}$ as in Theorem~\ref{thm:fullbalance-improvement}, let
\begin{equation}\label{eq:Cq}
C_q:=8\sqrt{C_{\mathrm{fb}}}.
\end{equation}
 
\begin{fulltheorem}{thm:fullbalance-improvement}\label{thm:fullbalance-improvement-full}
Under Assumption~\ref{assumption:improvement-bound-assumption}, with probability at least $1-\delta$,
\[
\Delta_{\mathrm{EO}}(\widehat h_P)-\Delta_{\mathrm{EO}}(\widehat h_{\overline P_{\mathrm{fb}}})
\ge
\frac{\kappa\,c_\Psi}{\bar\omega_A\bar\omega_B}\,h_{\mathrm{fb}}
-C_q\sqrt{\epsilon_{\mathrm{syn},\mathrm{fb}}}
-R_P(\delta/2)-(\overline C_A+\overline C_B)\,\Delta_{\overline P_{\mathrm{fb}}}.
\]
\end{fulltheorem}
 
The main paper Theorem~\ref{thm:fullbalance-improvement} is the population bound \eqref{eq:eo-hp-hpfb-population-lower} in the proof.

\begin{fulltheorem}{thm:maintain-accuracy-fb}\label{thm:maintain-accuracy-fb-full}
Under Assumption~\ref{assumption:accuracypreserveass}, with probability at least $1-\delta$,
\[
\Err_{\widetilde P_{\mathrm{fb}}}(\widehat h_{\overline P_{\mathrm{fb}}})
\le
\inf_h\Err_{\widetilde P_{\mathrm{fb}}}(h)
+2\epsilon_{\mathrm{syn},\mathrm{fb}}
+2\Delta_{\overline P_{\mathrm{fb}}}.
\]
In particular, the bound holds with $\inf_h\Err_{\widetilde P_{\mathrm{fb}}}(h)$ replaced by $\Err_{\widetilde P_{\mathrm{fb}}}(\widehat h_P)$.
\end{fulltheorem}

\subsection{Convergence}
 
\begin{fulltheorem}{thm:full-balance-convergence}\label{thm:full-balance-convergence-full}
Under Assumption~\ref{assumption:convergence}, with probability at least $1-\delta$,
\[
\bigl|\Delta_{\mathrm{EO}}(\widehat h_{\overline P_{\mathrm{fb}}})-\Delta_{\mathrm{EO}}(h^*_{\mathrm{bal}})\bigr|
\le
(C^*_A+C^*_B)\Bigl(e_n(\overline P_{\mathrm{fb}})+\frac{L\sqrt d\,B}{\mu_{\mathrm{syn}}}\,\epsilon_{\mathrm{syn},\mathrm{fb}}\Bigr).
\]
Here $C^*_A,C^*_B$ are the margin constants of $f^*_{\mathrm{bal}}:=f_{\theta^*_{\mathrm{bal}}}$ in Condition~3 of Assumption~\ref{assumption:convergence}, and $\mu_{\mathrm{syn}}$ is the curvature of the population loss $L_{\overline P_{\mathrm{fb}}}$ between $\theta^*_{\widetilde P_{\mathrm{fb}}}$ and $\theta^*_{\overline P_{\mathrm{fb}}}$, whereas $\mu_{\overline P_{\mathrm{fb}}}$ in $e_n$ is the curvature of the empirical loss.
\end{fulltheorem}

\subsection{Adaptive Augmentation}\label{sec:appendixadaptivedetail}

\paragraph{Fold notation.}
For fold $k$, let $n_c^{(k)}$ and $n_c^{(-k)}$ denote the validation and training cell sizes, $n_{\mathrm{val}}^{(k)}=\sum_cn_c^{(k)}$, $n_{\mathrm{tr}}^{(k)}=\sum_cn_c^{(-k)}$, with empirical cell proportions $\widehat p_c^{(k)}=n_c^{(k)}/n_{\mathrm{val}}^{(k)}$ and $\widehat p_c^{(-k)}=n_c^{(-k)}/n_{\mathrm{tr}}^{(k)}$. Define
\[
n_{s+}^{\mathrm{val}}=\min_{k}n_{s+}^{(k)},
\qquad
n_{\mathrm{val}}=\min_{k}n_{\mathrm{val}}^{(k)},
\qquad
n_{\mathrm{cell,tr}}=\min_{k,\,c}n_c^{(-k)}.
\]
Let $N_\omega=|\Lambda_\omega|$ and $m^{(-k)}(\boldsymbol\lambda)=\sum_cm_c^{(-k)}(\boldsymbol\lambda)$, the number of synthetic observations generated in the training portion of fold $k$. With $P^{(-k)}_{\mathrm{syn},c}$ the generator fitted on that portion, the fold-specific augmented distribution is
\[
\overline P_{\boldsymbol\lambda}^{(-k)}
:=
\frac{\sum_cn_c^{(-k)}P_c+\sum_cm_c^{(-k)}(\boldsymbol\lambda)P_{\mathrm{syn},c}^{(-k)}}{n_{\mathrm{tr}}^{(k)}+m^{(-k)}(\boldsymbol\lambda)},
\]
abbreviated $Q_{k,\boldsymbol\lambda}$ in the proofs.

\paragraph{Discrepancies.}
The discrepancy between the rounded fold-specific and the continuous full-sample augmented distributions is
\[
\begin{aligned}
\epsilon_{\mathrm{syn,CV}}
:=&
\sup_{\boldsymbol\lambda\in\Lambda}\max_{k}
\Bigg\{
\frac12\sum_c\left|\frac{n_c^{(-k)}}{n_{\mathrm{tr}}^{(k)}+m^{(-k)}(\boldsymbol\lambda)}-\frac{p_c}{1+r\sum_{c'}\lambda_{c'}}\right|
\\
&+\frac12\sum_c\left|\frac{m_c^{(-k)}(\boldsymbol\lambda)}{n_{\mathrm{tr}}^{(k)}+m^{(-k)}(\boldsymbol\lambda)}-\frac{r\lambda_c}{1+r\sum_{c'}\lambda_{c'}}\right|
+\sum_c\frac{m_c^{(-k)}(\boldsymbol\lambda)}{n_{\mathrm{tr}}^{(k)}+m^{(-k)}(\boldsymbol\lambda)}\left(\epsilon_c+\epsilon_c^{(-k)}\right)
\Bigg\},
\end{aligned}
\]
whose first two terms capture the real and synthetic mixture weights, including integer rounding, and whose last term captures the generator errors. The validation stratification discrepancy is $\epsilon_{\mathrm{strat,val}}:=\frac12\max_k\sum_c|\widehat p_c^{(k)}-p_c|$.

\paragraph{Cross-validation errors.}
Let $\mu$ be the loss-curvature constant and $C^\Lambda_A$, $C^\Lambda_B$, $C^\Lambda_P$ the margin constants of Assumption~\ref{assumption:adaptive_search}, uniform over the candidate allocations, and write
\[
\begin{aligned}
E_{\mathrm{tr}}(\delta):={}&
\sqrt{\frac{dB^2\log\bigl(32dKN_\omega(3n_{\mathrm{cell,tr}})^d/\delta\bigr)}{2n_{\mathrm{cell,tr}}}}
+\frac{2\sqrt dL_\nabla D_\Theta}{n_{\mathrm{cell,tr}}}
\\
&+\sup_{k,\,\boldsymbol\lambda\in\Lambda_\omega}\frac{m^{(-k)}(\boldsymbol\lambda)}{n_{\mathrm{tr}}^{(k)}+m^{(-k)}(\boldsymbol\lambda)}\sqrt{\frac{dB^2\log(8dKN_\omega/\delta)}{2\bigl(m^{(-k)}(\boldsymbol\lambda)\vee1\bigr)}}
\end{aligned}
\]
for the training-side estimation error, and
\[
e_{\mathrm{cv}}(\delta):=\frac{L}{\mu}\Bigl(E_{\mathrm{tr}}(\delta)+\sqrt d\,B\,\epsilon_{\mathrm{syn,CV}}\Bigr)
\]
for the resulting distance between the fold-specific trained scores and the full-sample population scores. The uniform cross-validation errors are
\begin{align*}
\epsilon_{\mathrm{fair}}(\delta)
&:=
\sqrt{\frac{\log(16KN_\omega/\delta)}{2n_{A+}^{\mathrm{val}}}}
+\sqrt{\frac{\log(16KN_\omega/\delta)}{2n_{B+}^{\mathrm{val}}}}
+(C^\Lambda_A+C^\Lambda_B)\,e_{\mathrm{cv}}(\delta),
\\
\epsilon_{\mathrm{acc}}(\delta)
&:=
2\left[
\sqrt{\frac{\log(8KN_\omega/\delta)}{2n_{\mathrm{val}}}}
+\epsilon_{\mathrm{strat,val}}
+C^\Lambda_P\,e_{\mathrm{cv}}(\delta)
\right].
\end{align*}

\paragraph{Grid terms and constants.}
Let $\mu_{\mathrm{aug}}$, $H_{\mathrm{aug}}$ and $G_{\mathrm{aug}}$ be lower and upper Hessian bounds and a gradient bound for $\boldsymbol\lambda\mapsto\Delta_{\mathrm{EO}}(h^*_{\overline P_{\boldsymbol\lambda}})$ near $\boldsymbol\lambda^*$. Let $G_{\mathrm{acc}}$, $H_{\mathrm{acc}}$ be gradient and Hessian bounds for the population error. Let $m_{\mathrm{acc}}$ be the rate at which the error can be decreased in some direction near the boundary of $\Lambda_{\mathrm{acc}}$, and $\epsilon_0$, $\rho_{\mathrm{acc}}$ be the tolerance and radius over which these conditions hold. Define the grid discretization error and the feasibility correction radius
\[
D_\omega:=G_{\mathrm{acc}}\omega+\frac{H_{\mathrm{acc}}}{2}\omega^2,
\qquad
R_\omega(\delta):=\omega+\frac{2}{m_{\mathrm{acc}}}\bigl[2\epsilon_{\mathrm{acc}}(\delta)+D_\omega\bigr].
\]
The conditions of Theorem~\ref{thm:adaptive_near_optimality} are
\begin{equation}\label{eq:acasa-conditions}
\epsilon_{\mathrm{acc}}(\delta)+D_\omega\le\epsilon_0,
\quad
2\epsilon_{\mathrm{acc}}(\delta)+D_\omega\le\frac{m_{\mathrm{acc}}^2}{2H_{\mathrm{acc}}},
\quad
R_\omega(\delta)\le\rho_{\mathrm{acc}},
\quad
\rho_{\mathrm{saf}}\ge\frac{G_{\mathrm{aug}}}{H_{\mathrm{aug}}}+R_\omega(\delta).
\end{equation}

\begin{fulltheorem}{thm:adaptive_near_optimality}\label{thm:adaptive_near_optimality-full}
Under Assumption~\ref{assumption:adaptive_search}, condition on the group--label cells $(S_i,Y_i)_{i=1}^{n_P}$ of the real observations and on the fold partition, which determine all count-dependent quantities above, and suppose that \eqref{eq:acasa-conditions} holds, with probability at least $1-\delta$,
\begin{align}
\label{eq:allocation_convergence}
\bigl\|\widehat{\boldsymbol\lambda}-\boldsymbol\lambda^*\bigr\|_2
\le
\left\{
\frac{2H_{\mathrm{aug}}}{\mu_{\mathrm{aug}}^2}
\left[
\xi_{\mathrm{fair}}+2\epsilon_{\mathrm{fair}}(\delta)+G_{\mathrm{aug}}R_\omega(\delta)+\frac{H_{\mathrm{aug}}}{2}R_\omega(\delta)^2
\right]
\right\}^{1/2}
+\sqrt{\frac{4G_{\mathrm{aug}}}{\mu_{\mathrm{aug}}m_{\mathrm{acc}}}\epsilon_{\mathrm{acc}}(\delta)}.
\end{align}
\end{fulltheorem}
The main paper Theorem~\ref{thm:adaptive_near_optimality} follows from \eqref{eq:allocation_convergence}. See the proof of Theorem~\ref{thm:adaptive_near_optimality} in Appendix~\ref{sec:proofs}.

\begin{proposition}[Excess Equal Opportunity gap]\label{prop:excess-eo}
Suppose Assumption~\ref{assumption:adaptive_search} holds with \eqref{eq:augmentation_hessian} replaced by $\nabla_{\boldsymbol\lambda}^2\Delta_{\mathrm{EO}}(h^*_{\overline P_{\boldsymbol\lambda}})\preceq H_{\mathrm{aug}}I_4$ for some $H_{\rm aug}>0$, and without requiring $\boldsymbol\lambda^*$ to be unique. Suppose also that, for some solution $\boldsymbol\lambda^*$ of \eqref{eq:oracle_allocation} and some constant $\beta>0$,
\begin{equation}\label{eq:weighted-convexity}
\Delta_{\mathrm{EO}}\bigl(h^*_{\overline P_{\boldsymbol\lambda}}\bigr)-\Delta_{\mathrm{EO}}\bigl(h^*_{\overline P_{\boldsymbol\lambda^*}}\bigr)
\le\beta\Bigl[\nabla_{\boldsymbol\lambda}\Delta_{\mathrm{EO}}\bigl(h^*_{\overline P_{\boldsymbol\lambda}}\bigr)^\top(\boldsymbol\lambda-\boldsymbol\lambda^*)\Bigr]_+
\end{equation}
for every $\boldsymbol\lambda\in\Lambda$ with $\Err_P(h^*_{\overline P_{\boldsymbol\lambda}})-\Err_P(h^*_P)\le\xi_{\mathrm{acc}}+\epsilon_0$. Condition on the group--label cells of the real observations and on the fold partition, and suppose that \eqref{eq:acasa-conditions} holds. Then, with probability at least $1-\delta$,
\[
\Err_P\bigl(h^*_{\overline P_{\widehat{\boldsymbol\lambda}}}\bigr)-\Err_P(h^*_P)\le\xi_{\mathrm{acc}}+\epsilon_{\mathrm{acc}}(\delta)
\]
and
\begin{align*}
&\Delta_{\mathrm{EO}}\bigl(h^*_{\overline P_{\widehat{\boldsymbol\lambda}}}\bigr)-\Delta_{\mathrm{EO}}\bigl(h^*_{\overline P_{\boldsymbol\lambda^*}}\bigr)\\
&\qquad\le\beta\Bigl(\mathrm{diam}(\Lambda)+\frac{G_{\mathrm{aug}}}{H_{\mathrm{aug}}}\Bigr)\Bigl\{2H_{\mathrm{aug}}\Bigl[\xi_{\mathrm{fair}}+2\epsilon_{\mathrm{fair}}(\delta)+G_{\mathrm{aug}}R_\omega(\delta)+\frac{H_{\mathrm{aug}}}{2}R_\omega(\delta)^2\Bigr]\Bigr\}^{1/2}.
\end{align*}
In particular, if $\boldsymbol\lambda^{\mathrm{fb}}\in\Lambda_{\mathrm{acc}}$, the second bound also holds with $\boldsymbol\lambda^*$ replaced by $\boldsymbol\lambda^{\mathrm{fb}}$ on the left-hand side.
\end{proposition}

Both bounds concern the population-risk classifier $h^*_{\overline P_{\hat{\boldsymbol{\lambda}}}}$ at the selected allocation; the deployed classifier $\hat h_{\overline P_{\hat{\boldsymbol{\lambda}}}}$ differs from it by an estimation error of the kind controlled in \eqref{eq:proof_empirical_fairness} and \eqref{eq:proof_empirical_accuracy}.
Condition \eqref{eq:weighted-convexity} replaces curvature in $\boldsymbol\lambda$ by convexity in the cell composition: by Appendix~\ref{sec:disc-curvature}, under exact generation it holds with $\beta=1+r\max_{\boldsymbol\lambda\in\Lambda}\sum_c\lambda_c$ whenever the Equal Opportunity gap is a convex function of the cell composition.

\section{Detailed Assumptions}\label{sec:assumptiondetails}

\subsection{Technical Notations}\label{sec:technicalnotations}
The densities $q_{Q,y}$, the log-ratios $\Phi_Q$ and $\Psi_Q=\log\rho_Q+\Phi_Q$, the positive-prediction region $\mathcal{X}_Q^+=\{x:\Psi_Q(x)\geq 0\}$, the group shares $\omega_B(x)$ and $\bar\omega_s$ among the positives, and the odds ratio $\mathrm{OR}$ are defined in Section~\ref{sec:imbalance}. We write $f_Q^*:=f_{\theta_Q^*}$ and $\widehat f_Q:=f_{\widehat\theta_Q}$ for the population and empirical scores, so that $h_Q^*$ and $\widehat h_Q$ threshold them at $1/2$.

For a reference distribution $\widetilde P$ (prevalence-, group- or full-balanced), the flip region
\(
\mathcal{D}:=\mathcal{X}_{\widetilde P}^+\setminus\mathcal{X}_P^+
\)
is the set classified as positive by $h_{\widetilde P}^\eta$ but not by $h_P^\eta$. For $L>0$, let
\[
\mathcal{N}(L):=\{x:-L\leq\Psi_P(x)<0\}
\]
be the set of points whose posterior log-odds under $P$ lie within $L$ below zero. If $\Psi_{\widetilde P}\leq\Psi_P+L$, then $\mathcal{D}\subseteq\mathcal{N}(L)$, because $\Psi_P<0\leq\Psi_{\widetilde P}$ on $\mathcal{D}$. The bands $\mathcal{N}$, $\mathcal{N}_{\mathrm g}$ and $\mathcal{N}_{\mathrm{fb}}$ below are $\mathcal{N}(L)$ with $L=\log\mathrm{OR}$, $L_{\mathrm g}$ and $L_{\mathrm{fb}}$, each an upper bound on $\Psi_{\widetilde P}-\Psi_P$ for the corresponding reference.
Subsets of $\mathcal{X}$ defined through densities, such as $\mathcal{X}_Q^+$, $\mathcal{D}$ and $\mathcal{N}(L)$, and inclusions between them, are understood up to $\nu$-null sets, with $\nu$ the dominating measure of Condition~1 in Appendix~\ref{section:assumption-1}. This does not affect any probability under $P$, since every $P_c$ has a density with respect to $\nu$.

\subsection{Assumption for Theorem \ref{Thm:fairness-improvement}}\label{section:assumption-1}

\begin{assumption}\label{assumption:prevalence-diff}
Let $(\mathcal X,\mathcal F)$ be the feature space. We consider the following conditions.
\begin{enumerate}
    \item \textbf{Positivity and ordering.} (a) Suppose $0<\alpha<1$ and $0<\pi_A,\pi_B<1$. Under $P$, the measures $P(X\in\cdot,S=s,Y=y)$ have densities $p(x,s,y)$ with respect to a common $\sigma$-finite measure $\nu$. Write $p(x,y):=\sum_{s}p(x,s,y)$. Assume $p(x,1)>0$ and $p(x,0)>0$ for $\nu$-almost every $x$, so that $\eta_P(x)=p(x,1)/\bigl(p(x,0)+p(x,1)\bigr)\in(0,1)$ and $\Psi_P(x)$ is finite. (b) $\pi_A>\pi_B$.
 
    \item \textbf{Loss conditions.} Assume $L_Q$ and $\widehat{L}_Q$ are differentiable and $\theta_Q^*$, $\widehat{\theta}_Q$ are interior minimizers. Assume that differentiation and expectation can be interchanged at
$\theta_Q^*$ for $\ell$. Convexity at the minimizer $\left\langle \nabla\widehat{L}_Q(\theta)-\nabla\widehat{L}_Q(\widehat{\theta}_Q), \theta-\widehat{\theta}_Q\right\rangle\geq \mu_Q\|\theta-\widehat{\theta}_Q\|^2$ for $Q=P$, $\mu_Q>0$ and $\forall \theta\in\Theta$ almost surely.
    Assume bounded range for $\left\|\nabla\ell(\theta_Q^*;x,y)\right\|_\infty\leq B/2$ a.s. for $Q=P$. Lipschitz score $\mathop{\sup}\limits_{x}\left|f_{\theta}(x)-f_{\theta'}(x)\right|\leq L\|\theta-\theta'\|$.
 
    \item \textbf{Calibration and margin.} (a)~The population score satisfies $\mathop{\sup}\limits_{x}\left|f_{P}^*(x)-\eta_P(x)\right|\leq \epsilon_{\mathrm{cal},P}$. (b)~There are constants $C_A,C_B>0$ such that, for $s\in\{A,B\}$ and all $\varepsilon>0$,
    \[
    P_{s+}\bigl(|\eta_P(X)-\tfrac12|\le\varepsilon\bigr)\le C_s\,\varepsilon .
    \]
 
    \item \textbf{Sign at the balanced boundary.} $\mathrm{TPR}_A(h^\eta_{\widetilde P})\ge\mathrm{TPR}_B(h^\eta_{\widetilde P})$.
 
    \item \textbf{Boundary dominance.} For some $\kappa\in(0,\bar\omega_A]$, $\omega_B(x)\ge\bar\omega_B+\kappa$ for $\nu$-almost every $x$ in the band
    \(
    \mathcal N:=\bigl\{x:\ -\log\mathrm{OR}\le\Psi_P(x)<0\bigr\}.
    \)
 
    \item \textbf{Mass at the boundary.} Let $h_\kappa:=\log\bigl(1+(\mathrm{OR}-1)(\bar\omega_B+\kappa)\bigr)$. On $[-h_\kappa,0)$, the law of $\Psi_P(X)$ given $Y=1$ has a density bounded below by some $c_\Psi>0$. That is, $P\bigl(\Psi_P(X)\in J\mid Y=1\bigr)\ge c_\Psi\,|J|$ for every Borel set $J\subseteq[-h_\kappa,0)$.
\end{enumerate}
\end{assumption}

\subsection{Assumption for Theorem \ref{thm:imbalance-driven-bound}}\label{sec:second-assumption}
 
Suppose group $A$ is overrepresented, $\alpha=P(S=A)>1/2$. For $w\in[0,1]$ let $Q_w$ be the distribution with
\[
Q_w(S=B)=w,\quad Q_w(S=A)=1-w,\quad Q_w(X,Y\mid S=s)=P(X,Y\mid S=s),\quad s\in\{A,B\}.
\]
Then $Q_{1-\alpha}=P$, and the group-balanced reference is now $\widetilde P_{\mathrm g}:=Q_{1/2}$. In this subsection and in the proof of Theorem~\ref{thm:imbalance-driven-bound}, $\widetilde P$ denotes $\widetilde P_{\mathrm g}$. Equivalently, $\widetilde P=(1-\lambda_{\mathrm g}^*)P+\lambda_{\mathrm g}^*P'_B$ with $P'_B:=\pi_BP_{B+}+(1-\pi_B)P_{B-}$ and $\lambda_{\mathrm g}^*=1-1/(2\alpha)$: oracle group-$B$ observations are added with the prevalence and feature distribution of group $B$, which changes the representation of the groups but not $P(X,Y\mid S=s)$. Let
\[
L_{\mathrm g}:=\log\frac{\alpha}{1-\alpha},\qquad
\mathcal N_{\mathrm g}:=\bigl\{x:\ -L_{\mathrm g}\le\Psi_P(x)<0\bigr\},\qquad
h_{\mathrm g}:=c'\Bigl(\alpha-\frac12\Bigr).
\]
 
\begin{assumption}\label{assumption:group-imb-assmp}
Conditions 1(a), 2 and 3 of Assumption~\ref{assumption:prevalence-diff} hold, and Conditions~4--6 hold with $\widetilde P=\widetilde P_{\mathrm g}$, the band $\mathcal N$ replaced by $\mathcal N_{\mathrm g}$, and $h_\kappa$ replaced by $h_{\mathrm g}$. In addition, for $\nu$-almost every $x$ and all $w\in[1-\alpha,1/2]$,
\[
\partial_w\Psi_{Q_w}(x)\ge c'>0\ \text{ if } x\in\mathcal N_{\mathrm g},
\qquad
\partial_w\Psi_{Q_w}(x)\ge 0\ \text{ if } 0\le\Psi_P(x)<L_{\mathrm g}.
\]
\end{assumption}

\subsection{Assumption for Theorem \ref{thm:fullbalance-improvement}}\label{sec:assumptionforfb}
 
\paragraph{Oracle decomposition for the analysis.}
The full-balance intervention is applied directly. The following oracle distributions only decompose its effect along
\[
P\;\longrightarrow\;\widetilde P^{(1)}\;\longrightarrow\;\widetilde P^{(2)}\;\longrightarrow\;\widetilde P_{\mathrm{fb}}\;\longrightarrow\;\overline P_{\mathrm{fb}},
\]
that is, prevalence balancing, group-size balancing, class balancing, and replacement of oracle by synthetic augmentation. First, $\widetilde P^{(1)}:=(1-\lambda_1^*)P+\lambda_1^*P_{B+}$ is the prevalence-balanced distribution $\widetilde P$ of Section~\ref{sec:prevalence-imbalance}. Under it both groups have prevalence $\pi_A$, $\rho_{\widetilde P^{(1)}}=\rho_A$, and the share of group $A$ is
\[
\alpha_1:=\widetilde P^{(1)}(S=A)=\frac{\alpha(1-\pi_A)}{1-\bar\pi}.
\]
Second, when $\alpha_1>1/2$, let $Q_w$, $w\in[0,1]$, be the group-share family of Appendix~\ref{sec:second-assumption} built from $\widetilde P^{(1)}$ instead of $P$, i.e.  $Q_w(S=B)=w$ and $Q_w(X,Y\mid S=s)=\widetilde P^{(1)}(X,Y\mid S=s)$, and let
\[
\widetilde P^{(2)}:=Q_{1/2}=(1-\lambda_2^*)\widetilde P^{(1)}+\lambda_2^*P_B^{(\pi_A)},
\quad
P_B^{(\pi_A)}:=\pi_AP_{B+}+(1-\pi_A)P_{B-},
\quad
\lambda_2^*:=1-\frac{1}{2\alpha_1}.
\]
Under $\widetilde P^{(2)}$ the groups have mass $1/2$ and common prevalence $\pi_A$. Third, when $\pi_A\le1/2$, let
\[
\widetilde P_{\mathrm{fb}}:=(1-\lambda_3^*)\widetilde P^{(2)}+\lambda_3^*P_+,
\qquad
P_+:=\tfrac12P_{A+}+\tfrac12P_{B+},
\qquad
\lambda_3^*:=\frac{\frac12-\pi_A}{1-\pi_A};
\]
it gives mass $1/4$ to each cell and coincides with $\widetilde P_{\mathrm{fb}}$ of Section~\ref{sec:full-balance} (Lemma~\ref{lemma:class-shift}). Let
\begin{gather*}
L_{\mathrm g,1}:=\log\frac{\alpha_1}{1-\alpha_1},
\quad
L_{\mathrm{fb}}:=\log\mathrm{OR}+L_{\mathrm g,1}+\log\frac{1}{\rho_A},
\quad
\mathcal N_{\mathrm{fb}}:=\bigl\{x:\ -L_{\mathrm{fb}}\le\Psi_P(x)<0\bigr\},
\end{gather*}
and let $h_{\mathrm{fb}}$ be as in Theorem~\ref{thm:fullbalance-improvement}.
 
\begin{assumption}\label{assumption:improvement-bound-assumption}
We consider the following conditions.
\begin{enumerate}
\item \textbf{Oracle path.} $\alpha_1>1/2$ and $0<\pi_B<\pi_A\le1/2$. Conditions~1--3 of Assumption~\ref{assumption:prevalence-diff} hold, and Conditions~4--6 hold with $\widetilde P=\widetilde P_{\mathrm{fb}}$, the band $\mathcal N$ replaced by $\mathcal N_{\mathrm{fb}}$, and $h_\kappa$ replaced by $h_{\mathrm{fb}}$. The group-share condition of Assumption~\ref{assumption:group-imb-assmp} holds for the family $Q_w$ built from $\widetilde P^{(1)}$: for $\nu$-almost every $x$ and all $w\in[1-\alpha_1,1/2]$,
\[
\partial_w\Psi_{Q_w}(x)\ge c'>0\ \text{ if } x\in\mathcal N_{\mathrm{fb}},
\qquad
\partial_w\Psi_{Q_w}(x)\ge0\ \text{ if } 0\le\Psi_P(x)<L_{\mathrm g,1}.
\]
 
\item \textbf{Balanced margins.} Conditions 2 and 3(a) of
Assumption~\ref{assumption:prevalence-diff} hold with
$P$ replaced by $\overline P_{\mathrm{fb}}$.
In addition, for $s\in\{A,B\}$ and all
$\varepsilon>0$, assume that
\[
P_{s+}\left(
\left|\eta_{\overline P_{\mathrm{fb}}}(X)
-\frac12\right|\le\varepsilon
\right)
\le \overline C_s\varepsilon,
\]
for constants $\overline C_A,\overline C_B>0$,
where $P_{s+}=P(X\in\cdot\mid S=s,Y=1)$
denotes the original positive-cell distribution.
In addition, the margin condition holds for $\eta_{\widetilde P_{\mathrm{fb}}}$ in every cell: for some $C_{\mathrm{fb}}>0$, all $c\in\mathcal C$ and all $\varepsilon>0$,
\[
P_c\bigl(|\eta_{\widetilde P_{\mathrm{fb}}}(X)-\tfrac12|\le\varepsilon\bigr)\le C_{\mathrm{fb}}\,\varepsilon .
\]
 
\item For $(s,y)\in \{A+, A-,B+,B-\}$, the cellwise synthetic generators satisfy
\begin{align*}
d_{\mathrm{TV}}(P_{\mathrm{syn},(s,y)}(X),P(X\mid S=s,Y=y))&\le \epsilon_{(s,y)}.
\end{align*}
 
\item For the empirical analysis, additionally assume that \(\Theta\subset\mathbb R^d\) is compact with diameter \(D_\Theta<\infty\). For every cell \(c\in\mathcal C\), the loss gradient satisfies, for all \(\theta,\theta'\in\Theta\),
\(
\sup_{\theta\in\Theta}\|\nabla_\theta\ell(\theta;X,Y)\|_\infty\leq \frac B2,
\)
and
\(
\|\nabla_\theta\ell(\theta;X,Y)-\nabla_\theta\ell(\theta';X,Y)\|_\infty
\leq L_\nabla\|\theta-\theta'\|_2,
\)
for some constants \(B>0\) and \(L_\nabla<\infty\). These conditions hold almost surely under each real cell distribution \(P_c\) and, conditional on the original training data, under each fitted synthetic distribution \(P_{\mathrm{syn},c}\). Conditional on the original training data, the synthetic observations are mutually independent draws from their respective fitted cellwise generator distributions. Throughout, conditioning on the original training observations, whether the full sample or the training portion of a fold, also conditions on any randomness used to fit the generators from them, so that the fitted generators are fixed. Every condition in Appendices \ref{sec:full-statements} and \ref{sec:assumptiondetails} that involves a fitted generator or a distribution built from fitted generators, such as $P_{{\rm syn},c}$, $\overline P_{\rm fb}$, $\overline P_{\boldsymbol\lambda}$ or $P^{(-k)}_{\boldsymbol\lambda}$, holds for every realization of these observations and of this randomness, with constants that do not depend on the realization.
\end{enumerate}
\end{assumption}

\subsection{Assumption for Theorem \ref{thm:maintain-accuracy-fb}}\label{sec:maintainaccuracyassumption}
\begin{assumption}\label{assumption:accuracypreserveass}
Conditions~3 and~4 of Assumption~\ref{assumption:improvement-bound-assumption} hold, and Conditions~2 and~3(a) of Assumption~\ref{assumption:prevalence-diff} hold with $P$ replaced by $\overline P_{\mathrm{fb}}$.
\end{assumption}
 
The main paper Theorem~\ref{thm:maintain-accuracy-fb} uses only Condition~3 of Assumption~\ref{assumption:improvement-bound-assumption}. The other conditions give the concentration bound of Lemma~\ref{lemma:newconcentrationtwoQ} for $\overline P_{\mathrm{fb}}$.

\subsection{Assumption for Theorem \ref{thm:full-balance-convergence}}\label{sec:assumptionforconvergencethm}
 
\begin{assumption}\label{assumption:convergence}
We consider the following conditions.
\begin{enumerate}
\item Conditions~3 and~4 of Assumption~\ref{assumption:improvement-bound-assumption} hold, and Condition~2 of Assumption~\ref{assumption:prevalence-diff} holds with $P$ replaced by $\overline P_{\mathrm{fb}}$.
 
\item \textbf{Population minimizers.} $L_{\widetilde P_{\mathrm{fb}}}$ has a unique minimizer. $L_{\widetilde P_{\mathrm{fb}}}$ and $L_{\overline P_{\mathrm{fb}}}$ are differentiable with interior minimizers. Differentiation and expectation can be interchanged at $\theta^*_{\widetilde P_{\mathrm{fb}}}$ under both $Q\in\{\widetilde P_{\mathrm{fb}},\overline P_{\mathrm{fb}}\}$, and for some $\mu_{\mathrm{syn}}>0$,
\[
\bigl\langle
\nabla L_{\overline P_{\mathrm{fb}}}(\theta^*_{\widetilde P_{\mathrm{fb}}})-\nabla L_{\overline P_{\mathrm{fb}}}(\theta^*_{\overline P_{\mathrm{fb}}}),\,
\theta^*_{\widetilde P_{\mathrm{fb}}}-\theta^*_{\overline P_{\mathrm{fb}}}
\bigr\rangle
\ge
\mu_{\mathrm{syn}}\bigl\|\theta^*_{\widetilde P_{\mathrm{fb}}}-\theta^*_{\overline P_{\mathrm{fb}}}\bigr\|^2 .
\]
 
\item \textbf{Margin at the balanced target.} Condition~3(b) of Assumption~\ref{assumption:prevalence-diff} holds with $\eta_P$ replaced by $f^*_{\mathrm{bal}}:=f_{\theta^*_{\mathrm{bal}}}$: for some $C^*_A,C^*_B>0$, all $s\in\{A,B\}$ and all $\varepsilon>0$,
\[
P_{s+}\bigl(|f^*_{\mathrm{bal}}(X)-\tfrac12|\le\varepsilon\bigr)\le C^*_s\,\varepsilon .
\]
\end{enumerate}
\end{assumption}

\subsection{Assumption for Theorem \ref{thm:minimax-EO-lower}}\label{sec:minimaxassumption}
 
For a distribution $P_0\in\mathcal P$ and a cell $c=(s,y)$, let $P_{c,0}:=P_0(\,\cdot\mid S=s,Y=y)$. Write $\theta^*_0:=\theta^*_{\mathrm{bal},P_0}$ and $H_0:=\nabla^2_\theta L_{\mathrm{bal},P_0}(\theta^*_0)$. For $\theta\in\Theta$, let $h_\theta(x):=\mathbf 1\{f_\theta(x)\ge1/2\}$, and define the signed Equal Opportunity map of the baseline
\begin{equation}\label{eq:EO-map}
\Gamma(\theta):=P_{B+,0}\bigl(f_\theta(X)\ge\tfrac12\bigr)-P_{A+,0}\bigl(f_\theta(X)\ge\tfrac12\bigr),
\end{equation}
so that $\Delta_{\mathrm{EO},P_0}(h_\theta)=|\Gamma(\theta)|$ and, in particular, $\Delta_{\mathrm{EO},P_0}(h^*_{\mathrm{bal},P_0})=|\Gamma(\theta^*_0)|$.
 
\begin{assumption}\label{assumption:minimax-EO}
There exist a baseline distribution $P_0\in\mathcal P$ and a negative cell $c_0\in\{A-,B-\}$ such that the following conditions hold.
\begin{enumerate}
\item \textbf{Isolated population optimum.} The population minimizer $\theta^*_0$ is unique and belongs to $\operatorname{int}(\Theta)$. For every cell $c$, the map $\theta\mapsto\ell(\theta;z)$ is continuous on $\Theta$ for $P_{c,0}$-almost every $z$, and for every compact $K\subset\Theta$ there is an integrable envelope $M_{0,c,K}$ with $\sup_{\theta\in K}|\ell(\theta;Z)|\le M_{0,c,K}(Z)$ almost surely under $P_{c,0}$. There is an open neighborhood $\mathcal N_0$ of $\theta^*_0$ on which $\ell(\theta;z)$ is twice continuously differentiable in $\theta$ for $P_{c,0}$-almost every $z$, with integrable envelopes
\[
\sup_{\theta\in\mathcal N_0}\|\nabla_\theta\ell(\theta;Z)\|\le M_{1,c}(Z),
\qquad
\sup_{\theta\in\mathcal N_0}\|\nabla^2_\theta\ell(\theta;Z)\|_{\mathrm{op}}\le M_{2,c}(Z)
\]
almost surely under $P_{c,0}$. These conditions imply that $L_{\mathrm{bal},P_0}$ is twice continuously differentiable near $\theta^*_0$. Also assume $H_0\succ0$.
 
\item \textbf{Differentiable Equal Opportunity map.} The map $\Gamma$ in \eqref{eq:EO-map} is differentiable at $\theta^*_0$. Write $\gamma_{\mathrm{EO}}:=\nabla_\theta\Gamma(\theta^*_0)$.
 
\item \textbf{Boundary sensitivity.} The function
\(
\varphi(z):=\gamma_{\mathrm{EO}}^\top H_0^{-1}\,\nabla_\theta\ell(\theta^*_0;z)
\)
is not $P_{c_0,0}$-almost surely constant.
 
\item \textbf{Bounded tangent-submodel richness and compact parameter space.} There exists a fixed linear class $\mathcal H_{c_0}\subset L^\infty(P_{c_0,0})$ containing the constant functions, which is measure-determining for finite signed measures absolutely continuous with respect to $P_{c_0,0}$: if $\mu\ll P_{c_0,0}$ is a finite signed measure with $\int h\,d\mu=0$ for every $h\in\mathcal H_{c_0}$, then $\mu=0$. Every centered direction in $\mathcal H_{c_0}$ is the tangent of an admissible regular local submodel: for every $h\in\mathcal H_{c_0}$ with $\mathbb E_{P_{c_0,0}}[h]=0$, there exist $\varepsilon_h>0$ and cell distributions $\{Q_{t,h}:|t|<\varepsilon_h\}$ with $Q_{0,h}=P_{c_0,0}$ and $Q_{t,h}\ll P_{c_0,0}$, such that $t\mapsto r_{t,h}:=dQ_{t,h}/dP_{c_0,0}$ is continuously differentiable from $(-\varepsilon_h,\varepsilon_h)$ into $L^\infty(P_{c_0,0})$, with $r_{0,h}=1$ and $\frac{d}{dt}r_{t,h}|_{t=0}=h$. For $|t|<\varepsilon_h$, the distribution $P_{t,h}$ obtained from $P_0$ by replacing only the conditional distribution of cell $c_0$ by $Q_{t,h}$, with all other cell-conditional distributions and all cell probabilities unchanged, belongs to $\mathcal P$. Finally, $\Theta$ is compact.
\end{enumerate}
\end{assumption}

\subsection{Assumption for Theorem \ref{thm:adaptive_near_optimality}}\label{sec:assumptionforacasa}

By definition, the total synthetic sample size is $m(\boldsymbol\lambda)=\sum_{c\in\mathcal C}m_c(\boldsymbol\lambda)$.
For the population analysis, let
\(
r:=\frac{n_{\max}}{n_P},
\)
and define the continuous oracle and synthetic augmented distributions
\[
\widetilde P_{\boldsymbol\lambda}
:=
\frac{
P+r\sum_{c\in\mathcal C}\lambda_cP_c
}{
1+r\sum_{c\in\mathcal C}\lambda_c
},
\qquad
\overline P_{\boldsymbol\lambda}
:=
\frac{
P+r\sum_{c\in\mathcal C}\lambda_cP_{\mathrm{syn},c}
}{
1+r\sum_{c\in\mathcal C}\lambda_c
}.
\]
These population distributions use the continuous allocation parameter
$\boldsymbol\lambda$. The implemented procedure uses the rounded integer counts
$m_c(\boldsymbol\lambda)$ above.
The population target is
\begin{equation}
\label{eq:adaptive_population_problem}
\min_{\boldsymbol{\lambda}\in\Lambda}
\Delta_{\mathrm{EO}}
\left(h^*_{\overline P_{\boldsymbol{\lambda}}}\right)
\quad
\text{subject to}
\quad
\mathrm{Err}_{P}
\left(h^*_{\overline P_{\boldsymbol{\lambda}}}\right)
-
\mathrm{Err}_{P}
\left(h^*_{P}\right)
\leq
\xi_{\mathrm{acc}},
\end{equation}
where $\xi_{\mathrm{acc}}\geq0$ is the maximum allowed increase in prediction
error. Exact full balancing is a special case, corresponding to
\[
\lambda_c^{\mathrm{fb}}
=
1-\frac{n_c}{n_{\max}},
\qquad
m_c(\boldsymbol{\lambda}^{\mathrm{fb}})
=
n_{\max}-n_c.
\]

Because the population criteria in
\eqref{eq:adaptive_population_problem} are unknown, candidate allocations are
evaluated by $K$-fold cross-validation. Within each fold, both the synthetic
generator and the downstream classifier are fitted using only the training
portion, while fairness and prediction error are evaluated only on real
validation observations. For fold $k$, let
\[
n_{\max}^{(-k)}
=
\max_c n_c^{(-k)},
\qquad
m_c^{(-k)}(\boldsymbol{\lambda})
=
\left\lfloor
\lambda_c n_{\max}^{(-k)}
\right\rfloor,
\]
so that the same relative augmentation allocation is applied across folds.
Let $\widehat h_{\boldsymbol\lambda}^{(-k)}$ denote the classifier trained on the
augmented training portion of fold $k$.
The cellwise generators $P^{(-k)}_{\mathrm{syn},c}$ are fitted once on the training portion of fold $k$ and shared by all allocations. For each $\boldsymbol{\lambda}\in\Lambda_\omega$, the synthetic sample of fold $k$ is generated only once, when $\boldsymbol{\lambda}$ is first evaluated, and $\widehat{h}^{(-k)}_{\boldsymbol{\lambda}}$ is fitted once. The validation criteria defined below are stored and reused whenever $\boldsymbol{\lambda}$ is evaluated again. Thus, conditional on the data and the generation randomness, $\widehat{\Delta}^{\mathrm{CV}}_{\mathrm{EO}}$ and $\widehat{\mathrm{Err}}^{\mathrm{CV}}$ are fixed functions on $\Lambda_\omega$.
For $s\in\{A,B\}$, define the
validation-fold true positive rate
\[
\widehat{\mathrm{TPR}}_{s}^{(k)}(\boldsymbol\lambda)
:=
\frac{1}{n_{s+}^{(k)}}
\sum_{i\in\mathcal D_{s+,\mathrm{val}}^{(k)}}
\mathbf 1\!\left\{
\widehat h_{\boldsymbol\lambda}^{(-k)}(X_i)=1
\right\},
\]
and the corresponding Equal Opportunity gap
\[
\widehat\Delta_{\mathrm{EO}}^{(k)}(\boldsymbol\lambda)
:=
\left|
\widehat{\mathrm{TPR}}_{A}^{(k)}(\boldsymbol\lambda)
-
\widehat{\mathrm{TPR}}_{B}^{(k)}(\boldsymbol\lambda)
\right|.
\]
Also define the validation-fold prediction error
\[
\widehat{\mathrm{Err}}^{(k)}(\boldsymbol\lambda)
:=
\frac{1}{n_{\mathrm{val}}^{(k)}}
\sum_{i\in\mathcal D_{\mathrm{val}}^{(k)}}
\mathbf 1\!\left\{
\widehat h_{\boldsymbol\lambda}^{(-k)}(X_i)\neq Y_i
\right\}.
\]
The cross-validated criteria are
\[
\widehat\Delta_{\mathrm{EO}}^{\mathrm{CV}}(\boldsymbol\lambda)
:=
\frac{1}{K}\sum_{k=1}^K
\widehat\Delta_{\mathrm{EO}}^{(k)}(\boldsymbol\lambda),
\qquad
\widehat{\mathrm{Err}}^{\mathrm{CV}}(\boldsymbol\lambda)
:=
\frac{1}{K}\sum_{k=1}^K
\widehat{\mathrm{Err}}^{(k)}(\boldsymbol\lambda).
\]
We search over a bounded allocation region
$\Lambda\subset\mathbb R_+^4$ containing $\mathbf 0$, using a deterministic
finite master grid $\Lambda_\omega\subset\Lambda$. 
At iteration $t$, we consider the admissible coordinate-move candidates
\[
\mathcal M_t
:=
\left\{
\boldsymbol\lambda^{(t-1)}+\gamma_{c,j}^{(t)}e_c:
c\in\{A+,A-,B+,B-\},\ j=1,\ldots,J
\right\}
\cap \Lambda_\omega,
\]
where $e_c$ is the coordinate direction for cell $c$ and
\(
\gamma_{c,1}^{(t)}<\cdots<\gamma_{c,J}^{(t)}.
\) Here the increments may be positive or negative, provided that the
resulting candidate belongs to $\Lambda_\omega$. Negative increments may therefore reduce a previously selected augmentation amount, but only down to zero. The procedure never removes observations from the original sample.
For each candidate, define the estimated fairness gain
\begin{equation*}
\widehat G_{c,j}^{(t)}
=
\widehat\Delta_{\mathrm{EO}}^{\mathrm{CV}}
\left(\boldsymbol{\lambda}^{(t-1)}\right)
-
\widehat\Delta_{\mathrm{EO}}^{\mathrm{CV}}
\left(\boldsymbol{\lambda}_{c,j}^{(t)}\right).
\end{equation*}
A candidate is feasible if
\begin{equation*}
\widehat{\mathrm{Err}}^{\mathrm{CV}}
\left(\boldsymbol{\lambda}_{c,j}^{(t)}\right)
-
\widehat{\mathrm{Err}}^{\mathrm{CV}}(\mathbf0)
\leq
\xi_{\mathrm{acc}}.
\end{equation*}
The error constraint is always measured relative to the unaugmented model,
preventing predictive losses from accumulating across iterations. Among all
feasible candidates, we select
\begin{equation*}
(\widehat c_t,\widehat j_t)
\in
\arg\max_{(c,j)}
\widehat G_{c,j}^{(t)}
\quad
\text{subject to \eqref{eq:cv_accuracy_constraint}}.
\end{equation*}
Equivalently, the selected candidate has the smallest cross-validated EO gap among feasible candidates, as in Section~\ref{sec:adaptive_method}.
Let $\xi_{\mathrm{fair}}\geq 0$ denote the minimum required
cross-validated fairness improvement for accepting a move. The selected move is accepted only if
\(
\widehat G_{\widehat c_t,\widehat j_t}^{(t)}
>
\xi_{\mathrm{fair}},
\)
in which case
\begin{equation*}
\boldsymbol{\lambda}^{(t)}
=
\boldsymbol{\lambda}^{(t-1)}
+
\gamma_{\widehat c_t,\widehat j_t}^{(t)}
e_{\widehat c_t}.
\end{equation*}
If no coordinate move satisfies both the accuracy constraint and the minimum
fairness improvement, we apply a local grid safeguard. For a prespecified
radius $\rho_{\mathrm{saf}}>0$, we evaluate
$\boldsymbol{\lambda}'\in\Lambda_\omega$ satisfying
\(
\left\|
\boldsymbol{\lambda}'
-
\boldsymbol{\lambda}^{(t-1)}
\right\|_2
\leq
\rho_{\mathrm{saf}}.
\)
Among points satisfying
\[
\widehat{\mathrm{Err}}^{\mathrm{CV}}(\boldsymbol{\lambda}')
-
\widehat{\mathrm{Err}}^{\mathrm{CV}}(\mathbf0)
\leq
\xi_{\mathrm{acc}},
\]
we move to the allocation with the largest fairness improvement if that
improvement exceeds $\xi_{\mathrm{fair}}$. Otherwise, the search terminates.
Every accepted move, coordinate or safeguard, decreases $\widehat{\Delta}^{\mathrm{CV}}_{\mathrm{EO}}$ by more than $\xi_{\mathrm{fair}}\ge 0$. Since $\widehat{\Delta}^{\mathrm{CV}}_{\mathrm{EO}}$ is a fixed function on the finite grid $\Lambda_\omega$, the accepted allocations are distinct, so the search terminates after at most $N_\omega-1$ accepted moves, where $N_\omega=|\Lambda_\omega|$.
Let
\(
\widehat{\boldsymbol{\lambda}}
=
\boldsymbol{\lambda}^{(T)}
\)
be the final allocation. The same cell may be selected multiple times, and
the resulting cell sizes need not be equal. We then refit the synthetic
generators on the full training sample, generate
\(
m_c(\widehat{\boldsymbol{\lambda}})
=
\left\lfloor
\widehat\lambda_c n_{\max}
\right\rfloor
\)
observations from each cell, and fit the final classifier
$\widehat h_{\overline P_{\widehat{\boldsymbol{\lambda}}}}$.

\begin{assumption}[Adaptive-search regularity]
\label{assumption:adaptive_search}
Suppose the following conditions hold.
 
\begin{enumerate}
 
\item \textbf{Candidate grid.}
Assume that
\(
0\in\Lambda_\omega\subset\Lambda,
\)
and that $\Lambda_\omega$ is an $\omega$-net of the full allocation region
$\Lambda$ in Euclidean norm such that
\(
\sup_{\boldsymbol{\lambda}\in\Lambda}
\inf_{\boldsymbol{\lambda}'\in\Lambda_\omega}
\left\|
\boldsymbol{\lambda}
-
\boldsymbol{\lambda}'
\right\|_2
\leq
\omega.
\)
 
\item \textbf{Uniform learning and boundary regularity.}
For every augmented distribution
\[
Q\in
\left\{
\overline P_{\boldsymbol\lambda},\,
\overline P_{\boldsymbol\lambda}^{(-k)}:
\boldsymbol\lambda\in\Lambda,\ k=1,\ldots,K
\right\},
\]
assume that $L_Q$ and $\widehat L_Q$ are differentiable with interior
minimizers, and that differentiation and expectation can be interchanged
at the relevant population minimizers. Assume that the empirical curvature condition in
Assumption~\ref{assumption:prevalence-diff} holds uniformly over all such $Q$ for each
fold-specific augmented empirical loss with curvature constant
at least $\mu>0$. Moreover, for every pair
\[
Q,Q'
\in
\left\{
\overline P_{\boldsymbol\lambda},\,
\overline P_{\boldsymbol\lambda}^{(-k)}
:
\boldsymbol\lambda\in\Lambda,\;
k=1,\ldots,K
\right\},
\]
assume that the population loss $L_Q$ satisfies
\[
\left\langle
\nabla L_Q(\theta_{Q'}^*)
-
\nabla L_Q(\theta_Q^*),
\theta_{Q'}^*-\theta_Q^*
\right\rangle
\ge
\mu
\left\|
\theta_{Q'}^*-\theta_Q^*
\right\|_2^2.
\]
Suppose Condition~4 in Assumption \ref{assumption:improvement-bound-assumption} holds.
The score satisfies
\(
\sup_x
|f_\theta(x)-f_{\theta'}(x)|
\leq
L\|\theta-\theta'\|.
\)
Moreover, the margin condition of Condition~3(b) in Assumption~\ref{assumption:prevalence-diff} holds for every candidate population score $f^*_{\overline P_{\boldsymbol\lambda}}:=f_{\theta^*_{\overline P_{\boldsymbol\lambda}}}$, uniformly over $\boldsymbol\lambda\in\Lambda$, both for the positive cells and under $P$: for some constants $C^\Lambda_A,C^\Lambda_B,C^\Lambda_P>0$, all $\boldsymbol\lambda\in\Lambda$, $s\in\{A,B\}$ and $\varepsilon>0$,
\[
P_{s+}\bigl(|f^*_{\overline P_{\boldsymbol\lambda}}(X)-\tfrac12|\le\varepsilon\bigr)\le C^\Lambda_s\,\varepsilon,
\qquad
P\bigl(|f^*_{\overline P_{\boldsymbol\lambda}}(X)-\tfrac12|\le\varepsilon\bigr)\le C^\Lambda_P\,\varepsilon .
\]

\item \textbf{Cellwise generator approximation.}
For each $c\in\{A+,A-,B+,B-\}$, the full-sample generator satisfies
\[
d_{\mathrm{TV}}
\left(
P_{\mathrm{syn},c},P_c
\right)
\leq
\epsilon_c,\qquad d_{\mathrm{TV}}
\left(
P_{\mathrm{syn},c}^{(-k)},P_c
\right)
\leq
\epsilon_c^{(-k)}.
\]
Conditional on the real training portion of fold $k$, the generated observations are
mutually independent, with cell-specific distributions
$P_{\mathrm{syn},c}^{(-k)}$.

\item \textbf{Local identification and smooth constrained geometry.}
There exist constants
$\epsilon_0>0$ and $\rho_{\mathrm{acc}}>0$ such that the mappings
\[
\boldsymbol{\lambda}
\mapsto
\Delta_{\mathrm{EO}}
\left(
h^*_{\overline P_{\boldsymbol{\lambda}}}
\right),\qquad 
\boldsymbol{\lambda}
\mapsto
\mathrm{Err}_{P}
\left(
h^*_{\overline P_{\boldsymbol{\lambda}}}
\right)
-
\mathrm{Err}_{P}
\left(
h^*_{P}
\right)
\]
are twice continuously differentiable on a convex neighborhood containing every point
within Euclidean distance $\rho_{\mathrm{acc}}$ of the accuracy sublevel sets with
tolerances in
$[\xi_{\mathrm{acc}}-\epsilon_0,\xi_{\mathrm{acc}}+\epsilon_0]$.
Suppose $\boldsymbol{\lambda}^*$ is unique and that, throughout this neighborhood, there
exist constants
$0<\mu_{\mathrm{aug}}\leq H_{\mathrm{aug}}<\infty$ such that
\begin{equation}
\label{eq:augmentation_hessian}
\mu_{\mathrm{aug}}I_4
\preceq
\nabla_{\boldsymbol{\lambda}}^2
\Delta_{\mathrm{EO}}
\left(
h^*_{\overline P_{\boldsymbol{\lambda}}}
\right)
\preceq
H_{\mathrm{aug}}I_4.
\end{equation}
Suppose also that, for some $G_{\mathrm{aug}}<\infty$,
\begin{equation}
\label{eq:augmentation_gradient_bound}
\left\|
\nabla_{\boldsymbol{\lambda}}
\Delta_{\mathrm{EO}}
\left(
h^*_{\overline P_{\boldsymbol{\lambda}}}
\right)
\right\|_2
\leq
G_{\mathrm{aug}}
\end{equation}
throughout the same neighborhood.
For the accuracy criterion, suppose that throughout this neighborhood for some $G_{\mathrm{acc}}<\infty$ and $H_{\mathrm{acc}}>0$,
\[ 
\left\|
\nabla_{\boldsymbol{\lambda}}
\mathrm{Err}_{P}
\left(
h^*_{\overline P_{\boldsymbol{\lambda}}}
\right)
\right\|_2
\leq
G_{\mathrm{acc}},\qquad
\left\|
\nabla_{\boldsymbol{\lambda}}^2
\mathrm{Err}_{P}
\left(
h^*_{\overline P_{\boldsymbol{\lambda}}}
\right)
\right\|_{\mathrm{op}}
\leq
H_{\mathrm{acc}}.
\]
Assume for every $\boldsymbol{\lambda}$ satisfying
\(
\left|
\mathrm{Err}_{P}
\left(
h^*_{\overline P_{\boldsymbol{\lambda}}}
\right)
-
\mathrm{Err}_{P}
\left(
h^*_{P}
\right)
-
\xi_{\mathrm{acc}}
\right|
\leq
\epsilon_0,
\)
there exists a unit direction $v_{\boldsymbol{\lambda}}$ such that
\begin{equation}
\label{eq:accuracy_boundary_regular}
\nabla_{\boldsymbol{\lambda}}
\mathrm{Err}_{P}
\left(
h^*_{\overline P_{\boldsymbol{\lambda}}}
\right)^\top
v_{\boldsymbol{\lambda}}
\leq
-m_{\mathrm{acc}}
\end{equation}
for some $m_{\mathrm{acc}}>0$, and
\(
\boldsymbol{\lambda}
+
t v_{\boldsymbol{\lambda}}
\in
\Lambda
\)
for every $0\leq t\leq\rho_{\mathrm{acc}}$. Finally, assume that the accuracy sublevel
sets corresponding to tolerances in
$[\xi_{\mathrm{acc}}-\epsilon_0,\xi_{\mathrm{acc}}+\epsilon_0]$
are closed and convex.
 
\end{enumerate}
\end{assumption}

\section{Additional Discussion}\label{sec:remarksonassumptions}

\subsection{Discussion of the Synthetic-Generator Approximation Assumption}
Assumptions \ref{assumption:improvement-bound-assumption}(3) and \ref{assumption:adaptive_search}(3) quantify synthetic-generator accuracy through total variation (TV) distance. We adopt this metric for analytical clarity, as it directly controls discrepancies in probabilities of measurable events. The essential requirement, however, is suitable control of the synthetic distribution's approximation error, rather than TV distance specifically. Similar fairness guarantees could potentially be developed using alternative discrepancies, such as KL-divergence or Wasserstein distance, under appropriate regularity conditions.

Total variation approximation guarantees have been established for several classes of generative methods. Classical nonparametric density estimators admit \(L^1\) convergence guarantees, which directly imply TV convergence \citep{goldenshluger2011bandwidth}. For GANs, Jensen–Shannon divergence bounds yield TV control \citep{puchkin2024rates}, while constructive approximation results for normalizing flows establish KL bounds that also imply TV control \citep{alvarez2025constructive}. Moreover, TV convergence guarantees have been established directly for continuous-state diffusion models \citep{lee2023convergence} and discrete-state diffusion models \citep{zhang2025convergence}. These results demonstrate that TV-based generator approximation is theoretically attainable across different generative frameworks under suitable conditions. On the other hand, the TV condition need not hold with small error for interpolation-based methods such as SMOTE. For a finite training sample, standard SMOTE generates observations along finitely many line segments, which can be singular with respect to a full-dimensional continuous target distribution. Consequently, our numerical experiments with SMOTE illustrate its empirical effectiveness without asserting that the TV approximation assumption holds for this particular generator.

\subsection{Discussion of the Local Smoothness and Curvature Assumptions}\label{sec:disc-curvature}
Assumption \ref{assumption:adaptive_search}(4) imposes smoothness and curvature conditions on the Equal Opportunity gap only in a neighborhood of the accuracy-feasible region, rather than globally over the entire augmentation space. Since the EO gap involves an absolute difference between group true positive rates, differentiability may fail where the signed difference crosses zero. However, when this difference varies smoothly and maintains a fixed sign in the relevant neighborhood, the EO gap can remain twice continuously differentiable. Thus, these assumptions describe the local behavior needed for our theoretical analysis without requiring the same regularity globally.

The additional local curvature condition ensures that the optimal augmentation allocation is well identified, but in one direction it can only come from the generator error. Write $\boldsymbol p:=(p_c)_{c\in\mathcal C}$. The cell composition of $\overline P_{\boldsymbol\lambda}$ is $(\boldsymbol p+r\boldsymbol\lambda)/(1+r\sum_c\lambda_c)$, and under exact generation $\overline P_{\boldsymbol\lambda}=\widetilde P_{\boldsymbol\lambda}$ is the mixture of the cell distributions $P_c$ with these weights. Since the composition is unchanged along each ray $\{\boldsymbol\lambda+t(\boldsymbol\lambda+\boldsymbol p/r):t>-1\}$, so is $\Delta_{\mathrm{EO}}(h^*_{\overline P_{\boldsymbol\lambda}})$. Differentiating this invariance gives $\nabla_{\boldsymbol{\lambda}}\Delta_{\rm EO}(h^*_{\overline P_\lambda})^\top(\boldsymbol\lambda+\boldsymbol p/r)=0$, and differentiating once more gives
\[
\nabla_{\boldsymbol\lambda}^2\Delta_{\mathrm{EO}}\bigl(h^*_{\overline P_{\boldsymbol\lambda}}\bigr)(\boldsymbol\lambda+\boldsymbol p/r)=-\nabla_{\boldsymbol\lambda}\Delta_{\mathrm{EO}}\bigl(h^*_{\overline P_{\boldsymbol\lambda}}\bigr),
\]
so \eqref{eq:augmentation_hessian} cannot hold under exact generation, and even a positive semidefinite Hessian would force the gradient to vanish. With approximate generators, the curvature along these rays comes only from the generator error. The same invariance makes $\boldsymbol\lambda^*$ generally non-unique under exact generation, which is why Proposition \ref{prop:excess-eo} does not require uniqueness.

Proposition~\ref{prop:excess-eo} bounding the excess EO gap therefore replaces curvature in $\boldsymbol\lambda$ by convexity in the cell composition. The composition map $\boldsymbol\lambda\mapsto(\boldsymbol p+r\boldsymbol\lambda)/(1+r\sum_c\lambda_c)$ is linear-fractional. It therefore maps the accuracy sublevel set on which \eqref{eq:weighted-convexity} is required, which is convex by Assumption~\ref{assumption:adaptive_search}(4), onto a convex set, and its derivative at $\boldsymbol\lambda$ maps $\boldsymbol x-\boldsymbol\lambda$ to $(1+r\sum_cx_c)/(1+r\sum_c\lambda_c)$ times the difference between the compositions at $\boldsymbol x$ and $\boldsymbol\lambda$. Hence, under exact generation, if $\Delta_{\mathrm{EO}}(h^*_{\sum_c\zeta_cP_c})$ is a convex and differentiable function of the composition $\boldsymbol\zeta$ on that image, the chain rule and the convexity inequality give, for all $\boldsymbol\lambda,\boldsymbol x$ in the sublevel set,
\[
\Delta_{\mathrm{EO}}\bigl(h^*_{\overline P_{\boldsymbol\lambda}}\bigr)-\Delta_{\mathrm{EO}}\bigl(h^*_{\overline P_{\boldsymbol x}}\bigr)
\le\frac{1+r\sum_c\lambda_c}{1+r\sum_cx_c}\,\nabla_{\boldsymbol\lambda}\Delta_{\mathrm{EO}}\bigl(h^*_{\overline P_{\boldsymbol\lambda}}\bigr)^\top(\boldsymbol\lambda-\boldsymbol x),
\]
so \eqref{eq:weighted-convexity} holds with $\beta=1+r\max_{\boldsymbol\lambda\in\Lambda}\sum_c\lambda_c$. Only the case $\boldsymbol x=\boldsymbol\lambda^*$ is needed, so convexity can be weakened to star-convexity about the composition $\boldsymbol\zeta^*$ of $\boldsymbol\lambda^*$: with $F(\boldsymbol\zeta):=\Delta_{\rm EO}(h^*_{\sum_c\zeta_cP_c})$, it suffices that $F(\boldsymbol\zeta^*)\ge F(\boldsymbol\zeta)+\nabla F(\boldsymbol\zeta)^\top(\boldsymbol\zeta^*-\boldsymbol\zeta)$ for every $\boldsymbol\zeta$ in that image. With approximate generators, $\Delta_{\mathrm{EO}}(h^*_{\overline P_{\boldsymbol\lambda}})$ differs from its oracle counterpart $\Delta_{\mathrm{EO}}(h^*_{\widetilde P_{\boldsymbol\lambda}})$ through the generator error, and \eqref{eq:weighted-convexity} is then an assumption on the synthetic criterion, as \eqref{eq:augmentation_hessian} is in Theorem~\ref{thm:adaptive_near_optimality}. Unlike \eqref{eq:augmentation_hessian},  under this condition it holds exactly in the exact-generation limit, with a constant $\beta$ that does not depend on the generator. Unless $\rho_{\rm saf}\ge\mathrm{diam}(\Lambda)$, so that the safeguard checks every grid point, some condition of this kind is needed: ACASA stops at an approximately stationary allocation, and if the population criterion had a spurious stationary point in the feasible region, the search could stop there for every sample size. If \eqref{eq:weighted-convexity} holds only up to an additive slack, the proof of Proposition~\ref{prop:excess-eo} adds the same slack to the bound.

\subsection{Discussion of the Logarithmic Factor in the Convergence Rate}
\label{sec:disc-independence-minimax}

Theorem \ref{thm:full-balance-convergence} establishes a finite-sample convergence rate of
order $\sqrt{\log n_P/n_P}$ for fixed dimension and confidence
level, apart from generator-related errors. Under the uniformity condition of Section \ref{sec:minimaxlowerbound}, this matches the
$n_P^{-1/2}$ minimax lower bound in Theorem \ref{thm:minimax-EO-lower} up to a
logarithmic factor. This factor is not intrinsic to synthetic augmentation. It comes from the concentration argument used to accommodate dependence between the original and synthetic observations.

Unlike much of the existing literature that requires independence between original
and synthetic training samples \citep{nakada2024synthetic, tian2025conditional, ma2026synthetic}, our analysis allows the
synthetic generators to be fitted using the same original
observations employed for classifier training. Consequently,
the population minimizer
$\theta^*_{\overline P_{\mathrm{fb}}}$ may depend on the
original sample, preventing a direct application of pointwise
Hoeffding concentration to the real-data empirical gradient
evaluated at this minimizer. Instead, the proof 
establishes uniform concentration over $\Theta$ through a
covering-number argument. This introduces
the logarithmic factor $\log((3n_P)^d)$ in the concentration
bound. 

If independence is imposed through methods like sample splitting, this
additional logarithmic factor can be avoided. Specifically,
one may divide the original observations into two independent
subsamples, using one to fit the synthetic generators and
the other, together with the generated observations, to
train the classifier. Conditional on the generator-fitting
subsample and the observed training-cell counts, the
population minimizer is fixed and independent of the
real observations used for classifier training. Moreover,
the generated observations are conditionally independent
under the fitted generators. Therefore, pointwise Hoeffding
concentration applies to both the real-data and synthetic-data
gradient deviations, without requiring a covering-number
argument. Under the corresponding regularity conditions,
when both subsample sizes are proportional to $n_P$, the
sampling error is of order
$O_{\mathbb P}(n_P^{-1/2})$ for fixed dimension, apart from
generator-related errors, recovering the sample-size
dependence of the minimax lower bound in Theorem \ref{thm:minimax-EO-lower}.

\subsection{Comparison with Existing Adaptive Augmentation Methods}
\label{sec:comparison-adaptive}

Our work is related to the Conditional Data Synthesis Augmentation (CoDSA) framework of \cite{tian2025conditional}, which adaptively determines synthetic sample volumes and allocations across underrepresented regions to improve statistical accuracy and generalization. The key distinction lies in the optimization objective. While CoDSA primarily focuses on predictive performance, ACASA explicitly minimizes the Equal Opportunity gap subject to a constraint on predictive-performance degradation. Accordingly, ACASA selects augmentation amounts across group--label cells using cross-validated fairness and accuracy estimates. The theoretical contributions also differ: CoDSA characterizes statistical accuracy gains from synthetic augmentation, whereas our analysis establishes conditions for fairness improvement through full balancing and provides a finite-sample guarantee comparing the ACASA allocation with the population-optimal accuracy-feasible allocation.

\subsection{Full Balancing Does Not Necessarily Imply Fairness}\label{sec:fb-not-fair}

Exact full balancing equalizes the probabilities of all group--label cells, but does not necessarily eliminate EO disparities. The EO gap of the oracle balanced-population classifier satisfies
\(
\Delta_{\mathrm{EO}}(h_{\mathrm{bal}}^*)
=
|P_{A+}(\mathcal X^*_{\rm bal})-P_{B+}(\mathcal X^*_{\rm bal})|\), where $\mathcal X^*_{\rm bal}=\{f_{\theta^*_{\rm bal}}\ge 1/2\}$. Although full balancing equalizes the cell probabilities, it preserves the underlying cell-conditional feature distributions. Consequently, differences between $P_{A+}$ and $P_{B+}$ may persist, and the resulting EO gap need not vanish.
Theorem \ref{thm:fullbalance-improvement} provides sufficient conditions under which full balancing reduces the EO gap relative to the original classifier, rather than guaranteeing exact EO. Similarly, Theorem \ref{thm:full-balance-convergence} establishes convergence to the EO gap of the oracle balanced-population classifier, which may remain positive. This distinction motivates ACASA, which directly minimizes the estimated EO disparity subject to a predictive-accuracy constraint, allowing adaptive allocations to improve the fairness--accuracy trade-off beyond exact full balancing.

\subsection{The Role of Minority-Cell Sample Size under Severe Imbalance}

The effectiveness of synthetic augmentation depends critically on the number of real observations available in each group--label cell. Let $p_{\min}=\min_{c\in\mathcal{C}}p_c$. The expected sample size of the smallest cell is $n_Pp_{\min}$, which governs the statistical difficulty of estimating its conditional feature distribution. Under moderate imbalance, sufficiently large cell sizes may allow accurate synthetic generation and effective fairness improvement. Under severe imbalance, however, estimation errors in minority-cell generators can become substantial and may propagate into the augmented classifier. In extremely sparse cells, reliable estimation of the underlying distributions and group-specific true positive rates may be infeasible.

This limitation is reflected in the cellwise synthetic-generation error terms in Theorems \ref{thm:fullbalance-improvement}--\ref{thm:full-balance-convergence}, which can offset the benefits of augmentation when the generators are inaccurate. Increasing the number of synthetic observations does not necessarily compensate for insufficient original data, since synthetic samples generated from the same training observations do not provide independent information about the underlying population. This observation is consistent with the numerical results, where augmentation methods exhibit deteriorating performance under extreme imbalance.

\subsection{Population Imbalance versus Sampling Imbalance}\label{sec:pop-vs-sampling}

Data imbalance may arise from two distinct sources: sampling imbalance and population imbalance. Sampling imbalance occurs when the observed training distribution differs from the target population due to unequal sampling or selection mechanisms. In this case, adjusting group--label proportions may help correct the sampling distortion, provided the target proportions are appropriately specified. In contrast, population imbalance reflects genuine differences in group sizes or label prevalences in the target population. In this setting, balancing deliberately modifies the training distribution to improve fairness rather than recover the original population distribution.
When correcting sampling imbalance, the augmentation allocation should account for the intended target population. When addressing population imbalance, the allocation instead reflects a fairness--accuracy trade-off, and predictive performance under the original distribution $P$ may differ from that under the oracle balanced distribution $\widetilde{P}_{\mathrm{fb}}$. As discussed in Remark~\ref{remark:choiceofvalidation}, the validation distribution can be chosen according to the application-specific objective.

\section{Data Generation and Implementation Details}\label{sec:numerical-detail}
\subsection{Simulation Design and Implementation}\label{sec:simulation-detail}
We consider a binary classification problem with groups $S\in\{A,B\}$ and labels $Y\in\{0,1\}$, giving the four group--label cells $A+$, $A-$, $B+$, and $B-$. The feature vector has $15$ coordinates. Within each cell $c$, the first $5$ coordinates  follow the mixture distribution
\[
X_{1:5}\mid c\sim (1-p_c)N(\mu_{c[1:5]},2.25I_5)+p_c\,\mathcal{L}(\mu_{c[1:5]}+15W),
\]
where $\mathcal{L}(\mu_{c[1:5]}+15W)$ denotes the distribution of $\mu_{c[1:5]}+15W$, with independent $t_3$ entries in $W$. The cell-specific mixture weights are $(p_{A+},p_{A-},p_{B+},p_{B-})=(0.09,0.04,0.12,0.06)$, and the means are
\[
\begin{aligned}
\mu_{A+}
&=(1.6,\,1.2,\,-0.8,\,0.4,\,1.6,\,
   0.9,\,0.9,\,0.9,\,
   0,\,0,\,0,\,0,\,0,\,0,\,0)^\top,\\
\mu_{A-}
&=(-1.2,\,2.4,\,-1.6,\,-1.2,\,-1.2,\,
   1.8,\,1.8,\,1.8,\,
   0,\,0,\,0,\,0,\,0,\,0,\,0)^\top,\\
\mu_{B+}
&=(0.4,\,-2.0,\,0.4,\,-0.8,\,0.4,\,
   2.4,\,2.4,\,2.4,\,
   0,\,0,\,0,\,0,\,0,\,0,\,0)^\top,\\
\mu_{B-}
&=(-0.8,\,-1.2,\,1.2,\,1.6,\,-0.8,\,
   2.7,\,2.7,\,2.7,\,
   0,\,0,\,0,\,0,\,0,\,0,\,0)^\top.
\end{aligned}
\]
The remaining $10$ coordinates have variance $9$ and are independent of $X_{[1:5]}$. Training and test observations are independently generated from the same cell-conditional distributions.

The independent test sample contains $5{,}000$ observations from each cell and is kept fixed in composition across the simulation experiments. For the main comparison of ACASA with raw training and full balancing, the training cell sizes are $(n_{A+},n_{A-},n_{B+},n_{B-})=(2{,}000,3{,}000,1{,}000,3{,}000)$. To study prevalence imbalance separately, both groups contain $3{,}000$ training observations, with $(n_{A+},n_{A-})=(1{,}500,1{,}500)$ and $n_{B+}$ decreasing from $1{,}500$ to $30$, while $n_{B-}=3{,}000-n_{B+}$. We use the sweep grid \[
\begin{aligned}
\mathcal{G}=\{&
30,40,50,60,75,90,100,125,150,175,200,225,250,275,300,\\&350,400,450,500,550,600,650,700,750,800,850,900,950,1000,\\&1050,1100,1150,1200,1250,1300,1350,1400,1450,1500\}
\end{aligned}
\]
for $n_{B+}$. We use a nonuniform grid with finer spacing near extreme imbalance to obtain roughly comparable changes in the imbalance ratio. To study group-size imbalance, we instead set $\rho_A=\rho_B=1$ and keep the total training sample size equal to $6{,}000$, while decreasing the size of group $B$ from $3{,}000$ to $60$ and increasing the size of group $A$ by the same amount. Here we use the grid
\[
\begin{aligned}
\mathcal{G}_{B}=\{&
60,80,100,120,150,180,200,250,300,350,400,450,500,550,600,\\
&700,800,900,1000,1100,1200,1300,1400,1500,1600,1700,1800,1900,2000,\\
&2100,2200,2300,2400,2500,2600,2700,2800,2900,3000\}.
\end{aligned}
\]
We also use a nonuniform grid to attain roughly comparable changes in the imbalance ratio.

Raw training, full balancing, and ACASA use the same logistic-regression specification. Full balancing raises each group--label cell to the size of the largest cell, using reweighting, SMOTE, or ADASYN. Synthetic observations are generated within the corresponding cell. ACASA evaluates cell-specific augmentation allocations by $3$-fold cross-validation, stratified for four cells. Here we use a balanced validation criterion. Although each validation fold retains the imbalanced cell proportions through stratified splitting, we compute the prediction error separately within each cell and then average the four cell-specific error rates with equal weights of $1/4$. We use allocation-grid spacing $0.2$, coordinate increments of $\pm 0.2$, a safeguard radius of $0.6$, $\xi_{\mathrm{acc}}=0.01$, and $\xi_{\mathrm{fair}}=0.003$, with the same settings for reweighting, SMOTE, and ADASYN. SMOTE and ADASYN both use $10$ nearest neighbors. 
 Results without post-processing use threshold $0.5$. For the fairness--performance analysis, each training cell is divided equally into fitting and calibration samples; augmentation selection and model fitting use only the fitting sample, whereas the calibration sample is used to select $(\tau_A,\tau_B)$ by minimizing prediction error subject to $|\mathrm{TPR}_A-\mathrm{TPR}_B|\leq\epsilon$. The selected thresholds are then evaluated on the independent test sample. Each sweep point and the main ACASA experiments use $200$ independent repetitions, and pointwise $95\%$ confidence intervals are obtained from $5{,}000$ bootstrap resamples of the $200$ replications. For Figure~\ref{fig:acasa-convergence}, different runs may terminate after different numbers of accepted ACASA iterations. At each iteration, runs that have terminated earlier retain the EO gap corresponding to their last accepted allocation. The reported mean and pointwise bootstrap confidence intervals are computed across all runs using these carried-forward values.

\subsection{COMPAS Data Implementation Details}
\label{sec:realdata-detail}

We use the COMPAS two-year recidivism dataset \citep{larson2016we, he2025fairness}. The response is $Y=1$ for recidivism within two years and $Y=0$ otherwise, with groups $A$ (male) and $B$ (female). We follow ProPublica's preprocessing protocol \citep{larson2016we}.
Specifically, we retain records with
$\texttt{days\_b\_screening\_arrest}\in[-30,30]$ to reduce
potential mismatches between the COMPAS assessment and the
associated arrest. We exclude records with $\texttt{is\_recid}=-1$,
which indicates that no corresponding COMPAS case could be found,
and records with $\texttt{c\_charge\_degree}=\texttt{O}$, which
represent ordinary traffic offenses that do not result in
incarceration. We also exclude records with
$\texttt{score\_text}=\texttt{N/A}$ to restrict the analysis to
individuals with an available COMPAS assessment, although the
assessment score itself is not used as a predictor. The six predictors are age, the number of prior offenses, juvenile felony count, juvenile misdemeanor count, other juvenile-offense count, and an indicator of a felony charge. Sex, race, and the original COMPAS risk score are excluded from the predictors.
Finally, we remove records with missing values in the six
predictors, the outcome, sex, or race. 

For each of $100$ random seeds, we independently split the observations within each group--label cell into fitting, calibration, and test sets. We allocate $30\%$ to testing and $30\%$ of the remaining observations to calibration, leaving approximately $49\%$ of the full sample for fitting. All methods use logistic regression with intercept $\theta_0$ and coefficient vector $\theta_1$, fitted by minimizing $\sum_i w_i\,\ell(\theta;X_i,Y_i)+(0.05/n)\|\theta_1\|_2^2$ over $\theta=(\theta_0,\theta_1)$, where $\ell$ is the cross-entropy loss, the sum runs over the real and synthetic training observations, and $w_i=1$ except under reweighting (Appendix~\ref{app:reweighting}). We compare raw training, full balancing, and ACASA using reweighting, SMOTE, and diffusion model. Full balancing increases the effective size of each group--label cell to that of the largest fitting cell. For SMOTE, we use $10$ nearest neighbors. The first five interpolated predictors are mapped to their nearest values observed in the fitting sample. For the binary felony-charge indicator, the interpolated value is rounded to the nearest binary value: values below $0.5$ are assigned $0$, and values above $0.5$ are assigned $1$. If the interpolated value is exactly $0.5$, we use the binary value of the neighbor randomly selected from the base observation's nearest neighbors for that interpolation.

For diffusion model, we fit a single Gaussian diffusion model to the six predictors, conditional on the group--label cell. The first five predictors are transformed using a quantile transformation estimated from the fitting observations. The model predicts Gaussian noise using two hidden layers of width $256$, with $1{,}000$ diffusion steps and a cosine noise schedule. We train the model for $3{,}000$ optimization steps using a batch size of $256$, an initial learning rate of $10^{-3}$ with linear decay, and weight decay $10^{-5}$. After generation, the first five predictors are transformed back to their original scale and assigned the nearest values observed in the fitting sample. Diffusion model is trained separately on the training observations of each cross-validation fold and on the full fitting sample for final model fitting. Neither calibration nor test observations are used to train the generator.

ACASA selects cell-specific augmentation amounts through $3$-fold cross-validation, minimizing the estimated EO gap subject to a predictive-performance constraint. We consider an accuracy-constrained and an MCC-constrained variant, with tolerances and safeguard radii set separately to account for the different scales of the two criteria. The accuracy-constrained procedure permits a decrease in cross-validated accuracy relative to raw training of at most $0.025$ for reweighting, SMOTE and diffusion model, while the MCC-constrained procedure permits a decrease in cross-validated MCC of at most $0.015$ for all methods. In both variants, predictive performance and the EO gap are evaluated at the common threshold $0.5$ within each fold. The coordinate increments are chosen from $\{0.1,0.2,0.3,0.4,0.5\}$, with a maximum allocation parameter of $1$, safeguard-grid spacing of $0.2$. The safeguard radius is $0.2$ under the accuracy constraint and $0.45$ under the MCC constraint. The minimum required EO-gap improvement for accepting an update is $0.003$ under the accuracy constraint, and $0.0005$ under the MCC constraint. 

We evaluate fairness before post-processing using the test-set EO gap $|\mathrm{TPR}_A-\mathrm{TPR}_B|$ at the common threshold $0.5$. For runs that terminate early, the final accepted allocation is retained for subsequent iteration counts. To evaluate predictive performance after post-processing, we select group-specific thresholds $(\tau_A,\tau_B)$ on the calibration set to maximize pooled accuracy or MCC, subject to the EO-gap constraint $|\mathrm{TPR}_A-\mathrm{TPR}_B|\leq\epsilon$. We consider $\epsilon\in\{0.05,0.075,0.10,0.125,0.15,0.25,0.50,1.00\}$. The thresholds are selected from a grid of $150$ equally spaced values on $[0,1]$, supplemented by $0.5$, and then evaluated on the test set. The case $\epsilon=1$ corresponds to unconstrained threshold optimization rather than the absence of post-processing. Reported results are averaged over $100$ runs, each using a different random seed to independently split the data into fitting, calibration, and test sets. Pointwise $95\%$ bootstrap confidence intervals are computed from $5{,}000$ resamples of the $100$ runs.

\subsection{Implementation of Reweighting}\label{app:reweighting}
The reweighting versions of full balancing and ACASA use observation weights in place of synthetic observations and are otherwise unchanged. Within each training portion (that of each cross-validation fold, or the full fitting sample for the final classifier), let $n_c$ be the number of real observations in cell $c$, $n=\sum_{c\in\mathcal{C}}n_c$ and $n_{\max}=\max_{c\in\mathcal{C}}n_c$, and let $m_c(\lambda)=\lfloor\lambda_c n_{\max}\rfloor$ be the number of synthetic observations that allocation $\lambda$ would add to cell $c$, with $m(\lambda)=\sum_{c\in\mathcal{C}}m_c(\lambda)$. Each real observation in cell $c$ receives weight $w_c(\lambda)=\{1+m_c(\lambda)/n_c\}\,n/\{n+m(\lambda)\}$, and the logistic regression is fitted with these weights. Cell $c$ thus carries the fraction $\{n_c+m_c(\lambda)\}/\{n+m(\lambda)\}$ of the total weight, its share of the augmented sample, so reweighting reproduces the cell composition of augmentation without generating new feature configurations (Remark~\ref{remark:augvsreweight}). The common factor $n/\{n+m(\lambda)\}$ leaves these fractions unchanged and keeps the total weight equal to $n$, so the regularization of the logistic regression has the same strength relative to the loss for every $\lambda$. For full balancing, $\lambda=\lambda^{\mathrm{fb}}$ and every cell receives total weight $n/4$.

\section{Proofs}\label{sec:proofs}

\subsection{Proof of Theorems and Propositions} 

\begin{proof}[Proof of Theorem \ref{Thm:fairness-improvement} and Theorem \ref{Thm:fairness-improvement-full}]
By definition, $h_Q^\eta=\mathbf 1\{\eta_Q\ge1/2\}=\mathbf 1\{\Psi_Q\ge0\}$, so $\mathrm{TPR}_s(h_Q^\eta)=P_{s+}(\mathcal X_Q^+)$ for $Q\in\{P,\widetilde P\}$. Write $d_Q:=\mathrm{TPR}_A(h_Q^\eta)-\mathrm{TPR}_B(h_Q^\eta)$, so that $\Delta_{\mathrm{EO}}(h_Q^\eta)=|d_Q|$. Let $h_\kappa$ and $\mathcal N$ be as in Assumption~\ref{assumption:prevalence-diff}. We have the following lemma.

\begin{lemma}[Pointwise log-odds shift]\label{lemma:log-odds-shift}
Assume Condition~1 of Assumption~\ref{assumption:prevalence-diff}. For $\lambda\in[0,1)$ let $P_\lambda:=(1-\lambda)P+\lambda P_{B+}$ and $p_{B+}:=P(S=B,Y=1)=(1-\alpha)\pi_B$. Then, for $\nu$-almost every $x$,
\[
\Psi_{P_\lambda}(x)-\Psi_P(x)=\log\Bigl(1+\frac{\lambda}{(1-\lambda)\,p_{B+}}\,\omega_B(x)\Bigr).
\]
For $\lambda=\lambda_1^*$ the constant $\lambda/\bigl((1-\lambda)p_{B+}\bigr)$ equals $\mathrm{OR}-1$. Hence
\[
\Psi_{\widetilde P}(x)-\Psi_P(x)=\log\bigl(1+(\mathrm{OR}-1)\,\omega_B(x)\bigr)\in[0,\log\mathrm{OR}],
\qquad
\mathcal X_P^+\subseteq\mathcal X_{\widetilde P}^+ .
\]
\end{lemma}
We then proceed with the proof in several steps.
\paragraph{Step 1: the flip region.}
By Lemma~\ref{lemma:log-odds-shift}, $\mathcal X_P^+\subseteq\mathcal X_{\widetilde P}^+$. Let $\mathcal D:=\mathcal X_{\widetilde P}^+\setminus\mathcal X_P^+$ be the set of observations that are classified as positive under $\widetilde P$ but not under $P$. Then for $s\in\{A,B\}$
\begin{equation}\label{eq:flip-tpr}
\mathrm{TPR}_s(h_{\widetilde P}^\eta)-\mathrm{TPR}_s(h_P^\eta)=P_{s+}(\mathcal D),
\end{equation}
and, by \eqref{eq:logodds-shift-main},
\[
\mathcal D=\bigl\{x:\ -\log\bigl(1+(\mathrm{OR}-1)\omega_B(x)\bigr)\le\Psi_P(x)<0\bigr\}\subseteq\mathcal N,
\]
because the shift is at most $\log\mathrm{OR}$.
 
\paragraph{Step 2: the Equal Opportunity gap difference.}
By \eqref{eq:flip-tpr},
\begin{equation}\label{eq:eo-decomp}
d_P=d_{\widetilde P}+P_{B+}(\mathcal D)-P_{A+}(\mathcal D).
\end{equation}
Since $P(X\in\cdot,S=s\mid Y=1)=\bar\omega_s\,P_{s+}$ and $\omega_B(x)=P(S=B\mid X=x,Y=1)$,
\[
P_{B+}(\mathcal D)-P_{A+}(\mathcal D)=\int_{\mathcal D}\Bigl(\frac{\omega_B(x)}{\bar\omega_B}-\frac{1-\omega_B(x)}{\bar\omega_A}\Bigr)P(dx\mid Y=1).
\]
The integrand is increasing in $\omega_B(x)$. On $\mathcal D\subseteq\mathcal N$, Condition~5 gives $\omega_B(x)\ge\bar\omega_B+\kappa$, and, using $\bar\omega_A+\bar\omega_B=1$,
\[
\frac{\bar\omega_B+\kappa}{\bar\omega_B}-\frac{\bar\omega_A-\kappa}{\bar\omega_A}=\frac{\kappa}{\bar\omega_B}+\frac{\kappa}{\bar\omega_A}=\frac{\kappa}{\bar\omega_A\bar\omega_B}.
\]
Therefore
\begin{equation}\label{eq:dominance}
P_{B+}(\mathcal D)-P_{A+}(\mathcal D)\;\ge\;\frac{\kappa}{\bar\omega_A\bar\omega_B}\,P(X\in\mathcal D\mid Y=1)\;\ge\;0 .
\end{equation}
By Condition~4, $d_{\widetilde P}\ge0$, and then $d_P\ge d_{\widetilde P}\ge0$ by \eqref{eq:eo-decomp} and \eqref{eq:dominance}. Hence $\Delta_{\mathrm{EO}}(h_P^\eta)=d_P$, $\Delta_{\mathrm{EO}}(h_{\widetilde P}^\eta)=d_{\widetilde P}$, and
\[
\Delta_{\mathrm{EO}}(h_P^\eta)-\Delta_{\mathrm{EO}}(h_{\widetilde P}^\eta)=P_{B+}(\mathcal D)-P_{A+}(\mathcal D).
\]
The first inequality of the theorem holds because $\Delta_{\mathrm{EO}}(h_{\widetilde P}^\eta)\ge0$.
 
\paragraph{Step 3: the flip region is not negligible.}
We show that $\{x:-h_\kappa\le\Psi_P(x)<0\}\subseteq\mathcal D$. Since $\bar\omega_B+\kappa\le\bar\omega_B+\bar\omega_A=1$, we have $h_\kappa\le\log\mathrm{OR}$, so any such $x$ lies in $\mathcal N$ and Condition~5 gives $\omega_B(x)\ge\bar\omega_B+\kappa$. By \eqref{eq:logodds-shift-main},
\[
\Psi_{\widetilde P}(x)=\Psi_P(x)+\log\bigl(1+(\mathrm{OR}-1)\omega_B(x)\bigr)\ge-h_\kappa+h_\kappa=0,
\]
so $x\in\mathcal X_{\widetilde P}^+\setminus\mathcal X_P^+=\mathcal D$. By Condition~6, applied with $J=[-h_\kappa,0)$,
\[
P(X\in\mathcal D\mid Y=1)\ge P\bigl(-h_\kappa\le\Psi_P(X)<0\mid Y=1\bigr)\ge c_\Psi\,h_\kappa .
\]
Combining this with Step 2 and \eqref{eq:dominance},
\begin{equation}\label{eq:eo-lower}
\Delta_{\mathrm{EO}}(h_P^\eta)\ge\Delta_{\mathrm{EO}}(h_P^\eta)-\Delta_{\mathrm{EO}}(h_{\widetilde P}^\eta)\ge\frac{\kappa\,c_\Psi}{\bar\omega_A\bar\omega_B}\,h_\kappa,
\end{equation}
which is the statement of the theorem.
 
\paragraph{Step 4: the prevalence term.}
The identity $\bar\omega_B=(1-\alpha)\pi_B/\bar\pi$ and $\mathrm{OR}-1=(\pi_A-\pi_B)/\bigl(\pi_B(1-\pi_A)\bigr)$ give
\[
(\mathrm{OR}-1)\,\bar\omega_B=\frac{(1-\alpha)(\pi_A-\pi_B)}{\bar\pi(1-\pi_A)}\ge4(1-\alpha)(\pi_A-\pi_B),
\]
because $\bar\pi\le\pi_A$ and $\pi_A(1-\pi_A)\le1/4$. As $\kappa>0$ and $\mathrm{OR}>1$, $h_\kappa\ge\log\bigl(1+(\mathrm{OR}-1)\bar\omega_B\bigr)\ge\log\bigl(1+4(1-\alpha)(\pi_A-\pi_B)\bigr)$, which is the lower bound quoted after the theorem. This finishes the proof of Theorem \ref{Thm:fairness-improvement}.
 
\paragraph{Step 5: calibration and finite-sample error.}
We now turn to the empirical classifier $\widehat h_P$ and prove Theorem~\ref{Thm:fairness-improvement-full}. Only $Q=P$ is involved. Consider the convexity of the minimizer and plug in the population minimizer $\theta_Q^*$:
\begin{equation}\label{eq:7}
\left\langle \nabla\widehat{L}_Q(\theta_Q^*)-\nabla\widehat{L}_Q(\widehat{\theta}_Q), \theta_Q^*-\widehat{\theta}_Q\right\rangle\geq \mu_Q\|\theta_Q^*-\widehat{\theta}_Q\|^2.
\end{equation}
By first order optimality,
\begin{equation}\label{eq:8}
\left\langle \nabla\widehat{L}_Q(\theta_Q^*)-\nabla\widehat{L}_Q(\widehat{\theta}_Q), \theta_Q^*-\widehat{\theta}_Q\right\rangle=\left\langle \nabla\widehat{L}_Q(\theta_Q^*), \theta_Q^*-\widehat{\theta}_Q\right\rangle\leq \left\| \nabla\widehat{L}_Q(\theta_Q^*)\right\|\left\|\theta_Q^*-\widehat{\theta}_Q\right\|.
\end{equation}
Combining \eqref{eq:7} and \eqref{eq:8},
\[
\left\|\theta_Q^*-\widehat{\theta}_Q\right\|\leq \frac{1}{\mu_Q}\left\| \nabla\widehat{L}_Q(\theta_Q^*)\right\|.
\]
Next, $\nabla\widehat{L}_Q(\theta_Q^*)=\frac{1}{n_Q}\sum_{i=1}^{n_Q}\nabla\ell(\theta_Q^*;x_i,y_i)$ has mean $\nabla L_Q(\theta_Q^*)=0$. By a union bound over the $d$ coordinates and Hoeffding's inequality, using $\|v\|_2>t\Rightarrow\exists j:|v_j|>t/\sqrt d$,
\begin{align*}
P\left(\left\| \nabla\widehat{L}_Q(\theta_Q^*)\right\|>t\right)
&\leq \sum_{j=1}^d P\left(\left|\Bigl[\frac{1}{n_Q}\sum_{i=1}^{n_Q}\nabla\ell(\theta_Q^*;x_i,y_i)\Bigr]_j-\Bigl[\bE_Q\nabla\ell(\theta_Q^*;x_i,y_i)\Bigr]_j\right|>\frac{t}{\sqrt{d}}\right)\\
&\leq 2d\exp\left\{-\frac{2t^2n_Q}{dB^2}\right\}.
\end{align*}
Choosing $t=\sqrt{\frac{dB^2\log(2d/\delta)}{2n_Q}}$, with probability at least $1-\delta$,
\[
\left\|\theta_Q^*-\widehat{\theta}_Q\right\| \leq \frac{1}{\mu_Q}\sqrt{\frac{dB^2\log(2d/\delta)}{2n_Q}}.
\]
By the Lipschitz property of the score and Condition~3(a), with probability at least $1-\delta$,
\begin{equation}\label{eq:empirical-analysis}
\mathop{\sup}\limits_{x}\left|\widehat{f}_{P}(x)-\eta_P(x)\right|\leq \epsilon_{\mathrm{cal},P}+\frac{L}{\mu_P}\sqrt{\frac{dB^2\log(2d/\delta)}{2n_P}}=\varepsilon_n(\delta).
\end{equation}
Recall $h_P^\eta(x)=\mathbf 1\{\eta_P(x)\ge1/2\}$ and $\widehat h_P(x)=\mathbf 1\{\widehat f_P(x)\ge1/2\}$. On the event \eqref{eq:empirical-analysis}, if $\widehat h_P(x)\ne h_P^\eta(x)$, then $\eta_P(x)$ and $\widehat f_P(x)$ lie on opposite sides of $1/2$, so $|\eta_P(x)-\tfrac12|\le|\widehat f_P(x)-\eta_P(x)|\le\varepsilon_n(\delta)$. Therefore, by the margin condition 3(b), for $s\in\{A,B\}$,
\begin{equation}\label{eq:empirical-TPR-error}
\left|\mathrm{TPR}_s(\widehat{h}_P)-\mathrm{TPR}_s(h_P^\eta)\right|
\le P_{s+}\bigl(\widehat{h}_P(X)\neq h_P^\eta(X)\bigr)
\le P_{s+}\bigl(|\eta_P(X)-\tfrac12|\le\varepsilon_n(\delta)\bigr)
\le C_s\,\varepsilon_n(\delta).
\end{equation}
Since $\bigl||a|-|b|\bigr|\le|a-b|$,
\begin{align*}
\left|\Delta_{\mathrm{EO}}(\widehat{h}_P)-\Delta_{\mathrm{EO}}(h_P^\eta)\right|
&\le\left|\mathrm{TPR}_A(\widehat{h}_P)-\mathrm{TPR}_A(h_P^\eta)\right|+\left|\mathrm{TPR}_B(\widehat{h}_P)-\mathrm{TPR}_B(h_P^\eta)\right|\\
&\le(C_A+C_B)\,\varepsilon_n(\delta)=R_P(\delta).
\end{align*}
Combining this with \eqref{eq:eo-lower}, with probability at least $1-\delta$,
\[
\Delta_{\mathrm{EO}}(\widehat{h}_P)\ge\frac{\kappa\,c_\Psi}{\bar\omega_A\bar\omega_B}\,\log\!\Bigl(1+(\mathrm{OR}-1)\bigl(\bar\omega_B+\kappa\bigr)\Bigr)-R_P(\delta).
\]
This finishes the proof of Theorem \ref{Thm:fairness-improvement-full}.
\end{proof}

\begin{proof}[Proof of Theorem \ref{thm:imbalance-driven-bound} and Theorem \ref{thm:imbalance-driven-bound-full}]
Let $\mathcal D:=\mathcal X_{\widetilde P}^+\setminus\mathcal X_P^+$. The proof of Theorem~\ref{Thm:fairness-improvement} used the shift \eqref{eq:logodds-shift-main} only through three facts: (a)~$\mathcal X_P^+\subseteq\mathcal X_{\widetilde P}^+$; (b)~$\mathcal D\subseteq\mathcal N$; (c)~$\{x:-h_\kappa\le\Psi_P(x)<0\}\subseteq\mathcal D$. We verify them with $(\mathcal N,h_\kappa)$ replaced by $(\mathcal N_{\mathrm g},h_{\mathrm g})$. We have the following lemma.
\begin{lemma}[Group-share shift]\label{lemma:group-shift}
Assume Condition~1(a) of Assumption~\ref{assumption:prevalence-diff}. For $s\in\{A,B\}$ let $g_s(x):=\sum_y p(x,s,y)/P(S=s)$ and $\eta_s(x):=p(x,s,1)/\sum_yp(x,s,y)$. Then $q_w(x,y)=(1-w)\,p(x,A,y)/\alpha+w\,p(x,B,y)/(1-\alpha)$, the map $w\mapsto\Psi_{Q_w}(x)$ is smooth on $(0,1)$ with derivative \eqref{eq:group-derivative}, and, for $\nu$-almost every $x$,
\[
\bigl|\Psi_{\widetilde P}(x)-\Psi_P(x)\bigr|\le L_{\mathrm g}.
\]
\end{lemma}
By Lemma~\ref{lemma:group-shift},
\[
\Psi_{\widetilde P}(x)-\Psi_P(x)=\int_{1-\alpha}^{1/2}\partial_w\Psi_{Q_w}(x)\,dw\in[-L_{\mathrm g},L_{\mathrm g}].
\]
(a)~If $\Psi_P(x)\ge L_{\mathrm g}$, then $\Psi_{\widetilde P}(x)\ge0$ by the bound; if $0\le\Psi_P(x)<L_{\mathrm g}$, the integrand is nonnegative by Assumption~\ref{assumption:group-imb-assmp}, so again $\Psi_{\widetilde P}(x)\ge\Psi_P(x)\ge0$. Hence $\mathcal{X}^+_P\subseteq\mathcal{X}^+_{\widetilde P}$.
(b)~If $x\in\mathcal D$, then $\Psi_P(x)<0\le\Psi_{\widetilde P}(x)\le\Psi_P(x)+L_{\mathrm g}$, so $x\in\mathcal N_{\mathrm g}$.
(c)~On $\mathcal N_{\mathrm g}$ the integrand is at least $c'$, so $\Psi_{\widetilde P}(x)\ge\Psi_P(x)+h_{\mathrm g}$. We first show that $h_{\mathrm g}\le L_{\mathrm g}$. Let $b:=\min\{h_{\mathrm g},L_{\mathrm g}\}>0$. By Condition~6 of Assumption~\ref{assumption:group-imb-assmp},
\[
P(-b\le\Psi_P(X)<0\mid Y=1)\ge c_\Psi b>0.
\]
Since $\{x:-b\le\Psi_P(x)<0\}\subseteq\mathcal N_{\mathrm g}$, the set $\mathcal N_{\mathrm g}$ has positive probability under $P(\cdot\mid Y=1)$. Hence, combining the lower shift bound on $\mathcal N_{\mathrm g}$ with Lemma~\ref{lemma:group-shift} gives $h_{\mathrm g}\le L_{\mathrm g}$. Therefore, every $x$ with $-h_{\mathrm g}\le\Psi_P(x)<0$ lies in $\mathcal N_{\mathrm g}$ and satisfies $\Psi_{\widetilde P}(x)\ge\Psi_P(x)+h_{\mathrm g}\ge0$, that is, $x\in\mathcal D$.

Given (a)--(c), the rest of Steps~1--3 in the proof of Theorem~\ref{Thm:fairness-improvement}
applies with $(\mathcal{N},h_\kappa)$ replaced by $(\mathcal{N}_g,h_g)$ and
Conditions~4--6 as specified in Assumption~\ref{assumption:group-imb-assmp}: (a) gives
\eqref{eq:flip-tpr} and \eqref{eq:eo-decomp}; (b) and Condition~5 give \eqref{eq:dominance};
Condition~4 then gives
$\Delta_{\mathrm{EO}}(h^{\eta}_{P})-\Delta_{\mathrm{EO}}(h^{\eta}_{\widetilde P})
=P_{B+}(\mathcal{D})-P_{A+}(\mathcal{D})$; and (c) with Condition~6 gives
$P(X\in\mathcal{D}\mid Y=1)\ge c_\Psi h_g$. Combining these as in Step~3,
\eqref{eq:eo-lower} becomes
\begin{equation}\label{eq:EO-population-gap}
\Delta_{\mathrm{EO}}(h_P^\eta)\ge\Delta_{\mathrm{EO}}(h_P^\eta)-\Delta_{\mathrm{EO}}(h_{\widetilde P}^\eta)\ge\frac{\kappa\,c_\Psi}{\bar\omega_A\bar\omega_B}\,h_{\mathrm g}=\frac{\kappa\,c_\Psi\,c'}{\bar\omega_A\bar\omega_B}\Bigl(\alpha-\frac12\Bigr).
\end{equation}
This finishes the proof of Theorem \ref{thm:imbalance-driven-bound}.
Step~5 of the proof of Theorem \ref{Thm:fairness-improvement} involves only $P$ and Conditions 1(a), 2 and 3, so $|\Delta_{\mathrm{EO}}(\widehat h_P)-\Delta_{\mathrm{EO}}(h_P^\eta)|\le R_P(\delta)$ with probability at least $1-\delta$; combined with \eqref{eq:EO-population-gap}, this proves Theorem~\ref{thm:imbalance-driven-bound-full}.
\end{proof}

\begin{proof}[Proof of Theorem \ref{thm:fullbalance-improvement} and Theorem \ref{thm:fullbalance-improvement-full}]
The three oracle steps shift the posterior log-odds pointwise. We need the following lemmas.
\begin{lemma}[Class-balancing shift]\label{lemma:class-shift}
Assume Condition~1 of Assumption~\ref{assumption:prevalence-diff}, $\alpha_1>1/2$ and $\pi_A\le1/2$. Then $\widetilde P_{\mathrm{fb}}$ gives mass $1/4$ to each cell, has cell-conditional distributions $P_c$, and, for $\nu$-almost every $x$,
\[
\Psi_{\widetilde P_{\mathrm{fb}}}(x)-\Psi_{\widetilde P^{(2)}}(x)=\log\frac{1}{\rho_A}.
\]
\end{lemma}
\begin{lemma}[Synthetic Bayes classifier]\label{lemma:syn-bayes}
Under Conditions~2--3 of Assumption~\ref{assumption:improvement-bound-assumption},
\[
\bigl|\Delta_{\mathrm{EO}}(h^\eta_{\widetilde P_{\mathrm{fb}}})-\Delta_{\mathrm{EO}}(h^\eta_{\overline P_{\mathrm{fb}}})\bigr|\le C_q\sqrt{\epsilon_{\mathrm{syn},\mathrm{fb}}}.
\]
\end{lemma}
 
\paragraph{Step 1: total log-odds shift.} For $\nu$-almost every $x$,
\[
\Psi_{\widetilde P_{\mathrm{fb}}}(x)-\Psi_P(x)=s_1(x)+s_2(x)+s_3,
\]
where $s_1:=\Psi_{\widetilde P^{(1)}}-\Psi_P=\log(1+(\mathrm{OR}-1)\omega_B)\in[0,\log\mathrm{OR}]$ by Lemma~\ref{lemma:log-odds-shift}; $s_2:=\Psi_{\widetilde P^{(2)}}-\Psi_{\widetilde P^{(1)}}=\int_{1-\alpha_1}^{1/2}\partial_w\Psi_{Q_w}\,dw\in[-L_{\mathrm g,1},L_{\mathrm g,1}]$ by Lemma~\ref{lemma:group-shift} applied to $\widetilde P^{(1)}$, whose densities are at least $(1-\lambda_1^*)$ times those of $P$, so that Condition~1(a) transfers; and $s_3:=\log(1/\rho_A)\ge0$ by Lemma~\ref{lemma:class-shift}. In particular the total shift is at most $L_{\mathrm{fb}}$.
 
\paragraph{Step 2: the oracle bound.} Let $\mathcal D:=\mathcal X^+_{\widetilde P_{\mathrm{fb}}}\setminus\mathcal X^+_P$. As in the proof of Theorem~\ref{thm:imbalance-driven-bound}, we verify (a)~$\mathcal X^+_P\subseteq\mathcal X^+_{\widetilde P_{\mathrm{fb}}}$, (b)~$\mathcal D\subseteq\mathcal N_{\mathrm{fb}}$ and (c)~$\{x:-h_{\mathrm{fb}}\le\Psi_P(x)<0\}\subseteq\mathcal D$ respectively:
 
(a)~Since $s_1,s_3\ge0$: if $\Psi_P(x)\ge L_{\mathrm g,1}$, then $\Psi_{\widetilde P_{\mathrm{fb}}}(x)\ge\Psi_P(x)-L_{\mathrm g,1}\ge0$. If $0\le\Psi_P(x)<L_{\mathrm g,1}$, then $s_2(x)\ge0$ by Condition~1 of Assumption \ref{assumption:improvement-bound-assumption}, so $\Psi_{\widetilde P_{\mathrm{fb}}}(x)\ge\Psi_P(x)\ge0$.

(b)~If $x\in\mathcal D$, then $\Psi_P(x)<0\le\Psi_{\widetilde P_{\mathrm{fb}}}(x)\le\Psi_P(x)+L_{\mathrm{fb}}$, so $x\in\mathcal N_{\mathrm{fb}}$.

(c)~On $\mathcal N_{\mathrm{fb}}$, Condition~5 gives $\omega_B\ge\bar\omega_B+\kappa$, hence $s_1\ge\log(1+(\mathrm{OR}-1)(\bar\omega_B+\kappa))$, and Condition~1 of Assumption \ref{assumption:improvement-bound-assumption} gives $s_2\ge c'(\alpha_1-\frac12)$. Thus the total shift is at least $h_{\mathrm{fb}}$ on $\mathcal N_{\mathrm{fb}}$. The argument in the proof of Theorem~\ref{thm:imbalance-driven-bound}, with Condition~6 applied on $[-\min\{h_{\mathrm{fb}},L_{\mathrm{fb}}\},0)$, gives $h_{\mathrm{fb}}\le L_{\mathrm{fb}}$. Hence every $x$ with $-h_{\mathrm{fb}}\le\Psi_P(x)<0$ lies in $\mathcal N_{\mathrm{fb}}$ and satisfies $\Psi_{\widetilde P_{\mathrm{fb}}}(x)\ge0$. That is, $x\in\mathcal D$.
 
Given (a)--(c), the rest of Steps 1–3 in the proof of Theorem~\ref{Thm:fairness-improvement} applies, as in the proof of Theorem \ref{thm:imbalance-driven-bound}, with $\widetilde P$, $\mathcal N$ and $h_\kappa$ replaced by $\widetilde P_{\mathrm{fb}}$, $\mathcal N_{\mathrm{fb}}$ and $h_{\mathrm{fb}}$, and gives
\[
\Delta_{\mathrm{EO}}(h^\eta_P)-\Delta_{\mathrm{EO}}(h^\eta_{\widetilde P_{\mathrm{fb}}})\ge\frac{\kappa\,c_\Psi}{\bar\omega_A\bar\omega_B}\,h_{\mathrm{fb}}.
\]
 
\paragraph{Step 3: synthetic generation.} Combining Step~2 with Lemma~\ref{lemma:syn-bayes},
\begin{equation}\label{eq:eo-hp-hpfb-population-lower}
\Delta_{\mathrm{EO}}(h_P^\eta)-\Delta_{\mathrm{EO}}(h_{\overline P_{\mathrm{fb}}}^\eta)
\ge\frac{\kappa\,c_\Psi}{\bar\omega_A\bar\omega_B}\,h_{\mathrm{fb}}-C_q\sqrt{\epsilon_{\mathrm{syn},\mathrm{fb}}},
\end{equation}
which is Theorem~\ref{thm:fullbalance-improvement}.
 
\paragraph{Step 4: empirical classifiers.} We use the following concentration lemma.
\begin{lemma}\label{lemma:newconcentrationtwoQ}
Under Conditions~2 and~3(a) of Assumption~\ref{assumption:prevalence-diff},
$\sup_x|\widehat f_P(x)-\eta_P(x)|\le\Delta_P$ with probability at least $1-\delta/2$.
Under Assumption~\ref{assumption:accuracypreserveass},
$\sup_x|\widehat f_{\overline P_{\mathrm{fb}}}(x)-\eta_{\overline P_{\mathrm{fb}}}(x)|\le\Delta_{\overline P_{\mathrm{fb}}}$
with probability at least $1-\delta/2$.
Under Assumption~\ref{assumption:improvement-bound-assumption}, both bounds hold simultaneously with probability at least $1-\delta$. Finally, under Condition~1 of Assumption~\ref{assumption:convergence}, which drops the calibration condition
from Assumption~\ref{assumption:accuracypreserveass}, $\sup_x|\widehat f_{\overline P_{\rm fb}}(x)-f^*_{\overline P_{\rm fb}}(x)|
\le e_n(\overline P_{\rm fb})$ with probability at least $1-\delta/2$.
\end{lemma}
On the event where the first two bounds of Lemma \ref{lemma:newconcentrationtwoQ} hold simultaneously, the disagreement argument leading to \eqref{eq:empirical-TPR-error}, applied with the margin constants of $P$ (Condition~3 of Assumption~\ref{assumption:prevalence-diff}) and of $\overline P_{\mathrm{fb}}$ (Condition~2 of Assumption~\ref{assumption:improvement-bound-assumption}), gives \[|\Delta_{\mathrm{EO}}(\widehat h_P)-\Delta_{\mathrm{EO}}(h^\eta_P)|\le(C_A+C_B)\Delta_P=R_P(\delta/2)\] and \[|\Delta_{\mathrm{EO}}(\widehat h_{\overline P_{\mathrm{fb}}})-\Delta_{\mathrm{EO}}(h^\eta_{\overline P_{\mathrm{fb}}})|\le(\overline C_A+\overline C_B)\Delta_{\overline P_{\mathrm{fb}}}.\] Combined with \eqref{eq:eo-hp-hpfb-population-lower}, this proves Theorem~\ref{thm:fullbalance-improvement-full}.
\end{proof}

\begin{proof}[Proof of Theorem~\ref{thm:maintain-accuracy-fb} and Theorem \ref{thm:maintain-accuracy-fb-full}] 
First, we have the following lemma.
\begin{lemma}[Bayes optimality]\label{lem:bayes-optimality-fb}
Let $Q$ be any distribution on $(X,Y)$ with $Y\in\{0,1\}$, $\eta_Q(x):=Q(Y=1\mid X=x)$ and $h_Q^\eta(x):=\mathbf 1\{\eta_Q(x)\ge1/2\}$. Then, for every measurable classifier $h$,
\[
\Err_Q(h)-\Err_Q(h^\eta_Q)=\mathbb E_Q\bigl[\,|2\eta_Q(X)-1|\;\mathbf 1\{h(X)\neq h^\eta_Q(X)\}\bigr]\;\ge\;0 .
\]
\end{lemma}
Since $\overline P_{\mathrm{fb}}-\widetilde P_{\mathrm{fb}}=\frac{1}{4p^*}\sum_cd_c(P_{\mathrm{syn},c}-P_c)$, Condition~3 of Assumption~\ref{assumption:improvement-bound-assumption} gives \[d_{\mathrm{TV}}(\overline P_{\mathrm{fb}},\widetilde P_{\mathrm{fb}})\le\frac{1}{4p^*}\sum_cd_c\epsilon_c=\epsilon_{\mathrm{syn},\mathrm{fb}}.\] As $\Err_Q(h)=\mathbb E_Q[\mathbf 1\{h(X)\neq Y\}]$, it follows that
\begin{equation}\label{eq:fixed-h-transfer-acc-revised}
\bigl|\Err_{\widetilde P_{\mathrm{fb}}}(h)-\Err_{\overline P_{\mathrm{fb}}}(h)\bigr|\le\epsilon_{\mathrm{syn},\mathrm{fb}}
\qquad\text{for every classifier }h.
\end{equation}
Since $h^\eta_{\overline P_{\mathrm{fb}}}$ is Bayes-optimal under $\overline P_{\mathrm{fb}}$ (Lemma~\ref{lem:bayes-optimality-fb}),
\[
\Err_{\widetilde P_{\mathrm{fb}}}(h^\eta_{\overline P_{\mathrm{fb}}})
\le\Err_{\overline P_{\mathrm{fb}}}(h^\eta_{\overline P_{\mathrm{fb}}})+\epsilon_{\mathrm{syn},\mathrm{fb}}
\le\Err_{\overline P_{\mathrm{fb}}}(h)+\epsilon_{\mathrm{syn},\mathrm{fb}}
\le\Err_{\widetilde P_{\mathrm{fb}}}(h)+2\epsilon_{\mathrm{syn},\mathrm{fb}},
\]
and taking the infimum over $h$ gives Theorem~\ref{thm:maintain-accuracy-fb}. 

Conditional on the real and synthetic training data, $\overline P_{\rm fb}$ and $\widehat h_{\overline P_{\rm fb}}$ are fixed, so Lemma \ref{lem:bayes-optimality-fb} and \eqref{eq:fixed-h-transfer-acc-revised} apply to them.
By Lemma~\ref{lemma:newconcentrationtwoQ} for $Q=\overline P_{\mathrm{fb}}$, with probability at least $1-\delta/2$, $\sup_x|\widehat f_{\overline P_{\mathrm{fb}}}(x)-\eta_{\overline P_{\mathrm{fb}}}(x)|\le\Delta_{\overline P_{\mathrm{fb}}}$. On this event, as in Step~5 of the proof of Theorem~\ref{Thm:fairness-improvement}, $\widehat h_{\overline P_{\mathrm{fb}}}(x)\neq h^\eta_{\overline P_{\mathrm{fb}}}(x)$ implies $|\eta_{\overline P_{\mathrm{fb}}}(x)-\frac12|\le\Delta_{\overline P_{\mathrm{fb}}}$, so Lemma~\ref{lem:bayes-optimality-fb} gives
\begin{equation}\label{eq:barPfb-vs-bayes-revised}
\Err_{\overline P_{\mathrm{fb}}}(\widehat h_{\overline P_{\mathrm{fb}}})-\Err_{\overline P_{\mathrm{fb}}}(h^\eta_{\overline P_{\mathrm{fb}}})
\le2\Delta_{\overline P_{\mathrm{fb}}}\cdot\overline P_{\mathrm{fb}}\bigl(\widehat h_{\overline P_{\mathrm{fb}}}\neq h^\eta_{\overline P_{\mathrm{fb}}}\bigr)
\le2\Delta_{\overline P_{\mathrm{fb}}}.
\end{equation}
For every classifier $h$, by \eqref{eq:fixed-h-transfer-acc-revised}, \eqref{eq:barPfb-vs-bayes-revised} and Lemma~\ref{lem:bayes-optimality-fb},
\[
\Err_{\widetilde P_{\mathrm{fb}}}(\widehat h_{\overline P_{\mathrm{fb}}})
\le\Err_{\overline P_{\mathrm{fb}}}(h^\eta_{\overline P_{\mathrm{fb}}})+2\Delta_{\overline P_{\mathrm{fb}}}+\epsilon_{\mathrm{syn},\mathrm{fb}}
\le\Err_{\widetilde P_{\mathrm{fb}}}(h)+2\epsilon_{\mathrm{syn},\mathrm{fb}}+2\Delta_{\overline P_{\mathrm{fb}}}.
\]
Taking the infimum over $h$ proves Theorem~\ref{thm:maintain-accuracy-fb-full}, and $h=\widehat h_P$ gives the special case. 
\end{proof}

\begin{proof}[Proof of Theorem~\ref{thm:full-balance-convergence} and Theorem \ref{thm:full-balance-convergence-full}]
By construction, $\widetilde P_{\mathrm{fb}}$ gives mass $1/4$ to each cell $c$ with cell-conditional distribution $P_c$, so $L_{\widetilde P_{\mathrm{fb}}}=L_{\mathrm{bal}}$; by uniqueness, $\theta^*_{\mathrm{bal}}=\theta^*_{\widetilde P_{\mathrm{fb}}}$ and $h^*_{\mathrm{bal}}=h^*_{\widetilde P_{\mathrm{fb}}}$. Write $\epsilon:=\epsilon_{\mathrm{syn},\mathrm{fb}}$.

\paragraph{Step 1: estimation.} By Lemma~\ref{lemma:newconcentrationtwoQ}, with probability at least $1-\delta/2$,
\[\sup_x|\widehat f_{\overline P_{\rm fb}}(x)-f^*_{\overline P_{\rm fb}}(x)|\le e_n(\overline P_{\rm fb}).\]

\paragraph{Step 2: synthetic generation.} By the proof of  Theorem~\ref{thm:maintain-accuracy-fb}, $d_{\mathrm{TV}}(\overline P_{\mathrm{fb}},\widetilde P_{\mathrm{fb}})\le\epsilon$. Since $\nabla L_{\widetilde P_{\mathrm{fb}}}(\theta^*_{\mathrm{bal}})=0$ and $|\partial_j\ell|\le B/2$ under both distributions (Condition~4 of Assumption~\ref{assumption:improvement-bound-assumption}), each coordinate of \[\nabla L_{\overline P_{\mathrm{fb}}}(\theta^*_{\mathrm{bal}})=\mathbb E_{\overline P_{\mathrm{fb}}}[\nabla\ell(\theta^*_{\mathrm{bal}};X,Y)]-\mathbb E_{\widetilde P_{\mathrm{fb}}}[\nabla\ell(\theta^*_{\mathrm{bal}};X,Y)]\] is at most $B\epsilon$ in absolute value, so $\|\nabla L_{\overline P_{\mathrm{fb}}}(\theta^*_{\mathrm{bal}})\|\le\sqrt dB\epsilon$. As $\nabla L_{\overline P_{\mathrm{fb}}}(\theta^*_{\overline P_{\mathrm{fb}}})=0$, Condition~2 of Assumption~\ref{assumption:convergence} and the Cauchy--Schwarz step of \eqref{eq:7}--\eqref{eq:8} give
\begin{equation}\label{eq:parameter-syn-error}
\bigl\|\theta^*_{\mathrm{bal}}-\theta^*_{\overline P_{\mathrm{fb}}}\bigr\|\le\frac{\sqrt dB}{\mu_{\mathrm{syn}}}\,\epsilon,
\qquad\text{hence}\qquad
\sup_x\bigl|f^*_{\mathrm{bal}}(x)-f^*_{\overline P_{\mathrm{fb}}}(x)\bigr|\le\delta_{\mathrm{syn}}:=\frac{L\sqrt dB}{\mu_{\mathrm{syn}}}\,\epsilon .
\end{equation}
 
\paragraph{Step 3: disagreement.} On the event of Step~1, \[\sup_x|\widehat f_{\overline P_{\mathrm{fb}}}-f^*_{\mathrm{bal}}|\le r:=e_n(\overline P_{\mathrm{fb}})+\delta_{\mathrm{syn}}.\] If $\widehat h_{\overline P_{\mathrm{fb}}}(x)\neq h^*_{\mathrm{bal}}(x)$, the two scores lie on opposite sides of $1/2$, so $|f^*_{\mathrm{bal}}(x)-\frac12|\le r$. As in \eqref{eq:empirical-TPR-error}, Condition~3 of Assumption~\ref{assumption:convergence} gives \[|\mathrm{TPR}_s(\widehat h_{\overline P_{\mathrm{fb}}})-\mathrm{TPR}_s(h^*_{\mathrm{bal}})|\le P_{s+}(|f^*_{\mathrm{bal}}(X)-\frac12|\le r)\le C^*_sr,\] and summing over $s$ proves Theorem~\ref{thm:full-balance-convergence-full}. The same argument with $f^*_{\overline P_{\mathrm{fb}}}$ in place of $\widehat f_{\overline P_{\mathrm{fb}}}$ gives the population term
\begin{equation}\label{eq:EO-diff-syn-error}
\bigl|\Delta_{\mathrm{EO}}(h^*_{\overline P_{\mathrm{fb}}})-\Delta_{\mathrm{EO}}(h^*_{\mathrm{bal}})\bigr|\le(C^*_A+C^*_B)\,\delta_{\mathrm{syn}},
\end{equation}
and with $\eta_{\widetilde P_{\rm fb}}$ in place of $\widehat f_{\overline P_{\rm fb}}$, whose distance to $f^*_{\rm bal}$ is at most $\epsilon_{{\rm cal},\widetilde P_{\rm fb}}$, it gives $|\Delta_{\rm EO}(h^\eta_{\widetilde P_{\rm fb}})-\Delta_{\rm EO}(h^*_{\rm bal})|\le (C^*_A+C^*_B)\,\epsilon_{{\rm cal},\widetilde P_{\rm fb}}$, the comparison of the two targets in Section~\ref{sec:balance-convergence}.

We then prove Theorem~\ref{thm:full-balance-convergence} in the main paper.
By Theorem \ref{thm:full-balance-convergence-full}, it suffices to bound the bracket in $e_n(\overline P_{\mathrm{fb}})$.
Let $r_n(\delta):=\sqrt{d\log(8d(3n_P)^d/\delta)/n_P}$, which equals the square root in the main text. Since $n_P\le n_{\overline P_{\mathrm{fb}}}$, we have $n_P/n_{\overline P_{\mathrm{fb}}}\le1$ and $\sqrt{n_{\overline P_{\mathrm{fb}}}-n_P}/n_{\overline P_{\mathrm{fb}}}\le n_P^{-1/2}$. Also $\log(8d/\delta)\le\log(8d(3n_P)^d/\delta)$ and $1/n_P\le r_n(\delta)$. Hence the bracket in $e_n(\overline P_{\mathrm{fb}})$ is at most $(\sqrt2B+2\sqrt dL_\nabla D_\Theta)\,r_n(\delta)+\sqrt dB\,\epsilon_{\mathrm{mix}}$, and the theorem holds with
\[
C_1:=\frac{(C^*_A+C^*_B)L}{\mu_{\overline P_{\mathrm{fb}}}}\max\bigl\{\sqrt2B+2\sqrt dL_\nabla D_\Theta,\ \sqrt dB\bigr\},
\qquad
C_2:=\frac{(C^*_A+C^*_B)L\sqrt dB}{\mu_{\mathrm{syn}}}.
\]
This finishes the proof of Theorem~\ref{thm:full-balance-convergence}. 
\end{proof}

\begin{proof}[Proof of Theorem \ref{thm:minimax-EO-lower}]
We use the notation of Appendix~\ref{sec:minimaxassumption} and write $\mathbb E:=\mathbb E_{P_{c_0,0}}$. The proof constructs a one-parameter family $\{P_t\}$ that changes only the negative cell $c_0$, shows that the Equal Opportunity gap of the balanced target moves at rate $|t|$ along it while the data distribution moves at rate $\sqrt n\,|t|$, and concludes with a two-point argument.
 
\paragraph{Step 1: an informative direction.} By Condition~1, we have \[|\varphi(Z)|\le\|H_0^{-1}\gamma_{\mathrm{EO}}\|\,M_{1,c_0}(Z)\] almost surely, so $\varphi\in L^1(P_{c_0,0})$. Let $\widetilde\varphi:=\varphi-\mathbb E[\varphi]$. It has mean zero and, by Condition~3, is not almost surely zero. Hence $\mu(dz):=\widetilde\varphi(z)\,P_{c_0,0}(dz)$ is a nonzero finite signed measure with $\mu\ll P_{c_0,0}$, and since $\mathcal H_{c_0}$ is measure-determining, some $h\in\mathcal H_{c_0}$ satisfies $\int h\,d\mu=\mathbb E[h\widetilde\varphi]\neq0$. Let $\bar h:=h-\mathbb E[h]$, which lies in $\mathcal H_{c_0}$ because the class is linear and contains the constants. As $\mathbb E[\widetilde\varphi]=0$, we have $\mathbb E[\bar h\widetilde\varphi]=\mathbb E[h\widetilde\varphi]\neq0$, so $\bar h$ is not almost surely zero, and we may set $u:=\bar h/\|\bar h\|_{L^\infty(P_{c_0,0})}$. Then $u\in\mathcal H_{c_0}$, $\mathbb E[u]=0$, $\|u\|_\infty=1$, and, using $\mathbb E[u]=0$ once more,
\begin{equation}\label{eq:uzphiz-exp-nonzero}
\mathbb E[u(Z)\varphi(Z)]=\mathbb E[u(Z)\widetilde\varphi(Z)]\neq0 .
\end{equation}
 
\paragraph{Step 2: the hard submodel and its Kullback--Leibler divergence.} Condition~4, applied to $u$, gives $\varepsilon_u>0$, likelihood ratios $r_t:=dP_{c_0,t}/dP_{c_0,0}$ with $r_0=1$ and $\frac{d}{dt}r_t|_{t=0}=u$ in $L^\infty(P_{c_0,0})$, and full distributions $P_t\in\mathcal P$ that differ from $P_0$ only in cell $c_0$. Since $\|(r_t-1)/t-u\|_\infty\to0$ as $t\to0$, there is $\delta_u\in(0,\varepsilon_u)$ with $\|(r_t-1)/t-u\|_\infty\le1$, and hence \[\|r_t-1\|_\infty\le(1+\|u\|_\infty)|t|=2|t|,\quad \text{for}~ |t|\le\delta_u.\] Set $t_0:=\min\{\delta_u,1/4\}$. Then, for $|t|\le t_0$,
\begin{equation}\label{eq:rt-up-low-bound}
P_t\in\mathcal P,
\qquad
\|r_t-1\|_\infty\le2|t|\le\tfrac12 .
\end{equation}
Because the cell probabilities are fixed and only cell $c_0$ changes, $dP_0/dP_t$ equals $1/r_t$ on cell $c_0$ and $1$ elsewhere, so $\mathrm{KL}(P_0\|P_t)=p_{c_0}\,\mathbb E[-\log r_t(Z)]$. Write $s_t:=r_t-1$. Since $r_t$ is a probability density with respect to $P_{c_0,0}$, $\mathbb E[s_t]=0$, and $|s_t|\le1/2$ by \eqref{eq:rt-up-low-bound}. Using $-\log(1+x)\le-x+x^2$ for $|x|\le1/2$,
\[
\mathrm{KL}(P_0\|P_t)\le p_{c_0}\bigl(-\mathbb E[s_t]+\mathbb E[s_t^2]\bigr)\le p_{c_0}\|s_t\|_\infty^2\le4p_{c_0}t^2 .
\]
Let $\mathbb P_t:=P_t^{\otimes n}\otimes\nu_U$ be the law of the sample and the auxiliary randomness $U$, whose law $\nu_U$ does not depend on $t$. By additivity of the KL-divergence over independent coordinates,
\begin{equation}\label{eq:KL-bound-nt2}
\mathrm{KL}(\mathbb P_0\|\mathbb P_t)=n\,\mathrm{KL}(P_0\|P_t)\le4np_{c_0}t^2,
\qquad |t|\le t_0 .
\end{equation}
 
\paragraph{Step 3: the balanced target moves smoothly.} Write $L_t:=L_{\mathrm{bal},P_t}$. Since $L_{\mathrm{bal},P}=\frac14\sum_c\mathbb E_{P_c}[\ell]$ and only cell $c_0$ changes,
\begin{equation}\label{eq:Lt-representation}
L_t(\theta)=L_0(\theta)+\tfrac14\,\mathbb E\bigl[(r_t(Z)-1)\,\ell(\theta;Z)\bigr].
\end{equation}
We then proceed with the proof in four small steps:

\emph{(a) Uniform closeness.} For every compact $K\subset\Theta$, \eqref{eq:rt-up-low-bound} and the envelope $M_{0,c_0,K}$ give
\[
\sup_{\theta\in K}|L_t(\theta)-L_0(\theta)|\le\frac{|t|}{2}\,\mathbb E[M_{0,c_0,K}]\longrightarrow0
\qquad\text{as }t\to0 .
\] Taking $K=\Theta$, which is compact by Condition~4, the convergence is uniform on $\Theta$.

\emph{(b) Consistency of the minimizers.} Each $L_t$ is continuous on the compact set $\Theta$ (by dominated convergence with the envelope), so it has a global minimizer. Let $t_k\to0$ and let $\theta_k$ minimize $L_{t_k}$. For any limit point $\bar\theta$ of $(\theta_k)$, along the corresponding subsequence,
\[
L_0(\theta_k)\le L_{t_k}(\theta_k)+\sup_\Theta|L_{t_k}-L_0|\le L_{t_k}(\theta^*_0)+\sup_\Theta|L_{t_k}-L_0|\le L_0(\theta^*_0)+2\sup_\Theta|L_{t_k}-L_0| ,
\]
so continuity of $L_0$ and (a) give $L_0(\bar\theta)\le L_0(\theta^*_0)$. By uniqueness of the baseline minimizer, $\bar\theta=\theta^*_0$. As $\Theta$ is compact, every global minimizer of $L_t$ therefore tends to $\theta^*_0$ as $t\to0$.
 
\emph{(c) Uniqueness.} Since $H_0\succ0$ and the Hessian of $L_0$ is continuous near the interior point $\theta^*_0$, there are a closed ball $\overline B\subset\mathcal N_0\cap\operatorname{int}(\Theta)$ centered at $\theta^*_0$ and $\lambda_0>0$ with $\lambda_{\min}(\nabla^2_\theta L_0)\ge\lambda_0$ on $\overline B$. Differentiating \eqref{eq:Lt-representation} under the expectation, which the envelopes $M_{1,c_0}$, $M_{2,c_0}$ permit,
\begin{equation}\label{eq:Lt-derivatives}
\nabla_\theta L_t(\theta)=\nabla_\theta L_0(\theta)+\tfrac14\,\mathbb E\bigl[(r_t-1)\nabla_\theta\ell(\theta;Z)\bigr],
\quad
\nabla^2_\theta L_t(\theta)=\nabla^2_\theta L_0(\theta)+\tfrac14\,\mathbb E\bigl[(r_t-1)\nabla^2_\theta\ell(\theta;Z)\bigr]
\end{equation}
on $\overline B$, so \[\sup_{\overline B}\|\nabla^2_\theta L_t-\nabla^2_\theta L_0\|_{\mathrm{op}}\le\frac{|t|}{2}\mathbb E[M_{2,c_0}].\] After decreasing $t_0$, $\lambda_{\min}(\nabla^2_\theta L_t)\ge\lambda_0/2$ on $\overline B$ for $|t|\le t_0$. Then $L_t$ is strictly convex on $\overline B$ and has at most one stationary point there. By (b), after decreasing $t_0$ further, every global minimizer of $L_t$ lies in $\overline B\subset\operatorname{int}(\Theta)$ and is therefore stationary. Hence the global minimizer is unique. Denote it by $\theta_t:=\theta^*_{\mathrm{bal},P_t}$.
 
\emph{(d) Smoothness and first-order movement.} By \eqref{eq:Lt-derivatives}, the map $(t,\theta)\mapsto\nabla_\theta L_t(\theta)$ is continuously differentiable near $(0,\theta^*_0)$, with \[\partial_\theta\nabla_\theta L_t(\theta)=\nabla^2_\theta L_t(\theta),\quad \text{and}\quad \partial_t\nabla_\theta L_t(\theta)=\frac14\mathbb E\left[\frac{d r_t}{dt}\nabla_\theta\ell(\theta;Z)\right],\] since $t\mapsto r_t$ is $C^1$ in $L^\infty(P_{c_0,0})$ and $\nabla_\theta\ell$ is continuous in $\theta$ with envelope $M_{1,c_0}$. At $(0,\theta^*_0)$ the equation $\nabla_\theta L_t(\theta)=0$ holds and its $\theta$-Jacobian $H_0$ is invertible. By the implicit-function theorem, its solutions near $(0,\theta^*_0)$ form the graph of a unique $C^1$ map $t\mapsto\widetilde\theta(t)$ with $\widetilde\theta(0)=\theta^*_0$. Since $\theta_t\to\theta^*_0$ by (b) and $\nabla_\theta L_t(\theta_t)=0$, we have $\theta_t=\widetilde\theta(t)$ for small $|t|$, so $t\mapsto\theta_t$ is $C^1$ near $0$. Differentiating $\nabla_\theta L_t(\theta_t)=0$ at $t=0$ gives $0=b_u+H_0\,\dot\theta_0$ with $b_u:=\frac14\mathbb E[u(Z)\nabla_\theta\ell(\theta^*_0;Z)]$, that is,
\begin{equation}\label{eq:theta-t-t-derivative-0}
\dot\theta_0:=\frac{d\theta_t}{dt}\Big|_{t=0}=-H_0^{-1}b_u .
\end{equation}
 
\paragraph{Step 4: the Equal Opportunity gap of the target moves at rate $|t|$.} Because $c_0$ is a negative cell, the positive-cell distributions $P_{A+,0}$ and $P_{B+,0}$ are the same under every $P_t$. Consequently, for every measurable rule $q:\mathcal X\to[0,1]$,
\begin{equation}\label{eq:EO-invariant}
\Delta_{\mathrm{EO},P_t}(q)=\Delta_{\mathrm{EO},P_0}(q)=:\mathcal E(q),
\end{equation}
and, with $\Gamma$ from \eqref{eq:EO-map}, $\Delta_{\mathrm{EO},P_t}(h^*_{\mathrm{bal},P_t})=|\Gamma(\theta_t)|$ for $|t|\le t_0$. By Condition~2 and Step~3(d), the chain rule shows that $t\mapsto\Gamma(\theta_t)$ is differentiable at $0$ with derivative
\[
\frac{d}{dt}\Gamma(\theta_t)\Big|_{t=0}
=\gamma_{\mathrm{EO}}^\top\dot\theta_0
=-\gamma_{\mathrm{EO}}^\top H_0^{-1}b_u
=-\tfrac14\,\mathbb E\bigl[u(Z)\,\gamma_{\mathrm{EO}}^\top H_0^{-1}\nabla_\theta\ell(\theta^*_0;Z)\bigr]
=-\tfrac14\,\mathbb E[u(Z)\varphi(Z)],
\]
which is nonzero by \eqref{eq:uzphiz-exp-nonzero}. Let $a:=\frac18|\mathbb E[u\varphi]|>0$, half the absolute value of this derivative. Since $(\Gamma(\theta_t)-\Gamma(\theta^*_0))/t$ converges to the derivative, whose absolute value is $2a$, there is $t_1\in(0,t_0]$ such that $|\Gamma(\theta_t)-\Gamma(\theta^*_0)|/|t|\ge a$, that is, \[|\Gamma(\theta_t)-\Gamma(\theta^*_0)|\ge a|t|,\quad  \text{for}~ 0<|t|\le t_1.\] It remains to remove the absolute values in $|\Gamma|$. If $\Gamma(\theta^*_0)=0$, then $\bigl||\Gamma(\theta_t)|-|\Gamma(\theta^*_0)|\bigr|=|\Gamma(\theta_t)-\Gamma(\theta^*_0)|$ trivially. If $\Gamma(\theta^*_0)\neq0$, then, since $t\mapsto\Gamma(\theta_t)$ is continuous at $0$, after decreasing $t_1$ the value $\Gamma(\theta_t)$ has the sign of $\Gamma(\theta^*_0)$, and the same identity holds. In both cases,
\begin{equation}\label{eq:deltaEOdiff-lower-t}
\bigl|\Delta_{\mathrm{EO},P_t}(h^*_{\mathrm{bal},P_t})-\Delta_{\mathrm{EO},P_0}(h^*_{\mathrm{bal},P_0})\bigr|\ge a\,|t|,
\qquad |t|\le t_1 .
\end{equation}
 
\paragraph{Step 5: two-point lower bound.} Let $\widehat q_n$ be any learning procedure and set
\[
t_n:=\min\bigl\{t_1,\,1/(8\sqrt{np_{c_0}})\bigr\}.
\] By \eqref{eq:KL-bound-nt2}, we have the KL-divergence
\[\mathrm{KL}(\mathbb P_0\|\mathbb P_{t_n})\le4np_{c_0}t_n^2\le1/16,\] and Pinsker's inequality gives $d_{\mathrm{TV}}(\mathbb P_0,\mathbb P_{t_n})\le\sqrt{1/32}<1/4$. Write $\Delta_0:=\Delta_{\mathrm{EO},P_0}(h^*_{\mathrm{bal},P_0})$ and $\Delta_1:=\Delta_{\mathrm{EO},P_{t_n}}(h^*_{\mathrm{bal},P_{t_n}})$. By \eqref{eq:EO-invariant}, the estimated gap $\widehat{\mathcal E}:=\mathcal E(\widehat q_n)$ equals $\Delta_{\mathrm{EO},P}(\widehat q_n)$ for both $P=P_0$ and $P=P_{t_n}$. It is a measurable function of the sample and $U$, by joint measurability and boundedness of $\widehat q_n$. Consider the test $\psi:=\mathbf 1\{|\widehat{\mathcal E}-\Delta_1|<|\widehat{\mathcal E}-\Delta_0|\}$. By the triangle inequality, $|\widehat{\mathcal E}-\Delta_0|\ge|\Delta_1-\Delta_0|/2$ when $\psi=1$, and $|\widehat{\mathcal E}-\Delta_1|\ge|\Delta_1-\Delta_0|/2$ when $\psi=0$. Hence
\begin{align*}
\mathbb E_{\mathbb P_0}|\widehat{\mathcal E}-\Delta_0|+\mathbb E_{\mathbb P_{t_n}}|\widehat{\mathcal E}-\Delta_1|
&\ge\frac{|\Delta_1-\Delta_0|}{2}\Bigl[\mathbb P_0(\psi=1)+\mathbb P_{t_n}(\psi=0)\Bigr]\\
&\ge\frac{|\Delta_1-\Delta_0|}{2}\bigl(1-d_{\mathrm{TV}}(\mathbb P_0,\mathbb P_{t_n})\bigr)
\ge\frac38\,|\Delta_1-\Delta_0| ,
\end{align*}
where the middle inequality holds for every test by the definition of total variation. The larger of the two expectations is at least half the sum, so by \eqref{eq:deltaEOdiff-lower-t},
\[
\sup_{P\in\mathcal P}\mathbb E_P\Bigl[\bigl|\Delta_{\mathrm{EO},P}(\widehat q_n)-\Delta_{\mathrm{EO},P}(h^*_{\mathrm{bal},P})\bigr|\Bigr]
\ge\max_{P\in\{P_0,P_{t_n}\}}\mathbb E_P\Bigl[\bigl|\Delta_{\mathrm{EO},P}(\widehat q_n)-\Delta_{\mathrm{EO},P}(h^*_{\mathrm{bal},P})\bigr|\Bigr]
\ge\frac{3}{16}\,a\,t_n .
\]
This holds for every procedure $\widehat q_n$, so
\[
\mathfrak R^{\mathrm{EO}}_n\ge\frac{3a}{16}\min\Bigl\{t_1,\frac{1}{8\sqrt{np_{c_0}}}\Bigr\}.
\]
Since $\min\{t,\kappa x\}\ge\min\{1,\kappa\}\min\{t,x\}$ for $\kappa,t,x>0$, the theorem follows with
\[
t_*:=t_1,
\qquad
C_{\mathrm{EO}}:=\frac{3a}{16}\min\Bigl\{1,\frac{1}{8\sqrt{p_{c_0}}}\Bigr\}. \qedhere
\]
\end{proof}

\begin{proof}[Proof of Theorem~\ref{thm:adaptive_near_optimality-full}]
The proof has four steps. Step~1 shows that the cross-validated fairness criterion is uniformly close to its population counterpart $\Delta_{\mathrm{EO}}(h^*_{\overline P_{\boldsymbol\lambda}})$ over the grid. Step~2 does the same for the accuracy criterion and derives its consequences for feasibility. Step~3 shows that, at termination, the selected allocation nearly satisfies the projected-gradient fixed-point condition of the population problem. Step~4 turns this into the distance bound \eqref{eq:allocation_convergence} using the curvature of the population fairness criterion. The trained classifiers enter only through Steps~1 and~2, which is where the margin condition of Assumption~\ref{assumption:adaptive_search}(2) is used. Steps~3 and~4 use only the resulting uniform bounds and the population conditions of Assumption~\ref{assumption:adaptive_search}(4).
By construction (Appendix~\ref{sec:assumptionforacasa}), each pair $(k,\boldsymbol{\lambda})\in\{1,\dots,K\}\times\Lambda_\omega$ uses a single synthetic sample drawn from generators fitted once per fold, so $\widehat{\Delta}^{\mathrm{CV}}_{\mathrm{EO}}$ and $\widehat{\mathrm{Err}}^{\mathrm{CV}}$ are fixed functions on $\Lambda_\omega$. Moreover, the search terminates, so $\widehat{\boldsymbol{\lambda}}$ is well defined. Every criterion value used by the algorithm, including those at $\widehat{\boldsymbol{\lambda}}$ and at the safeguard points in Step~3, is therefore covered by the bounds below, which hold uniformly over $k$ and $\boldsymbol{\lambda}\in\Lambda_\omega$.
Throughout, we condition on the group--label cells of the real observations and on the fold partition, which does not depend on the features. All probabilities below are conditional on them. Then $n^{\mathrm{val}}_{s+}$, $n_{\mathrm{val}}$, $n_{\mathrm{cell,tr}}$, $\epsilon_{\mathrm{strat,val}}$ and $\epsilon_{\mathrm{syn,CV}}$, and hence $\epsilon_{\mathrm{fair}}(\delta)$, $\epsilon_{\mathrm{acc}}(\delta)$ and \eqref{eq:acasa-conditions}, are fixed, and by Assumption~\ref{assumption:improvement-bound-assumption}(4) the constants of Assumption~\ref{assumption:adaptive_search} do not depend on the realization of the fitted generators.

\paragraph{Step 1: the cross-validated fairness criterion.}
Fix a fold $k$ and an allocation
$\boldsymbol{\lambda}\in\Lambda_\omega$.
For fold $k$, the real training sample contains $n_c^{(-k)}$
observations from cell $c$, while the augmentation rule generates
\(
m_c^{(-k)}(\boldsymbol\lambda)
=
\left\lfloor
\lambda_c n_{\max}^{(-k)}
\right\rfloor
\)
synthetic observations from $P_{\mathrm{syn},c}^{(-k)}$.
Hence the fold-specific population augmented distribution is
\begin{align}
Q_{k,\boldsymbol\lambda}
&=
\frac{
\displaystyle\sum_{c\in\mathcal C} n_c^{(-k)} P_c
+
\displaystyle\sum_{c\in\mathcal C}
m_c^{(-k)}(\boldsymbol\lambda)P_{\mathrm{syn},c}^{(-k)}
}{
n_{\mathrm{tr}}^{(k)}+m^{(-k)}(\boldsymbol\lambda)
}
=
\frac{
\displaystyle\sum_{c\in\mathcal C}\widehat p_c^{(-k)}P_c
+
\displaystyle\sum_{c\in\mathcal C}
\frac{m_c^{(-k)}(\boldsymbol\lambda)}{n_{\mathrm{tr}}^{(k)}}
P_{\mathrm{syn},c}^{(-k)}
}{
1+\frac{m^{(-k)}(\boldsymbol\lambda)}{n_{\mathrm{tr}}^{(k)}}
},
\label{eq:fold_population_distribution}
\end{align}
where
\(
\widehat p_c^{(-k)}
=
\frac{n_c^{(-k)}}{n_{\mathrm{tr}}^{(k)}}\).
Likewise, the continuous full-sample population distribution associated
with allocation $\boldsymbol\lambda$ is
\begin{align}
\overline P_{\boldsymbol\lambda}
=
\frac{
\displaystyle\sum_{c\in\mathcal C} p_c P_c
+
r\displaystyle\sum_{c\in\mathcal C}\lambda_c P_{\mathrm{syn},c}
}{
1+r\displaystyle\sum_{c\in\mathcal C}\lambda_c
},
\label{eq:full_population_distribution}
\end{align}
where 
\(
P=\displaystyle\sum_{c\in\mathcal C}p_cP_c,
\) and \(
r=\frac{n_{\max}}{n_P}.
\)
Let
\[
L_{Q_{k,\boldsymbol{\lambda}}}(\theta)
=
\bE_{Q_{k,\boldsymbol{\lambda}}}
\left[
\ell(\theta;X,Y)
\right],
\qquad
\theta^*_{Q_{k,\boldsymbol{\lambda}}}
\in
\arg\min_{\theta\in\Theta}
L_{Q_{k,\boldsymbol{\lambda}}}(\theta),
\]
and define
\[
f^*_{Q_{k,\boldsymbol{\lambda}}}(x)
=
f_{\theta^*_{Q_{k,\boldsymbol{\lambda}}}}(x),
\qquad
h^*_{Q_{k,\boldsymbol{\lambda}}}(x)
=
\mathbf 1
\left\{
f^*_{Q_{k,\boldsymbol{\lambda}}}(x)\geq\frac12
\right\}.
\]
For the corresponding finite augmented training sample in fold $k$, define
\[
\widehat L_{Q_{k,\boldsymbol{\lambda}}}(\theta)
=
\frac{
1
}{
n_{\mathrm{tr}}^{(k)}
+
m^{(-k)}(\boldsymbol{\lambda})
}\left\{
\sum_{i\in\mathcal D_{\mathrm{tr}}^{(k)}}
\ell(\theta;X_i,Y_i)
+
\sum_{j=1}^{m^{(-k)}(\boldsymbol{\lambda})}
\ell(\theta;\widetilde X_j,\widetilde Y_j)\right\},
\]
with
\[
\widehat\theta_{Q_{k,\boldsymbol{\lambda}}}
\in
\arg\min_{\theta\in\Theta}
\widehat L_{Q_{k,\boldsymbol{\lambda}}}(\theta),
\quad
\widehat f_{Q_{k,\boldsymbol{\lambda}}}(x)
=
f_{\widehat\theta_{Q_{k,\boldsymbol{\lambda}}}}(x),\quad \widehat h_{Q_{k,\boldsymbol{\lambda}}}(x)
=
\mathbf 1
\left\{
\widehat f_{Q_{k,\boldsymbol{\lambda}}}(x)\geq\frac12
\right\}.
\]
By the same curvature and first-order optimality argument used in the proof of
Theorem \ref{Thm:fairness-improvement}, equations \eqref{eq:7}--\eqref{eq:8}, we convert the empirical gradient error into a parameter
estimation error. Specifically, by the uniform curvature condition,
\[
\left\langle
\nabla\widehat L_{Q_{k,\boldsymbol{\lambda}}}(\theta_{Q_{k,\boldsymbol{\lambda}}}^*)
-
\nabla\widehat L_{Q_{k,\boldsymbol{\lambda}}}(\widehat\theta_{Q_{k,\boldsymbol{\lambda}}}),
\theta_{Q_{k,\boldsymbol{\lambda}}}^*-\widehat\theta_{Q_{k,\boldsymbol{\lambda}}}
\right\rangle
\geq
\mu
\left\|
\theta_{Q_{k,\boldsymbol{\lambda}}}^*-\widehat\theta_{Q_{k,\boldsymbol{\lambda}}}
\right\|_2^2.
\]
Since $\widehat\theta_{Q_{k,\boldsymbol{\lambda}}}$ is an interior minimizer of $\widehat L_{Q_{k,\boldsymbol{\lambda}}}$,
\(
\nabla\widehat L_{Q_{k,\boldsymbol{\lambda}}}(\widehat\theta_{Q_{k,\boldsymbol{\lambda}}})=0.
\)
Thus, 
\[
\begin{aligned}
\mu
\left\|
\theta_{Q_{k,\boldsymbol{\lambda}}}^*-\widehat\theta_{Q_{k,\boldsymbol{\lambda}}}
\right\|_2^2
&\leq
\left\langle
\nabla\widehat L_{Q_{k,\boldsymbol{\lambda}}}(\theta_{Q_{k,\boldsymbol{\lambda}}}^*),
\theta_{Q_{k,\boldsymbol{\lambda}}}^*-\widehat\theta_{Q_{k,\boldsymbol{\lambda}}}
\right\rangle
\leq
\left\|
\nabla\widehat L_{Q_{k,\boldsymbol{\lambda}}}(\theta_{Q_{k,\boldsymbol{\lambda}}}^*)
\right\|_2
\left\|
\theta_{Q_{k,\boldsymbol{\lambda}}}^*-\widehat\theta_{Q_{k,\boldsymbol{\lambda}}}
\right\|_2.
\end{aligned}
\]
Moreover, since $\theta_{Q_{k,\boldsymbol{\lambda}}}^*$ is an interior minimizer of the population risk,
$\nabla L_{Q_{k,\boldsymbol{\lambda}}}(\theta_{Q_{k,\boldsymbol{\lambda}}}^*)=0$, so
\begin{equation*}
\left\|
\widehat\theta_{Q_{k,\boldsymbol{\lambda}}}
-
\theta^*_{Q_{k,\boldsymbol{\lambda}}}
\right\|_2
\leq
\frac{1}{\mu}
\left\|
\nabla
\widehat L_{Q_{k,\boldsymbol{\lambda}}}
\left(
\theta^*_{Q_{k,\boldsymbol{\lambda}}}
\right)
-
\nabla
L_{Q_{k,\boldsymbol{\lambda}}}
\left(
\theta^*_{Q_{k,\boldsymbol{\lambda}}}
\right)
\right\|_2.
\end{equation*}
It therefore remains to control the empirical gradient error uniformly over folds and
candidate allocations.
Because the folds are stratified, the real training observations are not treated as an
i.i.d. sample from $P$. Instead, conditional on the cell counts, observations in cell
$c$ are i.i.d. from $P_c$. For each cell define its empirical gradient average in fold
$k$ by
\[
\frac{1}{n_c^{(-k)}}
\sum_{i\in\mathcal D_{c,\mathrm{tr}}^{(k)}}
\nabla\ell(\theta;X_i,Y_i).
\]
To obtain a concentration bound that is uniform over the continuous parameter
space $\Theta$, we first discretize $\Theta$ by a finite covering net. Recall
that $\Theta\subset\mathbb R^d$ is compact with Euclidean diameter
\(
D_\Theta
:=
\sup_{\theta,\theta'\in\Theta}
\|\theta-\theta'\|_2.
\)
Let
\(
\varepsilon_\Theta
:=
\frac{D_\Theta}{n_{\mathrm{cell,tr}}},
\)
and let $\mathcal N\subseteq\Theta$ be a maximal
$\varepsilon_\Theta$-separated subset of $\Theta$, so that
\(
\|\theta^\circ-\theta^{\circ\prime}\|_2
>
\varepsilon_\Theta
\)
for any distinct
$\theta^\circ,\theta^{\circ\prime}\in\mathcal N$.
By maximality, $\mathcal N$ is also an $\varepsilon_\Theta$-net of $\Theta$.
It remains to bound the cardinality of $\mathcal N$. Fix any
$\theta_0\in\Theta$. Since $\Theta$ has diameter $D_\Theta$,
every $\theta^\circ\in\mathcal N$ satisfies
\(
\|\theta^\circ-\theta_0\|_2\leq D_\Theta.
\)
Moreover, since $\mathcal N$ is $\varepsilon_\Theta$-separated, the Euclidean
balls
\[
\theta^\circ+\frac{\varepsilon_\Theta}{2}B_2^d,
\qquad
\theta^\circ\in\mathcal N,
\]
are pairwise disjoint. 
Hence
\[
\theta^\circ+\frac{\varepsilon_\Theta}{2}B_2^d
\subseteq
\theta_0+
\left(
D_\Theta+\frac{\varepsilon_\Theta}{2}
\right)B_2^d
\]
for every $\theta^\circ\in\mathcal N$.
Because the preceding balls are pairwise disjoint and are all contained in
the larger Euclidean ball,
\begin{align*}
|\mathcal N|\,
\operatorname{vol}
\left(
\frac{\varepsilon_\Theta}{2}B_2^d
\right)
&\leq
\operatorname{vol}
\left[
\left(
D_\Theta+\frac{\varepsilon_\Theta}{2}
\right)B_2^d
\right].
\end{align*}
Using
\(
\operatorname{vol}(rB_2^d)
=
r^d\operatorname{vol}(B_2^d),
\)
we obtain
\[
|\mathcal N|
\left(
\frac{\varepsilon_\Theta}{2}
\right)^d
\leq
\left(
D_\Theta+\frac{\varepsilon_\Theta}{2}
\right)^d.
\]
Therefore,
\[
|\mathcal N|
\leq
\left(
1+\frac{2D_\Theta}{\varepsilon_\Theta}
\right)^d.
\]
Substituting
$\varepsilon_\Theta=D_\Theta/n_{\mathrm{cell,tr}}$ gives
\[
|\mathcal N|
\leq
\left(
1+2n_{\mathrm{cell,tr}}
\right)^d
\leq
\left(
3n_{\mathrm{cell,tr}}
\right)^d.
\]
We next establish the empirical-gradient concentration bound on the covering net.
Fix a fold $k$, a cell
$c\in\{A+,A-,B+,B-\}$, a net point
$\theta^\circ\in\mathcal N$, and a coordinate
$j\in\{1,\ldots,d\}$.
Conditional on the realized cell counts, the observations
$\{(X_i,Y_i):i\in\mathcal D_{c,\mathrm{tr}}^{(k)}\}$
are i.i.d.\ from $P_c$. Moreover, by the bounded-gradient assumption,
\(
\left|
\partial_j\ell(\theta^\circ;X_i,Y_i)
\right|
\leq
\frac{B}{2}.
\)
Hence Hoeffding's inequality gives, for every $t>0$,
\begin{align}
&
P\left(
\left|
\frac{1}{n_c^{(-k)}}
\sum_{i\in\mathcal D_{c,\mathrm{tr}}^{(k)}}
\partial_j\ell(\theta^\circ;X_i,Y_i)
-
\bE_{P_c}
\left[
\partial_j\ell(\theta^\circ;X,Y)
\right]
\right|
>t
\right)
\nonumber\\
&\qquad\leq
2\exp\left(
-\frac{2n_c^{(-k)}t^2}{B^2}
\right)
\leq
2\exp\left(
-\frac{2n_{\mathrm{cell,tr}}t^2}{B^2}
\right),
\label{eq:cellwise_gradient_hoeffding}
\end{align}
where
$n_c^{(-k)}\geq n_{\mathrm{cell,tr}}$ by definition.
Set
\(
t_\delta
:=
B
\sqrt{
\frac{
\log\!\left(
32dKN_\omega|\mathcal N|/\delta
\right)
}{
2n_{\mathrm{cell,tr}}
}
}.
\)
Then \eqref{eq:cellwise_gradient_hoeffding} implies
\[
P\left(
\left|
\frac{1}{n_c^{(-k)}}
\sum_{i\in\mathcal D_{c,\mathrm{tr}}^{(k)}}
\partial_j\ell(\theta^\circ;X_i,Y_i)
-
\bE_{P_c}
\left[
\partial_j\ell(\theta^\circ;X,Y)
\right]
\right|
>t_\delta
\right)
\leq
\frac{\delta}
{16dKN_\omega|\mathcal N|}.
\]
For each fold $k$, cell $c$, coordinate
$j\in\{1,\ldots,d\}$, net point $\theta^\circ\in\mathcal N$, and candidate
allocation $\boldsymbol{\lambda}\in\Lambda_\omega$, define the bad event
\[
\mathcal E_{k,c,j,\theta^\circ,\boldsymbol{\lambda}}
:=
\left\{
\left|
\frac{1}{n_c^{(-k)}}
\sum_{i\in\mathcal D_{c,\mathrm{tr}}^{(k)}}
\partial_j\ell(\theta^\circ;X_i,Y_i)
-
\bE_{P_c}
\left[
\partial_j\ell(\theta^\circ;X,Y)
\right]
\right|
>t_\delta
\right\}.
\]
For every fixed choice of these indices, the previous Hoeffding bound gives an upper bound.
There are at most
\(
4dKN_\omega|\mathcal N|
\)
such indexed events, corresponding to the four cells, $d$ gradient coordinates,
$K$ folds, $N_\omega$ candidate allocations, and $|\mathcal N|$ covering-net
points. Therefore, by the union bound,
\begin{align*}
P\left(
\bigcup_{k,c,j,\theta^\circ,\boldsymbol{\lambda}}
\mathcal E_{k,c,j,\theta^\circ,\boldsymbol{\lambda}}
\right)
&\leq
\sum_{k,c,j,\theta^\circ,\boldsymbol{\lambda}}
P\left(
\mathcal E_{k,c,j,\theta^\circ,\boldsymbol{\lambda}}
\right)\leq
4dKN_\omega|\mathcal N|
\cdot
\frac{\delta}
{16dKN_\omega|\mathcal N|}
=
\frac{\delta}{4}.
\end{align*}
That is,
\[
P\left(
\max_{\substack{
1\leq k\leq K,\,
c,\,
1\leq j\leq d,\,
\theta^\circ\in\mathcal N
}}
\left|
\frac{1}{n_c^{(-k)}}
\sum_{i\in\mathcal D_{c,\mathrm{tr}}^{(k)}}
\partial_j\ell(\theta^\circ;X_i,Y_i)
-
\bE_{P_c}
\left[
\partial_j\ell(\theta^\circ;X,Y)
\right]
\right|
>t_\delta
\right)
\leq
\frac{\delta}{4}.
\]
On the complementary event, for every fold $k$ and every
$\theta^\circ\in\mathcal N$, each coordinate satisfies
\begin{align*}
\left|
\sum_c
\widehat p_c^{(-k)}
\left[
\frac{1}{n_c^{(-k)}}
\sum_{i\in\mathcal D_{c,\mathrm{tr}}^{(k)}}
\partial_j\ell(\theta^\circ;X_i,Y_i)
-
\bE_{P_c}
\partial_j\ell(\theta^\circ;X,Y)
\right]
\right|\leq
\sum_c
\widehat p_c^{(-k)}t_\delta
=
t_\delta,
\end{align*}
because
$\sum_c\widehat p_c^{(-k)}=1$.
Therefore, converting the coordinatewise bound to Euclidean norm gives
\begin{equation}
\label{eq:gradient_bound_on_net}
\sup_{\theta^\circ\in\mathcal N}
\left\|
\sum_c
\widehat p_c^{(-k)}
\left[
\frac{1}{n_c^{(-k)}}
\sum_{i\in\mathcal D_{c,\mathrm{tr}}^{(k)}}
\nabla\ell(\theta^\circ;X_i,Y_i)
-
\bE_{P_c}
\nabla\ell(\theta^\circ;X,Y)
\right]
\right\|_2
\leq
\sqrt d\,t_\delta.
\end{equation}
It remains to extend \eqref{eq:gradient_bound_on_net} from the finite covering net
$\mathcal N$ to every $\theta\in\Theta$. Fix an arbitrary $\theta\in\Theta$ and choose
$\theta^\circ\in\mathcal N$ such that
\(
\|\theta-\theta^\circ\|_2
\leq
\frac{D_\Theta}{n_{\mathrm{cell,tr}}}.
\)
For each cell $c$, add and subtract the corresponding empirical and population gradients
evaluated at $\theta^\circ$. Then
\begin{align*}
&\frac{1}{n_c^{(-k)}}
\sum_{i\in\mathcal D_{c,\mathrm{tr}}^{(k)}}
\nabla\ell(\theta;X_i,Y_i)
-
\bE_{P_c}
\left[
\nabla\ell(\theta;X,Y)
\right]
\\&=
\left\{
\frac{1}{n_c^{(-k)}}
\sum_{i\in\mathcal D_{c,\mathrm{tr}}^{(k)}}
\nabla\ell(\theta^\circ;X_i,Y_i)
-
\bE_{P_c}
\left[
\nabla\ell(\theta^\circ;X,Y)
\right]
\right\}
\\
&\qquad+
\frac{1}{n_c^{(-k)}}
\sum_{i\in\mathcal D_{c,\mathrm{tr}}^{(k)}}
\left[
\nabla\ell(\theta;X_i,Y_i)
-
\nabla\ell(\theta^\circ;X_i,Y_i)
\right]-
\bE_{P_c}
\left[
\nabla\ell(\theta;X,Y)
-
\nabla\ell(\theta^\circ;X,Y)
\right].
\end{align*}
By the Lipschitz assumption on the loss gradient,
\(
\left\|
\nabla\ell(\theta;X,Y)
-
\nabla\ell(\theta^\circ;X,Y)
\right\|_\infty
\leq
L_\nabla
\|\theta-\theta^\circ\|_2.
\)
Hence the empirical-gradient difference satisfies
\[
\left\|
\frac{1}{n_c^{(-k)}}
\sum_{i\in\mathcal D_{c,\mathrm{tr}}^{(k)}}
\left[
\nabla\ell(\theta;X_i,Y_i)
-
\nabla\ell(\theta^\circ;X_i,Y_i)
\right]
\right\|_\infty
\leq
L_\nabla
\|\theta-\theta^\circ\|_2,
\]
and, by Jensen's inequality, the same bound holds for the population-gradient
difference:
\[
\left\|
\bE_{P_c}
\left[
\nabla\ell(\theta;X,Y)
-
\nabla\ell(\theta^\circ;X,Y)
\right]
\right\|_\infty
\leq
L_\nabla
\|\theta-\theta^\circ\|_2.
\]
Therefore, by the triangle inequality,
\begin{align*}
&\left\|
\left[
\frac{1}{n_c^{(-k)}}
\sum_{i\in\mathcal D_{c,\mathrm{tr}}^{(k)}}
\nabla\ell(\theta;X_i,Y_i)
-
\bE_{P_c}\nabla\ell(\theta;X,Y)
\right]
\right.\\
&\qquad\qquad\qquad\left.-
\left[
\frac{1}{n_c^{(-k)}}
\sum_{i\in\mathcal D_{c,\mathrm{tr}}^{(k)}}
\nabla\ell(\theta^\circ;X_i,Y_i)
-
\bE_{P_c}\nabla\ell(\theta^\circ;X,Y)
\right]
\right\|_\infty
\\
&\qquad\qquad\qquad\leq
2L_\nabla
\|\theta-\theta^\circ\|_2
\leq
\frac{2L_\nabla D_\Theta}
{n_{\mathrm{cell,tr}}}.
\end{align*}
Since the cell weights satisfy
$\sum_c\widehat p_c^{(-k)}=1$, the same bound holds after taking the weighted sum over cells.
Finally, using
$\|v\|_2\leq\sqrt d\,\|v\|_\infty$, we obtain
\begin{align*}
&
\left\|
\sum_c
\widehat p_c^{(-k)}
\left[
\frac{1}{n_c^{(-k)}}
\sum_{i\in\mathcal D_{c,\mathrm{tr}}^{(k)}}
\nabla\ell(\theta;X_i,Y_i)
-
\bE_{P_c}
\left[
\nabla\ell(\theta;X,Y)
\right]
\right]
\right.
\nonumber\\
&\qquad\left.
-
\sum_c
\widehat p_c^{(-k)}
\left[
\frac{1}{n_c^{(-k)}}
\sum_{i\in\mathcal D_{c,\mathrm{tr}}^{(k)}}
\nabla\ell(\theta^\circ;X_i,Y_i)
-
\bE_{P_c}
\left[
\nabla\ell(\theta^\circ;X,Y)
\right]
\right]
\right\|_2
\leq
\frac{
2\sqrt d\,L_\nabla D_\Theta
}{
n_{\mathrm{cell,tr}}
}.
\end{align*}
Combining this approximation error with
\eqref{eq:gradient_bound_on_net}, and using
\(
|\mathcal N|
\leq
(3n_{\mathrm{cell,tr}})^d,
\)
we obtain, with probability at least $1-\delta/4$,
\begin{align}
\sup_{\theta\in\Theta}
\left\|
\sum_c
\widehat p_c^{(-k)}
\left[
\frac{1}{n_c^{(-k)}}
\sum_{i\in\mathcal D_{c,\mathrm{tr}}^{(k)}}
\nabla\ell(\theta;X_i,Y_i)
-
\bE_{P_c}
\left[
\nabla\ell(\theta;X,Y)
\right]
\right]
\right\|_2
\nonumber\\
\leq
\sqrt{
\frac{
dB^2
\log\!\left(
32dKN_\omega(3n_{\mathrm{cell,tr}})^d/\delta
\right)
}{
2n_{\mathrm{cell,tr}}
}
}
+
\frac{
2\sqrt d\,L_\nabla D_\Theta
}{
n_{\mathrm{cell,tr}}
}.
\label{eq:stratified_real_gradient_bound}
\end{align}
 
For the synthetic component, condition on the real training portion of fold $k$.
Conditional on these observations, the fitted generators
$P_{\mathrm{syn},c}^{(-k)}$, the fold-specific population distribution
$Q_{k,\boldsymbol{\lambda}}$, and hence
$\theta^*_{Q_{k,\boldsymbol{\lambda}}}$ are fixed.
The only remaining randomness comes from the newly generated synthetic observations.
Fix a fold $k$, an allocation
$\boldsymbol{\lambda}\in\Lambda_\omega$ such that
$m^{(-k)}(\boldsymbol{\lambda})>0$, and a gradient coordinate
$q\in\{1,\ldots,d\}$.
The
$m^{(-k)}(\boldsymbol{\lambda})$
synthetic observations are conditionally independent, with exactly
$m_c^{(-k)}(\boldsymbol{\lambda})$
observations generated from
$P_{\mathrm{syn},c}^{(-k)}$ for each cell $c$.
By the bounded-gradient assumption,
\(
\left|
\partial_q
\ell
\left(
\theta^*_{Q_{k,\boldsymbol{\lambda}}};
\widetilde X_j,\widetilde Y_j
\right)
\right|
\leq
\frac{B}{2},
\)
so every synthetic gradient coordinate takes values in an interval of length at most
$B$. Moreover, conditional on the real training sample, if the $j$th synthetic
observation is generated from cell $c$, then
\[
\bE
\left[
\left.
\partial_q
\ell
\left(
\theta^*_{Q_{k,\boldsymbol{\lambda}}};
\widetilde X_j,\widetilde Y_j
\right)
\right|
\mathcal D_{\mathrm{tr}}^{(k)}
\right]
=
\bE_{P_{\mathrm{syn},c}^{(-k)}}
\left[
\partial_q
\ell
\left(
\theta^*_{Q_{k,\boldsymbol{\lambda}}};
X,Y
\right)
\right].
\]
Therefore, averaging the conditional expectations over all generated observations gives
\begin{align*}
&
\frac{1}{
m^{(-k)}(\boldsymbol{\lambda})
}
\sum_{j=1}^{m^{(-k)}(\boldsymbol{\lambda})}
\bE
\left[
\left.
\partial_q
\ell
\left(
\theta^*_{Q_{k,\boldsymbol{\lambda}}};
\widetilde X_j,\widetilde Y_j
\right)
\right|
\mathcal D_{\mathrm{tr}}^{(k)}
\right]=
\sum_c
\frac{
m_c^{(-k)}(\boldsymbol{\lambda})
}{
m^{(-k)}(\boldsymbol{\lambda})
}
\bE_{P_{\mathrm{syn},c}^{(-k)}}
\left[
\partial_q
\ell
\left(
\theta^*_{Q_{k,\boldsymbol{\lambda}}};
X,Y
\right)
\right].
\end{align*}
Since the generated observations are conditionally independent and each corresponding
gradient coordinate is bounded in an interval of length at most $B$, Hoeffding's
inequality gives, for every $t>0$,
\begin{align*}
&P\Bigg(
\left.
\left|
\frac{1}{
m^{(-k)}(\boldsymbol{\lambda})
}
\sum_{j=1}^{m^{(-k)}(\boldsymbol{\lambda})}
\partial_q
\ell
\left(
\theta^*_{Q_{k,\boldsymbol{\lambda}}};
\widetilde X_j,\widetilde Y_j
\right)\right.\right.
\\
&\qquad\qquad\qquad\left.\left.-
\sum_c
\frac{
m_c^{(-k)}(\boldsymbol{\lambda})
}{
m^{(-k)}(\boldsymbol{\lambda})
}
\mathbb E_{P_{\mathrm{syn},c}^{(-k)}}
\left[
\partial_q
\ell
\left(
\theta^*_{Q_{k,\boldsymbol{\lambda}}};
X,Y
\right)
\right]
\right|
>t
\;\right|\;
\mathcal D_{\mathrm{tr}}^{(k)}
\Bigg)
\nonumber\\
&\qquad\qquad\qquad\leq
2
\exp
\left(
-\frac{
2m^{(-k)}(\boldsymbol{\lambda})t^2
}{
B^2
}
\right).
\end{align*}
Taking
\(
t
=
B
\sqrt{
\frac{
\log(8dKN_\omega/\delta)
}{
2m^{(-k)}(\boldsymbol{\lambda})
}
},
\)
we obtain
\begin{align*}
&
P\Bigg(
\left.
\left|
\frac{1}{
m^{(-k)}(\boldsymbol{\lambda})
}
\sum_{j=1}^{m^{(-k)}(\boldsymbol{\lambda})}
\partial_q
\ell
\left(
\theta^*_{Q_{k,\boldsymbol{\lambda}}};
\widetilde X_j,\widetilde Y_j
\right)
\right.
\right.
\\
&\qquad\left.
\left.
-
\sum_c
\frac{
m_c^{(-k)}(\boldsymbol{\lambda})
}{
m^{(-k)}(\boldsymbol{\lambda})
}
\mathbb E_{P_{\mathrm{syn},c}^{(-k)}}
\left[
\partial_q
\ell
\left(
\theta^*_{Q_{k,\boldsymbol{\lambda}}};
X,Y
\right)
\right]
\right|
>
B
\sqrt{
\frac{
\log(8dKN_\omega/\delta)
}{
2m^{(-k)}(\boldsymbol{\lambda})
}
}
\;\right|\;
\mathcal D_{\mathrm{tr}}^{(k)}
\Bigg)
\leq
\frac{\delta}{4dKN_\omega}.
\end{align*}
Taking a union bound over the $d$ gradient coordinates, the $K$ folds, and the
$N_\omega$ candidate allocations shows that, conditional on the real training samples,
the preceding bound holds simultaneously for all such indices with probability at least
$1-\delta/4$. Integrating over the real training samples gives the same 
probability bound.
On this event, all $d$ gradient coordinates satisfy the corresponding scalar bound.
Using
$\|v\|_2\leq\sqrt d\,\|v\|_\infty$, we obtain simultaneously over
$k$ and
$\boldsymbol{\lambda}\in\Lambda_\omega$, with probability at least $1-\frac{\delta}{4}$,
\begin{align}
\left\|
\frac{1}{
m^{(-k)}(\boldsymbol{\lambda})
}
\sum_{j=1}^{m^{(-k)}(\boldsymbol{\lambda})}
\nabla\ell
\left(
\theta^*_{Q_{k,\boldsymbol{\lambda}}};
\widetilde X_j,\widetilde Y_j
\right)
-
\sum_c
\frac{
m_c^{(-k)}(\boldsymbol{\lambda})
}{
m^{(-k)}(\boldsymbol{\lambda})
}
\mathbb E_{P_{\mathrm{syn},c}^{(-k)}}
\left[
\nabla\ell
\left(
\theta^*_{Q_{k,\boldsymbol{\lambda}}};
X,Y
\right)
\right]
\right\|_2
\nonumber\\
\leq
\sqrt{
\frac{
dB^2\log(8dKN_\omega/\delta)
}{
2\left(m^{(-k)}(\boldsymbol{\lambda})\vee 1\right)
}
}.
\label{eq:synthetic_gradient_bound}
\end{align}
When $m^{(-k)}(\boldsymbol{\lambda})=0$, there is no synthetic component in
the augmented empirical risk, and its contribution to the empirical-gradient
deviation is understood to be zero.

We now combine the real and synthetic empirical-gradient deviations according to their
weights in the augmented empirical risk. By definition,
\begin{align*}
\widehat L_{Q_{k,\boldsymbol{\lambda}}}(\theta)
=&
\frac{
n_{\mathrm{tr}}^{(k)}
}{
n_{\mathrm{tr}}^{(k)}
+
m^{(-k)}(\boldsymbol{\lambda})
}
\sum_c
\widehat p_c^{(-k)}
\frac{1}{n_c^{(-k)}}
\sum_{i\in\mathcal D_{c,\mathrm{tr}}^{(k)}}
\ell(\theta;X_i,Y_i)
\\
&\qquad\qquad+
\frac{
m^{(-k)}(\boldsymbol{\lambda})
}{
n_{\mathrm{tr}}^{(k)}
+
m^{(-k)}(\boldsymbol{\lambda})
}
\frac{1}{
m^{(-k)}(\boldsymbol{\lambda})
}
\sum_{j=1}^{m^{(-k)}(\boldsymbol{\lambda})}
\ell(\theta;\widetilde X_j,\widetilde Y_j),
\end{align*}
whereas the corresponding population risk can be written as
\begin{align*}
L_{Q_{k,\boldsymbol{\lambda}}}(\theta)
={}&
\frac{
n_{\mathrm{tr}}^{(k)}
}{
n_{\mathrm{tr}}^{(k)}
+
m^{(-k)}(\boldsymbol{\lambda})
}
\sum_c
\widehat p_c^{(-k)}
\bE_{P_c}
\left[
\ell(\theta;X,Y)
\right]
\\
&\qquad\qquad+
\frac{
m^{(-k)}(\boldsymbol{\lambda})
}{
n_{\mathrm{tr}}^{(k)}
+
m^{(-k)}(\boldsymbol{\lambda})
}
\sum_c
\frac{
m_c^{(-k)}(\boldsymbol{\lambda})
}{
m^{(-k)}(\boldsymbol{\lambda})
}
\bE_{P_{\mathrm{syn},c}^{(-k)}}
\left[
\ell(\theta;X,Y)
\right].
\end{align*}
Therefore, evaluating the gradients at
$\theta^*_{Q_{k,\boldsymbol{\lambda}}}$ and applying the triangle inequality gives
\begin{align*}
&
\left\|
\nabla\widehat L_{Q_{k,\boldsymbol{\lambda}}}
\left(
\theta^*_{Q_{k,\boldsymbol{\lambda}}}
\right)
-
\nabla L_{Q_{k,\boldsymbol{\lambda}}}
\left(
\theta^*_{Q_{k,\boldsymbol{\lambda}}}
\right)
\right\|_2
\\
&\leq
\frac{
n_{\mathrm{tr}}^{(k)}
}{
n_{\mathrm{tr}}^{(k)}
+
m^{(-k)}(\boldsymbol{\lambda})
}
\left\|
\sum_c
\widehat p_c^{(-k)}
\left[
\frac{1}{n_c^{(-k)}}
\sum_{i\in\mathcal D_{c,\mathrm{tr}}^{(k)}}
\nabla\ell
\left(
\theta^*_{Q_{k,\boldsymbol{\lambda}}};X_i,Y_i
\right)
-
\bE_{P_c}
\left[
\nabla\ell
\left(
\theta^*_{Q_{k,\boldsymbol{\lambda}}};X,Y
\right)
\right]
\right]
\right\|_2
\\
&+
\frac{
m^{(-k)}(\boldsymbol{\lambda})
}{
n_{\mathrm{tr}}^{(k)}
+
m^{(-k)}(\boldsymbol{\lambda})
}
\left\|
\frac{1}{
m^{(-k)}(\boldsymbol{\lambda})
}
\sum_{j=1}^{m^{(-k)}(\boldsymbol{\lambda})}
\nabla\ell
\left(
\theta^*_{Q_{k,\boldsymbol{\lambda}}};
\widetilde X_j,\widetilde Y_j
\right)
-
\sum_c
\frac{
m_c^{(-k)}(\boldsymbol{\lambda})
}{
m^{(-k)}(\boldsymbol{\lambda})
}
\bE_{P_{\mathrm{syn},c}^{(-k)}}
\left[
\nabla\ell
\left(
\theta^*_{Q_{k,\boldsymbol{\lambda}}};X,Y
\right)
\right]
\right\|_2.
\end{align*}
Using the real-data bound \eqref{eq:stratified_real_gradient_bound} and the synthetic-data
bound \eqref{eq:synthetic_gradient_bound}, together with
\(
\frac{
n_{\mathrm{tr}}^{(k)}
}{
n_{\mathrm{tr}}^{(k)}
+
m^{(-k)}(\boldsymbol{\lambda})
}
\leq1,
\)
we obtain
\begin{align}
\left\|
\widehat\theta_{Q_{k,\boldsymbol{\lambda}}}
-
\theta^*_{Q_{k,\boldsymbol{\lambda}}}
\right\|_2&\leq
\frac{1}{\mu}
\Bigg[
\sqrt{
\frac{
dB^2
\log\!\left(
32dKN_\omega(3n_{\mathrm{cell,tr}})^d/\delta
\right)
}{
2n_{\mathrm{cell,tr}}
}
}
+
\frac{
2\sqrt d\,L_\nabla D_\Theta
}{
n_{\mathrm{cell,tr}}
}
\nonumber\\
&\hspace{0.7in}
+
\sup_{\substack{
1\leq k\leq K\\
\boldsymbol{\lambda}\in\Lambda_\omega
}}
\frac{
m^{(-k)}(\boldsymbol{\lambda})
}{
n_{\mathrm{tr}}^{(k)}
+
m^{(-k)}(\boldsymbol{\lambda})
}
\sqrt{
\frac{
dB^2
\log(8dKN_\omega/\delta)
}{
2
\left(m^{(-k)}(\boldsymbol{\lambda})\vee 1\right)
}
}
\Bigg].
\label{eq:parameter_cv_bound}
\end{align}
With $E_{\mathrm{tr}}(\delta)$ as in Appendix~\ref{sec:full-statements}, \eqref{eq:parameter_cv_bound} states that $\|\widehat\theta_{Q_{k,\boldsymbol\lambda}}-\theta^*_{Q_{k,\boldsymbol\lambda}}\|_2\le E_{\mathrm{tr}}(\delta)/\mu$ for all $k$ and $\boldsymbol\lambda\in\Lambda_\omega$. The trained fold-specific classifier must, however, be compared with the full-sample target $h^*_{\overline P_{\boldsymbol\lambda}}$, so we first bound the distance between the two population minimizers $\theta^*_{Q_{k,\boldsymbol\lambda}}$ and $\theta^*_{\overline P_{\boldsymbol\lambda}}$, and then pass from parameters to the two criteria in a single step.
 
We next compare the fold-specific augmented population $Q_{k,\boldsymbol\lambda}$
with the continuous full-sample augmented population $\overline P_{\boldsymbol\lambda}$.
The two distributions can differ both through their mixture weights and
through the fold-specific synthetic generators. To separate these effects,
add and subtract the intermediate mixture
\[
\frac{
\displaystyle\sum_{c\in\mathcal C} n_c^{(-k)}P_c
+
\sum_{c\in\mathcal C} m_c^{(-k)}(\boldsymbol\lambda)P_{\mathrm{syn},c}
}{
n_{\mathrm{tr}}^{(k)}+m^{(-k)}(\boldsymbol\lambda)
},
\]
which uses the exact rounded fold-specific mixture weights but the
full-sample synthetic generators. By the triangle inequality for total
variation,
\[
\begin{aligned}
&d_{\mathrm{TV}}(Q_{k,\boldsymbol\lambda},\overline P_{\boldsymbol\lambda})
\le
\sum_{c\in\mathcal C}
\frac{m_c^{(-k)}(\boldsymbol\lambda)}
     {n_{\mathrm{tr}}^{(k)}+m^{(-k)}(\boldsymbol\lambda)}
d_{\mathrm{TV}}
\!\left(
P_{\mathrm{syn},c}^{(-k)},P_{\mathrm{syn},c}
\right)
\\
&+
\frac12\sum_{c\in\mathcal C}
\left|
\frac{n_c^{(-k)}}
     {n_{\mathrm{tr}}^{(k)}+m^{(-k)}(\boldsymbol\lambda)}
-
\frac{p_c}
     {1+r\sum_{c'\in\mathcal C}\lambda_{c'}}
\right|
+
\frac12\sum_{c\in\mathcal C}
\left|
\frac{m_c^{(-k)}(\boldsymbol\lambda)}
     {n_{\mathrm{tr}}^{(k)}+m^{(-k)}(\boldsymbol\lambda)}
-
\frac{r\lambda_c}
     {1+r\sum_{c'\in\mathcal C}\lambda_{c'}}
\right|.
\end{aligned}
\]
Moreover, by the triangle inequality through the oracle cell distribution
$P_c$,
\[
d_{\mathrm{TV}}
\!\left(
P_{\mathrm{syn},c}^{(-k)},P_{\mathrm{syn},c}
\right)
\le
d_{\mathrm{TV}}
\!\left(
P_{\mathrm{syn},c}^{(-k)},P_c
\right)
+
d_{\mathrm{TV}}
\!\left(
P_c,P_{\mathrm{syn},c}
\right)
\le
\epsilon_c^{(-k)}+\epsilon_c.
\]
Therefore, by the definition of $\epsilon_{\mathrm{syn,CV}}$,
\[
d_{\mathrm{TV}}(Q_{k,\boldsymbol\lambda},\overline P_{\boldsymbol\lambda})
\le
\epsilon_{\mathrm{syn,CV}}.
\]
We now turn this into a bound on the population minimizers. The argument is that of Step~2 in the proof of Theorem~\ref{thm:full-balance-convergence}. Since $\theta^*_{Q_{k,\boldsymbol\lambda}}$ is an interior minimizer of $L_{Q_{k,\boldsymbol\lambda}}$, we have $\nabla L_{Q_{k,\boldsymbol\lambda}}(\theta^*_{Q_{k,\boldsymbol\lambda}})=0$, and hence
\[
\nabla L_{\overline P_{\boldsymbol\lambda}}(\theta^*_{Q_{k,\boldsymbol\lambda}})
=
\bE_{\overline P_{\boldsymbol\lambda}}\bigl[\nabla\ell(\theta^*_{Q_{k,\boldsymbol\lambda}};X,Y)\bigr]
-
\bE_{Q_{k,\boldsymbol\lambda}}\bigl[\nabla\ell(\theta^*_{Q_{k,\boldsymbol\lambda}};X,Y)\bigr].
\]
Each coordinate of $\nabla\ell$ is bounded by $B/2$ under every real and synthetic cell distribution (Condition~4 of Assumption~\ref{assumption:improvement-bound-assumption}), and hence under both mixtures. Therefore each coordinate of the right-hand side is at most $B\,d_{\mathrm{TV}}(Q_{k,\boldsymbol\lambda},\overline P_{\boldsymbol\lambda})\le B\epsilon_{\mathrm{syn,CV}}$ in absolute value, and
\(
\|\nabla L_{\overline P_{\boldsymbol\lambda}}(\theta^*_{Q_{k,\boldsymbol\lambda}})\|_2\le\sqrt d\,B\,\epsilon_{\mathrm{syn,CV}}.
\)
Apply the pairwise curvature condition of Assumption~\ref{assumption:adaptive_search}(2) with $(Q,Q')=(\overline P_{\boldsymbol\lambda},Q_{k,\boldsymbol\lambda})$. Since $\nabla L_{\overline P_{\boldsymbol\lambda}}(\theta^*_{\overline P_{\boldsymbol\lambda}})=0$, the Cauchy--Schwarz inequality gives
\[
\mu\bigl\|\theta^*_{Q_{k,\boldsymbol\lambda}}-\theta^*_{\overline P_{\boldsymbol\lambda}}\bigr\|_2^2
\le
\bigl\langle
\nabla L_{\overline P_{\boldsymbol\lambda}}(\theta^*_{Q_{k,\boldsymbol\lambda}}),\,
\theta^*_{Q_{k,\boldsymbol\lambda}}-\theta^*_{\overline P_{\boldsymbol\lambda}}
\bigr\rangle
\le
\sqrt d\,B\,\epsilon_{\mathrm{syn,CV}}
\bigl\|\theta^*_{Q_{k,\boldsymbol\lambda}}-\theta^*_{\overline P_{\boldsymbol\lambda}}\bigr\|_2,
\]
that is,
\begin{equation}
\label{eq:fold_full_parameter}
\bigl\|\theta^*_{Q_{k,\boldsymbol\lambda}}-\theta^*_{\overline P_{\boldsymbol\lambda}}\bigr\|_2
\le
\frac{\sqrt d\,B}{\mu}\,\epsilon_{\mathrm{syn,CV}} .
\end{equation}
 
We now pass from parameters to the two criteria. This is the disagreement argument of Step~3 in the proof of Theorem~\ref{thm:full-balance-convergence}, applied to each fold and each grid point. Write
\[
e_{k,\boldsymbol\lambda}
:=
\sup_x\bigl|\widehat f_{Q_{k,\boldsymbol\lambda}}(x)-f^*_{\overline P_{\boldsymbol\lambda}}(x)\bigr|
\]
for the distance between the trained fold-specific score and the full-sample population score. By score Lipschitzness, \eqref{eq:parameter_cv_bound} and \eqref{eq:fold_full_parameter},
\begin{equation}
\label{eq:score_cv_bound}
e_{k,\boldsymbol\lambda}
\le
L\bigl\|\widehat\theta_{Q_{k,\boldsymbol\lambda}}-\theta^*_{Q_{k,\boldsymbol\lambda}}\bigr\|_2
+L\bigl\|\theta^*_{Q_{k,\boldsymbol\lambda}}-\theta^*_{\overline P_{\boldsymbol\lambda}}\bigr\|_2
\le
\frac{L}{\mu}\Bigl(E_{\mathrm{tr}}(\delta)+\sqrt d\,B\,\epsilon_{\mathrm{syn,CV}}\Bigr)
=e_{\mathrm{cv}}(\delta).
\end{equation}
If $\widehat h_{Q_{k,\boldsymbol\lambda}}(x)\neq h^*_{\overline P_{\boldsymbol\lambda}}(x)$, then $\widehat f_{Q_{k,\boldsymbol\lambda}}(x)$ and $f^*_{\overline P_{\boldsymbol\lambda}}(x)$ lie on opposite sides of $1/2$, so \[|f^*_{\overline P_{\boldsymbol\lambda}}(x)-\frac12|\le|\widehat f_{Q_{k,\boldsymbol\lambda}}(x)-f^*_{\overline P_{\boldsymbol\lambda}}(x)|\le e_{k,\boldsymbol\lambda}.\] That is,
\begin{equation}
\label{eq:cv_disagreement}
\bigl\{x:\widehat h_{Q_{k,\boldsymbol\lambda}}(x)\neq h^*_{\overline P_{\boldsymbol\lambda}}(x)\bigr\}
\subseteq
\bigl\{x:|f^*_{\overline P_{\boldsymbol\lambda}}(x)-\tfrac12|\le e_{k,\boldsymbol\lambda}\bigr\}.
\end{equation}
Conditional on the training-side data, $\widehat h_{Q_{k,\boldsymbol\lambda}}$ is a fixed classifier, and its true positive rates and prediction error refer to an independent observation from $P$. Two indicators $\mathbf 1\{h(x)=1\}$, or two losses $\mathbf 1\{h(x)\neq y\}$, can differ only where the two classifiers disagree. Hence, by \eqref{eq:cv_disagreement} and the margin condition of Assumption~\ref{assumption:adaptive_search}(2), for $s\in\{A,B\}$,
\begin{align*}
\bigl|\mathrm{TPR}_s(\widehat h_{Q_{k,\boldsymbol\lambda}})-\mathrm{TPR}_s(h^*_{\overline P_{\boldsymbol\lambda}})\bigr|
&\le
P_{s+}\bigl(|f^*_{\overline P_{\boldsymbol\lambda}}(X)-\tfrac12|\le e_{k,\boldsymbol\lambda}\bigr)
\le
C^\Lambda_s\,e_{k,\boldsymbol\lambda},
\\
\bigl|\mathrm{Err}_P(\widehat h_{Q_{k,\boldsymbol\lambda}})-\mathrm{Err}_P(h^*_{\overline P_{\boldsymbol\lambda}})\bigr|
&\le
P\bigl(|f^*_{\overline P_{\boldsymbol\lambda}}(X)-\tfrac12|\le e_{k,\boldsymbol\lambda}\bigr)
\le
C^\Lambda_P\,e_{k,\boldsymbol\lambda}.
\end{align*}
Using $\bigl||a|-|b|\bigr|\le|a-b|$ for the Equal Opportunity gap, together with \eqref{eq:score_cv_bound}, we obtain, on the event on which \eqref{eq:parameter_cv_bound} holds and uniformly over $k$ and $\boldsymbol\lambda\in\Lambda_\omega$,
\begin{gather}
\bigl|\Delta_{\mathrm{EO}}(\widehat h_{Q_{k,\boldsymbol\lambda}})-\Delta_{\mathrm{EO}}(h^*_{\overline P_{\boldsymbol\lambda}})\bigr|
\le
(C^\Lambda_A+C^\Lambda_B)\,e_{\mathrm{cv}}(\delta),
\label{eq:proof_empirical_fairness}
\\
\bigl|\mathrm{Err}_P(\widehat h_{Q_{k,\boldsymbol\lambda}})-\mathrm{Err}_P(h^*_{\overline P_{\boldsymbol\lambda}})\bigr|
\le
C^\Lambda_P\,e_{\mathrm{cv}}(\delta).
\label{eq:proof_empirical_accuracy}
\end{gather}
We next control the validation-fold estimation error. Fix
$s\in\{A,B\}$, a fold $k$, and
$\boldsymbol{\lambda}\in\Lambda_\omega$. 
Conditional on the real training data outside fold $k$ and the synthetic
observations generated to train $\widehat h_{Q_{k,\boldsymbol\lambda}}$, the classifier
$\widehat h_{Q_{k,\boldsymbol\lambda}}$ is fixed. The validation observations
remain independent of this training-side randomness. Moreover, by the stratified
construction of the folds, the $n_{s+}^{(k)}$ validation observations in cell
$(s,+)$ are i.i.d. from $P_{s+}$. Hence
\[
\widehat{\mathrm{TPR}}_s^{(k)}
=
\frac{1}{n_{s+}^{(k)}}
\sum_{i\in\mathcal D_{s+,\mathrm{val}}^{(k)}}
\mathbf 1
\left\{
\widehat h_{Q_{k,\boldsymbol{\lambda}}}(X_i)=1
\right\},
\]
where the dependence of
$\widehat{\mathrm{TPR}}_s^{(k)}$ on
$\boldsymbol{\lambda}$ is suppressed for notational simplicity. Conditional on the
training data  and synthetic training-side randomness, together denoted by $\mathcal D_{\mathrm{tr}}^{(k)}$ for brevity,
\[
\bE\left[
\left.
\widehat{\mathrm{TPR}}_s^{(k)}
\right|
\mathcal D_{\mathrm{tr}}^{(k)}
\right]
=
\mathrm{TPR}_s
\left(
\widehat h_{Q_{k,\boldsymbol{\lambda}}}
\right).
\]
Since the summands are conditionally independent Bernoulli random variables,
Hoeffding's inequality gives, for every $t>0$,
\[
P\left(
\left.
\left|
\widehat{\mathrm{TPR}}_s^{(k)}
-
\mathrm{TPR}_s
\left(
\widehat h_{Q_{k,\boldsymbol{\lambda}}}
\right)
\right|
>t
\;\right|\;
\mathcal D_{\mathrm{tr}}^{(k)}
\right)
\leq
2\exp\left(-2n_{s+}^{(k)}t^2\right).
\]
By definition,
$n_{s+}^{(k)}\geq n_{s+}^{\mathrm{val}}$. Therefore, taking
\(
t
=
\sqrt{
\frac{
\log(16KN_\omega/\delta)
}{
2n_{s+}^{\mathrm{val}}
}
}
\)
gives
\[
P\left(
\left.
\left|
\widehat{\mathrm{TPR}}_s^{(k)}
-
\mathrm{TPR}_s
\left(
\widehat h_{Q_{k,\boldsymbol{\lambda}}}
\right)
\right|
>
\sqrt{
\frac{
\log(16KN_\omega/\delta)
}{
2n_{s+}^{\mathrm{val}}
}
}
\;\right|\;
\mathcal D_{\mathrm{tr}}^{(k)}
\right)
\leq
\frac{\delta}{8KN_\omega}.
\]
There are $2KN_\omega$ choices of
$(s,k,\boldsymbol{\lambda})$. Thus, by a union bound, with probability at least
$1-\delta/4$, simultaneously for
$s\in\{A,B\}$, $1\leq k\leq K$, and
$\boldsymbol{\lambda}\in\Lambda_\omega$,
\begin{equation}
\left|
\widehat{\mathrm{TPR}}_s^{(k)}
-
\mathrm{TPR}_s
\left(
\widehat h_{Q_{k,\boldsymbol{\lambda}}}
\right)
\right|
\leq
\sqrt{
\frac{
\log(16KN_\omega/\delta)
}{
2n_{s+}^{\mathrm{val}}
}
},
\label{eq:validation_tpr_concentration}
\end{equation}
where integrating over the real and synthetic training-side randomness gives the
same probability bound. By the definition of
$\Delta_{\mathrm{EO}}$, equation~\eqref{eq:validation_tpr_concentration} implies
\begin{align}
\left|
\widehat\Delta_{\mathrm{EO}}^{(k)}
(\boldsymbol{\lambda})
-
\Delta_{\mathrm{EO}}
\left(
\widehat h_{Q_{k,\boldsymbol{\lambda}}}
\right)
\right|&\leq
\left|
\widehat{\mathrm{TPR}}_A^{(k)}
-
\mathrm{TPR}_A
\left(
\widehat h_{Q_{k,\boldsymbol{\lambda}}}
\right)
\right|
+
\left|
\widehat{\mathrm{TPR}}_B^{(k)}
-
\mathrm{TPR}_B
\left(
\widehat h_{Q_{k,\boldsymbol{\lambda}}}
\right)
\right|
\nonumber\\
&\qquad\leq
\sqrt{
\frac{
\log(16KN_\omega/\delta)
}{
2n_{A+}^{\mathrm{val}}
}
}
+
\sqrt{
\frac{
\log(16KN_\omega/\delta)
}{
2n_{B+}^{\mathrm{val}}
}
}.
\label{eq:validation_fairness_concentration}
\end{align}
We now combine the two sources of error. By the triangle inequality,
\begin{align}
\left|
\widehat\Delta_{\mathrm{EO}}^{(k)}(\boldsymbol{\lambda})
-
\Delta_{\mathrm{EO}}\left(h^*_{\overline P_{\boldsymbol{\lambda}}}\right)
\right|
\leq
\left|
\widehat\Delta_{\mathrm{EO}}^{(k)}(\boldsymbol{\lambda})
-
\Delta_{\mathrm{EO}}\left(\widehat h_{Q_{k,\boldsymbol{\lambda}}}\right)
\right|
+
\left|
\Delta_{\mathrm{EO}}\left(\widehat h_{Q_{k,\boldsymbol{\lambda}}}\right)
-
\Delta_{\mathrm{EO}}\left(h^*_{\overline P_{\boldsymbol{\lambda}}}\right)
\right|.
\label{eq:cv_fairness_decomposition}
\end{align}
The first term is the validation error, controlled by \eqref{eq:validation_fairness_concentration}. The second combines the training-side estimation error with the discrepancy between the fold-specific and full-sample augmented populations, and is controlled by \eqref{eq:proof_empirical_fairness}. By the definition of $\epsilon_{\mathrm{fair}}(\delta)$, uniformly over $k$ and $\boldsymbol{\lambda}\in\Lambda_\omega$,
\begin{align*}
\left|
\widehat\Delta_{\mathrm{EO}}^{(k)}(\boldsymbol{\lambda})
-
\Delta_{\mathrm{EO}}\left(h^*_{\overline P_{\boldsymbol{\lambda}}}\right)
\right|
\leq
\epsilon_{\mathrm{fair}}(\delta).
\end{align*}
Finally, since
\(
\widehat\Delta_{\mathrm{EO}}^{\mathrm{CV}}
(\boldsymbol{\lambda})
=
\frac1K
\sum_{k=1}^K
\widehat\Delta_{\mathrm{EO}}^{(k)}
(\boldsymbol{\lambda}),
\)
the triangle inequality gives
\begin{align*}
&
\left|
\widehat\Delta_{\mathrm{EO}}^{\mathrm{CV}}
(\boldsymbol{\lambda})
-
\Delta_{\mathrm{EO}}
\left(
h^*_{\overline P_{\boldsymbol{\lambda}}}
\right)
\right|
\leq
\frac1K
\sum_{k=1}^K
\left|
\widehat\Delta_{\mathrm{EO}}^{(k)}
(\boldsymbol{\lambda})
-
\Delta_{\mathrm{EO}}
\left(
h^*_{\overline P_{\boldsymbol{\lambda}}}
\right)
\right|
\leq
\epsilon_{\mathrm{fair}}(\delta).
\end{align*}
Taking the supremum over
$\boldsymbol{\lambda}\in\Lambda_\omega$ therefore yields
\begin{equation}
\sup_{\boldsymbol{\lambda}\in\Lambda_\omega}
\left|
\widehat\Delta_{\mathrm{EO}}^{\mathrm{CV}}
(\boldsymbol{\lambda})
-
\Delta_{\mathrm{EO}}
\left(
h^*_{\overline P_{\boldsymbol{\lambda}}}
\right)
\right|
\leq
\epsilon_{\mathrm{fair}}(\delta).
\label{eq:uniform_cv_fairness_derived}
\end{equation}

\paragraph{Step 2: the accuracy criterion and feasibility.}
We first control the validation-fold estimation error. 
As in the validation-TPR concentration argument, condition on the real
training data outside fold $k$ and the synthetic observations generated to
train $\widehat h_{Q_{k,\boldsymbol\lambda}}$. Then
$\widehat h_{Q_{k,\boldsymbol\lambda}}$ is fixed, while the observations in the
validation fold remain independent.
The fold-k error is
\[
\widehat{\mathrm{Err}}^{(k)}(\boldsymbol{\lambda})
:=
\frac{1}{n_{\mathrm{val}}^{(k)}}
\sum_{c}
\sum_{i\in\mathcal D_{c,\mathrm{val}}^{(k)}}
\mathbf 1
\left\{
\widehat h_{Q_{k,\boldsymbol{\lambda}}}(X_i)\neq Y_i
\right\}.
\]
Because the validation fold is stratified, an observation in cell $c$ is distributed
according to $P_c$. 
Hence, conditional on the real and synthetic training-side randomness,
\begin{align*}
\mathbb E\left[
\left.
\widehat{\mathrm{Err}}^{(k)}(\boldsymbol{\lambda})
\right|
\mathcal D_{\mathrm{tr}}^{(k)}
\right]
&=
\frac{1}{n_{\mathrm{val}}^{(k)}}
\sum_c
n_c^{(k)}
\mathrm{Err}_{P_c}
\left(
\widehat h_{Q_{k,\boldsymbol{\lambda}}}
\right)
=
\sum_c
\widehat p_c^{(k)}
\mathrm{Err}_{P_c}
\left(
\widehat h_{Q_{k,\boldsymbol{\lambda}}}
\right).
\end{align*}
The error indicators are independent Bernoulli random variables conditional
on the real and synthetic training-side randomness, although they need not
be identically distributed across cells.
Therefore, by Hoeffding's inequality and the same argument as the union bound argument for validation TPR step,
over the
$K$ folds and $N_\omega$ candidate allocations, we have
\[
\left|
\widehat{\mathrm{Err}}^{(k)}(\boldsymbol{\lambda})
-
\sum_c
\widehat p_c^{(k)}
\mathrm{Err}_{P_c}
\left(
\widehat h_{Q_{k,\boldsymbol{\lambda}}}
\right)
\right|
\leq
\sqrt{
\frac{
\log(8KN_\omega/\delta)
}{
2n_{\mathrm{val}}
}
},
\]
with probability at least $1-\delta/4$
simultaneously over $k$ and
$\boldsymbol{\lambda}\in\Lambda_\omega$. For the remainder of the proof, work on the intersection of the preceding
concentration events. By a union bound, these bounds hold simultaneously
with probability at least $1-\delta$.
Since $\mathrm{Err}_P(h)=\sum_c p_c\,\mathrm{Err}_{P_c}(h)$ and $\sum_c(\widehat p^{(k)}_c-p_c)=0$, for every classifier $h$,
\[
\Big|\sum_c\widehat p^{(k)}_c\,\mathrm{Err}_{P_c}(h)-\mathrm{Err}_P(h)\Big|
=\Big|\sum_c\big(\widehat p^{(k)}_c-p_c\big)\big(\mathrm{Err}_{P_c}(h)-\tfrac12\big)\Big|
\le\epsilon_{\mathrm{strat,val}},
\]
because $|\mathrm{Err}_{P_c}(h)-\frac12|\le\frac12$, where $\epsilon_{\mathrm{strat,val}}:=\frac12\max_{1\le k\le K}\sum_c|\widehat p^{(k)}_c-p_c|$.
 
The remaining comparison, between $\mathrm{Err}_P(\widehat h_{Q_{k,\boldsymbol\lambda}})$ and $\mathrm{Err}_P(h^*_{\overline P_{\boldsymbol\lambda}})$, is \eqref{eq:proof_empirical_accuracy} from Step~1. Combining the validation bound, the stratification bound and \eqref{eq:proof_empirical_accuracy} by the triangle inequality gives, for every $k$ and $\boldsymbol\lambda\in\Lambda_\omega$,
\[
\Bigl|
\widehat{\mathrm{Err}}^{(k)}(\boldsymbol\lambda)
-
\mathrm{Err}_P\bigl(h^*_{\overline P_{\boldsymbol\lambda}}\bigr)
\Bigr|
\le
\sqrt{\frac{\log(8KN_\omega/\delta)}{2n_{\mathrm{val}}}}
+\epsilon_{\mathrm{strat,val}}
+C^\Lambda_P\,e_{\mathrm{cv}}(\delta)
=
\frac{\epsilon_{\mathrm{acc}}(\delta)}{2}.
\]
Averaging over the $K$ folds, the same bound holds with $\widehat{\mathrm{Err}}^{(k)}$ replaced by $\widehat{\mathrm{Err}}^{\mathrm{CV}}$. Since $\mathbf 0\in\Lambda_\omega$ by Assumption~\ref{assumption:adaptive_search}(1) and $\overline P_{\mathbf 0}=P$, it holds in particular at $\boldsymbol\lambda=\mathbf 0$, where $h^*_{\overline P_{\mathbf 0}}=h^*_P$. Subtracting the bounds at $\boldsymbol\lambda$ and at $\mathbf 0$ yields
\begin{equation}
\label{eq:uniform_cv_accuracy_derived}
\begin{aligned}
\sup_{\boldsymbol{\lambda}\in\Lambda_\omega}
\Bigg|
&
\left[
\widehat{\mathrm{Err}}^{\mathrm{CV}}
(\boldsymbol{\lambda})
-
\widehat{\mathrm{Err}}^{\mathrm{CV}}(\mathbf0)
\right]
-
\left[
\mathrm{Err}_{P}
\left(
h^*_{\overline P_{\boldsymbol{\lambda}}}
\right)
-
\mathrm{Err}_{P}
\left(
h^*_{P}
\right)
\right]
\Bigg|
\leq
\epsilon_{\mathrm{acc}}(\delta).
\end{aligned}
\end{equation}
By previous analysis, every CV-feasible allocation is feasible for the relaxed population constraint
\(
\mathrm{Err}_{P}
\left(
h^*_{\overline P_{\boldsymbol{\lambda}}}
\right)
-
\mathrm{Err}_{P}
\left(
h^*_{P}
\right)
\leq
\xi_{\mathrm{acc}}
+
\epsilon_{\mathrm{acc}}(\delta),
\)
whereas every allocation satisfying
\(
\mathrm{Err}_{P}
\left(
h^*_{\overline P_{\boldsymbol{\lambda}}}
\right)
-
\mathrm{Err}_{P}
\left(
h^*_{P}
\right)
\leq
\xi_{\mathrm{acc}}
-
\epsilon_{\mathrm{acc}}(\delta)
\)
is necessarily CV feasible.

Let \[\Lambda_{\mathrm{acc}}^{+}:=\{\boldsymbol\lambda\in\Lambda:\mathrm{Err}_P(h^*_{\overline P_{\boldsymbol\lambda}})-\mathrm{Err}_P(h^*_P)\le\xi_{\mathrm{acc}}+\epsilon_{\mathrm{acc}}(\delta)\}\] be the relaxed population-feasible set. Since $\epsilon_{\mathrm{acc}}(\delta)\le\epsilon_0$ by \eqref{eq:acasa-conditions}, $\Lambda_{\mathrm{acc}}$ and $\Lambda_{\mathrm{acc}}^{+}$ are accuracy sublevel sets with tolerances in $[\xi_{\mathrm{acc}}-\epsilon_0,\xi_{\mathrm{acc}}+\epsilon_0]$. By Assumption~\ref{assumption:adaptive_search}(4), they are therefore closed and convex, and every point within distance $\rho_{\mathrm{acc}}$ of them lies in the neighborhood on which both population criteria are twice continuously differentiable and \eqref{eq:augmentation_hessian}, \eqref{eq:augmentation_gradient_bound} and the gradient and Hessian bounds for the accuracy criterion hold. Every Taylor expansion below is taken along a segment that either starts in $\Lambda_{\mathrm{acc}}^{+}$ and has length at most $R_\omega(\delta)\le\rho_{\mathrm{acc}}$, or joins two points of the convex set $\Lambda_{\mathrm{acc}}^{+}$, so these bounds hold along the whole segment.
We next quantify the distance from the relaxed population-feasible set to a grid point
that is guaranteed to remain feasible under cross-validation. Let
$\boldsymbol{\lambda}_0$ satisfy
\[
\mathrm{Err}_{P}
\left(
h^*_{\overline P_{\boldsymbol{\lambda}_0}}
\right)
-
\mathrm{Err}_{P}
\left(
h^*_{P}
\right)
\leq
\xi_{\mathrm{acc}}
+
\epsilon_{\mathrm{acc}}(\delta).
\]
If
\[
\mathrm{Err}_{P}
\left(
h^*_{\overline P_{\boldsymbol{\lambda}_0}}
\right)
-
\mathrm{Err}_{P}
\left(
h^*_{P}
\right)
\leq
\xi_{\mathrm{acc}}
-
\epsilon_{\mathrm{acc}}(\delta)
-
G_{\mathrm{acc}}\omega
-
\frac{H_{\mathrm{acc}}}{2}\omega^2,
\]
then choose any
$\boldsymbol{\lambda}_{0,\omega}\in\Lambda_\omega$ satisfying
\(
\left\|
\boldsymbol{\lambda}_{0,\omega}
-
\boldsymbol{\lambda}_0
\right\|_2
\leq
\omega.
\)
By Taylor's expansion and the accuracy gradient and Hessian bounds,
\[
\begin{aligned}
&
\mathrm{Err}_{P}
\left(
h^*_{\overline P_{\boldsymbol{\lambda}_{0,\omega}}}
\right)
-
\mathrm{Err}_{P}
\left(
h^*_{P}
\right)
\leq
\xi_{\mathrm{acc}}
-
\epsilon_{\mathrm{acc}}(\delta).
\end{aligned}
\]
Hence \eqref{eq:uniform_cv_accuracy_derived} guarantees that
$\boldsymbol{\lambda}_{0,\omega}$ is CV feasible.
Otherwise, the theorem condition
\[
\epsilon_{\mathrm{acc}}(\delta)
+
G_{\mathrm{acc}}\omega
+
\frac{H_{\mathrm{acc}}}{2}\omega^2
\leq
\epsilon_0
\]
ensures that $\boldsymbol{\lambda}_0$ lies in the boundary region covered by
assumption. By
\eqref{eq:accuracy_boundary_regular}, there exists a unit direction
$v_{\boldsymbol{\lambda}_0}$ satisfying the required inward directional derivative.
Set
\[
t
=
\frac{2}{m_{\mathrm{acc}}}
\left[
2\epsilon_{\mathrm{acc}}(\delta)
+
G_{\mathrm{acc}}\omega
+
\frac{H_{\mathrm{acc}}}{2}\omega^2
\right].
\]
Define, for
$0\leq s\leq t$,
\[
\phi(s)
:=
\mathrm{Err}_{P}
\left(
h^*_{\overline P_{\boldsymbol{\lambda}_0+
s v_{\boldsymbol{\lambda}_0}}}
\right)
-
\mathrm{Err}_{P}(h_P^*).
\]
Since $t\leq\rho_{\mathrm{acc}}$, the entire segment lies in the neighborhood on
which the accuracy criterion is twice continuously differentiable. Hence, by
Taylor's theorem, for some $\widetilde s\in(0,t)$,
\[
\phi(t)
=
\phi(0)
+
t\phi'(0)
+
\frac{t^2}{2}\phi''(\widetilde s).
\]
By the chain rule,
\[
\phi'(0)
=
\nabla_{\boldsymbol{\lambda}}
\mathrm{Err}_{P}
\left(
h^*_{\overline P_{\boldsymbol{\lambda}_0}}
\right)^\top
v_{\boldsymbol{\lambda}_0}
\leq
-m_{\mathrm{acc}}
\]
by \eqref{eq:accuracy_boundary_regular}. Moreover,
\[
\phi''(\widetilde s)
=
v_{\boldsymbol{\lambda}_0}^{\top}
\nabla_{\boldsymbol{\lambda}}^2
\mathrm{Err}_{P}
\left(
h^*_{\overline P_{\boldsymbol{\lambda}_0+
\widetilde s v_{\boldsymbol{\lambda}_0}}}
\right)
v_{\boldsymbol{\lambda}_0}
\leq
H_{\mathrm{acc}},
\]
where we used
$\|v_{\boldsymbol{\lambda}_0}\|_2=1$ and the uniform Hessian
operator-norm bound. Therefore,
\[
\phi(t)
\leq
\phi(0)
-
m_{\mathrm{acc}}t
+
\frac{H_{\mathrm{acc}}}{2}t^2.
\]
Since $\boldsymbol{\lambda}_0$ satisfies the relaxed population constraint,
$\phi(0)\leq\xi_{\mathrm{acc}}+\epsilon_{\mathrm{acc}}(\delta)$, and hence
\[
\mathrm{Err}_{P}
\left(
h^*_{\overline P_{\boldsymbol{\lambda}_0+
t v_{\boldsymbol{\lambda}_0}}}
\right)
-
\mathrm{Err}_{P}(h_P^*)
\leq
\xi_{\mathrm{acc}}
+
\epsilon_{\mathrm{acc}}(\delta)
-
m_{\mathrm{acc}}t
+
\frac{H_{\mathrm{acc}}}{2}t^2.
\]
Since the condition
\(
2\epsilon_{\mathrm{acc}}(\delta)
+
G_{\mathrm{acc}}\omega
+
\frac{H_{\mathrm{acc}}}{2}\omega^2
\leq
\frac{m_{\mathrm{acc}}^2}{2H_{\mathrm{acc}}},
\)
the last display implies
\[
\begin{aligned}
&
\mathrm{Err}_{P}
\left(
h^*_{\overline P_{\boldsymbol{\lambda}_0
+t v_{\boldsymbol{\lambda}_0}}}
\right)
-
\mathrm{Err}_{P}
\left(
h^*_{P}
\right)\leq
\xi_{\mathrm{acc}}
-
\epsilon_{\mathrm{acc}}(\delta)
-
G_{\mathrm{acc}}\omega
-
\frac{H_{\mathrm{acc}}}{2}\omega^2.
\end{aligned}
\]
Now choose
$\boldsymbol{\lambda}_{0,\omega}\in\Lambda_\omega$ with
\(
\left\|
\boldsymbol{\lambda}_{0,\omega}
-
\left(
\boldsymbol{\lambda}_0
+
t v_{\boldsymbol{\lambda}_0}
\right)
\right\|_2
\leq
\omega.
\)
A second Taylor expansion gives
\[
\begin{aligned}
&
\mathrm{Err}_{P}
\left(
h^*_{\overline P_{\boldsymbol{\lambda}_{0,\omega}}}
\right)
-
\mathrm{Err}_{P}
\left(
h^*_{P}
\right)
\leq
\xi_{\mathrm{acc}}
-
\epsilon_{\mathrm{acc}}(\delta),
\end{aligned}
\]
so \eqref{eq:uniform_cv_accuracy_derived} again implies that
$\boldsymbol{\lambda}_{0,\omega}$ is CV feasible. In either case, every point satisfying
the relaxed population constraint has a CV-feasible master-grid point satisfying
\begin{equation}
\label{eq:relaxed_to_cv_grid_distance}
\left\|
\boldsymbol{\lambda}_{0,\omega}
-
\boldsymbol{\lambda}_0
\right\|_2
\leq
\omega
+
\frac{2}{m_{\mathrm{acc}}}
\left[
2\epsilon_{\mathrm{acc}}(\delta)
+
G_{\mathrm{acc}}\omega
+
\frac{H_{\mathrm{acc}}}{2}\omega^2
\right].
\end{equation}
This establishes simultaneously the required metric-regularity correction and the
preservation of CV feasibility after discretization.

Similarly, let $\boldsymbol\lambda_0\in\Lambda_{\mathrm{acc}}^{+}$ and define $\phi$ as above. If $\boldsymbol\lambda_0\notin\Lambda_{\mathrm{acc}}$, then $0<\phi(0)-\xi_{\mathrm{acc}}\le\epsilon_{\mathrm{acc}}(\delta)\le\epsilon_0$, so \eqref{eq:accuracy_boundary_regular} applies at $\boldsymbol\lambda_0$. Take $t=2\epsilon_{\mathrm{acc}}(\delta)/m_{\mathrm{acc}}$. Since $R_\omega(\delta)\ge4\epsilon_{\mathrm{acc}}(\delta)/m_{\mathrm{acc}}$, we have $t\le R_\omega(\delta)\le\rho_{\mathrm{acc}}$, so $\boldsymbol\lambda_0+t\boldsymbol v_{\boldsymbol\lambda_0}\in\Lambda$, and the expansion used for \eqref{eq:relaxed_to_cv_grid_distance} gives
\[
\phi(t)\le\xi_{\mathrm{acc}}+\epsilon_{\mathrm{acc}}(\delta)-m_{\mathrm{acc}}t+\frac{H_{\mathrm{acc}}}{2}t^2
=\xi_{\mathrm{acc}}-\epsilon_{\mathrm{acc}}(\delta)+\frac{2H_{\mathrm{acc}}\,\epsilon_{\mathrm{acc}}(\delta)^2}{m_{\mathrm{acc}}^2}
\le\xi_{\mathrm{acc}},
\]
where the last inequality uses $\epsilon_{\mathrm{acc}}(\delta)\le m_{\mathrm{acc}}^2/(4H_{\mathrm{acc}})$, which follows from the second condition in \eqref{eq:acasa-conditions}. Hence $\boldsymbol\lambda_0+t\boldsymbol v_{\boldsymbol\lambda_0}\in\Lambda_{\mathrm{acc}}$, and in either case
\begin{equation}
\label{eq:relaxed_to_exact_distance}
\inf_{\boldsymbol{\lambda}\in\Lambda_{\mathrm{acc}}}
\left\|
\boldsymbol{\lambda}
-
\boldsymbol{\lambda}_0
\right\|_2
\leq
\frac{
2\epsilon_{\mathrm{acc}}(\delta)
}{
m_{\mathrm{acc}}
}.
\end{equation}

\paragraph{Step 3: projected-gradient control at termination.}
Recall the relaxed population-feasible set $\Lambda_{\mathrm{acc}}^{+}$ from Step~2, which is closed and convex. For any nonempty closed convex set
$\mathcal K\subset\mathbb R^4$, let
\(
\Pi_{\mathcal K}(z)
:=
\arg\min_{\boldsymbol{\lambda}\in\mathcal K}
\left\|
\boldsymbol{\lambda}-z
\right\|_2
\)
denote the Euclidean projection of $z$ onto $\mathcal K$. The projection is unique
because $\mathcal K$ is closed and convex.
At the final allocation $\widehat{\boldsymbol{\lambda}}$, define the population
projected-gradient point by
\begin{equation}
\boldsymbol{\lambda}^{\mathrm{pg}}
:=
\Pi_{\Lambda_{\mathrm{acc}}^{+}}
\left[
\widehat{\boldsymbol{\lambda}}
-
\frac{1}{H_{\mathrm{aug}}}
\left.
\nabla_{\boldsymbol{\lambda}}
\Delta_{\mathrm{EO}}
\left(
h^*_{\overline P_{\boldsymbol{\lambda}}}
\right)
\right|_{\boldsymbol{\lambda}
=
\widehat{\boldsymbol{\lambda}}}
\right].
\label{eq:population_projected_gradient}
\end{equation}
Because $\widehat{\boldsymbol{\lambda}}$ is CV feasible,
\eqref{eq:uniform_cv_accuracy_derived} implies
$\widehat{\boldsymbol{\lambda}}\in\Lambda_{\mathrm{acc}}^{+}$.
Since Euclidean projection onto a closed convex set is nonexpansive and
$\widehat{\boldsymbol{\lambda}}
=
\Pi_{\Lambda_{\mathrm{acc}}^{+}}
(\widehat{\boldsymbol{\lambda}})$,
\[
\begin{aligned}
\left\|
\boldsymbol{\lambda}^{\mathrm{pg}}
-
\widehat{\boldsymbol{\lambda}}
\right\|_2
&\leq
\frac{1}{H_{\mathrm{aug}}}
\left\|
\nabla_{\boldsymbol{\lambda}}
\Delta_{\mathrm{EO}}
\left(
h^*_{\overline P_{\widehat{\boldsymbol{\lambda}}}}
\right)
\right\|_2
\leq
\frac{G_{\mathrm{aug}}}{H_{\mathrm{aug}}}.
\end{aligned}
\]
Applying \eqref{eq:relaxed_to_cv_grid_distance} to
$\boldsymbol{\lambda}^{\mathrm{pg}}$ yields a point
$\widetilde{\boldsymbol{\lambda}}^{\mathrm{pg}}\in\Lambda_\omega$ that is guaranteed to
be CV feasible and satisfies
\begin{equation}
\label{eq:projected_candidate_distance}
\left\|
\widetilde{\boldsymbol{\lambda}}^{\mathrm{pg}}
-
\boldsymbol{\lambda}^{\mathrm{pg}}
\right\|_2
\leq
\omega
+
\frac{2}{m_{\mathrm{acc}}}
\left[
2\epsilon_{\mathrm{acc}}(\delta)
+
G_{\mathrm{acc}}\omega
+
\frac{H_{\mathrm{acc}}}{2}\omega^2
\right].
\end{equation}
Therefore,
\[
\begin{aligned}
\left\|
\widetilde{\boldsymbol{\lambda}}^{\mathrm{pg}}
-
\widehat{\boldsymbol{\lambda}}
\right\|_2
\leq\;&
\frac{G_{\mathrm{aug}}}{H_{\mathrm{aug}}}
+
\omega
+
\frac{2}{m_{\mathrm{acc}}}
\left[
2\epsilon_{\mathrm{acc}}(\delta)
+
G_{\mathrm{acc}}\omega
+
\frac{H_{\mathrm{acc}}}{2}\omega^2
\right]
\leq
\rho_{\mathrm{saf}},
\end{aligned}
\]
where the final inequality is the safeguard-radius condition in the theorem.
Consequently, although the algorithm never computes
$\boldsymbol{\lambda}^{\mathrm{pg}}$, the empirical safeguard necessarily evaluates the
CV-feasible grid point
$\widetilde{\boldsymbol{\lambda}}^{\mathrm{pg}}$ before terminating.
 
The algorithm terminates only if no CV-feasible safeguard point improves the
cross-validated fairness criterion by more than $\xi_{\mathrm{fair}}$. Also, $\widetilde{\boldsymbol{\lambda}}^{\mathrm{pg}}$ is one of the candidate safeguard points, so
\[
\widehat\Delta_{\mathrm{EO}}^{\mathrm{CV}}
\left(
\widehat{\boldsymbol{\lambda}}
\right)
-
\widehat\Delta_{\mathrm{EO}}^{\mathrm{CV}}
\left(
\widetilde{\boldsymbol{\lambda}}^{\mathrm{pg}}
\right)
\leq
\xi_{\mathrm{fair}}.
\]
Moreover, by \eqref{eq:uniform_cv_fairness_derived},
\[
\Delta_{\mathrm{EO}}
\left(
h^*_{\overline P_{\widehat{\boldsymbol{\lambda}}}}
\right)
\leq
\widehat\Delta_{\mathrm{EO}}^{\mathrm{CV}}
\left(
\widehat{\boldsymbol{\lambda}}
\right)
+
\epsilon_{\mathrm{fair}}(\delta),
\qquad
\Delta_{\mathrm{EO}}
\left(
h^*_{\overline P_{\widetilde{\boldsymbol{\lambda}}^{\mathrm{pg}}}}
\right)
\geq
\widehat\Delta_{\mathrm{EO}}^{\mathrm{CV}}
\left(
\widetilde{\boldsymbol{\lambda}}^{\mathrm{pg}}
\right)
-
\epsilon_{\mathrm{fair}}(\delta).
\]
Therefore,
\begin{align}
&
\Delta_{\mathrm{EO}}
\left(
h^*_{\overline P_{\widehat{\boldsymbol{\lambda}}}}
\right)
-
\Delta_{\mathrm{EO}}
\left(
h^*_{\overline P_{\widetilde{\boldsymbol{\lambda}}^{\mathrm{pg}}}}
\right)\leq
\xi_{\mathrm{fair}}
+
2\epsilon_{\mathrm{fair}}(\delta).
\label{eq:projected_empirical_stop}
\end{align}
By Taylor's theorem around
$\boldsymbol{\lambda}^{\mathrm{pg}}$, for some
$\boldsymbol{\lambda}^{\circ}$ on the line segment joining
$\boldsymbol{\lambda}^{\mathrm{pg}}$ and
$\widetilde{\boldsymbol{\lambda}}^{\mathrm{pg}}$,
\begin{align*}
\Delta_{\mathrm{EO}}
\left(
h^*_{\overline P_{\widetilde{\boldsymbol{\lambda}}^{\mathrm{pg}}}}
\right)
-
\Delta_{\mathrm{EO}}
\left(
h^*_{\overline P_{\boldsymbol{\lambda}^{\mathrm{pg}}}}
\right)
=
\nabla_{\boldsymbol{\lambda}}
\Delta_{\mathrm{EO}}
\left(
h^*_{\overline P_{\boldsymbol{\lambda}^{\mathrm{pg}}}}
\right)^\top
\left(
\widetilde{\boldsymbol{\lambda}}^{\mathrm{pg}}
-
\boldsymbol{\lambda}^{\mathrm{pg}}
\right)
\\+
\frac12
\left(
\widetilde{\boldsymbol{\lambda}}^{\mathrm{pg}}
-
\boldsymbol{\lambda}^{\mathrm{pg}}
\right)^\top
\nabla_{\boldsymbol{\lambda}}^2
\Delta_{\mathrm{EO}}
\left(
h^*_{\overline P_{\boldsymbol{\lambda}^{\circ}}}
\right)
\left(
\widetilde{\boldsymbol{\lambda}}^{\mathrm{pg}}
-
\boldsymbol{\lambda}^{\mathrm{pg}}
\right).
\end{align*}
By the gradient bound and the upper Hessian bound in the assumption,
\begin{align*}
&
\Delta_{\mathrm{EO}}
\left(
h^*_{\overline P_{\widetilde{\boldsymbol{\lambda}}^{\mathrm{pg}}}}
\right)
-
\Delta_{\mathrm{EO}}
\left(
h^*_{\overline P_{\boldsymbol{\lambda}^{\mathrm{pg}}}}
\right)
\leq
G_{\mathrm{aug}}
\left\|
\widetilde{\boldsymbol{\lambda}}^{\mathrm{pg}}
-
\boldsymbol{\lambda}^{\mathrm{pg}}
\right\|_2
+
\frac{H_{\mathrm{aug}}}{2}
\left\|
\widetilde{\boldsymbol{\lambda}}^{\mathrm{pg}}
-
\boldsymbol{\lambda}^{\mathrm{pg}}
\right\|_2^2.
\end{align*}
Using the distance bound in equation \eqref{eq:projected_candidate_distance},
we obtain
\begin{align}
\Delta_{\mathrm{EO}}
\left(
h^*_{\overline P_{\widetilde{\boldsymbol{\lambda}}^{\mathrm{pg}}}}
\right)
-
\Delta_{\mathrm{EO}}
\left(
h^*_{\overline P_{\boldsymbol{\lambda}^{\mathrm{pg}}}}
\right)
\leq
G_{\mathrm{aug}}
\Bigg\{
\omega
+
\frac{2}{m_{\mathrm{acc}}}
\left[
2\epsilon_{\mathrm{acc}}(\delta)
+
G_{\mathrm{acc}}\omega
+
\frac{H_{\mathrm{acc}}}{2}\omega^2
\right]
\Bigg\}
\nonumber\\
+
\frac{H_{\mathrm{aug}}}{2}
\Bigg\{
\omega
+
\frac{2}{m_{\mathrm{acc}}}
\left[
2\epsilon_{\mathrm{acc}}(\delta)
+
G_{\mathrm{acc}}\omega
+
\frac{H_{\mathrm{acc}}}{2}\omega^2
\right]
\Bigg\}^{2}.
\label{eq:projected_discretization}
\end{align}
Combining
\eqref{eq:projected_empirical_stop} and
\eqref{eq:projected_discretization} yields
\begin{align}
\nonumber\Delta_{\mathrm{EO}}
\left(
h^*_{\overline P_{\widehat{\boldsymbol{\lambda}}}}
\right)
-
\Delta_{\mathrm{EO}}
\left(
h^*_{\overline P_{\boldsymbol{\lambda}^{\mathrm{pg}}}}
\right)
\leq
\xi_{\mathrm{fair}}
+
2\epsilon_{\mathrm{fair}}(\delta)+
G_{\mathrm{aug}}
\Bigg\{
\omega
+
\frac{2}{m_{\mathrm{acc}}}
\left[
2\epsilon_{\mathrm{acc}}(\delta)
+
G_{\mathrm{acc}}\omega
+
\frac{H_{\mathrm{acc}}}{2}\omega^2
\right]
\Bigg\}
\\+
\frac{H_{\mathrm{aug}}}{2}
\Bigg\{
\omega
+
\frac{2}{m_{\mathrm{acc}}}
\left[
2\epsilon_{\mathrm{acc}}(\delta)
+
G_{\mathrm{acc}}\omega
+
\frac{H_{\mathrm{acc}}}{2}\omega^2
\right]
\Bigg\}^2.
\label{eq:projected_population_improvement}
\end{align}
By the definition of Euclidean projection onto a closed convex set, if
$y=\Pi_{\mathcal K}(z)$, then
\[
\langle z-y,u-y\rangle\leq0
\qquad
\text{for every }u\in\mathcal K.
\]
Since
$\widehat{\boldsymbol{\lambda}}\in\Lambda_{\mathrm{acc}}^{+}$,
we may apply this property with
$u=\widehat{\boldsymbol{\lambda}}$. Hence
\begin{align*}
0
&\geq
\Bigg\langle
\widehat{\boldsymbol{\lambda}}
-
\frac{1}{H_{\mathrm{aug}}}
\left.
\nabla_{\boldsymbol{\lambda}}
\Delta_{\mathrm{EO}}
\left(
h^*_{\overline P_{\boldsymbol{\lambda}}}
\right)
\right|_{\boldsymbol{\lambda}
=
\widehat{\boldsymbol{\lambda}}}
-
\boldsymbol{\lambda}^{\mathrm{pg}},
\,
\widehat{\boldsymbol{\lambda}}
-
\boldsymbol{\lambda}^{\mathrm{pg}}
\Bigg\rangle
\\
&=
\left\|
\boldsymbol{\lambda}^{\mathrm{pg}}
-
\widehat{\boldsymbol{\lambda}}
\right\|_2^2
+
\frac{1}{H_{\mathrm{aug}}}
\left.
\nabla_{\boldsymbol{\lambda}}
\Delta_{\mathrm{EO}}
\left(
h^*_{\overline P_{\boldsymbol{\lambda}}}
\right)^\top
\right|_{\boldsymbol{\lambda}
=
\widehat{\boldsymbol{\lambda}}}
\left(
\boldsymbol{\lambda}^{\mathrm{pg}}
-
\widehat{\boldsymbol{\lambda}}
\right).
\end{align*}
Therefore,
\[
\left.
\nabla_{\boldsymbol{\lambda}}
\Delta_{\mathrm{EO}}
\left(
h^*_{\overline P_{\boldsymbol{\lambda}}}
\right)^\top
\right|_{\boldsymbol{\lambda}
=
\widehat{\boldsymbol{\lambda}}}
\left(
\boldsymbol{\lambda}^{\mathrm{pg}}
-
\widehat{\boldsymbol{\lambda}}
\right)
\leq
-
H_{\mathrm{aug}}
\left\|
\boldsymbol{\lambda}^{\mathrm{pg}}
-
\widehat{\boldsymbol{\lambda}}
\right\|_2^2.
\]
Next, by Taylor's theorem and the upper Hessian bound in assumption,
\begin{align*}
\Delta_{\mathrm{EO}}
\left(
h^*_{\overline P_{\boldsymbol{\lambda}^{\mathrm{pg}}}}
\right)
&\leq
\Delta_{\mathrm{EO}}
\left(
h^*_{\overline P_{\widehat{\boldsymbol{\lambda}}}}
\right)
+
\left.
\nabla_{\boldsymbol{\lambda}}
\Delta_{\mathrm{EO}}
\left(
h^*_{\overline P_{\boldsymbol{\lambda}}}
\right)^\top
\right|_{\boldsymbol{\lambda}
=
\widehat{\boldsymbol{\lambda}}}
\left(
\boldsymbol{\lambda}^{\mathrm{pg}}
-
\widehat{\boldsymbol{\lambda}}
\right)
+
\frac{H_{\mathrm{aug}}}{2}
\left\|
\boldsymbol{\lambda}^{\mathrm{pg}}
-
\widehat{\boldsymbol{\lambda}}
\right\|_2^2
\\
&\leq
\Delta_{\mathrm{EO}}
\left(
h^*_{\overline P_{\widehat{\boldsymbol{\lambda}}}}
\right)
-
\frac{H_{\mathrm{aug}}}{2}
\left\|
\boldsymbol{\lambda}^{\mathrm{pg}}
-
\widehat{\boldsymbol{\lambda}}
\right\|_2^2.
\end{align*}
Combining this with
\eqref{eq:projected_population_improvement} gives
\begin{align}
\left\|
\boldsymbol{\lambda}^{\mathrm{pg}}
-
\widehat{\boldsymbol{\lambda}}
\right\|_2^2
\leq\;&
\frac{2}{H_{\mathrm{aug}}}
\Bigg[
\xi_{\mathrm{fair}}
+
2\epsilon_{\mathrm{fair}}(\delta)
+
G_{\mathrm{aug}}
\Bigg\{
\omega
+
\frac{2}{m_{\mathrm{acc}}}
\left[
2\epsilon_{\mathrm{acc}}(\delta)
+
G_{\mathrm{acc}}\omega
+
\frac{H_{\mathrm{acc}}}{2}\omega^2
\right]
\Bigg\}
\nonumber\\
&+
\frac{H_{\mathrm{aug}}}{2}
\Bigg\{
\omega
+
\frac{2}{m_{\mathrm{acc}}}
\left[
2\epsilon_{\mathrm{acc}}(\delta)
+
G_{\mathrm{acc}}\omega
+
\frac{H_{\mathrm{acc}}}{2}\omega^2
\right]
\Bigg\}^2
\Bigg].
\label{eq:projected_mapping_bound}
\end{align}

\paragraph{Step 4: distance to the oracle allocation.}
We first bound the distance from $\widehat{\boldsymbol{\lambda}}$ to the minimizer of the population fairness criterion over the relaxed feasible set, and then compare that minimizer with $\boldsymbol{\lambda}^*$.

The set $\Lambda_{\mathrm{acc}}^{+}$ is nonempty (it contains $\mathbf 0$, since $\overline P_{\mathbf 0}=P$), closed, and contained in the bounded set $\Lambda$, hence compact. The population fairness criterion $\boldsymbol\lambda\mapsto\Delta_{\mathrm{EO}}(h^*_{\overline P_{\boldsymbol\lambda}})$ is continuous on it, so it attains its minimum over $\Lambda_{\mathrm{acc}}^{+}$. Let $\boldsymbol\lambda^{*,+}$ denote a minimizer.
By the Hessian bounds assumption, the population fairness criterion is
$\mu_{\mathrm{aug}}$-strongly convex and has
$H_{\mathrm{aug}}$-Lipschitz gradient on the relevant convex neighborhood.
Hence its minimizer over the convex set
$\Lambda_{\mathrm{acc}}^{+}$ is unique.
Because $\boldsymbol{\lambda}^{*,+}$ minimizes the population fairness criterion
over the convex set $\Lambda_{\mathrm{acc}}^{+}$, take any
$\boldsymbol{\lambda}\in\Lambda_{\mathrm{acc}}^{+}$. By convexity,
\[
\boldsymbol{\lambda}^{*,+}
+t\left(
\boldsymbol{\lambda}-\boldsymbol{\lambda}^{*,+}
\right)
\in\Lambda_{\mathrm{acc}}^{+},
\qquad 0\leq t\leq1.
\]
Define
\[
\phi(t)
:=
\Delta_{\mathrm{EO}}
\left(
h^*_{\overline P_{
\boldsymbol{\lambda}^{*,+}
+t(\boldsymbol{\lambda}-\boldsymbol{\lambda}^{*,+})
}}
\right).
\]
Since $t=0$ minimizes $\phi(t)$ over $[0,1]$, its right derivative satisfies
$\phi'(0)\geq0$. By the chain rule,
\[
\phi'(0)
=\nabla_{\boldsymbol{\lambda}}
\Delta_{\mathrm{EO}}
\left(
h^*_{\overline P_{\boldsymbol{\lambda}^{*,+}}}
\right)^\top
\left(
\boldsymbol{\lambda}
-
\boldsymbol{\lambda}^{*,+}
\right)
\geq0,
\qquad
\text{for every }
\boldsymbol{\lambda}\in\Lambda_{\mathrm{acc}}^{+}.
\]
Recall that, for a nonempty closed convex set $\mathcal K$,
\[
y=\Pi_{\mathcal K}(z)
\quad\Longleftrightarrow\quad
\langle z-y,u-y\rangle\leq0
\qquad
\text{for every }u\in\mathcal K.
\]
Set
\(
\mathcal K=\Lambda_{\mathrm{acc}}^{+}
\), \(
y=\boldsymbol{\lambda}^{*,+},
\)
and
\(
z
=
\boldsymbol{\lambda}^{*,+}
-
\frac{1}{H_{\mathrm{aug}}}
\nabla_{\boldsymbol{\lambda}}
\Delta_{\mathrm{EO}}
\left(
h^*_{\overline P_{\boldsymbol{\lambda}^{*,+}}}
\right).
\)
Since $\boldsymbol{\lambda}^{*,+}\in\Lambda_{\mathrm{acc}}^{+}$, and for every
$\boldsymbol{\lambda}\in\Lambda_{\mathrm{acc}}^{+}$,
\begin{align*}
\left\langle
z-\boldsymbol{\lambda}^{*,+},
\boldsymbol{\lambda}-\boldsymbol{\lambda}^{*,+}
\right\rangle
&=
-\frac{1}{H_{\mathrm{aug}}}
\nabla_{\boldsymbol{\lambda}}
\Delta_{\mathrm{EO}}
\left(
h^*_{\overline P_{\boldsymbol{\lambda}^{*,+}}}
\right)^\top
\left(
\boldsymbol{\lambda}
-
\boldsymbol{\lambda}^{*,+}
\right)
\leq0,
\end{align*}
the projection characterization implies
\begin{equation}
\label{eq:kkt_fixed_point}
\boldsymbol{\lambda}^{*,+}
=
\Pi_{\Lambda_{\mathrm{acc}}^{+}}
\left[
\boldsymbol{\lambda}^{*,+}
-
\frac{1}{H_{\mathrm{aug}}}
\nabla_{\boldsymbol{\lambda}}
\Delta_{\mathrm{EO}}
\left(
h^*_{\overline P_{\boldsymbol{\lambda}^{*,+}}}
\right)
\right].
\end{equation}
By the Hessian bounds assumption, the ordinary
gradient step is contractive on $\Lambda_{\mathrm{acc}}^{+}$.
To see this explicitly, take any
$\boldsymbol{\lambda}_1,\boldsymbol{\lambda}_2
\in\Lambda_{\mathrm{acc}}^{+}$. Since the relevant neighborhood is convex,
the line segment joining these two points remains in the region. By the fundamental theorem of calculus,
\begin{align*}
&
\nabla_{\boldsymbol{\lambda}}
\Delta_{\mathrm{EO}}
\left(
h^*_{\overline P_{\boldsymbol{\lambda}_1}}
\right)
-
\nabla_{\boldsymbol{\lambda}}
\Delta_{\mathrm{EO}}
\left(
h^*_{\overline P_{\boldsymbol{\lambda}_2}}
\right)
=
\left[
\int_0^1
\nabla_{\boldsymbol{\lambda}}^2
\Delta_{\mathrm{EO}}
\left(
h^*_{\overline P_{
\boldsymbol{\lambda}_2+
s(\boldsymbol{\lambda}_1-\boldsymbol{\lambda}_2)
}}
\right)
\,ds
\right]
(\boldsymbol{\lambda}_1-\boldsymbol{\lambda}_2).
\end{align*}
The Hessian bounds imply
\[
\mu_{\mathrm{aug}}I_4
\preceq
\int_0^1
\nabla_{\boldsymbol{\lambda}}^2
\Delta_{\mathrm{EO}}
\left(
h^*_{\overline P_{
\boldsymbol{\lambda}_2+
s(\boldsymbol{\lambda}_1-\boldsymbol{\lambda}_2)
}}
\right)
\,ds
\preceq
H_{\mathrm{aug}}I_4.
\]
Hence
\begin{align*}
&
\Bigg\|
\left[
\boldsymbol{\lambda}_1
-
\frac{1}{H_{\mathrm{aug}}}
\nabla_{\boldsymbol{\lambda}}
\Delta_{\mathrm{EO}}
\left(
h^*_{\overline P_{\boldsymbol{\lambda}_1}}
\right)
\right]
-
\left[
\boldsymbol{\lambda}_2
-
\frac{1}{H_{\mathrm{aug}}}
\nabla_{\boldsymbol{\lambda}}
\Delta_{\mathrm{EO}}
\left(
h^*_{\overline P_{\boldsymbol{\lambda}_2}}
\right)
\right]
\Bigg\|_2
\leq
\left(
1-\frac{\mu_{\mathrm{aug}}}{H_{\mathrm{aug}}}
\right)
\left\|
\boldsymbol{\lambda}_1-\boldsymbol{\lambda}_2
\right\|_2.
\end{align*}
Moreover, Euclidean projection onto a closed convex set satisfies
\(
\left\|
\Pi_{\Lambda_{\mathrm{acc}}^{+}}(x)
-
\Pi_{\Lambda_{\mathrm{acc}}^{+}}(y)
\right\|_2
\leq
\|x-y\|_2.
\)
Applying this inequality to the projected-gradient definitions at
$\widehat{\boldsymbol{\lambda}}$ and
$\boldsymbol{\lambda}^{*,+}$, and using
equation
\eqref{eq:kkt_fixed_point}, 
\begin{align*}
\left\|
\boldsymbol{\lambda}^{\mathrm{pg}}
-
\boldsymbol{\lambda}^{*,+}
\right\|_2
&\leq
\Bigg\|
\left[
\widehat{\boldsymbol{\lambda}}
-
\frac{1}{H_{\mathrm{aug}}}
\nabla_{\boldsymbol{\lambda}}
\Delta_{\mathrm{EO}}
\left(
h^*_{\overline P_{\widehat{\boldsymbol{\lambda}}}}
\right)
\right]
-
\left[
\boldsymbol{\lambda}^{*,+}
-
\frac{1}{H_{\mathrm{aug}}}
\nabla_{\boldsymbol{\lambda}}
\Delta_{\mathrm{EO}}
\left(
h^*_{\overline P_{\boldsymbol{\lambda}^{*,+}}}
\right)
\right]
\Bigg\|_2
\\
&\leq
\left(
1-\frac{\mu_{\mathrm{aug}}}{H_{\mathrm{aug}}}
\right)
\left\|
\widehat{\boldsymbol{\lambda}}
-
\boldsymbol{\lambda}^{*,+}
\right\|_2.
\end{align*}
Therefore,
\[
\begin{aligned}
\left\|
\widehat{\boldsymbol{\lambda}}
-
\boldsymbol{\lambda}^{*,+}
\right\|_2
&\leq
\left\|
\widehat{\boldsymbol{\lambda}}
-
\boldsymbol{\lambda}^{\mathrm{pg}}
\right\|_2
+
\left\|
\boldsymbol{\lambda}^{\mathrm{pg}}
-
\boldsymbol{\lambda}^{*,+}
\right\|_2
\leq
\left\|
\widehat{\boldsymbol{\lambda}}
-
\boldsymbol{\lambda}^{\mathrm{pg}}
\right\|_2
+
\left(
1-
\frac{\mu_{\mathrm{aug}}}{H_{\mathrm{aug}}}
\right)
\left\|
\widehat{\boldsymbol{\lambda}}
-
\boldsymbol{\lambda}^{*,+}
\right\|_2.
\end{aligned}
\]
Rearranging,
\[
\left\|
\widehat{\boldsymbol{\lambda}}
-
\boldsymbol{\lambda}^{*,+}
\right\|_2
\leq
\frac{
H_{\mathrm{aug}}
}{
\mu_{\mathrm{aug}}
}
\left\|
\widehat{\boldsymbol{\lambda}}
-
\boldsymbol{\lambda}^{\mathrm{pg}}
\right\|_2.
\]
Using equation \eqref{eq:projected_mapping_bound} and the definition of $R_\omega(\delta)$ in Appendix~\ref{sec:appendixadaptivedetail},
\begin{align}
\big\|\widehat{\boldsymbol\lambda}-\boldsymbol\lambda^{*,+}\big\|_2
\le\left\{\frac{2H_{\mathrm{aug}}}{\mu_{\mathrm{aug}}^2}\Big[\xi_{\mathrm{fair}}+2\epsilon_{\mathrm{fair}}(\delta)+G_{\mathrm{aug}}R_\omega(\delta)+\frac{H_{\mathrm{aug}}}{2}R_\omega(\delta)^2\Big]\right\}^{1/2}.
\label{eq:distance_to_relaxed_oracle}
\end{align}
It remains to compare
$\boldsymbol{\lambda}^{*,+}$ with the exact oracle
$\boldsymbol{\lambda}^*$.
If
$\boldsymbol{\lambda}^{*,+}\in\Lambda_{\mathrm{acc}}$,
then uniqueness gives
$\boldsymbol{\lambda}^{*,+}=\boldsymbol{\lambda}^*$.
Otherwise, by \eqref{eq:relaxed_to_exact_distance}, there exists
$\widetilde{\boldsymbol{\lambda}}\in\Lambda_{\mathrm{acc}}$ satisfying
\(
\left\|
\widetilde{\boldsymbol{\lambda}}
-
\boldsymbol{\lambda}^{*,+}
\right\|_2
\leq
\frac{
2\epsilon_{\mathrm{acc}}(\delta)
}{
m_{\mathrm{acc}}
}.
\)
By the fundamental theorem of calculus along the line segment joining
$\boldsymbol{\lambda}^{*,+}$ and
$\widetilde{\boldsymbol{\lambda}}$, together with the gradient bound in assumption,
\begin{align*}
\Delta_{\mathrm{EO}}
\left(
h^*_{\overline P_{\widetilde{\boldsymbol{\lambda}}}}
\right)
-
\Delta_{\mathrm{EO}}
\left(
h^*_{\overline P_{\boldsymbol{\lambda}^{*,+}}}
\right)
&=
\int_0^1
\nabla_{\boldsymbol{\lambda}}
\Delta_{\mathrm{EO}}
\left(
h^*_{\overline P_{
\boldsymbol{\lambda}^{*,+}
+t(\widetilde{\boldsymbol{\lambda}}
-\boldsymbol{\lambda}^{*,+})
}}
\right)^\top
\left(
\widetilde{\boldsymbol{\lambda}}
-
\boldsymbol{\lambda}^{*,+}
\right)
\,dt
\\
&\leq
G_{\mathrm{aug}}
\left\|
\widetilde{\boldsymbol{\lambda}}
-
\boldsymbol{\lambda}^{*,+}
\right\|_2
\leq
\frac{2G_{\mathrm{aug}}}{m_{\mathrm{acc}}}
\epsilon_{\mathrm{acc}}(\delta).
\end{align*}
Since $\boldsymbol{\lambda}^*$ minimizes the population fairness criterion over
$\Lambda_{\mathrm{acc}}$ and
$\widetilde{\boldsymbol{\lambda}}\in\Lambda_{\mathrm{acc}}$,
\[
\Delta_{\mathrm{EO}}
\left(
h^*_{\overline P_{\boldsymbol{\lambda}^*}}
\right)
\leq
\Delta_{\mathrm{EO}}
\left(
h^*_{\overline P_{\widetilde{\boldsymbol{\lambda}}}}
\right).
\]
Therefore,
\[
\Delta_{\mathrm{EO}}
\left(
h^*_{\overline P_{\boldsymbol{\lambda}^*}}
\right)
-
\Delta_{\mathrm{EO}}
\left(
h^*_{\overline P_{\boldsymbol{\lambda}^{*,+}}}
\right)
\leq
\frac{2G_{\mathrm{aug}}}{m_{\mathrm{acc}}}
\epsilon_{\mathrm{acc}}(\delta).
\]
Moreover, because
$\Lambda_{\mathrm{acc}}\subseteq\Lambda_{\mathrm{acc}}^{+}$,
$\boldsymbol{\lambda}^*$ is feasible for the relaxed problem, and optimality of
$\boldsymbol{\lambda}^{*,+}$ gives
\[
\Delta_{\mathrm{EO}}
\left(
h^*_{\overline P_{\boldsymbol{\lambda}^{*,+}}}
\right)
\leq
\Delta_{\mathrm{EO}}
\left(
h^*_{\overline P_{\boldsymbol{\lambda}^{*}}}
\right).
\]
Consequently,
\[
0
\leq
\Delta_{\mathrm{EO}}
\left(
h^*_{\overline P_{\boldsymbol{\lambda}^{*}}}
\right)
-
\Delta_{\mathrm{EO}}
\left(
h^*_{\overline P_{\boldsymbol{\lambda}^{*,+}}}
\right)
\leq
\frac{2G_{\mathrm{aug}}}{m_{\mathrm{acc}}}
\epsilon_{\mathrm{acc}}(\delta).
\]
Because
$\Lambda_{\mathrm{acc}}\subseteq\Lambda_{\mathrm{acc}}^{+}$,
the exact oracle
$\boldsymbol{\lambda}^*$ is also feasible for the relaxed problem.
By Taylor's theorem applied around
$\boldsymbol{\lambda}^{*,+}$, for some point
$\widetilde{\boldsymbol{\lambda}}$ on the line segment joining
$\boldsymbol{\lambda}^{*,+}$ and $\boldsymbol{\lambda}^*$,
\begin{align*}
\Delta_{\mathrm{EO}}
\left(
h^*_{\overline P_{\boldsymbol{\lambda}^*}}
\right)
-
\Delta_{\mathrm{EO}}
\left(
h^*_{\overline P_{\boldsymbol{\lambda}^{*,+}}}
\right)
&=
\nabla_{\boldsymbol{\lambda}}
\Delta_{\mathrm{EO}}
\left(
h^*_{\overline P_{\boldsymbol{\lambda}^{*,+}}}
\right)^\top
\left(
\boldsymbol{\lambda}^*
-
\boldsymbol{\lambda}^{*,+}
\right)
\\
&\quad+
\frac{1}{2}
\left(
\boldsymbol{\lambda}^*
-
\boldsymbol{\lambda}^{*,+}
\right)^\top
\nabla_{\boldsymbol{\lambda}}^2
\Delta_{\mathrm{EO}}
\left(
h^*_{\overline P_{\widetilde{\boldsymbol{\lambda}}}}
\right)
\left(
\boldsymbol{\lambda}^*
-
\boldsymbol{\lambda}^{*,+}
\right).
\end{align*}
By the assumed lower Hessian bound,
\begin{align*}
&
\Delta_{\mathrm{EO}}
\left(
h^*_{\overline P_{\boldsymbol{\lambda}^*}}
\right)
-
\Delta_{\mathrm{EO}}
\left(
h^*_{\overline P_{\boldsymbol{\lambda}^{*,+}}}
\right)
\geq
\nabla_{\boldsymbol{\lambda}}
\Delta_{\mathrm{EO}}
\left(
h^*_{\overline P_{\boldsymbol{\lambda}^{*,+}}}
\right)^\top
\left(
\boldsymbol{\lambda}^*
-
\boldsymbol{\lambda}^{*,+}
\right)
+
\frac{\mu_{\mathrm{aug}}}{2}
\left\|
\boldsymbol{\lambda}^*
-
\boldsymbol{\lambda}^{*,+}
\right\|_2^2.
\end{align*}
Since
$\boldsymbol{\lambda}^{*,+}$ minimizes the fairness criterion over the convex
set $\Lambda_{\mathrm{acc}}^{+}$, its first-order optimality condition gives
\[
\nabla_{\boldsymbol{\lambda}}
\Delta_{\mathrm{EO}}
\left(
h^*_{\overline P_{\boldsymbol{\lambda}^{*,+}}}
\right)^\top
\left(
\boldsymbol{\lambda}
-
\boldsymbol{\lambda}^{*,+}
\right)
\geq0,
\qquad
\text{for every }
\boldsymbol{\lambda}\in\Lambda_{\mathrm{acc}}^{+}.
\]
Because
$\boldsymbol{\lambda}^*\in\Lambda_{\mathrm{acc}}
\subseteq\Lambda_{\mathrm{acc}}^{+}$,
taking
$\boldsymbol{\lambda}=\boldsymbol{\lambda}^*$ gives
\[
\nabla_{\boldsymbol{\lambda}}
\Delta_{\mathrm{EO}}
\left(
h^*_{\overline P_{\boldsymbol{\lambda}^{*,+}}}
\right)^\top
\left(
\boldsymbol{\lambda}^*
-
\boldsymbol{\lambda}^{*,+}
\right)
\geq0.
\]
Consequently,
\[
\Delta_{\mathrm{EO}}
\left(
h^*_{\overline P_{\boldsymbol{\lambda}^*}}
\right)
-
\Delta_{\mathrm{EO}}
\left(
h^*_{\overline P_{\boldsymbol{\lambda}^{*,+}}}
\right)
\geq
\frac{\mu_{\mathrm{aug}}}{2}
\left\|
\boldsymbol{\lambda}^{*,+}
-
\boldsymbol{\lambda}^*
\right\|_2^2.
\]
Therefore,
\begin{equation}
\label{eq:relaxed_exact_oracle_distance}
\left\|
\boldsymbol{\lambda}^{*,+}
-
\boldsymbol{\lambda}^*
\right\|_2
\leq
\sqrt{
\frac{
4G_{\mathrm{aug}}
}{
\mu_{\mathrm{aug}}m_{\mathrm{acc}}
}
\epsilon_{\mathrm{acc}}(\delta)
}.
\end{equation}
Finally, the triangle inequality together with
\eqref{eq:distance_to_relaxed_oracle} and
\eqref{eq:relaxed_exact_oracle_distance} yields
\eqref{eq:allocation_convergence}.
\end{proof}

\begin{proof}
[Proof of Theorem~\ref{thm:adaptive_near_optimality}] 
By Theorem~\ref{thm:adaptive_near_optimality-full}, it suffices to bound the right-hand side of \eqref{eq:allocation_convergence} by $C_\lambda\sqrt{\xi_{\mathrm{fair}}+\epsilon_{\mathrm{fair}}(\delta)+\epsilon_{\mathrm{acc}}(\delta)+\omega}$.
By \eqref{eq:acasa-conditions}, $\omega\le R_\omega(\delta)\le\rho_{\mathrm{acc}}$. Hence $G_{\mathrm{aug}}R_\omega+\frac{H_{\mathrm{aug}}}{2}R_\omega^2\le g\,R_\omega$ with \[g:=G_{\mathrm{aug}}+H_{\mathrm{aug}}\rho_{\mathrm{acc}}/2, \quad\text{and}\quad D_\omega\le(G_{\mathrm{acc}}+H_{\mathrm{acc}}\rho_{\mathrm{acc}}/2)\omega,\] so $R_\omega(\delta)\le a_1\omega+a_2\epsilon_{\mathrm{acc}}(\delta)$ with \[a_1:=1+\frac{2}{m_{\mathrm{acc}}}(G_{\mathrm{acc}}+H_{\mathrm{acc}}\rho_{\mathrm{acc}}/2),\quad \text{and} \quad a_2:=4/m_{\mathrm{acc}}.\] With \[K_0:=\max\{2,\,g\,a_1,\,g\,a_2\},\quad \text{and}\quad \Sigma:=\xi_{\mathrm{fair}}+\epsilon_{\mathrm{fair}}(\delta)+\epsilon_{\mathrm{acc}}(\delta)+\omega,\] the first term of \eqref{eq:allocation_convergence} is at most $\{(2H_{\mathrm{aug}}/\mu_{\mathrm{aug}}^2)K_0\Sigma\}^{1/2}$ and the second at most $\{(4G_{\mathrm{aug}}/(\mu_{\mathrm{aug}}m_{\mathrm{acc}}))\Sigma\}^{1/2}$. Using $\sqrt x+\sqrt y\le\sqrt{2(x+y)}$, the main-text bound holds with
\[
C_{\boldsymbol\lambda}:=\left\{2\left(\frac{2H_{\mathrm{aug}}K_0}{\mu_{\mathrm{aug}}^2}+\frac{4G_{\mathrm{aug}}}{\mu_{\mathrm{aug}}m_{\mathrm{acc}}}\right)\right\}^{1/2}.
\]
We thus finish the proof.
\end{proof}

\begin{proof}[Proof of Proposition~\ref{prop:excess-eo}]
Steps~1--3 of the proof of Theorem~\ref{thm:adaptive_near_optimality-full} use neither the lower bound in \eqref{eq:augmentation_hessian} nor the uniqueness of $\boldsymbol\lambda^*$, so \eqref{eq:uniform_cv_accuracy_derived} and \eqref{eq:projected_mapping_bound} hold on the same event. Since $\widehat{\boldsymbol\lambda}$ is CV feasible, \eqref{eq:uniform_cv_accuracy_derived} gives the first claim, that is, $\widehat{\boldsymbol\lambda}\in\Lambda_{\mathrm{acc}}^{+}$ with $\Lambda_{\mathrm{acc}}^{+}$ as in Step~2. As $\epsilon_{\mathrm{acc}}(\delta)\le\epsilon_0$ by \eqref{eq:acasa-conditions}, $\widehat{\boldsymbol\lambda}$ also lies in the set on which \eqref{eq:weighted-convexity} is required. Since $\boldsymbol\lambda^{\mathrm{pg}}$ in \eqref{eq:population_projected_gradient} is a projection onto $\Lambda_{\mathrm{acc}}^{+}$ and $\boldsymbol\lambda^*\in\Lambda_{\mathrm{acc}}\subseteq\Lambda_{\mathrm{acc}}^{+}$, the projection characterization used in Step~3 gives
\[
\nabla_{\boldsymbol\lambda}\Delta_{\mathrm{EO}}\bigl(h^*_{\overline P_{\widehat{\boldsymbol\lambda}}}\bigr)^\top(\boldsymbol\lambda^{\mathrm{pg}}-\boldsymbol\lambda^*)
\le H_{\mathrm{aug}}(\widehat{\boldsymbol\lambda}-\boldsymbol\lambda^{\mathrm{pg}})^\top(\boldsymbol\lambda^{\mathrm{pg}}-\boldsymbol\lambda^*)
\le H_{\mathrm{aug}}\,\mathrm{diam}(\Lambda)\,\|\widehat{\boldsymbol\lambda}-\boldsymbol\lambda^{\mathrm{pg}}\|_2,
\]
and \eqref{eq:augmentation_gradient_bound} gives $\nabla_{\boldsymbol\lambda}\Delta_{\mathrm{EO}}(h^*_{\overline P_{\widehat{\boldsymbol\lambda}}})^\top(\widehat{\boldsymbol\lambda}-\boldsymbol\lambda^{\mathrm{pg}})\le G_{\mathrm{aug}}\|\widehat{\boldsymbol\lambda}-\boldsymbol\lambda^{\mathrm{pg}}\|_2$. By \eqref{eq:projected_mapping_bound} and the definition of $R_\omega(\delta)$ in Appendix~\ref{sec:appendixadaptivedetail},
\[
\|\widehat{\boldsymbol\lambda}-\boldsymbol\lambda^{\mathrm{pg}}\|_2\le\Bigl\{\frac{2}{H_{\mathrm{aug}}}\Bigl[\xi_{\mathrm{fair}}+2\epsilon_{\mathrm{fair}}(\delta)+G_{\mathrm{aug}}R_\omega(\delta)+\frac{H_{\mathrm{aug}}}{2}R_\omega(\delta)^2\Bigr]\Bigr\}^{1/2}.
\]
Adding the two bounds gives
\begin{align*}
&\nabla_{\boldsymbol\lambda}\Delta_{\mathrm{EO}}\bigl(h^*_{\overline P_{\widehat{\boldsymbol\lambda}}}\bigr)^\top(\widehat{\boldsymbol\lambda}-\boldsymbol\lambda^*)\\
&\qquad\le\Bigl(\mathrm{diam}(\Lambda)+\frac{G_{\mathrm{aug}}}{H_{\mathrm{aug}}}\Bigr)\Bigl\{2H_{\mathrm{aug}}\Bigl[\xi_{\mathrm{fair}}+2\epsilon_{\mathrm{fair}}(\delta)+G_{\mathrm{aug}}R_\omega(\delta)+\frac{H_{\mathrm{aug}}}{2}R_\omega(\delta)^2\Bigr]\Bigr\}^{1/2},
\end{align*}
and applying \eqref{eq:weighted-convexity} at $\boldsymbol\lambda=\widehat{\boldsymbol\lambda}$ proves the second claim. The last statement follows because $\Delta_{\mathrm{EO}}(h^*_{\overline P_{\boldsymbol\lambda^*}})\le\Delta_{\mathrm{EO}}(h^*_{\overline P_{\boldsymbol\lambda^{\mathrm{fb}}}})$ when $\boldsymbol\lambda^{\mathrm{fb}}\in\Lambda_{\mathrm{acc}}$.
\end{proof}

\subsection{Proof of Technical Lemmas}

\begin{proof}[Proof of Lemma \ref{lemma:log-odds-shift}]
The cell distribution $P_{B+}$ puts all its mass on $\{Y=1\}$ and has density $p(x,B,1)/p_{B+}$ with respect to $\nu$. Therefore, under $P_\lambda$ the joint densities of $(X,Y)$ are
\[
p_\lambda(x,1)=(1-\lambda)p(x,1)+\lambda\,\frac{p(x,B,1)}{p_{B+}},
\qquad
p_\lambda(x,0)=(1-\lambda)p(x,0),
\]
and $\Psi_{P_\lambda}(x)=\log\bigl(p_\lambda(x,1)/p_\lambda(x,0)\bigr)$, $\Psi_P(x)=\log\bigl(p(x,1)/p(x,0)\bigr)$. Subtracting, the factor $p(x,0)$ cancels and
\[
\Psi_{P_\lambda}(x)-\Psi_P(x)=\log\frac{p_\lambda(x,1)}{(1-\lambda)p(x,1)}
=\log\left(1+\frac{\lambda}{(1-\lambda)p_{B+}}\cdot\frac{p(x,B,1)}{p(x,1)}\right),
\]
and $p(x,B,1)/p(x,1)=\omega_B(x)$. For the constant, note that $\bar\pi=\pi_A-(1-\alpha)(\pi_A-\pi_B)$, so $1-\bar\pi-(1-\alpha)(\pi_A-\pi_B)=1-\pi_A$ and $1-\lambda_1^*=(1-\pi_A)/(1-\bar\pi)$. Hence
\[
\frac{\lambda_1^*}{(1-\lambda_1^*)\,p_{B+}}
=\frac{(1-\alpha)(\pi_A-\pi_B)}{(1-\pi_A)\,(1-\alpha)\pi_B}
=\frac{\pi_A-\pi_B}{\pi_B(1-\pi_A)}
=\frac{\pi_A(1-\pi_B)}{\pi_B(1-\pi_A)}-1=\mathrm{OR}-1 .
\]
Since $\omega_B(x)\in[0,1]$ and $\mathrm{OR}>1$, the shift lies in $[0,\log\mathrm{OR}]$. A nonnegative shift gives $\Psi_{\widetilde P}\ge\Psi_P$ and so $\mathcal X_P^+\subseteq\mathcal X_{\widetilde P}^+$.
\end{proof}

\begin{proof}[Proof of Lemma \ref{lemma:group-shift}]
The form of $q_w$ follows from $Q_w(X,Y\mid S=s)=P(X,Y\mid S=s)$. Write $a_y:=p(x,A,y)/\alpha$ and $b_y:=p(x,B,y)/(1-\alpha)$, so $q_w(x,y)=(1-w)a_y+wb_y$ and
\[
\partial_w\Psi_{Q_w}(x)=\frac{b_1-a_1}{q_w(x,1)}-\frac{b_0-a_0}{q_w(x,0)}
=\frac{a_0b_1-a_1b_0}{q_w(x,0)\,q_w(x,1)},
\]
which is \eqref{eq:group-derivative} because $a_1=g_A\eta_A$, $a_0=g_A(1-\eta_A)$, $b_1=g_B\eta_B$ and $b_0=g_B(1-\eta_B)$. For the bound, $q_{1/2}(x,y)/q_{1-\alpha}(x,y)$ is a ratio of mixtures of $a_y,b_y$ whose weights change by the factors $1/(2\alpha)$ and $1/(2(1-\alpha))$, so for $y=0,1$ it lies in
\[
\Bigl[\tfrac{1}{2\alpha},\ \tfrac{1}{2(1-\alpha)}\Bigr].
\] Taking the logarithm of the ratio for $y=1$ over that for $y=0$ gives the claim.
\end{proof}

\begin{proof}[Proof of Lemma \ref{lemma:class-shift}]
Under $\widetilde P^{(2)}$ each group has mass $1/2$, prevalence $\pi_A$ and cell-conditional distributions $P_c$, so $\widetilde P^{(2)}(X\in\cdot,Y=1)=\pi_AP_+$ and $\widetilde P^{(2)}(X\in\cdot,Y=0)=(1-\pi_A)(\frac12P_{A-}+\frac12P_{B-})$. Adding $\lambda_3^*P_+$ changes only the positives: writing $p^{(2)}$, $p_{\mathrm{fb}}$ and $p_+$ for the densities,
\[
p_{\mathrm{fb}}(x,1)=\bigl((1-\lambda_3^*)\pi_A+\lambda_3^*\bigr)p_+(x)=\tfrac12\,p_+(x),
\qquad
p_{\mathrm{fb}}(x,0)=(1-\lambda_3^*)\,p^{(2)}(x,0),
\]
with $1-\lambda_3^*=1/(2(1-\pi_A))$. Each cell therefore has mass $1/4$ with cell-conditional distribution $P_c$, as for $\frac{1}{4p^*}(P+\sum_cd_cP_c)$. Moreover, \[\Psi_{\widetilde P_{\mathrm{fb}}}-\Psi_{\widetilde P^{(2)}}=\log\frac{1/2}{\pi_A}-\log\frac{1/2}{1-\pi_A}=\log\frac{1}{\rho_A}.\]
We thus complete the proof of the lemma.
\end{proof}
 
\begin{proof}[Proof of Lemma \ref{lemma:syn-bayes}]

Write $\tilde\eta:=\eta_{\widetilde P_{\mathrm{fb}}}$, $\epsilon:=\epsilon_{\mathrm{syn},\mathrm{fb}}$ and $G:=\{h^\eta_{\widetilde P_{\mathrm{fb}}}\neq h^\eta_{\overline P_{\mathrm{fb}}}\}$. By Lemma~\ref{lem:bayes-optimality-fb} for $Q\in\{\widetilde P_{\mathrm{fb}},\overline P_{\mathrm{fb}}\}$ and \eqref{eq:fixed-h-transfer-acc-revised},
\[
\mathbb E_{\widetilde P_{\mathrm{fb}}}\bigl[|2\tilde\eta(X)-1|\,\mathbf 1_G(X)\bigr]
=\Err_{\widetilde P_{\mathrm{fb}}}(h^\eta_{\overline P_{\mathrm{fb}}})-\Err_{\widetilde P_{\mathrm{fb}}}(h^\eta_{\widetilde P_{\mathrm{fb}}})\le2\epsilon .
\]
Since $\widetilde P_{\mathrm{fb}}=\frac14\sum_cP_c$, Condition~2 gives $\widetilde P_{\mathrm{fb}}(|\tilde\eta-\frac12|\le t)\le C_{\mathrm{fb}}t$, so for every $t>0$,
\[
\widetilde P_{\mathrm{fb}}(G)\le\widetilde P_{\mathrm{fb}}\bigl(|\tilde\eta-\tfrac12|\le t\bigr)+\frac{2\epsilon}{2t}\le C_{\mathrm{fb}}\,t+\frac{\epsilon}{t},
\]
and $t=\sqrt{\epsilon/C_{\mathrm{fb}}}$ (or $t\downarrow0$ if $\epsilon=0$) gives $\widetilde P_{\mathrm{fb}}(G)\le2\sqrt{C_{\mathrm{fb}}\epsilon}$. As $\widetilde P_{\mathrm{fb}}(S=s,Y=1)=\frac14$ and $\widetilde P_{\mathrm{fb}}(\cdot\mid S=s,Y=1)=P_{s+}$, we get \[\sum_sP_{s+}(G)=4\widetilde P_{\mathrm{fb}}(X\in G,Y=1)\le8\sqrt{C_{\mathrm{fb}}\epsilon}.\] Finally, \[|\Delta_{\mathrm{EO}}(h^\eta_{\widetilde P_{\mathrm{fb}}})-\Delta_{\mathrm{EO}}(h^\eta_{\overline P_{\mathrm{fb}}})|\le\sum_s|\mathrm{TPR}_s(h^\eta_{\widetilde P_{\mathrm{fb}}})-\mathrm{TPR}_s(h^\eta_{\overline P_{\mathrm{fb}}})|\le\sum_sP_{s+}(G),\] and $C_q=8\sqrt{C_{\mathrm{fb}}}$.
\end{proof}

\begin{proof}[Proof of Lemma \ref{lemma:newconcentrationtwoQ}]
For \(Q=P\), the concentration result follows from the argument leading to (\ref{eq:empirical-analysis}), with failure probability at most \(\delta/2\). For \(Q=\overline P_{\mathrm{fb}}\), we decompose the empirical-gradient deviation into real-data and synthetic-data components. Following the proof of Theorem \ref{thm:adaptive_near_optimality-full}, particularly (\ref{eq:stratified_real_gradient_bound})–(\ref{eq:synthetic_gradient_bound}), the real-data component is controlled uniformly over \(\Theta\) using a covering argument, conditional on the observed cell counts, while the synthetic-data component is controlled conditionally on the original observations using Hoeffding's inequality. The two concentration events are combined using a union bound without requiring independence between the original and synthetic datasets.

For the finite-sample full-balance procedure, each
cell $c\in\mathcal C$ contains $n_c$ original
observations and $n_{\max}-n_c$ synthetic
observations, where
$n_{\max}=\max_{c\in\mathcal C}n_c$.
Throughout this proof,
$\widehat L_{\overline P_{\mathrm{fb}}}$
denotes the empirical loss over the resulting
$n_{\overline P_{\mathrm{fb}}}=4n_{\max}$
observations, rather than an i.i.d.\ sample
drawn directly from $\overline P_{\mathrm{fb}}$.
The corresponding count-based population mixture
differs from $\overline P_{\mathrm{fb}}$ as follows:
\[
\begin{aligned}
&\frac{1}{n_{\overline P_{\mathrm{fb}}}}
\sum_{c\in\mathcal C}
\left[n_cP_c+(n_{\max}-n_c)P_{\mathrm{syn},c}\right]
-\overline P_{\mathrm{fb}}=
\sum_{c\in\mathcal C}
\left(
\frac{n_c}{n_{\overline P_{\mathrm{fb}}}}
-\frac{p_c}{4p^*}
\right)(P_c-P_{\mathrm{syn},c}).
\end{aligned}
\]
Denote $Q_D=\frac{1}{n_{\overline P_{\mathrm{fb}}}}
\sum_{c\in\mathcal C}
\left[n_cP_c+(n_{\max}-n_c)P_{\mathrm{syn},c}\right]$ for brevity.
By the definition of TV distance,
\(
d_{\mathrm{TV}}(Q_D,\overline P_{\mathrm{fb}})
=
\sup_{G}|Q_D(G)-\overline P_{\mathrm{fb}}(G)|.
\)
Substituting the identity gives
\[
\begin{aligned}
d_{\mathrm{TV}}&(Q_D,\overline P_{\mathrm{fb}})
=
\sup_{G}\left|
\sum_{c\in\mathcal C}\left(\frac{n_c}{n_{\overline P_{\mathrm{fb}}}}
-\frac{p_c}{4p^*}\right)
\bigl(P_c(G)-P_{\mathrm{syn},c}(G)\bigr)
\right|\\
&\leq
\sum_{c\in\mathcal C}\left|\frac{n_c}{n_{\overline P_{\mathrm{fb}}}}
-\frac{p_c}{4p^*}\right|
\sup_{G}|P_c(G)-P_{\mathrm{syn},c}(G)|=
\sum_{c\in\mathcal C}\left|\frac{n_c}{n_{\overline P_{\mathrm{fb}}}}
-\frac{p_c}{4p^*}\right|
d_{\mathrm{TV}}(P_c,P_{\mathrm{syn},c}).
\end{aligned}
\]
Assumption~\ref{assumption:improvement-bound-assumption} states that
\(
d_{\mathrm{TV}}(P_c,P_{\mathrm{syn},c})
\leq\epsilon_c.
\)
Therefore,
\[
d_{\mathrm{TV}}\left(
\frac{\sum_c[n_cP_c+(n_{\max}-n_c)P_{\mathrm{syn},c}]}
{n_{\overline P_{\mathrm{fb}}}},
\overline P_{\mathrm{fb}}
\right)
\leq\epsilon_{\mathrm{mix}}.
\]
To control the empirical gradient, let $\mathcal I_c$
denote the indices of the original observations in
cell $c$. Define the real-data empirical-gradient
deviation by
\begin{align*}
E_{\mathrm{real}}(\theta)
:=
\frac{1}{n_P}
\sum_{c\in\mathcal C}
\sum_{i\in\mathcal I_c}
\Big[
&\nabla_\theta\ell(\theta;X_i,Y_i)-\mathbb E_{P_c}
  [\nabla_\theta\ell(\theta;X,Y)]
\Big].
\end{align*}
When \(n_{\overline P_{\mathrm{fb}}}>n_P\), define the
synthetic-data empirical-gradient deviation by
\begin{align*}
E_{\mathrm{syn}}(\theta)
:=
\frac{1}{n_{\overline P_{\mathrm{fb}}}-n_P}
\sum_{c\in\mathcal C}
\sum_{j=1}^{n_{\max}-n_c}
\Big[
&\nabla_\theta\ell(
  \theta;\widetilde X_{c,j},\widetilde Y_{c,j})-\mathbb E_{P_{\mathrm{syn},c}}
  [\nabla_\theta\ell(\theta;X,Y)]
\Big].
\end{align*}
Otherwise, its weighted contribution is zero. Adding and subtracting the population gradient under
$Q_D$, we obtain
\begin{align*}
&\nabla\widehat L_{\overline P_{\mathrm{fb}}}
 \left(\theta^*_{\overline P_{\mathrm{fb}}}\right)
-\nabla L_{\overline P_{\mathrm{fb}}}
 \left(\theta^*_{\overline P_{\mathrm{fb}}}\right)\\
&\quad=
\frac{n_P}{n_{\overline P_{\mathrm{fb}}}}
E_{\mathrm{real}}
 \left(\theta^*_{\overline P_{\mathrm{fb}}}\right)+
\frac{n_{\overline P_{\mathrm{fb}}}-n_P}
     {n_{\overline P_{\mathrm{fb}}}}
E_{\mathrm{syn}}
 \left(\theta^*_{\overline P_{\mathrm{fb}}}\right)+
\nabla L_{Q_D}
 \left(\theta^*_{\overline P_{\mathrm{fb}}}\right)
-\nabla L_{\overline P_{\mathrm{fb}}}
 \left(\theta^*_{\overline P_{\mathrm{fb}}}\right).
\end{align*}

We control these terms separately.
For real-data concentration, conditional on the observed cell counts, the original
observations are independent and follow their respective
cell-conditional distributions. Applying the uniform
covering-number argument leading to (\ref{eq:stratified_real_gradient_bound}) in the proof
of Theorem \ref{thm:adaptive_near_optimality-full} to the pooled original sample, with
$n_P$ in place of $n_{\mathrm{cell,tr}}$, gives
\begin{align*}
&\sup_{\theta\in\Theta}
\|E_{\mathrm{real}}(\theta)\|_2\leq
B\sqrt{
\frac{d\log(8d(3n_P)^d/\delta)}
     {2n_P}}
+\frac{2\sqrt d L_\nabla D_\Theta}{n_P}
\end{align*}
with probability at least $1-\delta/4$.
The uniformity over $\Theta$ allows this bound to be
evaluated at the possibly data-dependent minimizer
$\theta^*_{\overline P_{\mathrm{fb}}}$.
Unlike (\ref{eq:stratified_real_gradient_bound}), the present argument applies Hoeffding's inequality directly to the pooled sum of the \(n_P\) original observations. Consequently, only a union bound over gradient coordinates and covering points is needed, yielding the logarithmic factor \(\log(8d(3n_P)^d/\delta)\).

For synthetic data concentration,
conditional on the original observations, the fitted
generator distributions and
$\theta^*_{\overline P_{\mathrm{fb}}}$
are fixed, while the synthetic observations are mutually
independent. Applying the conditional Hoeffding argument
leading to (\ref{eq:synthetic_gradient_bound}) in the proof of Theorem \ref{thm:adaptive_near_optimality-full} to the single
full-balance allocation yields
\begin{align*}
&\left\|
E_{\mathrm{syn}}
\left(\theta^*_{\overline P_{\mathrm{fb}}}\right)
\right\|_2\leq
B\sqrt{
\frac{d\log(8d/\delta)}
     {2(n_{\overline P_{\mathrm{fb}}}-n_P)}
}
\end{align*}
with conditional probability at least $1-\delta/4$,
whenever $n_{\overline P_{\mathrm{fb}}}>n_P$.
Integrating over the original sample gives the same
unconditional probability bound.

For the mixture discrepancy,
by the bounded-gradient condition and the preceding total variation bound,
\begin{align*}
&\big\|
\nabla L_{Q_D}
 \left(\theta^*_{\overline P_{\mathrm{fb}}}\right)
-\nabla L_{\overline P_{\mathrm{fb}}}
 \left(\theta^*_{\overline P_{\mathrm{fb}}}\right)
\big\|_2\leq
\sqrt d B\,
d_{\mathrm{TV}}(Q_D,\overline P_{\mathrm{fb}})\leq
\sqrt d B\epsilon_{\mathrm{mix}}.
\end{align*}

Combining the three bounds by the triangle inequality
and a union bound, without requiring independence
between the original and synthetic datasets, we obtain
with probability at least $1-\delta/2$,
\begin{align*}
\left\|
\nabla\widehat L_{\overline P_{\mathrm{fb}}}
\left(\theta^*_{\overline P_{\mathrm{fb}}}\right)
\right\|_2&\leq
\frac{n_P}{n_{\overline P_{\mathrm{fb}}}}
\left[
B\sqrt{
\frac{d\log(8d(3n_P)^d/\delta)}
     {2n_P}}
+\frac{2\sqrt d L_\nabla D_\Theta}{n_P}
\right]\\
&\qquad+
\frac{1}
     {n_{\overline P_{\mathrm{fb}}}}
B\sqrt{
\frac{(n_{\overline P_{\mathrm{fb}}}-n_P)d\log(8d/\delta)}
     {2}
}+
\sqrt d B\epsilon_{\mathrm{mix}},
\end{align*}
where we use
$\nabla L_{\overline P_{\mathrm{fb}}}
(\theta^*_{\overline P_{\mathrm{fb}}})=0$
by the interior population optimality condition.

We now convert the preceding gradient bound into a
score approximation bound, following the argument
leading to (\ref{eq:7})--(\ref{eq:empirical-analysis}) in the proof of Theorem \ref{Thm:fairness-improvement}.
By the empirical curvature condition, as used in (\ref{eq:7}), we have
\begin{align*}
&\mu_{\overline P_{\mathrm{fb}}}
\left\|
\widehat\theta_{\overline P_{\mathrm{fb}}}
-\theta^*_{\overline P_{\mathrm{fb}}}
\right\|_2^2\leq
\Big\langle
\nabla\widehat L_{\overline P_{\mathrm{fb}}}
\left(\theta^*_{\overline P_{\mathrm{fb}}}\right)
-
\nabla\widehat L_{\overline P_{\mathrm{fb}}}
\left(\widehat\theta_{\overline P_{\mathrm{fb}}}\right),
\theta^*_{\overline P_{\mathrm{fb}}}
-\widehat\theta_{\overline P_{\mathrm{fb}}}
\Big\rangle.
\end{align*}
Since the empirical minimizer is interior,
$\nabla\widehat L_{\overline P_{\mathrm{fb}}}
(\widehat\theta_{\overline P_{\mathrm{fb}}})=0$.
Therefore, by Cauchy--Schwarz, as in (\ref{eq:8}),
\begin{align*}
&\mu_{\overline P_{\mathrm{fb}}}
\left\|
\widehat\theta_{\overline P_{\mathrm{fb}}}
-\theta^*_{\overline P_{\mathrm{fb}}}
\right\|_2^2\leq
\left\|
\nabla\widehat L_{\overline P_{\mathrm{fb}}}
\left(\theta^*_{\overline P_{\mathrm{fb}}}\right)
\right\|_2
\left\|
\widehat\theta_{\overline P_{\mathrm{fb}}}
-\theta^*_{\overline P_{\mathrm{fb}}}
\right\|_2.
\end{align*}
Consequently,
\[
\left\|
\widehat\theta_{\overline P_{\mathrm{fb}}}
-\theta^*_{\overline P_{\mathrm{fb}}}
\right\|_2
\leq
\frac{1}
{\mu_{\overline P_{\mathrm{fb}}}}\left\|
\nabla\widehat L_{\overline P_{\mathrm{fb}}}
\left(\theta^*_{\overline P_{\mathrm{fb}}}\right)
\right\|_2.
\]

By the definition of $e_n$ in Appendix~\ref{sec:fullbalancingfullstatement}, the right-hand side of the gradient bound
above equals $(\mu_{\overline P_{\rm fb}}/L)\,e_n(\overline P_{\rm fb})$, so by the Lipschitz
property of the score, $\sup_x|\widehat f_{\overline P_{\rm fb}}(x)-f^*_{\overline P_{\rm fb}}(x)|
\le L\|\widehat\theta_{\overline P_{\rm fb}}-\theta^*_{\overline P_{\rm fb}}\|_2\le e_n(\overline P_{\rm fb})$, which proves the last claim.

Next, applying score Lipschitzness and the
population-score approximation condition as in the derivation
of (\ref{eq:empirical-analysis}), yields
\begin{align*}
&\sup_x
\left|
\widehat f_{\overline P_{\mathrm{fb}}}(x)
-\eta_{\overline P_{\mathrm{fb}}}(x)
\right|\leq
\sup_x
\left|
\widehat f_{\overline P_{\mathrm{fb}}}(x)
-f^*_{\overline P_{\mathrm{fb}}}(x)
\right|
+
\sup_x
\left|
f^*_{\overline P_{\mathrm{fb}}}(x)
-\eta_{\overline P_{\mathrm{fb}}}(x)
\right|\\
&\qquad\qquad\leq
L\left\|
\widehat\theta_{\overline P_{\mathrm{fb}}}
-\theta^*_{\overline P_{\mathrm{fb}}}
\right\|_2
+
\epsilon_{\mathrm{cal},\overline P_{\mathrm{fb}}}\leq
\epsilon_{\mathrm{cal},\overline P_{\mathrm{fb}}}
+
\frac{L\left\|
\nabla\widehat L_{\overline P_{\mathrm{fb}}}
\left(\theta^*_{\overline P_{\mathrm{fb}}}\right)
\right\|_2}
{\mu_{\overline P_{\mathrm{fb}}}}
\leq
\Delta_{\overline P_{\mathrm{fb}}}.
\end{align*}

This bound holds with probability at least
$1-\delta/2$ by the preceding concentration
argument. For $Q=P$, applying (\ref{eq:empirical-analysis}) with
$\delta$ replaced by $\delta/2$ gives
\(
\sup_x|\widehat f_P(x)-\eta_P(x)|
\leq \Delta_P
\)
with probability at least $1-\delta/2$.
A union bound therefore establishes the
conclusion simultaneously for
$Q\in\{P,\overline P_{\mathrm{fb}}\}$ with
probability at least $1-\delta$.
\end{proof}

\begin{proof}[Proof of Lemma \ref{lem:bayes-optimality-fb}]
Given $X=x$, predicting $1$ errs with probability $1-\eta_Q(x)$ and predicting $0$ with probability $\eta_Q(x)$. The classifier $h^\eta_Q$ picks the smaller of the two. Hence the conditional errors of $h$ and $h^\eta_Q$ coincide where $h(x)=h^\eta_Q(x)$ and otherwise differ by $|2\eta_Q(x)-1|$. Taking the expectation over $X$ gives the identity.
\end{proof}

\end{sloppypar}

\end{document}